\documentclass[11pt,a4paper]{article}
\usepackage[T1]{fontenc}
\usepackage[margin=2.5cm]{geometry}
\usepackage{amsmath,amssymb}
\usepackage{graphicx}
\usepackage{booktabs}
\usepackage{longtable}
\usepackage{array}
\usepackage{enumitem}
\usepackage{float}
\usepackage{caption}
\usepackage{xcolor}
\usepackage{fancyhdr}
\usepackage[colorlinks=true,linkcolor=blue,citecolor=blue,urlcolor=blue]{hyperref}

\setlist{nosep,leftmargin=*}
\title{\bfseries A Comprehensive Framework for All Stages of Large-Model Training: From Data and Tokenization, through Pretraining, to Post-Training Alignment (2017--2026)}
\author{Xiangjing An\textsuperscript{1,*} \quad Zhibin Jiao\textsuperscript{1} \quad Aiting Chen\textsuperscript{1}\\[2pt]
\small \textsuperscript{1}Independent Researcher Team \\
\small \textsuperscript{*}Corresponding author: \href{mailto:axjing.ai@gmail.com}{axjing.ai@gmail.com}
\thanks{Emails: Zhibin Jiao (jzb1631@163.com), Aiting Chen (ChengAiTiny@163.com).}}
\date{2026}

\begin{document}
\maketitle

\begin{abstract}
The capabilities of large language models (LLMs) and multimodal large language models (MLLMs) arise from the synergy of the entire pipeline---"data and tokenization $\rightarrow$ pretraining $\rightarrow$ post-training (SFT $\rightarrow$ preference alignment / reinforcement learning $\rightarrow$ continual optimization)"---rather than from the contribution of any single stage. However, existing surveys mostly adopt a single-stage or single-technique-family perspective: data selection [392], pretraining data construction [393], discrete tokenization [394], post-training [395], RL post-training [396][399], preference optimization [397][398], and multimodality [402][403][404][405] each form their own systems, lacking a unified stage division and a horizontally comparable basis. This paper proposes a unified stage division and a six-dimensional analytical framework (data, objective function, optimization strategy, system constraints, evaluation, and risk and governance), and on this basis systematically reviews 395 core references, covering 2017--2026 (with a few foundational earlier works), and presents a conceptual map of cross-stage effects---the forward constraint that pretraining imposes on post-training and the reverse empowerment that post-training exerts on data / pretraining. We find the following: first, the comparability of scaling laws (Kaplan [119] and Chinchilla [16]) depends on data quality and data mixture, a long-neglected variable; second, preference alignment (RLHF [15], DPO [32]) and reasoning RL (verifiable reward) are redefining the boundaries of "post-training," but the debate between Pass@1 and Pass@K has yet to yield a systematic conclusion; third, multimodal training poses structural challenges to the pure-text paradigm in terms of tokenizer, objective, and evaluation. The full text provides a cross-stage effects diagram, a method-family map, a practical decision tree, and $\geq$10 verifiable open problems, and reserves for each stage an extensible skeleton for focused surveys (sub-surveys). As of the retrieval date, peer-reviewed evidence (A+B) accounts for about 79\% of the 395 references.

\end{abstract}

\tableofcontents
\newpage

\section{Introduction}

\subsection{Motivation: Why a "General Framework for All Stages" Is Needed}

Conclusion first: the stages of the full training pipeline are not independent of one another; they exhibit strong coupling and feedback, and therefore any single-stage survey cannot fully explain where capabilities come from. The objective function and data mixture determined during pretraining directly set the attainable ceiling of "alignment" in the post-training stage---a model pretrained on a low-quality, narrowly distributed corpus will struggle to recover the knowledge and reasoning abilities it never acquired, no matter how extensive the instruction tuning. Conversely, the failure modes exposed during post-training (such as over-refusal, hallucination, and format non-compliance) in turn drive revisions to the data side and to the pretraining recipe (such as curriculum, two-stage data mixture, and tokenizer design). This bidirectional relationship of "forward constraint + reverse empowerment" is the starting point that distinguishes this paper from existing stage-specific surveys.

The rapid overlapping of training paradigms over the past three years has further intensified the risk of fragmentation. Taking 2022 as the dividing line, the Chinchilla training-type scaling law [16] re-estimated the data-to-parameter ratio, making "more tokens, better data" the mainstream orientation; subsequently, instruction tuning [163][172], RLHF [15], DPO [32], and reasoning RL with verifiable rewards expanded "post-training" from a lightweight finishing step into a capability-shaping stage on a par with pretraining [395][396]. At the same time, multimodal models [402][404] reopened questions of tokenization, objective functions, and evaluation. If these developments are read separately, researchers easily arrive at mutually incompatible experiences and conclusions; only by comparing them under a unified set of stages and a unified set of dimensions can one judge which are universal regularities and which are products of specific recipes.

The positioning of this paper is that of a "general framework" rather than a "most comprehensive survey": we strive to provide, for each stage, a structure sufficient to guide practice (definition---boundaries---method families---controversies---recommendations), with evidence strength explicitly annotated, rather than attempting to exhaust every variant of a given technique. The value of this approach lies in the following: when readers need to delve into a particular subfield, this paper offers an extensible skeleton and a unified terminology that can be expanded into focused surveys, rather than replacing dedicated surveys.

\subsection{Differences from Existing Surveys}

Existing high-quality surveys cover multiple "localities" of this topic but lack a unified framework that stitches them together. On the data dimension, Albalak et al. [392] systematically reviewed data selection methods, with an emphasis on pretraining and including chapters on SFT / alignment / ICL, but they do not connect the full chain of "data$\rightarrow$objective$\rightarrow$system$\rightarrow$evaluation$\rightarrow$governance," nor do they address multimodal tokenization; Parmar et al. [393] focused on full-pipeline ablations of pretraining data construction, lacking post-training and system constraints; Jia et al. [394] focused on the mechanisms and applications of discrete tokenizers without discussing the coupling between tokenization and training stages. On the pretraining dimension, existing work either takes the scaling law as its main thread or focuses on system efficiency, and rarely discusses objective functions, architecture, optimization stability, and distributed systems side by side. On the post-training dimension, Tie et al. [395] take five post-training paradigms (fine-tuning, alignment, reasoning, efficiency, and integration) as their sole main axis, without connecting backward to the pretraining empowerment relationship; the RL post-training surveys [396][399] and the reasoning RL surveys [400][401] cover intra-algorithm details in greater depth but have a weaker cross-stage constraint and system / governance perspective; the DPO survey series [397][398] and the reward model survey [408] focus on the preference optimization family. On the multimodal dimension, MM-LLMs [402], the data-centric multimodal survey [403], Omni-MLLM [404], and the efficient multimodal survey [405] each form their own systems and are not compared with pure-text training along the same six dimensions. In addition, although works such as data-mixture inference [406] and two-stage pretraining [407] touch upon cross-stage mixtures, they remain single-point methodological studies.

By contrast, the differences of this paper can be summarized in four points: a unified stage division (explicitly distinguishing the broad / narrow senses of "post-training"); a unified six-dimensional analytical framework with mutually orthogonal dimensions; an explicit cross-stage influence diagram (forward constraint and reverse empowerment); and a consistent evidence grading and Cited-by-evidence conclusion-strength annotation throughout. Accordingly, this paper does not compete with the above surveys in coverage; rather, it provides a "meta-framework" that can align each of them to the same coordinate system.

\subsection{List of Contributions}

The main contributions of this paper are as follows:

\begin{enumerate}[leftmargin=*,nosep]
\item \textbf{Unified stage division and terminological conventions}: We specify five stages---"data and tokenization$\rightarrow$pretraining$\rightarrow$post-training (SFT$\rightarrow$preference alignment/RL$\rightarrow$continual optimization)"---and explicitly clarify the broad and narrow ambiguities of "post-training" (Chapter 2).
\item \textbf{Six-dimensional analytical framework}: We characterize any training method along six orthogonal dimensions---data, objective function, optimization strategy, system constraints, evaluation, and risk and governance---enabling textual and multimodal methods, and pretraining and post-training, to be compared horizontally (Chapters 3--4, Chapter 11).
\item \textbf{Cross-stage influence diagram}: We present the explicit relationships of forward constraint (pretraining$\rightarrow$post-training) and reverse empowerment (post-training/evaluation$\rightarrow$data and pretraining design), with evidence strength annotated (Chapter 11).
\item \textbf{Uniformly graded comparison and conclusion strength}: We grade the 391 core references from A to D, and every comparison reports the parameter count, token count, and compute basis, avoiding direct comparisons of numbers across datasets (throughout the book, Appendices A/E).
\item \textbf{Practical decision tree and failure-mode checklist}: We provide candidate routes and trade-offs oriented toward budget, modality, context length, and compliance constraints (Chapter 13).
\item \textbf{$\geq$10 verifiable open problems}: We transform field disagreements into experimentally testable research questions (Chapters 14--15, with the complete pool in Appendix F).
\item \textbf{Extensible focused-survey skeleton}: Each stage chapter can be independently expanded into a focused survey and reuses a uniform writing template (Appendices A--F).
\end{enumerate}

\subsection{Scope, Audience, and Structure}

\textbf{Scope}: This paper focuses on the training process of large language models and multimodal large language models, with retrieval up to 2026-10-24. Included: data and tokenization, pretraining, instruction tuning (SFT/PEFT), preference alignment and reinforcement learning, continual optimization (continued pretraining, long-context extension, distillation/pruning/quantization, model editing and unlearning), a cross-cutting view of multimodal training, and evaluation and governance. \textbf{Excluded}: inference and serving deployment engineering, industry implementation cases, traditional non-LLM training, pure reinforcement learning theory, and hardware circuitry and cognitive science mechanisms.

\textbf{Audience}: Researchers and engineers engaged in model training, graduate students who need to understand training trade-offs across stages, and readers outside the field who are interested in the evolution of training paradigms. We assume readers possess basic knowledge of deep learning and Transformers (if needed, see [108]).

\textbf{Structure}: Chapter 2 presents the glossary and an overview of training stages; Chapter 3 formalizes the language and multimodal modeling, stage splitting, and evaluation notation; Chapter 4 presents the taxonomy; Chapters 5--9 discuss data and tokenization, pretraining, SFT, preference alignment/RL, and continual optimization, respectively; Chapter 10 discusses multimodal training from a cross-cutting perspective; Chapter 11 makes cross-stage and cross-modal comparisons; Chapter 12 reviews datasets and evaluation; Chapter 13 provides a practical guide; Chapters 14--16 discuss challenges, future directions, and conclusions; and Appendices A--F provide the literature matrix, hyperparameter and compute comparisons, reproducibility checklists, retrieval logs, and focused-survey outlines.

\subsection{Methodology and Evidence Conventions}

Conclusion first: the credibility of this survey comes not from its coverage but from the explicit binding of "conclusion---evidence"; to this end, this paper establishes verifiable conventions at three levels.

\textbf{Retrieval and screening.} The methodology follows the PRISMA process: we searched arXiv and Semantic Scholar with 11 topic clusters, identified approximately 640 records, screened them down to approximately 470 according to inclusion/exclusion criteria, and finally verified 391 core references (deduplicating 59 duplicate identifiers). A few foundational works predate the 2017--2026 window. The search terms, time periods, and inclusion/exclusion criteria are summarized in Appendix E; the reference-assembly scripts and metadata cache are provided as supplementary material and support offline recomputation.

\textbf{Evidence grading (A--D).} This paper annotates each cited work with its source reliability: \textbf{A} for peer-reviewed with public reproducibility artifacts; \textbf{B} for peer-reviewed; \textbf{C} for preprint/technical report/unconfirmed publication status; \textbf{D} for non-peer-reviewed sources such as blogs and model cards. In the current literature base, A+B account for approximately 79\%, C for approximately 21\%, with no ungraded items. Wherever judgment of conclusion strength is involved, this paper refers back to this grading (distribution in Appendix E).

\textbf{Comparison discipline.} Whenever works are compared across studies, the bases of parameter count $N$, training token count $D$, and compute $C$ must be given; when unavailable, the marker \texttt{[to be verified]} is used, and no conclusion of numerical superiority or inferiority is drawn. Citation verbs are strictly distinguished---"propose," "claim," "prove," and "report"---to reflect the evidential status of the original literature; technical reports and blogs are annotated with "non-peer-reviewed + version date," and marketing conclusions with "(claimed)." The purpose of these conventions is to allow readers to independently judge which conclusion can be safely reused and which can serve only as a hypothesis (see Chapter 12 and Appendix E for details).

\subsection{Reading Guide}

This paper can be read in three ways. \textbf{By stage}: read Chapters 5--9 in sequence to obtain the full training workflow; \textbf{by problem}: first read the practical guide in Chapter 13 to locate decision points, then return to the corresponding chapters for evidence and costs; \textbf{by comparison}: first read the horizontal comparison in Chapter 11 and the open problems in Chapter 14, then delve deeper as needed.

The book includes five core figures and tables: Figure 1, the overall diagram of the full training workflow (Chapter 2); Figure 2, the stage relationship table (Chapter 4); Figure 3, the method-family map (with four colors for classic/mainstream/frontier/controversial, Chapter 4); Figure 4, the practical decision tree (Chapter 13); and Figure 5, the evidence map and open-problems table (Chapter 11). The "representative work table" and "key controversies" sections of each chapter are separable minimal reading units; Appendices A--F provide the raw matrices, templates, and focused-survey skeleton for readers who wish to further verify or expand.

\textbf{How to judge the strength of a conclusion.} When reading, we recommend a three-step check: \textcircled{\small 1} Does the conclusion have A/B-level evidence, or does it come only from a single technical report (C)? \textcircled{\small 2} Does the comparison give the $(N, D, C)$ bases, or does it compare numbers directly across datasets? \textcircled{\small 3} Does the conclusion distinguish "mechanistic hypotheses" from "recipe details"---the former are generally transferable, while the latter are highly dependent on specific settings. A conclusion that passes all three steps can be safely reused; any that fails can serve only as a hypothesis pending verification. The "traffic-light" interpretation method given later in Chapter 12 is precisely the concretization of this check in the evaluation context.

\section{Glossary and Overview of Training Stages}

\subsection{Glossary}

Conclusion first: terminological ambiguity is the primary cause of failures in cross-stage comparison, so this section fixes the vocabulary used throughout the survey. The core terms are as follows:

\begin{itemize}[leftmargin=*,nosep]
\item \textbf{Autoregressive language modeling (AR / NTP)}: a training paradigm whose objective is to "predict the next token"; it is the dominant objective for current decoder-only LLMs [121].
\item \textbf{Masked language modeling (MLM)}: randomly masks input tokens and predicts the masked ones, exemplified by BERT [47].
\item \textbf{Sequence-to-sequence (seq2seq)}: unifies encoder--decoder into "text-to-text", exemplified by T5 [49].
\item \textbf{Tokenizer}: a component that splits raw text into subword units; representative methods include BPE [1] and Unigram / SentencePiece [5].
\item \textbf{Vocabulary}: a finite set of tokens; its size and coverage determine the boundary of fairness for multilingual and multimodal settings.
\item \textbf{Pretraining}: the stage that acquires general representations and knowledge from large-scale unlabeled corpora via self-supervised objectives.
\item \textbf{Supervised fine-tuning (SFT) / instruction tuning (IFT)}: supervised training on "instruction--response" pairs so that the model follows instructions [163][172].
\item \textbf{Parameter-efficient fine-tuning (PEFT)}: a family of fine-tuning methods that update only a small number of parameters (e.g., LoRA [12], QLoRA [30], Adapter, Prefix/P-Tuning [161][164]).
\item \textbf{Post-training}: in the \textbf{broad sense}, all training after pretraining, including SFT, preference alignment/RL, and continual optimization; in the \textbf{narrow sense}, only the preference alignment and reinforcement learning after SFT [395][396]. This survey defaults to the broad sense and explicitly marks every use of the narrow sense.
\item \textbf{Reinforcement learning from human feedback (RLHF)}: first trains a reward model, then optimizes the policy with RL (e.g., PPO) [15].
\item \textbf{Reinforcement learning from AI feedback (RLAIF)}: uses AI (models) to replace part of the human annotation in providing feedback.
\item \textbf{Direct preference optimization (DPO)}: a method that optimizes the policy directly from preference pairs without an explicit reward model or RL loop [32].
\item \textbf{Reinforcement learning with verifiable rewards (RLVR)}: applies RL with rule-based rewards on tasks such as mathematics and code whose answers can be verified automatically.
\item \textbf{Scaling laws}: empirical regularities of loss with respect to parameter count, data volume, and compute, exemplified by Kaplan et al. [119] and Hoffmann et al. (Chinchilla) [16].
\item \textbf{Mixture-of-experts (MoE)}: replaces the dense feed-forward with sparsely activated expert subnetworks, exemplified by Switch Transformer [9].
\item \textbf{Length extrapolation}: training short and inferring long, exemplified by ALiBi [129], RoPE and its extensions [11].
\item \textbf{Multimodal large language model (MLLM)}: a large model that receives and (potentially) generates multiple modalities; representative architectures include CLIP [10], Flamingo [17], BLIP-2 [265], and LLaVA [28].
\item \textbf{Evidence grading (A--D)}: this survey's annotation of the reliability of literature sources, where A = peer-reviewed with public reproducibility artifacts, B = peer-reviewed, C = preprint/technical report/unconfirmed status, D = blog/model card/non-peer-reviewed source.
\end{itemize}

\subsection{Overall Map of the Training Pipeline}

\begin{figure}[htbp]
\centering
\includegraphics[width=\linewidth]{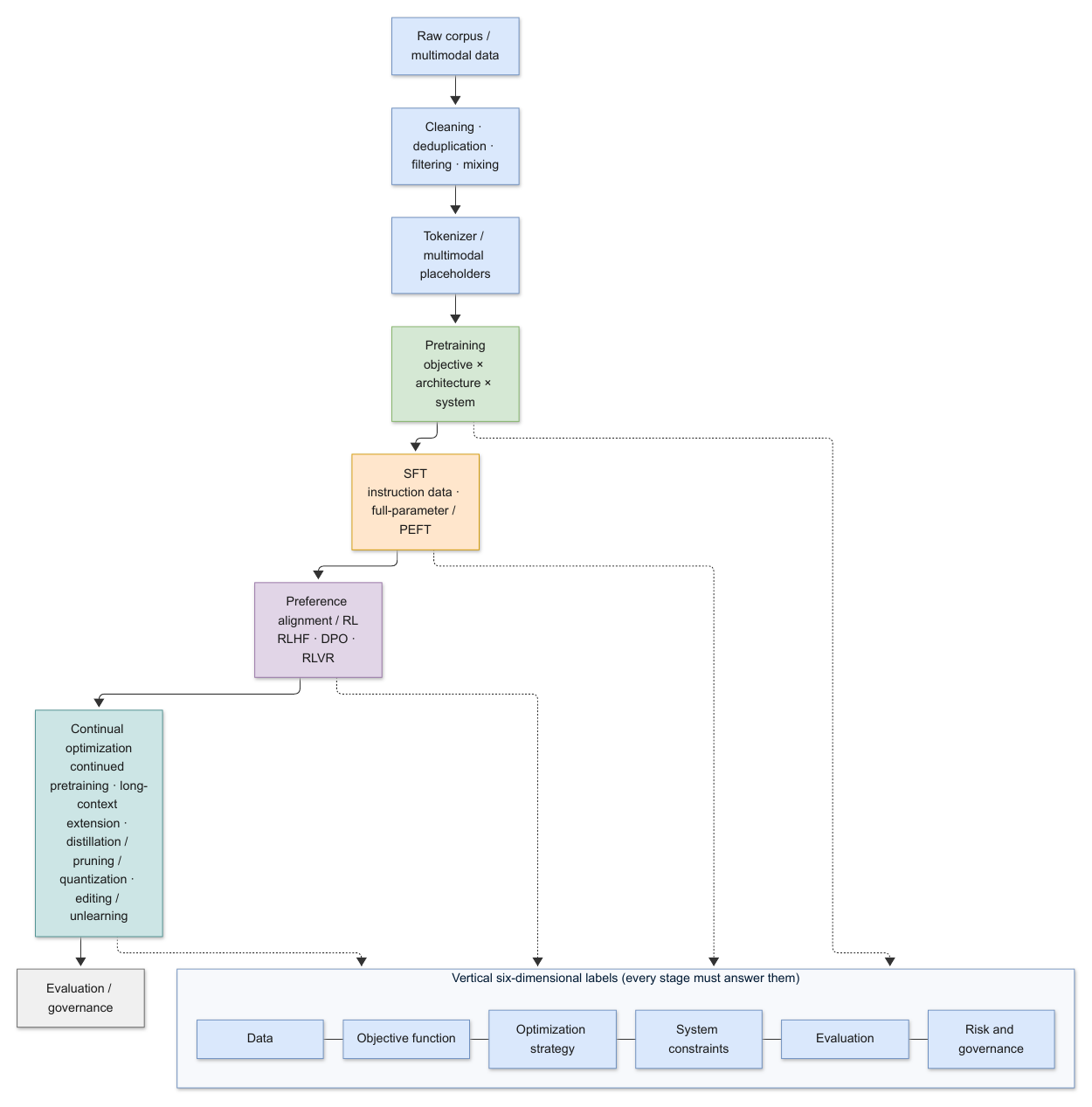}
\caption*{Figure 1: The full training pipeline---a linear backbone plus vertical six-dimensional labels (applied across all stages)}
\end{figure}

The key message of this figure is that the six dimensions are not exclusive to any stage but are questions that every stage must answer; stages interface with one another through "artifacts" (data $\rightarrow$ token sequence $\rightarrow$ base model $\rightarrow$ aligned model $\rightarrow$ deployment-optimized model). When reading the survey, any method can be located as "a choice made along one dimension of one stage".

\subsection{The Semantic Ambiguity of "Post-Training" and the Convention Adopted Here}

In practice the term "post-training" has two usages, and mixing them leads to inaccurate literature comparisons. The \textbf{broad} usage (adopted by [395][396]) refers to all optimization after pretraining, including SFT; the \textbf{narrow} usage refers only to the preference alignment and reinforcement learning after SFT. The difference between the two is not a word game: if SFT is counted as post-training, the "share of compute spent on post-training" is significantly inflated, and the causal contributions of SFT and preference alignment become hard to separate. This survey adopts the following convention: the broad sense is used by default, but SFT is listed separately as Chapter 7 and distinguished from preference alignment/RL (Chapter 8); whenever the "post-training" conclusions of others are cited, their semantic scope is first determined before being included in comparisons, and annotated in the text when necessary. This convention runs through Chapters 7--9 and 11.

\subsection{Overview of Training Stages (Six-Dimension $\times$ Stage Matrix)}

Conclusion first: crossing the stage definitions of Section 1 with the six dimensions of Section 3 yields the "backbone table" of this survey---it serves both as a structural map of the whole book and as the minimal common framework for horizontal comparison.

\begingroup\scriptsize\setlength{\tabcolsep}{4pt}
\noindent\textbf{Table 1: The six-dimension $\times$ stage matrix (read stages across rows, dimensions down columns)}\par\medskip
\begin{longtable}{@{}>{\raggedright\arraybackslash}p{0.1193\textwidth}>{\raggedright\arraybackslash}p{0.1988\textwidth}>{\raggedright\arraybackslash}p{0.1193\textwidth}>{\raggedright\arraybackslash}p{0.1193\textwidth}>{\raggedright\arraybackslash}p{0.1093\textwidth}>{\raggedright\arraybackslash}p{0.1193\textwidth}>{\raggedright\arraybackslash}p{0.1093\textwidth}@{}}
\toprule
Stage \textbackslash{} Dimension & Data & Objective Function & Optimization Strategy & System Constraints & Evaluation & Risk and Governance \\
\midrule
\endhead
Data and tokenization & Collection/\allowbreak{}cleaning/\allowbreak{}deduplication/\allowbreak{}mixing & -\allowbreak{}-\allowbreak{}-\allowbreak{} & -\allowbreak{}-\allowbreak{}-\allowbreak{} & Vocabulary size vs throughput & Compression rate,\allowbreak{} coverage,\allowbreak{} fairness & Copyright,\allowbreak{} privacy,\allowbreak{} toxicity \\
Pretraining & Trillions of tokens,\allowbreak{} mixture and curriculum & NTP/\allowbreak{}MLM/\allowbreak{}denoising/\allowbreak{}contrastive & Scaling laws,\allowbreak{} learning rate,\allowbreak{} numerical precision & 3D parallelism,\allowbreak{} ZeRO/\allowbreak{}FSDP & Perplexity,\allowbreak{} downstream zero/\allowbreak{}few-\allowbreak{}shot & Data provenance,\allowbreak{} energy consumption \\
SFT & Instruction-\allowbreak{}-\allowbreak{}response pairs,\allowbreak{} templates & Conditional LM +\allowbreak{} loss masking & Full-\allowbreak{}parameter /\allowbreak{} PEFT & Multi-\allowbreak{}turn dialogue GPU memory & Instruction following,\allowbreak{} format & Synthetic data bias \\
Preference alignment /\allowbreak{} RL & Preference pairs,\allowbreak{} rewards,\allowbreak{} verifiable environments & BT loss,\allowbreak{} KL constraint,\allowbreak{} GRPO & PPO stability,\allowbreak{} rollout cost & Sampling throughput & Win rate,\allowbreak{} RewardBench & Jailbreak,\allowbreak{} sycophancy,\allowbreak{} reward hacking \\
Continual optimization & Domain/\allowbreak{}long-\allowbreak{}context data & Continued training,\allowbreak{} distillation,\allowbreak{} quantization objectives & Low-\allowbreak{}rank,\allowbreak{} merging algorithms & Deployment latency and GPU memory & Retention,\allowbreak{} catastrophic forgetting & Unlearning ethics,\allowbreak{} licensing \\
\bottomrule
\end{longtable}
\endgroup

The key to reading this table is to "read stages horizontally and dimensions vertically": any new method is a filling-in or an improvement of one cell in the table. The intentional blanks in the matrix (the "objective" and "optimization" of the data stage) indicate that the dimension is not an independent problem at that stage rather than an omission; Chapter 4 builds on this to give finer definitions of the classification dimensions.

\subsection{Checklist of Common Terminological Ambiguities}

The following ambiguities repeatedly cause comparison errors in practice; this survey unifies the conventions and recommends that readers annotate them as well when citing:

\begin{itemize}[leftmargin=*,nosep]
\item \textbf{"Post-training"}: the broad sense includes SFT, the narrow sense does not (Section 2.3).
\item \textbf{"Alignment"}: sometimes used loosely for all post-training; in this survey it refers only to the optimization toward human values and safety in the preference/RL stage.
\item \textbf{"Parameter count"}: for MoE, total parameters and activated parameters must be distinguished; otherwise the model is not comparable to dense models [9][151].
\item \textbf{"Number of tokens"}: whether multilingual/code/multimodal tokens are included and whether the figure is post-deduplication directly affect conclusions about data scale.
\item \textbf{"Capability vs behavior"}: capability refers to the knowledge and skills that can be elicited by zero-shot/few-shot prompting, while behavior refers to output style and compliance; post-training mainly changes behavior and indirectly affects the accessibility of capability (running through Chapter 11).
\item \textbf{"Safety / alignment tax"}: safety alignment improves refusal and honesty but may impair general capabilities; the "tax" is this capability loss (Chapters 8 and 14).
\end{itemize}

The purpose of fixing these conventions is not to pursue literal uniformity but to expose comparisons of "same name, different meaning" explicitly, so that they can be corrected item by item.

\subsection{Quick-Reference Table for Terms and Evidence Grading}

To facilitate lookup-style reading, this section condenses the core concepts used repeatedly throughout the survey into a single quick-reference table:

\begingroup\small\setlength{\tabcolsep}{6pt}
\noindent\textbf{Table 2: Quick reference for terminology and evidence grading}\par\medskip
\begin{longtable}{@{}>{\raggedright\arraybackslash}p{0.2977\textwidth}>{\raggedright\arraybackslash}p{0.4330\textwidth}>{\raggedright\arraybackslash}p{0.2165\textwidth}@{}}
\toprule
Concept & One-\allowbreak{}sentence definition & Key chapters \\
\midrule
\endhead
Five stages & Data and tokenization $\rightarrow$ pretraining $\rightarrow$ SFT $\rightarrow$ preference alignment/\allowbreak{}RL $\rightarrow$ continual optimization & Chapters 5-\allowbreak{}-\allowbreak{}9 \\
Six dimensions & Data,\allowbreak{} objective,\allowbreak{} optimization,\allowbreak{} system,\allowbreak{} evaluation,\allowbreak{} governance & Sections 2.4 and 4.2 \\
Capability vs behavior & Capability =\allowbreak{} elicitable knowledge and skills; behavior =\allowbreak{} output style and compliance & Sections 3.2 and Chapter 11 \\
Forward constraint & Earlier stages limit the attainable upper bound of later stages & Sections 3.2 and 11.5 \\
Reverse empowerment & Post-\allowbreak{}training/\allowbreak{}evaluation failures feed back into data and pretraining design & Chapters 11 and 13 \\
Scaling laws & Empirical regularities of loss with respect to $N, D, C$ (\allowbreak{}Kaplan/\allowbreak{}Chinchilla) & Section 6.3 \\
Alignment tax & Loss of general capabilities caused by safety alignment & Chapters 8 and 14 \\
Reward hacking & The policy exploits loopholes in the proxy reward instead of completing the true goal & Section 8.8 \\
Evidence grading & A (\allowbreak{}peer-\allowbreak{}reviewed +\allowbreak{} reproducibility artifacts)/\allowbreak{}B (\allowbreak{}peer-\allowbreak{}reviewed)/\allowbreak{}C (\allowbreak{}preprint\textperiodcentered{}report)/\allowbreak{}D (\allowbreak{}blog) & Appendix E \\
Traffic-\allowbreak{}light reading & Green (\allowbreak{}reusable)/\allowbreak{}Yellow (\allowbreak{}caution needed)/\allowbreak{}Red (\allowbreak{}not comparable) & Section 12.11 \\
\bottomrule
\end{longtable}
\endgroup

The table is meant to be used as follows: when expressions such as "strength of conclusion" or "insufficient evidence" appear later, first return to this table to locate their definitions, then judge whether they apply to the setting you care about. Terminological stability is a precondition for valid cross-stage comparison, and is also why this survey places Chapter 2 before the methodological chapters.

\section{Background and Formalization}

\subsection{Formalization of Language Modeling and Multimodal Modeling}

To state the conclusion first: different objective functions (NTP, MLM, denoising, contrastive, matching) are essentially different choices regarding "where the supervision signal comes from and over which tokens/modalities the loss is computed"; understanding this point is a prerequisite for cross-stage comparison.

\textbf{Autoregressive language modeling (NTP)}. Given a token sequence $x = (x_1, \dots, x_T)$, minimize

\[
\mathcal{L}_{\text{NTP}}(\theta) = -\sum_{t=1}^{T} \log p_\theta(x_t \mid x_{<t}).
\]

The GPT series [121][141] and mainstream decoder-only models all adopt this objective. Its advantage is that the objective is consistent with inference (autoregressive generation); its disadvantage is that it exploits only unidirectional context.

\textbf{Masked language modeling (MLM)}. Randomly masking a set of positions $M$, minimize $\mathcal{L}_{\text{MLM}} = -\sum_{t \in M} \log p_\theta(x_t \mid x_{\setminus M})$. BERT [47] uses it to obtain bidirectional representations, but the objective is inconsistent with generation, so it has since been used mostly for understanding-type tasks or as an encoder component. RoBERTa [48] shows that the effectiveness of MLM depends heavily on training details (dynamic masking, longer training, removing NSP).

\textbf{Sequence-to-sequence (seq2seq)}. T5 [49] unifies all tasks into "text-to-text"; the objective is still conditional NTP, but it distinguishes an encoder and a decoder.

\textbf{Contrastive and matching objectives}. Multimodal alignment often employs contrastive learning (InfoNCE-type) and image-text matching. CLIP [10] uses a contrastive objective to align image and text embeddings; Flamingo [17] connects a frozen or semi-frozen visual encoder with a language model through cross-attention; LLaVA [28] uses a projector to map visual features into the LLM's token space before performing (multimodal) NTP. It can be seen that multimodality did not invent an entirely new loss; rather, it combines NTP/contrastive/matching in cross-modal settings, which determines that the core issue of Chapter 10 remains the consistency problem of "objective--data--alignment".

\textbf{Denoising and multi-task objectives}. The BART family unifies understanding and generation via denoising autoencoding (reconstructing after shuffling/masking), and T5 [49] compresses multiple tasks into a unified format via "text-to-text". The value of such objectives lies in merging heterogeneous tasks into a single loss, at the cost of introducing an additional variable---task prefixes and format design---which is precisely the precursor of the "template and loss masking" issue in Chapter 7.

\textbf{A summary of objective functions}. The above objectives can be arranged along a spectrum according to the "source of the supervision signal": purely self-supervised (NTP/MLM/denoising) $\rightarrow$ weakly supervised contrast (image-text contrast) $\rightarrow$ strongly supervised (instructions, preferences, verifiable rewards). The further along the spectrum, the closer the signal is to the target behavior, but the higher the acquisition cost and the more easily bias is injected. The controversies in the post-training chapters (synthetic data bias, reward hacking) are essentially all the cost of "a signal that is both strong and biased".

\subsection{Formal Segmentation of Training Stages and Notation System}

This paper uses a unified quadruple $\langle \mathcal{D}, \mathcal{L}, \mathcal{O}, \mathcal{C} \rangle$ to describe any training stage: $\mathcal{D}$ is the data distribution (including mixture and filtering), $\mathcal{L}$ is the loss/reward, $\mathcal{O}$ is the optimizer and schedule, and $\mathcal{C}$ is the compute and system constraints. Accordingly, the five stages can be written as:

\[
\text{Data} \xrightarrow{\mathcal{D}} \text{Pretrain} \xrightarrow{\theta_0} \text{SFT} \xrightarrow{\theta_1} \text{Align/RL} \xrightarrow{\theta_2} \text{Continual} \xrightarrow{\theta_3}.
\]

where $\theta_0$ is the base model parameters, and $\theta_1, \theta_2, \theta_3$ are the parameters after SFT, alignment, and continual optimization, respectively. The forward constraint can be formalized as: the attainable performance upper bound of $\theta_{k}$ is a function of $\theta_{k-1}$ and the current stage's $(\mathcal{D}, \mathcal{L})$, that is,

\[
\text{Cap}(\theta_k) \le f_k\big(\text{Cap}(\theta_{k-1});\, \mathcal{D}_k, \mathcal{L}_k\big).
\]

This inequality serves as a reminder that blindly increasing the number of post-training steps without simultaneously improving $\mathcal{D}_k$ (higher-quality alignment data) or $\theta_{k-1}$ (a stronger base) will cause returns to saturate rapidly.

\textbf{Formalization of the objective in the preference/alignment stage}. The goal of preference alignment is to bias the output distribution of the policy $\pi_\theta$ toward preferred responses while not deviating excessively from the reference policy $\pi_{\text{ref}}$. RLHF jointly optimizes a reward $r_\phi$ and a KL regularization term:

\[
\max_\theta\ \mathbb{E}_{x,\,y\sim\pi_\theta}\big[r_\phi(x,y)\big] - \beta\, \mathrm{KL}\!\big(\pi_\theta \,\|\, \pi_{\text{ref}}\big),
\]

where $\beta$ controls the trade-off between "staying close to preferences" and "not drifting". RL-free methods such as DPO prove that this objective can be equivalently reduced to an implicit reward comparison over preference pairs, thereby eliminating the explicit reward model and the sampling loop [32]. This formalization uniformly explains the core tension of Chapter 8: the \textbf{verifiability} of the reward determines signal quality, while $\beta$ and the sampling budget determine "how much is learned and how much drift occurs".

\textbf{Behavioral dimension and capability dimension}. Let $\text{Beh}(\theta)$ denote the output behavior (style, compliance, refusal tendency); then post-training mainly changes $\text{Beh}$ and indirectly affects $\text{Cap}$ through the coupling of "behavior--attainable capability": good instruction following can "unlock" knowledge that the base model already possesses but has not been elicited, whereas over-alignment shrinks the exploration space and lowers the sampling upper bound (Pass@K). Accordingly, Chapter 11 proposes a "capability-behavior orthogonal axis" as an observational framework for cross-stage comparison.

\textbf{Formalization of the continual optimization stage}. Continued pretraining approximates restarting the pretraining of $\theta_2$ on a domain data distribution $\mathcal{D}_{\text{dom}}$, and it must constrain general capabilities from degrading ($\text{Cap}(\theta_3) \gtrsim \text{Cap}(\theta_2)$, preventing catastrophic forgetting). Compression (quantization/pruning/distillation) can be viewed as minimizing the output distribution shift $\mathrm{KL}(p_{\theta_3} \| p_{\theta_2})$ under parameter or activation precision constraints; model editing attempts to directly rewrite parameters on a specific set of knowledge while controlling "ripple effects" (untargeted rewriting). The common formal problem of these three types of operations is \textbf{transformation under preservation constraints}---which is also the basis for Chapter 9's unified use of "preservation" as the primary axis of evaluation.

\subsection{Families of Evaluation Metrics and Dataset Classification}

Evaluation can be divided along two axes: \textbf{intrinsic metrics} versus \textbf{extrinsic metrics}, and \textbf{understanding-type} versus \textbf{generation-type}. Intrinsic metrics (perplexity, bits-per-byte) directly measure the model distribution; extrinsic metrics (accuracy, pass rate, human-evaluated win rate) measure downstream task performance. HELM [22] advocates "holistic evaluation" across multiple scenarios and metrics, and MMLU [7] covers knowledge across 57 disciplines; the two respectively represent the "broad coverage" and "knowledge-intensive" categories of benchmarks. Generation-type tasks (summarization, dialogue, code) usually require human or model review and are sensitive to prompts and decoding parameters.

Key reminder: scores from different benchmarks cannot be directly added or directly compared across models unless data contamination control, prompts, and decoding protocols have been unified. Multimodal evaluation (e.g., visual question answering, image captioning) additionally superimposes modality alignment and hallucination issues, so the strength of its conclusions is usually lower than that of pure-text benchmarks (see Chapter 12 and Appendix E for details).

\textbf{The boundaries of intrinsic metrics}. Intrinsic metrics such as perplexity and bits-per-byte facilitate comparison across runs but do not necessarily correspond to downstream capabilities: the data distribution, tokenizer granularity, and deduplication protocol all change their absolute values, so that "lower perplexity" does not imply "more usable". Therefore, intrinsic metrics are suitable for "relative comparison within the same setting" but not for absolute assertions across works (Sections 6.3 and 12.2).

\textbf{Metric families for alignment and safety evaluation}. Post-training introduces three categories of distinctive metrics: \textbf{preference win rate} (human- or model-evaluated; see the evaluation methodology of [22]), \textbf{safety metrics} (jailbreak success rate, over-refusal rate, honesty), and \textbf{verifiable reasoning metrics} (Pass@1 and Pass@K). The common difficulty of these three categories is that "the judge itself is biased": model judges may favor models from the same source, human evaluation is costly and of limited consistency, and Pass@K is sensitive to the sampling budget. Therefore, alignment conclusions especially need to report the judging protocol and sampling settings; otherwise, the values are not comparable (Chapters 8 and 12).

\textbf{Stage dependence of metric selection}. The same model focuses on different metrics at different stages: pretraining looks at perplexity and downstream zero/few-shot; SFT looks at instruction following and format; preference/RL looks at win rate, safety, and Pass@K; continual optimization looks at preservation, throughput, and safety regression. Applying the metrics of one stage to another is a common source of evaluation mismatch (Chapter 11 will build a cross-stage metric comparison on this basis).

\textbf{Multilingual and fairness evaluation}. The "average" of benchmark scores often masks the large variance across languages/groups: aggregate values dominated by English benchmarks overestimate multilingual capability, and likewise overall accuracy masks the degradation of low-resource groups. Therefore, any conclusion that claims "general capability" should report breakdown metrics by language/by group rather than only a mean (Section 12.7). This is also why this paper includes "fairness" as an evaluation dimension---it is not an add-on but a necessary condition for judging whether a conclusion holds.

\subsection{Notation for Compute and Cost}

To support cross-work comparison, this paper uniformly adopts the following protocol: the parameter count $N$ (distinguishing total parameters from activated parameters, the latter being especially critical for MoE [9][151]); the number of training tokens $D$; the approximate training FLOPs $C \approx 6ND$ (a commonly used approximation for autoregressive training); and wall-clock time / GPU hours and peak GPU memory. Wherever cross-work comparison is involved, the main text provides the $(N, D, C)$ triple protocol as far as possible; when unavailable, it is marked \texttt{[to be verified]}. This convention serves to suppress the most common methodological error of "directly comparing numbers across datasets and compute budgets".

At the system level, dense training scales equivalently through data parallelism, tensor parallelism, pipeline parallelism, and ZeRO [6]; the attention implementation (FlashAttention [135]) determines the memory and throughput ceiling for long sequences. These system constraints in turn limit the available batch size, sequence length, and parallelism strategy, thereby affecting optimization stability---which is precisely why the six-dimensional framework treats "system constraints" as a separate dimension.

\textbf{Origin and boundaries of the compute approximation}. The origin of $C \approx 6ND$ is that the forward pass is approximately $2ND$ and the backward pass approximately $4ND$ (the backward computation is about twice the forward). This approximation applies to dense, autoregressive, single-pass training; for MoE, it must be replaced with the activated parameters $N_{\text{act}}$; for multimodal models, the visual encoder and projector must be counted separately; and for long context, the quadratic term of attention must be included. Therefore, $C$ should be regarded as an "order-of-magnitude estimate" rather than an exact value---which also explains why this paper insists on reporting compute as a "magnitude-range" rather than a single value (Chapters 11 and 13).

\textbf{Effective compute and MFU}. From an engineering perspective, what matters more is "effective compute", i.e., the ratio of hardware peak to actual throughput (MFU). MFU depends on the parallelism strategy, batch size, sequence length, and communication overhead; the same model can differ by several times in MFU across different clusters. Therefore, "how many GPU-days were used for training" is not comparable across works unless MFU and the parallelism configuration are reported together (Section 6.6).

\textbf{Inference compute and test-time compute}. Beyond training compute, inference/sampling compute (especially RL rollout and long chain-of-thought reasoning) can no longer be ignored in the post-training era; denoting the sampling cost of RL as $C_{\text{rollout}}$, the total cost of post-training also depends on the policy length, the number of samples, and the reward evaluation overhead. This item is the formalization background of Open Problem 1 in Chapter 14.

\subsection{Formalization Assumptions and Limitations}

The notation system in this section rests on several simplifying assumptions; making them explicit is intended to prevent readers from mistaking the formal notation for a complete theory:

First, \textbf{the assumption that capability can be scalarized}. Using $\text{Cap}(\cdot)$ to represent model capability and assuming its upper bound is constrained by preceding stages (Equation (3.2)) is an analytical tool rather than a proven law; "capability" is in fact multidimensional and task-dependent, and an upper bound along any single dimension is more credible than a single scalar.

Second, \textbf{the assumption that stages are separable}. Dividing training into five stages facilitates analysis, but the interfaces between stages (data mixture, templates, objectives) often interpenetrate; this paper explicitly annotates such coupling during cross-stage comparison (Chapter 11).

Third, \textbf{the transferability assumption of scaling laws}. $C \approx 6ND$ and the Kaplan/Chinchilla extrapolations both assume a relatively stable distribution; when data quality, mixture, or architecture changes significantly, the extrapolation fails (Section 6.3). Therefore, this paper annotates the applicable protocol for all scaling-law conclusions rather than treating them as universal constants.

Fourth, \textbf{the reproducibility assumption}. All "comparisons" presuppose that protocols can be aligned; when alignment is impossible, this paper only states the method without giving numerical rankings (Section 4.3). Acknowledging these limitations, rather than hiding them, is more consistent with this paper's "auditable" stance: formalization is meant to make assumptions testable, not to provide a sense of certainty.

\textbf{Formalization notation for multimodality}. To facilitate the comparison in Chapter 10, this paper denotes the multimodal input as a set of modalities $\{x^{(1)}, \dots, x^{(M)}\}$ (text, image, audio, etc.), whose unified representation is composed of a three-part "encoder--connector--backbone" composite: the encoder $E_m$ maps each modality to continuous features, the connector $g_m$ projects it into the backbone's token space (or discretizes it and appends it to the vocabulary), and the backbone $\pi_\theta$ then generates the target modality via (conditional) NTP. From this, the three major structural choices of multimodality can be formalized: \textbf{whether the encoder is frozen} (training cost vs alignment quality), \textbf{whether the connector is discrete or continuous} (compression rate vs flexibility), and \textbf{whether the backbone is shared across modalities} (parameter efficiency vs interference). These three choices constitute the formal basis for the conclusion in Chapter 10 regarding "modality peripherals on a shared backbone", and they also show that multimodality does not depart from the five-stage framework of this paper but merely introduces an additional modality variable at each stage.

\section{Taxonomy}

\subsection{Overview: Stage $\times$ Six Dimensions $\times$ Modality/Architecture}

Conclusion first: the taxonomy in this paper is a three-dimensional coordinate system---\textbf{stage} (five stages), \textbf{six dimensions} (data, objective, optimization, systems, evaluation, governance), and \textbf{modality and architecture} (text / multimodal; dense / sparse). Any method can be uniquely (or, at most, in a small number of places) positioned within this coordinate system, which makes heterogeneous work comparable across the board. \textbf{Figure 2}: Stage-level relationships---input / output / objective / key methods / principal risks / principal evaluation

\begingroup\tiny\setlength{\tabcolsep}{4pt}
\begin{longtable}{@{}>{\raggedright\arraybackslash}p{0.1118\textwidth}>{\raggedright\arraybackslash}p{0.1864\textwidth}>{\raggedright\arraybackslash}p{0.0932\textwidth}>{\raggedright\arraybackslash}p{0.1398\textwidth}>{\raggedright\arraybackslash}p{0.1211\textwidth}>{\raggedright\arraybackslash}p{0.1211\textwidth}>{\raggedright\arraybackslash}p{0.1211\textwidth}@{}}
\toprule
Stage & Input & Output & Objective & Key Methods & Principal Risks & Principal Evaluation \\
\midrule
\endhead
Data and Tokenization (\allowbreak{}Chapter 5) & Raw corpus /\allowbreak{} multimodal data & Cleaned dataset +\allowbreak{} tokenizer & High quality,\allowbreak{} leak-\allowbreak{}free,\allowbreak{} well-\allowbreak{}proportioned & Deduplication,\allowbreak{} filtering [392]\allowbreak{},\allowbreak{} mixing,\allowbreak{} BPE/\allowbreak{}Unigram [1]\allowbreak{}[5]\allowbreak{} & Contamination,\allowbreak{} toxicity,\allowbreak{} copyright,\allowbreak{} vocabulary bias & Deduplication rate,\allowbreak{} coverage,\allowbreak{} downstream gains \\
Pretraining (\allowbreak{}Chapter 6) & token sequences & Base model $\theta_0$ & Acquisition of general representations and knowledge & NTP/\allowbreak{}MLM [47]\allowbreak{},\allowbreak{} scaling laws [119]\allowbreak{}[16]\allowbreak{},\allowbreak{} MoE [9]\allowbreak{} & loss spike,\allowbreak{} data-\allowbreak{}constrained,\allowbreak{} compute imbalance & Perplexity,\allowbreak{} downstream zero/\allowbreak{}few-\allowbreak{}shot [121]\allowbreak{} \\
SFT (\allowbreak{}Chapter 7) & instruction-\allowbreak{}-\allowbreak{}response pairs & Instruct model $\theta_1$ & Instruction following,\allowbreak{} formatting & Full-\allowbreak{}parameter /\allowbreak{} LoRA [12]\allowbreak{} /\allowbreak{} QLoRA [30]\allowbreak{} & Overfitting,\allowbreak{} capability forgetting,\allowbreak{} template leakage & Instruction following,\allowbreak{} human /\allowbreak{} model evaluation \\
Preference Alignment /\allowbreak{} RL (\allowbreak{}Chapter 8) & preference pairs /\allowbreak{} verifiable environments & Aligned model $\theta_2$ & Preference alignment,\allowbreak{} suppression of harmful outputs & RLHF [15]\allowbreak{},\allowbreak{} DPO [32]\allowbreak{},\allowbreak{} RLVR & Reward hacking,\allowbreak{} sycophancy,\allowbreak{} capability narrowing & Win rate,\allowbreak{} safety,\allowbreak{} reasoning Pass@1/\allowbreak{}K \\
Continual Optimization (\allowbreak{}Chapter 9) & base /\allowbreak{} aligned model +\allowbreak{} domain data & Deployment-\allowbreak{}optimized model $\theta_3$ & Domain adaptation,\allowbreak{} efficiency,\allowbreak{} long context & Continued pretraining,\allowbreak{} long-\allowbreak{}context extension [11]\allowbreak{}[129]\allowbreak{},\allowbreak{} distillation/\allowbreak{}pruning/\allowbreak{}quantization [21]\allowbreak{} & Alignment tax,\allowbreak{} forgetting,\allowbreak{} quantization damage & Domain metrics,\allowbreak{} throughput,\allowbreak{} safety regression \\
\bottomrule
\end{longtable}
\endgroup

\subsection{Definition of Taxonomy Dimensions and Notes on Orthogonality}

\textbf{The stage dimension} is a division by time / dependency order, answering "at which step of the training pipeline." \textbf{The six dimensions} are analytical perspectives internal to each stage, answering "which kind of choice was made at that step": data refers to the source and proportion of the training signal; objective refers to the loss / reward; optimization refers to the optimizer, schedule, and stability measures; systems refers to parallelism, GPU memory, and throughput constraints; evaluation refers to the means of measurement; and governance refers to safety, compliance, and fairness. \textbf{The modality / architecture dimension} characterizes the form of the model.

Two places of non-strict orthogonality must be noted. First, "data" and "objective" are highly coupled in self-supervised stages (the masking strategy is both a data transformation and an objective design), and this paper explicitly points out the coupling points when comparing. Second, "optimization" and "systems" mutually constrain each other through batch size, precision (BF16/FP16/FP8), and parallelism granularity. Apart from these two places, the six dimensions may be regarded as orthogonal, so a "six-dimensional score" can be assigned to any method.

\textbf{Definition of the modality/architecture dimension.} This dimension characterizes "what the model processes and with what structure": modality includes text, image, video, audio, speech, and interleaved multimodal; architecture includes dense and sparse (MoE), decoder-only and encoder-decoder. Both are continuous spectra rather than binary values---the interleaved-data ratio, the number of experts and routing sparsity, and whether visual/language parameters are shared all take continuous values. Therefore, when actually positioning a method, one should record the specific "modality combination + structural parameters" rather than merely attaching a "multimodal/sparse" label (Chapter 10 further subdivides multimodal architecture paradigms accordingly).

\textbf{An example of six-dimensional scoring.} Taking some SFT method as an example, its scale may be: data (source and scale of the instruction set), objective (conditional NTP + masking strategy), optimization (full-parameter or PEFT, learning rate), systems (GPU memory and multi-turn data layout), evaluation (instruction-following benchmarks), governance (synthetic-data risks). Scoring is not intended to "rate methods as high or low," but to expose which questions each work has actually answered when comparing---dimensions left unanswered are evidence gaps.

\subsection{Usage Notes: How to Compare Across Works}

When using this taxonomy to compare two works, the following order is recommended: first align the \textbf{stage} (avoid directly juxtaposing SFT results with RL results); then align the \textbf{modality / architecture} (text vs multimodal, dense vs sparse); then compare under the same \textbf{six dimensions} (e.g., both look at the "objective function" or both look at "evaluation"); and finally verify whether the $(N, D, C)$ conventions are comparable. Wherever the conventions cannot be aligned, this paper uniformly refrains from giving numerical superiority/inferiority conclusions and instead states only the method and the evidence. Chapter 11 will use this procedure to systematically compare three pairs of objects---"pretraining vs post-training," "text vs multimodal," and "LLM vs MLLM"---and provide an evidence-strength map (Figure 5).

\subsection{Relationship to Existing Taxonomy Frameworks}

Conclusion first: the taxonomy in this paper does not compete with existing frameworks, but rather provides a common coordinate system for them. The survey of five post-training paradigms [395] takes "fine-tuning/alignment/reasoning/efficiency/integration" as its axes, covering the interior of post-training but lacking cross-stage scope; the surveys of RL post-training [396][399] take algorithm families as their axes, going deep into a single stage; the data-centric multimodal survey [403] and the efficient multimodal survey [405] take modality and efficiency as their axes. The difference of this paper lies in explicitly introducing a \textbf{stage axis} and a \textbf{six-dimensional axis}, which allows all of the above frameworks to be "positioned"---for example, the "reasoning" paradigm in [395] can be positioned to the RLVR family in Chapter 8 of this paper, and the modality issues in [403] can be positioned to Chapter 10 and the data dimension.

The practical value of this relationship is twofold: first, readers can use the coordinates of this paper to juxtapose the conclusions of several specialized surveys without confusing the stages; second, when different surveys give different conclusions about the same phenomenon, one can check whether they are genuinely comparable in terms of "stage/modality/compute conventions"---most apparent conflicts stem from this.

\subsection{Usage Boundaries and Common Misuses}

A taxonomy is an analytical tool, and misuse produces a spurious order. Three common misuses warrant vigilance:

\textbf{Misuse one: pushing orthogonality to an extreme.} Section 4.2 has already noted two places of non-strict orthogonality (data--objective, optimization--systems); if one insists that the six dimensions do not affect one another, one arrives at the erroneous conclusion of "each independently optimal"---in reality, a change in data mixture alters the optimal vocabulary, and a change in parallelism granularity alters the usable batch size and thereby affects stability, both of which are cross-dimensional couplings.

\textbf{Misuse two: substituting stage division for causal attribution.} Placing a method in the "pretraining" or "post-training" cell only indicates its position in the pipeline, not which variable its gains come from. Cross-stage gains are often jointly determined by data, objective, and compute, and attribution requires convention alignment and ablation following the procedure in Section 4.3.

\textbf{Misuse three: treating the modality/architecture dimension as a binary label.} Text/multimodal and dense/sparse are not mutually exclusive binary values but continuous spectra (the interleaved-data ratio, the number of experts, and routing sparsity are all continuous). Binarizing continuous variables discards a great many real cases such as "partially multimodal" and "mildly sparse."

\subsection{From Taxonomy to Method-Family Map (Figure 3)}

Expanding the coordinates of the taxonomy into temporal evolution yields the method-family map (Figure 3), which annotates the maturity of method families in four colors: \textbf{classic} (foundational methods that have been widely validated), \textbf{mainstream} (the current default choice), \textbf{frontier} (promising but with newer evidence), and \textbf{disputed} (with unresolved disagreements). Examples by stage:

\begin{itemize}[leftmargin=*,nosep]
\item \textbf{Data}: heuristic filtering (classic) $\rightarrow$ DoReMi/DSIR data selection (mainstream) $\rightarrow$ data-mixture inference (frontier) $\rightarrow$ tokenizer saturation and synthetic-data collapse (disputed).
\item \textbf{Pretraining}: NTP (classic) $\rightarrow$ decoder-only + Chinchilla proportions (mainstream) $\rightarrow$ MoE/FP8/long context (frontier) $\rightarrow$ Kaplan vs Chinchilla comparability (disputed).
\item \textbf{SFT}: human IFT (classic) $\rightarrow$ Self/Evol-Instruct (mainstream) $\rightarrow$ multimodal IFT (frontier).
\item \textbf{Preference/RL}: RLHF/PPO (classic) $\rightarrow$ the DPO family (mainstream) $\rightarrow$ GRPO/RLVR (frontier) $\rightarrow$ the debate over the Pass@K reasoning boundary (disputed).
\item \textbf{Continual optimization}: distillation/pruning (classic) $\rightarrow$ quantization-aware and model merging (mainstream) $\rightarrow$ model editing/machine unlearning (frontier/disputed).
\end{itemize}

\begin{figure}[htbp]
\centering
\includegraphics[width=\linewidth]{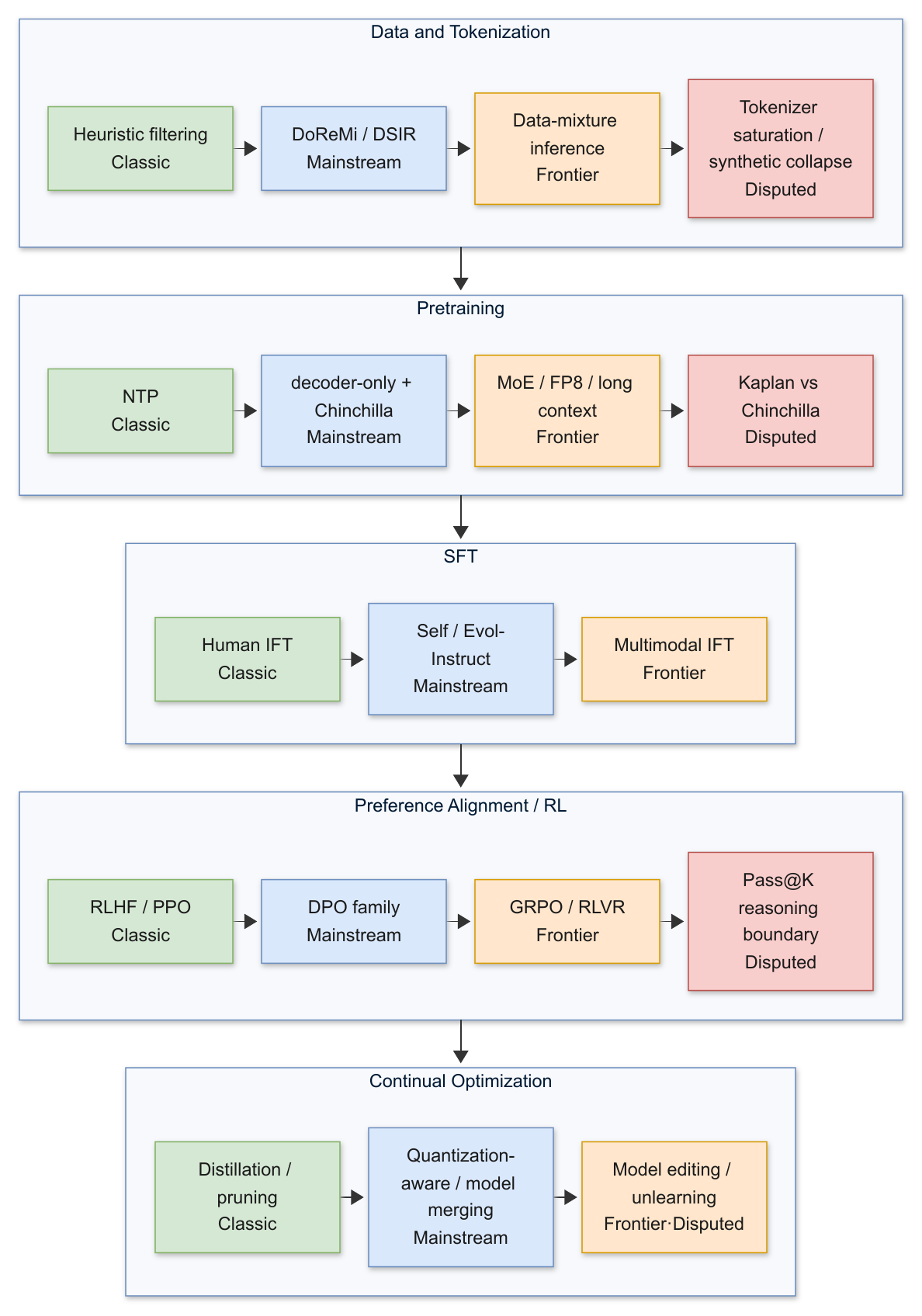}
\caption*{Figure 3: Method-family map---maturity annotated in four Morandi tones (sage = classic, misty blue = mainstream, clay = frontier, dusty rose = disputed)}
\end{figure}

The continuation of Figure 3 is "from top-left to bottom-right, with disputed items always retaining a distinct color": it does not claim that the newest is the best, but rather that each family should be read together with its level of evidence. This map is used in tandem with the "evidence map (Figure 5)" in Chapter 11---the former gives the genealogy of methods, the latter gives the strength and gaps of the evidence.

In summary, Chapters 2--4 together establish the analytical foundation of the entire book: terminology eliminates ambiguity, formalization provides notation and the boundaries of assumptions, and the taxonomy supplies comparison coordinates. Without any one of the three, the cross-sectional comparisons in subsequent chapters would lose their basis; readers are therefore advised not to skip these three chapters even if they skim the method chapters.

\section{Data and Tokenization}

\begin{quote}\small
This chapter, together with Ch6, serves as a pilot for the dual-form "general framework + focused survey" format. Its full length unfolds according to the Stage 2 Section 8 template: definition and boundaries $\rightarrow$ inputs, outputs, and core problems $\rightarrow$ method families $\rightarrow$ subword tokenization $\rightarrow$ vocabulary and special tokens $\rightarrow$ multimodal tokenizer $\rightarrow$ representative-work table $\rightarrow$ key controversies $\rightarrow$ practical recommendations $\rightarrow$ cross-stage connections and open problems. Target length: 8,000--9,000 characters.
\end{quote}

\subsection{Definition and Boundaries}

Conclusion first: the data and tokenization stage addresses "what to feed, how to segment it, and in what proportions to feed it." It determines the information ceiling of pretraining and post-training, but does not directly determine optimization dynamics. \textbf{Included}: data collection and cleaning, exact/approximate/semantic deduplication, quality and safety filtering, multilingual and domain mixing, data selection, curriculum and two-stage mixing, subword tokenizers (BPE/WordPiece/Unigram/SentencePiece/tiktoken) and vocabulary design, and multimodal tokenizers. \textbf{Excluded}: the human processes of annotation quality control (which belong to engineering management), and pure storage and retrieval systems. The boundary criterion is: anything that changes "the content of the training signal or the way it is segmented" is included; anything that only changes "how the signal is scheduled onto devices" belongs to the systems dimension (Chapter 6).

\textbf{Three categories of configuration for this stage (the minimal comparability standard)}: \textcircled{\small 1} data sources and cleaning pipelines (composition of the raw corpus, deduplication strategy, filtering thresholds, whether synthetic data is included); \textcircled{\small 2} mixing and curriculum (the weights of each source/domain, whether it is staged, whether it varies with scale); \textcircled{\small 3} tokenization and vocabulary (algorithm, vocabulary size, special tokens, multimodal placeholder scheme). If any one of these categories is missing, cross-work comparisons of data conclusions will fail---because a "dataset with the same name" differs enormously under different cleaning regimes (the crux of the controversy in Section 5.8). Each subsequent subsection of this chapter returns to these three configuration categories to annotate the evidence standard.

\textbf{Why the data dimension is the most easily underestimated.} Models and optimization have clear mathematical forms and are convenient to publish; data pipelines, by contrast, are often engineering details that are hard to reproduce and are not disclosed. The result is a systematic bias: papers showcase models and objectives while concealing data recipes, making "performance improvements" difficult to attribute. This survey places data first among the six dimensions precisely because it is currently the largest and least transparent variable, and the place where the evidence gap is most concentrated (Appendix C/E).

\textbf{Interface with continual learning/domain adaptation}: data and tokenization determine not only general pretraining, but also the difficulty of "continued pretraining and domain adaptation" in Chapter 9. Domain corpora often need re-cleaning, deduplication, and re-mixing, and catastrophic forgetting must be prevented; this makes the data pipeline an activity that runs through the model's entire lifecycle rather than a one-time upstream step (reverse empowerment, Figure 1).

\textbf{Calibration under the six-dimensional framework}: the data dimension simultaneously touches the other five dimensions---objective (different objectives yield different marginal returns on different data, 6.2), optimization (data ordering affects the loss curve and stability, 6.6), systems (throughput bottlenecks in data loading and shuffling), evaluation (contamination and fairness audits, Chapter 12), and risk governance (PII/copyright/toxicity, Chapter 14). Data decisions are therefore inherently cross-dimensional decisions and cannot be optimized in isolation within the data dimension alone.

\textbf{Terminological clarification}: in this chapter, "data" specifically refers to the training signal (tokens/samples) and does not include model parameters; "tokenization" refers to the process of mapping the raw signal into discrete/continuous tokens; what is called the "data recipe" downstream is precisely the complete combination of the three categories of configuration in Section 5.1.

\subsection{Inputs and Outputs, Core Problems}

The inputs are the raw corpus (web pages, books, code, dialogues, image-text pairs, video, etc.); the outputs are the cleaned dataset, the mixing scheme, and the tokenizer (including the vocabulary and merge rules). There are four core problems. First, how to trade off between scale and quality (the stricter the filtering, the cleaner the data but the lower the diversity) [71][87]. Second, how to remove overlap between the training set and the evaluation set (contamination) without erroneously deleting legitimate long-tail items [52]. Third, how to allocate vocabulary and mixing proportions for multiple languages and modalities, avoiding the systematic squeezing out of low-resource languages/modalities [51]. Fourth, whether the mixing should change dynamically with model scale and training stage [86][407].

\textbf{Data-constrained vs. data-abundant} is the premise that divides all data strategies. General English web corpora have been fully mined, and frontier training is mostly in the "data-constrained" regime (the growth rate of high-quality tokens lags behind the growth rate of compute); here the strategic focus shifts from "scaling up" to "improving quality, deduplicating, synthesizing, and reusing." In low-resource languages, specialized domains, and multimodal settings, however, the situation remains "data-scarce," with the focus on collection and construction. The optimal strategies of the two are opposite, and mixing them leads to erroneous conclusions---this is the root of Section 5.8 and open problem 2.

\textbf{The data--evaluation loop}: data decisions (filtering, mixing, deduplication) can only be validated through evaluation, and evaluation itself depends on data (benchmarks, contamination audits). Hence the "correctness" of the data dimension is relative and depends on the evaluation standard; without the contamination and unfairness analysis of Chapter 12, one cannot judge the true value of a data recipe.

\textbf{Two types of contamination}: \textbf{direct overlap} between the training set and the evaluation set (the same text) and \textbf{indirect contamination} (question style/answer patterns are implicitly learned, or the evaluation set is generated by a model and then misused). Direct overlap can be detected by exact/approximate deduplication [52]; indirect contamination is difficult to audit and intensifies with model scale, making it the category with the weakest current evidence yet the most severe impact. Therefore, any conclusion that "a model improves on benchmark X" must pass a contamination audit before it can be accepted (Section 12.3).

\textbf{The priority of the four questions shifts with the stage}: in the pretraining stage, "scale vs. quality" is the main line; in the post-training stage, "contamination and fairness" gain weight (because instruction data is closer to the evaluation distribution); in the multimodal stage, "vocabulary/modality mixing" becomes a newly added main line. Thus there is no one-time optimal data scheme, only local optima oriented toward a specific stage and evaluation objective.

\textbf{Realistic estimates of data volume}: both the total order of magnitude of available high-quality public text and the actual number of tokens trained by various models suffer from a large number of inconsistent standards (whether duplicates are included, whether multiple languages are counted, whether code is counted), causing conclusions such as "A uses more data than B" to be frequently incomparable. When comparing data volume across works, one should standardize on the "deduplicated, vocabulary-based token count" standard and note whether multimodal tokens are included.

\textbf{The conflict between deduplication and mixing}: aggressive deduplication reduces the usable volume of certain sources (forums, templated text, etc.), thereby forcing the mixing to tilt toward "less-duplicated but more official" sources, potentially losing colloquial and diverse styles. Therefore the deduplication threshold and the mixing weights should be evaluated jointly; tuning either in isolation sacrifices the other objective (yet another instance of six-dimensional coupling).

\textbf{The quantitative relationship between scale and quality}: under data constraints, the marginal return of scaling up diminishes while the marginal return of improving quality rises; the crossover point moves with the compute budget ([87]), so "add data or clean data" should be treated as a continuous trade-off rather than a binary choice.

\subsection{Classification of Method Families}

Along the six dimensions, data and tokenization methods can be grouped into five families:

\begin{itemize}[leftmargin=*,nosep]
\item \textbf{Cleaning and deduplication}: heuristic filtering (language identification, perplexity, repeated n-gram removal), exact deduplication (hashing), approximate deduplication (MinHash/LSH), and semantic deduplication. Representative evidence: the C4 documentation study reveals the commercial and bias costs of heuristic filtering [56]; exact/approximate deduplication is shown to reduce memorization and improve downstream performance [13].
\item \textbf{Quality and safety filtering}: quality classifiers, toxicity filtering, and privacy (PII) and copyright handling. The datasheet work of Pile emphasizes the auditability of data sources and uses [61]; the quantification of the effects of quality, timeliness, domain, and toxicity on training is given in [71].
\item \textbf{Mixing and data selection}: weighting by source/domain, importance resampling DSIR [67], learnability prioritization (such as the RHO-LOSS idea) [63], the minimax mixing optimization of DoReMi [70], data mixing laws [86], skill-driven mixing Skill-it [74], the compute-dependent scaling law of data filtering [87], and two-stage (general first, then high-quality) mixing [407].
\item \textbf{Tokenizers and vocabulary}: BPE [1], WordPiece (BERT) [47], Unigram/SentencePiece [4][5], token-free (Canine [55], ByT5 [57]); and inferring the training data mixture from the merge table [406].
\item \textbf{Multimodal tokenizers}: discretization of vision/audio (VQ-VAE [3], VQGAN [8]), visual encoders (CLIP [10], SigLIP [26], SigLIP2 [46], DINOv2 [27]), and multimodal placeholders/alignment (Flamingo [17], LLaVA [28], Qwen-VL [277]).
\end{itemize}

\textbf{The five families are not a pipeline but a coupling}: the granularity of deduplication affects the number of usable samples, thereby changing the optimal mixing; the filtering intensity changes the data distribution, thereby changing the optimal vocabulary (subword frequencies); the vocabulary in turn affects the information per token, and hence the "effective token count" and the mixing budget. Treating data work as "one-time preprocessing" is therefore a common misconception---in reality it is a closed loop repeatedly adjusted along with training and evaluation (the reverse-empowerment arrow in Figure 1).

\textbf{Safety, privacy, and copyright}: quality filtering simultaneously performs a compliance function---removing PII, respecting robots/licensing, and handling copyrighted text. These processes affect the usable corpus volume and distribution, yet there is almost no standardized public reporting standard; differences across jurisdictions also blur the boundary of "trainable data." A datasheet such as Pile's [61] provides an auditable starting point, but transparency in large-scale actual execution remains low (a governance issue, Chapter 14).

\textbf{Differences in evidence strength}: deduplication [13], mixing laws [86][87], and tokenizer effects [51] have relatively strong Grade-A evidence; the "optimal thresholds" of quality classifiers and semantic deduplication mostly come from single-point studies (B/C); and the "largest-scale real data recipes" are almost entirely undisclosed (C or blank). This chapter can therefore draw mechanistic conclusions, but recipe-level conclusions are not transferable.

\textbf{The genealogy of deduplication techniques}: exact deduplication uses document-level hashing (SHA on normalized text, etc.) to remove exact duplicates; approximate deduplication uses MinHash + LSH to find near-duplicates with high Jaccard similarity in sublinear time; semantic deduplication uses embedding similarity to capture samples that are worded differently but semantically duplicated. The granularity (document/passage/sentence) and thresholds of the three jointly determine "how much to delete and what to keep": too coarse misses passage-level duplication in long documents (a contamination risk), while too fine erroneously deletes legitimate templated text (code, formulas). The empirical practice is "multi-granularity combination + per-source thresholds," but public threshold--benefit curves are lacking, making this Grade-B/C evidence.

\textbf{Technical differences among mixing methods}: DoReMi [70] uses a minimax (group DRO) idea to automatically adjust per-domain weights on a proxy model, minimizing the worst-domain loss; DSIR [67] resamples by the importance of hashed n-grams, tilting a general corpus toward its target distribution without training a model; Skill-it [74] builds a mixing order from the learning dependencies among skills, favoring curriculum learning; and data mixing laws [86] directly fit the "mixing--validation loss" relation as a predictor for extrapolation across scales. The four address different facets of the same problem (weights, direction, order, predictability) and are often combined in practice, but public evidence on the optimal combination is lacking.

\textbf{The composition of heuristic filtering}: a typical pipeline performs, in sequence, language identification, boilerplate removal (navigation bars/disclaimers), removal of overly short/overly long samples, a repeated n-gram ratio threshold, and perplexity filtering (scored by a reference LM). Each item is simultaneously an operation of "purification" and "loss of diversity," and the thresholds are highly empirical; the C4 analysis [56] shows that such filtering systematically affects topic and group distributions, so the filtering configuration should be regarded as a design choice requiring a fairness audit rather than a neutral cleaning step.

\textbf{Curriculum learning and two-stage training}: curriculum learning (easy first, then hard; ordered by skill dependencies) and two-stage mixing (first building a foundation with general corpora, then finishing with high-quality corpora) represent data strategies along the "time" axis. Skill-it [74] and the two-stage report [407] provide directional evidence, but "when to switch, and whether the benefit of switching is sensitive to model scale" lacks systematic study (mostly Grade B/C). One interpretable mechanism is that later high-quality data can "cleanse" the distribution bias introduced by earlier low-quality data, approximating a form of implicit fine-tuning.

\textbf{Why the optimal mixing is hard to reproduce}: mixing experiments are highly sensitive to proxy model scale, training steps, evaluation set, and random seed, and the optimal mixing at small scale may reverse at large scale (data mixing laws [86] were proposed precisely to provide an extrapolation framework). Thus "we used an A:B:C mixing" is not itself a reusable conclusion; it must be reported together with the extrapolation model and the validation standard.

\textbf{How to evaluate a data decision}: because of the data--evaluation loop (5.2), evaluating a data strategy requires three types of control: \textcircled{\small 1} fixing the model and compute and changing only the data (ablation); \textcircled{\small 2} fixing the data and sweeping the scale (testing the scale robustness of the conclusion); \textcircled{\small 3} validating under multiple evaluation sets and contamination audits (to prevent overfitting to a single benchmark). If any one is missing, the data conclusion may be "accidentally valid under a specific setting"---this is the minimum threshold for advancing data work from experience to accumulable knowledge.

\subsection{Subword Tokenization Details}

BPE [1] builds the vocabulary by merging high-frequency byte/character pairs and is the default choice of the GPT series; it is fairly robust to multiple languages and unknown words, but its coverage is uneven for morphologically rich languages and languages sensitive to typography (no-space segmentation). WordPiece (BERT [47]) selects merge pairs by likelihood gain rather than frequency. Unigram / SentencePiece [4][5] model directly on raw text (including spaces), avoiding the unfairness of pre-tokenization toward CJK and multilingual text, and support subword regularization (randomly sampling tokenizations during training to mitigate vocabulary bias). Empirical studies show that the choice of tokenizer has a substantial and source-dependent effect on the monolingual performance of multilingual models [51], indicating that the assumption "tokenization is neutral preprocessing" does not hold.

\textbf{Variants and engineering details of BPE}: byte-level BPE (since GPT-2) uses 256 bytes as the initial alphabet and naturally handles arbitrary Unicode; the cost is that the same character may produce multiple tokens due to encoding, which is especially uneconomical for CJK (a single Chinese character may occupy 2--3 bytes). Implementations such as tiktoken accelerate BPE to the point of handling massive corpora, allowing the vocabulary to be trained directly on large-scale data. There are also engineering conventions regarding merge order, case handling, and digit splitting; although these conventions do not change the algorithm, they significantly change token efficiency and downstream performance---yet another source of opacity.

\textbf{Per-token compression rate (fertility)}: the number of tokens produced by the same sentence under different languages/tokenizers can differ by several times, directly determining the "effective context" and inference cost under an equal budget. Low-resource languages have higher fertility due to insufficient vocabulary coverage, so under an equal token budget they carry less actual information---a numerical manifestation of the multilingual fairness problem and the mechanistic explanation of the conclusion in [51]. When evaluating a tokenization scheme, the fertility of each language should be reported explicitly rather than only the global average.

\textbf{Theoretical differences among the three subword algorithms}: BPE is a greedy frequency merge aiming at the shortest encoding (it does not explicitly model probabilities); WordPiece uses the criterion of "largest likelihood gain after merging," approaching the local optimum of a language model; Unigram first posits a subword probability model, then prunes the vocabulary via EM, theoretically yielding the maximum a posteriori segmentation and naturally supporting subword regularization. The difference among the three is not one of algorithmic superiority, but of "whether the segmentation probability is explicitly modeled"---this explains why low-resource languages are more sensitive to the choice of algorithm ([51][4][5]).

\textbf{The segmentation challenge of CJK and morphologically rich languages}: CJK has no spaces and blurry word boundaries, so byte-level BPE often splits common Chinese characters into multiple bytes and splits words into single characters; morphologically rich languages (such as Finnish and Turkish) suffer from overly long subwords that force fallback. Both raise fertility and squeeze the effective context. The native-text modeling and subword regularization of SentencePiece/Unigram can mitigate this, but the optimal strategy still depends on the specific language and data volume, and no universal conclusion exists yet [51][4][5].

\textbf{The token-free route.} ByT5 [57] trains directly on UTF-8 bytes, and Canine [55] uses a character-level encoder; both avoid vocabulary bias and trade longer sequences for language neutrality, at the cost of increased sequence length and compute. This constitutes a long-standing axis of controversy---"discrete subwords vs. bytes/characters" (5.8).

\textbf{Inferring the data mixture from the merge table.} Recent work shows that BPE's merge rules and vocabulary carry a "fingerprint" of the training corpus distribution, from which the data mixture of an undisclosed model can be reverse-engineered [406]. This has a dual implication for data transparency auditing: it is both a tool for verifiability and a hint that the vocabulary itself may leak the composition of the training data. Its methodological significance extends beyond tokenization itself---it provides an operational entry point for "data auditing of closed-source models," and is directly related to the contamination issues of Chapter 12 and the governance issues of Chapter 14.

\subsection{Vocabulary Design, Multilingual Fairness, and Dialogue/Tool Tokens}

The size and composition of the vocabulary directly affect the information per token and the inference cost for multiple languages: too small a vocabulary leads to frequent fallback to bytes, while too large a vocabulary dilutes statistics and enlarges the output layer. Multilingual fairness is therefore partly determined by the vocabulary, not solely by the data mixing [51]. In dialogue and agent scenarios, additional conventions are needed for the role/separator control tokens in the \textbf{chat template} and for \textbf{tool-call tokens} (such as function names and argument boundary markers); the design of these special tokens affects the stability and generalization of the SFT and RL stages, and improper design can cause format compliance to fail when migrating across templates (see Chapters 7 and 12 for details).

\textbf{The trade-off in vocabulary size.} Enlarging the vocabulary shortens the sequence length and reduces the pressure on attention and positional encoding, but the output projection layer ($V \times d$) and the embedding table grow linearly in parameter count with the vocabulary, and rare subwords in data-constrained languages are insufficiently trained. There exists an implicit optimum related to data scale and language distribution, but public systematic evidence is limited (mostly Grade-B/C single-point reports). The common practice in multilingual settings is to "allocate a token budget per language and explicitly measure the token compression rate and monolingual performance of each language" [51], rather than pursuing a single global vocabulary optimum.

\textbf{Multilingual vocabulary construction practice}: a common approach is to allocate vocabulary capacity according to the expected data proportion of each language/script, and after joint training to re-check using the per-language token compression rate and monolingual perplexity; warm-starting (initializing from an existing vocabulary) reduces training cost but inherits its bias. The key risk is the "silent degradation of majority languages"---global average metrics mask the severe squeezing of low-resource languages, so evaluation must report per language (Chapter 12).

\textbf{Vocabulary and inference economics}: a larger vocabulary improves the compression rate and reduces the number of generation steps, but drives up the cost of softmax and embeddings and may harm the sampling quality of rare subwords. On-device deployment is especially sensitive to vocabulary size (the embedding table consumes GPU memory/RAM). The vocabulary is therefore the intersection of "training efficiency---inference cost---multilingual fairness," and must be decided jointly under the target deployment conditions rather than by maximizing a single metric in isolation.

\textbf{The contractualization of special tokens.} Chat templates and tool-call tokens are essentially "interfaces between the model and external protocols"; once fixed in the pretraining stage, they are difficult to change: changing them breaks already-trained patterns and causes cross-version incompatibility. They should therefore be treated as \textbf{versioned contracts} rather than ordinary configuration, maintaining a single source of truth across the data pipeline, SFT data, and the evaluation toolchain; cross-stage inconsistency is a common root cause of format-compliance failure and degraded instruction generalization (Chapters 7 and 12).

\textbf{The semantic burden of tool-call tokens}: encoding function names and argument boundaries as ordinary subwords (rather than dedicated tokens) leads to unstable parsing, especially when arguments contain arbitrary strings. A more robust approach is to use structured separators + explicit schema constraints, and to treat the stability of the schema as part of the data contract (Chapters 7 and 13).

\subsection{Multimodal Tokenizer}

Multimodal training expands the tokenizer problem from "text segmentation" to "cross-modal discrete/continuous representation." In vision there are two routes: \textbf{discrete tokens} (VQ-VAE [3], VQGAN [8], which quantize images into a codebook) and \textbf{continuous features + projection} (CLIP [10], SigLIP2 [46], DINOv2 [27] encoding followed by a projector into the LLM, as in LLaVA [28], Flamingo [17]). Audio/speech likewise has both discrete codec and continuous feature routes. Interleaved image-text data work such as OBELISC [273] shows that the "placeholder + cross-attention + alignment objective" design of multimodal tokens is strongly coupled with the data format. Unifying the optimization objectives of discrete tokens and continuous tokens remains an open problem (Chapter 10, open problem 3).

\textbf{Resolution and token budget.} The "token economy" of images/video is a core constraint of multimodal training: higher resolution yields finer visual understanding but a surge in token count (roughly growing with area), and video even more so (adding linearly with the number of frames). This has given rise to schemes that trade token efficiency for capability, such as "arbitrary resolution + dynamic segmentation" (Qwen2-VL [288]) and lightweight compressors (Q-Former, BLIP-2 [265]). The trade-off of these schemes between "fine-grained recognition" and "long-video understanding" currently lacks unified evaluation, with evidence mostly at Grade C/B.

\textbf{Data format is the tokenizer.} Multimodal token schemes cannot be optimized independently of the data format: the character equivalence of image-text pairs, the proportion of interleaved image-text, and the granularity of inter-modal alignment all change "how much information one visual token represents." This again shows that data and tokenization are two sides of the same problem (Chapter 10).

\textbf{The token economy of video/audio is even more extreme}: the token count of video grows in a coupled manner with the frame count and resolution, while audio faces a trade-off between "frame rate vs. semantic units" (continuous features vs. discrete codebooks). Current mainstream multimodal models are trained with "images as the mainstay + a small amount of video," and both long-range temporal modeling of video and fine-grained alignment of audio lack public scaling laws, constituting an evidence gap (Chapter 10).

\textbf{Attempts at a unified multimodal vocabulary}: encoding images/audio into a token space isomorphic to text (a unified vocabulary) can simplify sequence modeling and reuse the optimization and system stack on the text side, but discretizing a continuous signal without losing detail has undetermined costs and benefits (the two lines of VQ-type [3][8] and continuous projection [10][28] coexist). This is one of the core open problems of Chapter 10.

\textbf{The quality dimension of multimodal alignment data}: the quality of image-text pairs is not just clarity and resolution, but also the strength of image-text relevance (weakly related pairs teach the model that "image and text do not constrain each other"), annotation bias (regional/cultural bias is amplified), and safety content. Large-scale automatic filtering of images and text (CLIP scores, NSFW, caption quality) itself introduces bias, and combined with the multimodal evaluation sensitivity of Chapter 12, makes multimodal data problems harder to audit than pure text.

\textbf{The coupling between multimodal token schemes and objectives}: the discrete-token route (VQ [3][8]) naturally fits the autoregressive objective of "predicting the next visual token" and can be unified with text NTP; the continuous-projection route ([10][28]) requires an additional alignment loss (contrastive/matching) in parallel with text NTP. Thus "which token to choose" is in fact choosing "which objective to train with," and the two cannot be optimized separately---again confirming the judgment that "data format is the tokenizer."

\subsection{Representative-Work Table}

\begingroup\small\setlength{\tabcolsep}{6pt}
\noindent\textbf{Table 3: Chapter 5 representative works---data and tokenization}\par\medskip
\begin{longtable}{@{}>{\raggedright\arraybackslash}p{0.1762\textwidth}>{\raggedright\arraybackslash}p{0.1898\textwidth}>{\raggedright\arraybackslash}p{0.2575\textwidth}>{\raggedright\arraybackslash}p{0.1084\textwidth}>{\raggedright\arraybackslash}p{0.1626\textwidth}@{}}
\toprule
Direction & Representative work & Core idea & Evidence & Notes \\
\midrule
\endhead
Deduplication & Lee et al. [13]\allowbreak{} & Exact/\allowbreak{}approximate deduplication reduces memorization and improves performance & A & Widely reproduced \\
Data selection & DSIR [67]\allowbreak{} & Importance resampling to match the target distribution & A & Requires target distribution \\
Learnability selection & RHO-\allowbreak{}LOSS-\allowbreak{}type [63]\allowbreak{} & Sampling prioritized by learnability & A & Requires a reference model \\
Mixing optimization & DoReMi [70]\allowbreak{} & Minimax domain-\allowbreak{}weight optimization & A & Requires domain labels \\
Mixing laws & Data mixing laws [86]\allowbreak{} & Predict the mixing-\allowbreak{}-\allowbreak{}loss relation & A & Extrapolate with caution \\
Filtering compute & Filtering scaling law [87]\allowbreak{} & Filtering optimality depends on the compute budget & A & Against "one-\allowbreak{}size-\allowbreak{}fits-\allowbreak{}all" \\
Skill mixing & Skill-\allowbreak{}it [74]\allowbreak{} & Order data by skill learning dependencies & A & Requires skill annotations \\
Two-\allowbreak{}stage & [407]\allowbreak{} & General first,\allowbreak{} then high-\allowbreak{}quality mixing & C & Report,\allowbreak{} not peer-\allowbreak{}reviewed \\
Cleaning cost & C4 analysis [56]\allowbreak{} & Bias and commercial cost of heuristic filtering & A & Requires audit \\
Quality/\allowbreak{}toxicity & [71]\allowbreak{} & Quantifying the effects of quality,\allowbreak{} timeliness,\allowbreak{} and domain & A & Multi-\allowbreak{}factor \\
Auditability & Pile datasheet [61]\allowbreak{} & Traceable data sources and uses & A & Governance \\
Contamination & [52]\allowbreak{} & Detection and impact of training-\allowbreak{}-\allowbreak{}evaluation overlap & A & Key \\
Mixture inference & [406]\allowbreak{} & Inferring the data mixture from BPE & A & Auditing tool \\
Tokenizer evaluation & [51]\allowbreak{} & Tokenizer affects monolingual performance & A & Multilingual \\
Token-\allowbreak{}free & ByT5 [57]\allowbreak{} /\allowbreak{} Canine [55]\allowbreak{} & Byte/\allowbreak{}character-\allowbreak{}level modeling & A & Variable-\allowbreak{}length sequences \\
Multimodal discrete & VQGAN [8]\allowbreak{} & Learned discrete codebook & A & Image generation \\
Multimodal alignment & CLIP [10]\allowbreak{} /\allowbreak{} LLaVA [28]\allowbreak{} & Contrastive pretraining /\allowbreak{} projection connection & A & Mainstream paradigm \\
\bottomrule
\end{longtable}
\endgroup

\textbf{Reading guide}: the "Evidence" column of the table uses this survey's unified grading; any "recipe-level" conclusion (a specific threshold, a specific mixing value) cannot be directly transferred across datasets even if it comes from a Grade-A paper (Section 3.4). The table should be used as a "mechanism index"---choose methods by mechanism, choose parameters by local experiments.

\subsection{Key Controversies: Tokenizer Neutrality, Data-Quality Transferability, and Synthetic Bootstrapping}

This chapter converges on three controversies, each annotated as "consensus vs. remaining disagreement."

First, \textbf{whether tokenization is merely lossless preprocessing}. The consensus is that tokenization significantly changes token efficiency and multilingual performance ([51][406]). The disagreement concerns "whether tokenization can be decoupled": one side holds that the tokenizer must be co-designed with the objective and data [51][406]; the other side's token-free work [55][57] shows that byte/character-level modeling can achieve comparable or even better robustness, suggesting that the reliance on subwords is partly an engineering choice. There is currently \textbf{insufficient evidence} to determine "which route is generally better under a given compute budget," but it can be confirmed that for low-resource and morphologically complex languages, differences in vocabulary and tokenization are amplified into measurable performance gaps [51]. This is the direct source of open problem 4.

Second, \textbf{the transferability of data quality} (echoing Section 6.4). The consensus is that quality affects convergence; the disagreement concerns "whether high-quality data allows reaching an equivalent loss with fewer tokens," i.e., whether the Chinchilla constant shifts globally with data quality. The filtering scaling law [87] and data mixing laws [86] support the idea that "optimality depends on compute/quality," but there is as yet no extrapolation formula that takes "data quality" as an explicit variable and covers multiple scales. Existing conclusions mostly come from single-point studies (Grade B/C), so any claim of the form "data quality can substitute for 2$\times$ the token count" should be marked as pending verification.

Third, \textbf{the bootstrapping boundary of synthetic data}. The consensus is that synthetic data can supplement scarce domains and "missing intermediate steps" (such as reasoning traces); the disagreement concerns whether long-term iteration leads to \textbf{capability/diversity collapse} (the model-collapse risk). Shumailov et al. [411] provide rigorous theoretical and empirical evidence showing that unconstrained recursive training on synthetic data inexorably triggers early loss of distribution tails and irreversible variance collapse (Model Collapse, Grade A evidence), establishing the indispensable role of real-data anchors; determining safe mixture ratios and quality filters in industrial pipelines remains an active challenge (Open Problem 11). Any claim that "purely synthetic data can losslessly replace real data" should therefore be treated with caution.

\textbf{The common root of the three controversies}: tokenizer neutrality, quality transferability, and synthetic bootstrapping essentially all point to the same question---\textbf{at what granularity data carries information, and how the value of that information is quantified}. If "effective information content" could be treated as a first-class quantity alongside the token count (the interface of Section 5.10), all three controversies could be expressed within a unified framework as "the benefit curve per unit of effective information."

\textbf{The practical implications of the controversies}: the three controversies are not purely academic---they directly determine budget allocation: if the tokenization effect is underestimated, insufficient investment is made in the vocabulary for multiple languages; if quality transferability is overestimated, one may over-filter and lose diversity; if synthetic bootstrapping is overestimated, one may fall into a self-loop of low-quality data. The "pending verification" annotation of this section is therefore not conservatism, but a guard against erroneous resource investment.

\subsection{Practical Recommendations}

\begin{enumerate}[leftmargin=*,nosep]
\item \textbf{Deduplication takes priority over adding data}: when the budget is limited, the marginal return of approximate deduplication and contamination cleaning is usually higher than continuing to expand homogeneous corpora [13][52].
\item \textbf{Filtering is not better the stricter it is}: the optimal filtering intensity varies with the compute budget [87][71], and filtering should be treated as a tunable item on par with $N, D$.
\item \textbf{Mixing must be dynamic}: a single static mixing will be defeated by differences in scale and stage; consider two-stage or mixing-law-guided mixing [86][407][74].
\item \textbf{Orient the vocabulary toward the target language/modality}: in multilingual settings use SentencePiece/Unigram and explicitly evaluate the per-language token cost [51]; in multimodal settings first decide "discrete vs. continuous," then design placeholders and alignment objectives [10][28].
\item \textbf{Contractualize special tokens}: chat templates and tool-call tokens require version management (a single source of truth) to avoid cross-stage inconsistency (Chapters 7 and 12).
\item \textbf{Treat data as an auditable asset}: register sources, filtering thresholds, deduplication methods, and mixing (Appendix D), and perform contamination audits on key benchmarks [61][52].
\item \textbf{When data-constrained, improve quality before synthesizing}: under the data-constrained regime, prioritize deduplication, quality improvement, and domain rebalancing, and carefully assess the bootstrapping benefit and capability-collapse risk of synthetic data (open problem 11).
\item \textbf{Publish a data card}: use a structure similar to the Pile datasheet to register sources, licenses, filtering rules, deduplication methods, and known biases [61], making data decisions auditable, reusable, and accountable.
\item \textbf{Isolate and version the vocabulary and data contract}: bind the tokenizer, vocabulary, chat template, and data snapshot in a versioned way, so that any single training run can be traced back to the exact input contract (the reproducibility checklist in Appendix D).
\end{enumerate}

\subsection{Cross-Stage Connections and Open Problems}

\textbf{Connections}: the quality, mixing, and vocabulary of data and tokenization directly determine the achievable loss of pretraining in Chapter 6 and the data foundation for alignment in Chapters 7--8; conversely, the contamination and unfairness exposed by evaluation in Chapter 12 flow back to correct this chapter's methods (reverse empowerment). \textbf{Open problems}: how data quality changes the optimal token/parameter ratio (does Chinchilla shift with data quality, problem 2); whether synthetic-data bootstrapping has a capability-collapse critical point (problem 11); the quantifiable causality between vocabulary design and multilingual fairness (problem 4); whether a scaling law exists for the mixing of the four multimodal modalities (image/text/video/audio) (newly added, evidence gap). See Appendix F for details.

\textbf{An actionable research design example}: for problem 2 (quality transferability), one can fix the token budget, pass a corpus through multiple levels of filtering intensity to generate multiple equal-token subsets, and measure whether the loss--compute curve shifts as a whole; for problem 11 (synthetic collapse), one can, at a controlled small scale, sweep the "real--synthetic" iteration ratio and examine when diversity decline and capability regression appear; for problem 4 (vocabulary causality), one can fix the data and compute, change only the tokenizer and vocabulary, and measure monolingual performance and per-token compression rate on several low-resource languages. All three can be completed within moderate compute, with reporting standards following Appendix D.

\textbf{Summary of the evidence distribution in this chapter}: for Ch5, the method-type work (deduplication, mixing, tokenization) has relatively thick Grade-A/B evidence, but "the largest-scale real data recipes," "multimodal mixing laws," and "long-term effects of synthetic data" are almost entirely Grade C or blank. This distribution means that this chapter can give robust \textbf{mechanistic conclusions} (deduplication is effective, filtering depends on compute, tokenization affects fairness), but \textbf{specific recipes are not transferable}---any claim of the form "some model's secret recipe" should return to evidence grading and standard re-verification.

\textbf{The quantitative interface with Chapter 6}: data quality enters the scaling law through the "effective token count"---under the same nominal tokens, deduplication and quality improvement raise the effective information density, equivalent to relaxing the data budget. The data recipe and the optimal $(N, D)$ mixing are therefore a \textbf{joint optimization} problem and cannot be optimized separately (Section 6.4). When evaluating any data strategy, the "effective token" standard should be reported rather than only nominal tokens; otherwise the data volume will be systematically overestimated and quality problems underestimated.

\textbf{The interface with post-training data in Chapters 7--8}: this chapter deals with self-supervised data; the instruction/preference data of Chapters 7--8 is superimposed on top of it. The two share the vocabulary, chat template, and contamination risk, and the distribution of post-training data (task type, difficulty, safety) should be constructed in a targeted way based on the gaps exposed by pretraining (reverse empowerment, Figure 1). The data pipeline should therefore run through "pretraining---SFT---alignment" rather than being optimized in isolated segments. This chapter, together with Ch6, forms the data foundation of the subsequent chapters, and its strength of conclusions (mechanistically robust, recipes non-transferable) also determines the prudence with which subsequent chapters should cite data-type claims.

\section{Pretraining}

\begin{quote}\small
This chapter is a pilot of the "general framework + focused survey (sub-survey)" dual form: the full length unfolds according to the Phase 2, Section 8 template, so that Ch5/Ch8/Ch9 can directly reuse this writing style and length yardstick. Target length: 8,000--9,000 characters.
\end{quote}

\subsection{Definition and Boundaries}

Conclusion first: pretraining acquires general representations and knowledge on large-scale corpora via self-supervised objectives, producing a "base model" $\theta_0$; it determines the upper bound of post-training capability, rather than final alignment quality. \textbf{Included}: objective-function families (6.2), architecture and MoE (6.3), scaling laws (6.4), optimization and stability (6.5), distributed systems (6.6), long-context training and extrapolation (6.7), and multimodal pretraining (6.8). \textbf{Not included}: data cleaning and tokenization (Chapter 5), instruction and preference data (Chapters 7--8), and pre-deployment compression (Chapter 9). The boundary criterion is "whether it proceeds without task-label signals": whenever the loss comes from unlabeled data itself, it is pretraining.

\textbf{Inputs and outputs.} The input is a token sequence (text) or multimodal tokens/features; the output is the base-model parameters $\theta_0$ together with the accompanying vocabulary, positional encoding, and (optionally) expert routing. The commonly used proxies for evaluating pretraining quality are downstream zero/few-shot performance [121] and perplexity, but the correlation between the two is affected by evaluation contamination (Chapter 12).

\textbf{Core questions.} This chapter answers four key questions: how the choice of objective function affects subsequent transferability (6.2); what the optimal ratio of parameters to tokens is under a given compute budget, and how data quality perturbs that ratio (6.4); whether sparsification (MoE), long context, and multimodality are merely "incremental engineering" or will change the scaling laws (6.3/6.7/6.8); and how the joint design of optimization and systems determines training stability (6.5/6.6).

\subsection{Objective-Function Families}

Autoregressive language modeling (NTP) is currently the absolute mainstream [121][140][141], defined in Section 3.1: token-by-token conditional maximum likelihood. NTP's three advantages determine its dominance: it is fully consistent with autoregressive inference tasks and requires no additional objective adaptation; it can turn arbitrary text (code, dialogue, scripts) into training signals, so data scaling has no bottleneck; and the optimization itself is a stable maximum-likelihood procedure, facilitating scaling extrapolation. Its cost is unidirectional context and the short-sightedness of the "next token" objective, which partly gave rise to the competing lines of masked and denoising objectives.

Masked language modeling (MLM) targets random masking and bidirectional prediction [47]; its advantage is obtaining bidirectional representations, which benefits understanding-type tasks, but its drawbacks are equally clear: the objective is inconsistent with generation, and the predicted-token ratio and masking strategy introduce hyperparameters. RoBERTa's [48] systematic experiments show that MLM's performance depends heavily on training details (dynamic masking, longer training, removing NSP), which indicates both that MLM has a large tuning space and that its conclusions have higher "setting sensitivity" than NTP.

The sequence-to-sequence (seq2seq) line, represented by T5 [49], unifies all tasks as "text-to-text"; it is essentially still conditional NTP but explicitly separates encoder and decoder. It still has structural advantages in scenarios requiring explicit conditioning of the input (summarization, translation, controlled generation), but in general dialogue and tool-use scenarios it yields to decoder-only.

\textbf{Multimodal objectives}: multimodal pretraining does not invent new losses but combines "contrastive + matching + NTP." CLIP [10] aligns the embedding space with an image-text contrastive objective, establishing the "representation pretraining" paradigm; Flamingo [17] places vision into the language model via a pretrained vision encoder + cross-attention + gating; LLaVA [28], BLIP-2 [265], and Qwen-VL [277] map visual features into the LLM's token space via a projector and then continue NTP. These three modes are essentially different ratios of "alignment signal to language loss" (Chapter 10), so discussing "NTP vs. contrastive" across modalities is a false dichotomy.

\textbf{Summary}: the choice of objective function is not merely "what to predict"; it also determines what kind of data yields the greatest marginal benefit and what kind of evaluation can reflect training quality. When comparing different objectives across works, one must fix the data scale and level of purification; otherwise the differences (mostly Grade-B evidence) will be drowned by confounding variables.

\textbf{Practical guidance for objective selection}: when bidirectional understanding representations are needed (retrieval, classification, embedding), contrastive or MLM-type objectives still have value; when the goal is to align with human instructions and support general generation and tool use, autoregressive NTP is the default and necessary foundation---because post-training (Chapters 7--8) is built on a generative conditional distribution. In other words, the choice of pretraining objective is partly constrained by the downstream alignment paradigm, another example of the "objective--evaluation--downstream" coupling.

Beyond NTP, MLM, and seq2seq, several transitional objectives also exist. \textbf{Prefix LM} lets the prefix of a sequence be predicted with bidirectional context and the suffix autoregressively, lying between MLM and NTP; it was once used in multi-task pretraining to accommodate both generation and understanding. \textbf{Denoising autoencoding} takes "restoring text under arbitrary noise (span masking, text dropout, order shuffling)" as its objective, a general framework for the span-restoration objective of the T5 [49] family; it also shows that the actual implementation of the seq2seq objective depends heavily on the form of the corruption function. The significance of understanding these transitional objectives is that they all adjust two continuous variables---\textbf{the visible range of context} and \textbf{the granularity of prediction}---and these two variables directly rewrite the marginal benefit of data: this is precisely the mechanism coupling the "objective function" with "data mixture" in Chapter 5 (a coupling point of the six-dimensional framework).

\textbf{Objective--evaluation consistency} is a hidden constraint when choosing an objective: whether an objective succeeds is ultimately still defined by downstream evaluation (Chapter 12). In the MLM era, "high scores on understanding benchmarks" defined success; in the NTP era, "few-shot and instruction following" defined success. This brings a methodological reminder: if one evaluates only on benchmarks that are "co-origin" with the objective, one will systematically overestimate that objective. Cross-objective comparisons must use third-party evaluations independent of the objective; otherwise one falls into circular reasoning.

\subsection{Architecture: decoder-only, encoder-decoder, and MoE}

The decoder-only structure (the GPT family [121][140]) has become mainstream because of objective consistency and engineering simplicity; encoder-decoder (T5 [49]) retains advantages in explicit conditional generation. The trade-off between the two was discussed in 6.2. This section focuses on the sparsely activated \textbf{MoE}, which represents the key shift of "scaling" from dense to sparse.

The core idea of MoE is to use gated routing to assign each token to a small number of "experts" (subnetworks), making the number of activated parameters far smaller than the total. \textbf{Key variables}: routing strategy (Top-K, soft routing), capacity factor and load-balancing loss, expert granularity and count, and the communication overhead of expert parallelism.

Milestone works established MoE practice along five dimensions: Switch Transformer [9] proved that single-expert routing (Top-1) is stable and efficient at extreme scale and provided an engineering recipe for capacity factor and load balancing; GShard [122] proposed automatic sharding and a conditional-computation framework, coupling sparse scaling with system scheduling; GLaM [130] matched the performance of a large dense model with roughly 1/3 of the activated compute, key evidence for "compute-efficiency-oriented scaling"; ST-MoE [131] systematically examined the stability of sparse-expert training (regularization, router initialization) and the transferability of "sparse-to-dense" fine-tuning; Mixtral [151] demonstrated the trainability of an open-source 8-expert sparse MoE, freeing the sparse architecture from institution-specific engineering. A survey caveat: MoE comparisons must distinguish \textbf{total parameters} from \textbf{activated parameters}, otherwise the "parameter count" discussion loses meaning.

\textbf{Limitations and openness}: the training stability of sparse models (loss spikes, expert collapse) is more sensitive than that of dense models; evidence on the interaction of MoE with long context and low-precision training is still dominated by internal reports (Grade C). Whether MoE routing is stable during the RL/alignment stage currently lacks systematic study (open problem, see 6.12).

\textbf{The essential trade-off between dense and sparse}: a dense model uses all parameters in every forward pass, with strong interpretability and stable training, but diminishing returns per unit of compute; a sparse model trades additional total parameters for "higher capacity at the same activated compute," at the cost of system complexity for routing and load balancing, plus the risk of expert imbalance under long-tail distributions and unstable router convergence. Available evidence supports "at comparable activated compute, sparse MoE can match or exceed a large dense model" ([130][151]), but third-party replicable evidence for this conclusion under RL post-training and fine-grained instruction following is still scarce.

\textbf{Router details}: the choice between Top-K routing (usually K=1 or 2) and soft routing, the capacity factor (the maximum multiple of tokens passed through), and the load-balancing auxiliary loss determine the stability and result quality of MoE training---too small a K loses information, too large degrades to dense; too small a capacity factor causes token dropping (loss fluctuation), too large introduces wasted computation. ST-MoE's [131] comparison of these variables shows that MoE's gains are not free: it shifts part of the "stability budget" to routing hyperparameters. Most consistent conclusions come from medium scale; recent frontier advances (such as MiMo-V2.6 [412], a 1.02T total / 42B active omni-modal MoE) disclose industrial recipes of freezing MoE routers and adopting hybrid sliding-window attention (hybrid-SWA) to stably scale context up to 1 million tokens under massive-batch asynchronous RL (Grade C), offering a viable reference for mitigating expert routing drift.

\textbf{Capacity and "memory vs. reasoning"}: whether dense or sparse, the parameters of the feed-forward layers carry most of the factual memory. This explains why "scaling parameters" yields significant gains on knowledge-type tasks, and also why MoE gains are pronounced on knowledge-intensive tasks but limited on pure reasoning/format-type tasks---the latter depend more on data and objectives than on capacity. This observation is indirectly supported by several works (Grade B), but controlled experiments that cleanly separate "memory" from "reasoning" are lacking.

\subsection{Scaling Laws: The Kaplan vs. Chinchilla Controversy and Reconciliation}

The scaling law characterizes the empirical regularity of "loss with respect to parameter count, data amount, and compute," and is the quantitative basis for pretraining budget decisions. Kaplan et al. [119] gave a power law under a relatively narrow setting, arguing that "a larger model is more cost-effective than more data." Chinchilla [16], through about 400 controlled training runs covering 70M--16B parameters and 5B--500B tokens, proposed that under a given compute budget the optimum should be trained at roughly \textbf{\textasciitilde{}20 tokens per parameter}, revealing that models such as GPT-3 [121] before it were clearly "undertrained," and shifting the optimal implementation from "parameter-first" to "data-first." This conclusion directly rewrote the (N, D) selection of open-source models such as LLaMA [140] and formed the current mainstream.

\textbf{Why the two conflict}: the divergence is not a mathematical contradiction but stems from differences in extrapolation settings---the learning-rate schedule and the allocation of the token budget, and whether extrapolation is to configurations outside the training range [119][16]. Thus "which scaling law is correct" is a false question; the real question is: under a \textbf{given data quality and a given budget}, where is the marginal return. Gopher's [60] large-scale empirics support the "data-first" intuition, but likewise depend on specific tuning details.

\textbf{How data quality perturbs scale}: recent work brings the data dimension into the scaling law. The scaling law of data filtering shows that \textbf{the optimal filtering strength depends on the compute budget}---when compute is limited, noise should be filtered more aggressively; when compute is ample, retaining more diversity is instead better [87]---which directly undermines the simplified understanding that "the Chinchilla ratio is a universal constant." The data-mixture law [86] models the interaction of mixture and scale with predictive loss, supporting dynamic rather than static fixed mixtures. The common implication of the two: \textbf{treat data as a first-class citizen first, then discuss the (N, D) optimum}---this is the interface between Chapter 5 and this chapter, and the source of open problem 2.

\textbf{Strength of evidence}: among Grade A/B evidence, the scaling law is one of the lines with more controlled experiments ([119][16][60]), but the largest-scale extrapolations still rely on internal reports; whenever a work across the literature cites an "optimal-ratio constant," this paper requires that data quality and compute scope be stated (Section 3.4).

\textbf{Distinguishing two regimes}: the scaling law is not a single-point optimum but a family of surfaces that move with (N, D, data quality). When \textbf{data-limited} (the total amount of obtainable high-quality data is fixed, as in vertical domains or low-resource languages), the optimal strategy is to "overtrain" on limited data---scaling parameters beyond the Chinchilla constant. This scenario is especially common in domain continued pretraining (Chapter 9), showing that "data-first" applies only to general corpora with abundant data. Moreover, the scaling law has not yet been jointly closed-formulated with the post-training budget: given a total budget, when is it more cost-effective to spend part of it on RL/alignment (Chapter 8) rather than continuing pretraining---an extrapolatable joint model is still lacking (open problem 1).

\textbf{Numerical anchors (scope-illustrative examples, not universal conclusions)}: Chinchilla [16] was trained with about 70B parameters and about 1.4T tokens (a data-to-parameter ratio of about 20:1), while LLaMA-65B [140] used about 65B parameters and about 1.4T tokens, adopting a similar ratio. These numbers are only \textbf{anchors under a specific data quality}, not portable constants---the controversy in 6.10 is, to a large extent, precisely the mistaking of anchors for an optimal law. When citing such numbers, one should simultaneously give the premises of data cleaning and mixture (Chapter 5); otherwise it may mislead compute decisions.

\textbf{The practice of budget estimation}: given a compute budget $C$ (roughly estimable as $\approx 6ND$), one can first use a scaling law to estimate an initial $(N, D)$ ratio, then perform a small-scale mixture sweep on the real data (e.g., at the $10^{18}$--$10^{20}$ FLOPs level) for local correction [16][86]. Engineering-wise, "data preparation (Chapter 5)---pretraining---post-training (Chapters 7/8)" should be regarded as a pipeline sharing a total budget: any single point's "optimum," if it ignores the upstream and downstream, will be suboptimal along the entire pipeline. This is also the motivation for open problem 1.

\subsection{Optimization and Stability}

The optimizer is the first gate of pretraining stability. Adam [105] and AdamW [110] with decoupled weight decay are the de facto standard, and their default recipe of "large learning rate + warmup + cosine decay" has been widely copied. Recent exploration unfolds along three lines: \textbf{low-memory/high-efficiency}---Lion [25], discovered by symbolic-optimizer search, uses a momentum-sign update to eliminate second moments, markedly reducing GPU memory and RAM in large-model training while performing comparably to AdamW; \textbf{second-order approximation}---Sophia [31] uses a diagonal Hessian approximation to accelerate pretraining convergence, but evidence for its robustness at extreme scale is limited; \textbf{hyperparameter scaling}---muP [132], via tensor-program theory, gives scaling rules for the learning rate and others with respect to width, enabling zero-shot transfer of hyperparameters from small to large models and reducing the cost of "tuning once" at scale.

\textbf{Stability engineering}: besides the optimizer, key means for pretraining stability include: learning-rate warmup and scheduling (insufficient warmup easily triggers loss spikes); gradient clipping; appropriate weight initialization (including scaling of residual branches); and \textbf{online monitoring of loss spikes with checkpoint rollback}. Spikes often correlate with a too-high learning rate and abnormal data batches (e.g., duplicates or encoding errors); there is no unified theoretical explanation yet, and engineering practice relies mainly on detect-and-rollback.

\textbf{Precision and numerical boundaries}: FP16/BF16 have become the default training precision, with BF16 being more robust to spikes due to its wider exponent range; FP8 and lower precisions improve throughput while significantly altering the numerical-stability boundary, and their interaction when combined with MoE, long sequences, and small batch sizes is unknown. \textbf{Precision is both an optimization problem and a systems problem}, a typical "coupling point" in the six-dimensional framework.

\textbf{Evidence and limitations}: optimizer comparisons have relatively strong Grade A/B evidence at controlled medium scale ([25][31]), but at true frontier scale the vast majority of comparisons lack public, replicable third-party experiments, and the conclusions are mostly single-institution reports (Grade C). Comparing any "X\% speedup" across works must state the precision, batch, sequence, and schedule scope.

\textbf{Rules of thumb for scheduling and monitoring}: the sensitivity of scheduling details to scaling extrapolation is precisely one technical source of the Kaplan [119] vs. Chinchilla [16] divergence. The rules of thumb commonly used in practice (e.g., scaling the learning rate with the square root of the batch, adjusting the schedule according to the effective batch and token progress) mostly come from internal experience and are unpublicly verified (Grade C); when citing across works, one must state the original schedule and perform a small-scale re-test before transfer. For stability monitoring, the \textbf{gradient norm} (rather than only the loss) can expose numerical divergence earlier; its interaction with training precision (especially pronounced under FP8) and batch size is an active area of current stability research.

\textbf{Three grades of precision (scope reminder)}: FP32 is the numerical safety baseline but incurs high GPU-memory and throughput overhead; BF16, with its 8-bit exponent, approaches FP32's stability at medium scale and has become the training mainstream; FP16 has a narrow magnitude range and requires dynamic-scaling tricks, being unstable in some implementations; FP8 (E4M3/E5M2) has been used in some ultra-large-scale training, but gradient accumulation and master weights usually fall back to higher precision. Precision choice is thus an explicit trade of "throughput vs. stability," and must be reported together with batch size, sequence, and degree of parallelism; otherwise any "speedup" conclusion cannot be third-party replicated.

\textbf{Effective batch and gradient noise}: batch size is not an isolated choice---it determines, together with gradient noise and learning rate, the "information efficiency per token." Empirically there exists a "critical batch size" beyond which training efficiency no longer improves linearly, and this critical value grows with model scale; but this is still an empirical regularity (Grade C). Many production recipes set the batch directly according to the GPU-memory limit, leaving a hidden coupling of "learning rate scaled accordingly"; when comparing optimization configurations across works, if the effective batch is not disclosed (the effective batch may differ from the nominal batch due to gradient accumulation), the conclusions are hard to transfer.

\subsection{Distribution and Systems}

Pretraining scalability is determined by the combination of parallelism strategies. Four types of parallelism constitute the mainstream toolbox: \textbf{data parallelism (DP)} shards data and all-reduces gradients; \textbf{tensor parallelism (TP)} splits intra-layer weight matrices (Megatron-LM [118] gives a reference implementation); \textbf{pipeline parallelism (PP)} segments layers and pipelines them with micro-batches (introduced by GPipe [115], including recomputation to reduce activation memory); \textbf{expert parallelism (EP)} works with MoE to distribute experts across devices (GShard [122]). ZeRO [6], by progressively sharding optimizer states, gradients, and parameters, eliminates the GPU-memory redundancy of data parallelism, enabling "very large batch + large model" to be trained without TP; FSDP [142] engineers this idea into a general framework and reports ultra-large-scale experience.

\textbf{The attention implementation} determines the upper bound for long sequences. FlashAttention [135] uses IO-aware block-wise exact attention (tiling + recomputation) to significantly reduce GPU memory and improve throughput, turning long-context training from infeasible into routine; the batch of variants that followed basically follow the same IO-optimization route.

\textbf{The coupling of systems and optimization}: the available batch, sequence length, and degree of parallelism are not free variables but are determined by GPU memory, bandwidth, and communication topology; they in turn constrain the learning-rate range, loss stability, and token throughput. Therefore, \textbf{the "hyperparameter tuning" of 6.5 must be performed jointly within the "systems" of 6.6}; otherwise the optimal learning rate and the optimal degree of parallelism will contradict each other. The compute-comparison table in Appendix C provides a template for registering such scope.

\textbf{Communication, memory, and fault tolerance}: the second set of system-level variables concerns communication and memory optimization---bandwidth optimization of gradient all-reduce (bucketing, latency hiding), minimization of bubbles in pipeline parallelism, activation recomputation/checkpointing (the classic trade-off of trading compute for GPU memory), and periodic state checkpointing and fault recovery (long pretraining runs cannot be interrupted, so fault tolerance directly determines whether one can return to the schedule trajectory where the optimal learning rate lies). This forms a \textbf{"throughput--GPU memory--wall-clock time" three-way budget game}: for the same FLOPs budget, different parallelism and recomputation configurations give different wall-clock times and solvable sequence upper bounds. Cross-work reporting must give all three scopes simultaneously (the Appendix C template is designed for this purpose).

\textbf{Intuition for common combinations (not a recipe)}: several-B to tens-of-B parameters commonly use DP + ZeRO/FSDP [6][142]; in medium-scale single-node multi-GPU scenarios, TP cuts intra-layer weights so that a single card can hold a larger layer [118]; hundreds-of-B to trillion-level scenarios require 3D parallelism (TP $\times$ PP $\times$ DP) combined with EP to handle MoE experts [122][151]. An auditable report should include: the total number of GPUs and the interconnect topology, each dimension of TP/PP/DP/EP, batch and sequence length, the layer granularity of activation recomputation, and measured throughput or MFU. Without these numbers, "cumulative GPU hours" by itself has almost no comparable value.

\textbf{Data order and shuffling}: "sample order," which appears to belong to the data dimension (Chapter 5), is doubly constrained by systems and optimization in pretraining: the randomness of shuffling, whether epochs are reused, and whether the order of cross-device shards is consistent all affect the loss curve and the occurrence of spikes. Cross-work reports also rarely disclose data order (random seed, shuffle, number of epochs), a hidden source of replication failure that should be included in the Appendix D reproducibility checklist.

\textbf{Cost scope}: a system comparison should give at least one of three: per-step wall-clock time, per-token throughput, or MLU/MFU (compute-utilization efficiency). These metrics are highly sensitive to precision, sequence length, and parallelism configuration, so any "we are faster" conclusion should be measured under the same software, hardware, and configuration. The compute table in Appendix C registers along the three dimensions of $(N, D, C)$ and configuration, precisely to avoid "throughputs under different scopes" being compared side by side.

\subsection{Long-Context Training and Length Extrapolation}

Long context has two distinct goals: \textbf{training long context} (requiring long-text data and corresponding compute) and \textbf{extrapolation} (supporting longer sequences at inference with a short training sequence). Positional encoding is the key to achieving both. RoPE [11], with rotary positional encoding, preserves relative positional relationships and has become the mainstream choice; ALiBi [129] achieves zero-shot length extrapolation with a linear distance bias, a representative scheme predating the interpolation era; positional interpolation (PI) [143] extends the window by scaling positions and reusing pretrained weights, with low cost but requiring long-text continued training for calibration; YaRN [37] improves the frequency scaling and attention temperature of interpolation, enlarging the effective window and reducing perplexity loss under limited long-text data.

Another line is \textbf{linear-time sequence models}: Mamba [149], with selective state spaces, achieves linear complexity in sequence length, avoiding the quadratic overhead of attention and offering a candidate path toward training-free ultra-long sequences, but its gap from attention on long-range retrieval-type tasks remains an open problem.

\textbf{Evaluation reminder}: many measures of "successful extrapolation" decay on real long texts, so one must distinguish "can read" from "can use" (e.g., multi-hop reasoning and localized retrieval within long texts). Whether "training short, extrapolating long is universally feasible" remains unsettled (open problem 12): existing successes mostly rely on interpolation/transformation tricks [143][37] and a small amount of long-text calibration, without a general scheme that requires no long-text training at all.

\textbf{Long-context training recipes}: the recipe has three input lines---long-text data (which must be extracted from massive corpora or reordered and concatenated), positional-encoding or interpolation transformations ([143][37]), and length-related loss weighting. The three are often used jointly and are difficult to attribute individually, one reason why long-context conclusions are mostly Grade-C reports. For evaluation, one should register both "can read" (perplexity, precise localized retrieval) and "can use" (long-text multi-hop reasoning, cross-paragraph information composition) metrics: the former is friendly to interpolation methods, while the latter is closer to real use; it suffices to register according to the evaluation matrix in Appendix E.

\textbf{The implementation spectrum of "train short, extrapolate long" (layered by cost--capability)}: pure extrapolation---no change to training, relying on the inherent properties of the positional encoding (ALiBi's linear bias [129], RoPE's periodicity [11]); interpolation-style---scaling positional coordinates after training (PI [143]), requiring a small amount of calibration; transformation-enhanced---adjusting frequency and attention temperature (YaRN [37]), often with a very small amount of long-text calibration; direct long-text training---lengthening the training sequence, the most costly but closest to "true long-text capability." The four range from low to high cost and from "transposition-style extrapolation" to "persistent usable long-range," and reporting the "method--cost--capability" trio together avoids the misleading conclusion of "successful extrapolation."

\textbf{Contamination in long-text evaluation}: the "longness" of long-text benchmarks may itself make contamination more covert---long documents in the pretraining corpus have a higher probability of overlapping with evaluation long texts, while existing deduplication tools are mostly sentence-level and struggle to cover paragraph-level semantic overlap. Thus, both before and after long-context extension, a separate contamination audit should be performed on long-text benchmarks (Section 12.3); otherwise "successful extension" may be partly memorization rather than generalization.

\textbf{The cost structure of long context}: the cost of long-text training includes not only the near-quadratic attention overhead of the sequence itself but also data construction (long documents are scarce and require reordering/concatenation), the long-text samples needed for calibration, and the linear storage growth of the KV cache at inference. Sequence length has different cost-curve shapes for training and inference (training near-quadratic, inference near-linear), so "supporting 128K during training" and "serving 128K at low cost" are two different things. In reporting, the training-sequence upper bound should be stated separately from the servable "effective context"---most claimed context windows are "can process," not "reliably usable."

\subsection{Multimodal Pretraining}

Multimodal pretraining can be architecturally divided into three types (detailed in Chapter 10): encoder-projector-LLM (LLaVA [28], BLIP-2 [265], Qwen-VL [277]), cross-attention fusion (Flamingo [17]), and unified arbitrary-resolution and multi-task (Qwen2-VL [288]). The data side relies on large-scale image-text pairs (LAION-5B [64]) and interleaved image-text (OBELISC [273]).

The essential differences from pure-text pretraining (also the core of Chapter 10): first, \textbf{token heterogeneity}---visual tokens and text tokens are asymmetric in information density, quantity, and meaning; second, \textbf{multiple alignment objectives}---contrastive, matching, and NTP objectives may conflict; third, \textbf{evaluation sensitivity}---multimodal benchmarks are more sensitive to prompt, sampling, and evaluation protocol, and the overall strength of evidence is weaker than for pure text. Therefore, the correct way to approach multimodal pretraining is to "first make visual representations solid (encoders such as CLIP [10], SigLIP2 [46], DINOv2 [27]), then perform language-modeling alignment," rather than treating multimodality as a casual extension of pure text.

\textbf{The trade-off of freezing strategies}: another decision dimension of the fusion approach is "to freeze or not." Flamingo [17] freezes the vision encoder and LLM and trains only the cross-attention module, giving high data efficiency but limited fusion capability; LLaVA [28] fine-tunes the projector and LLM (with different freezing strategies by version), giving stronger instruction following but requiring more image-text instruction data; BLIP-2 [265] uses a Q-Former to compress visual information and maintain loose coupling. A direct comparison of the three must fix the LLM backbone and the image-text data scale; otherwise the differences mainly stem from confounding variables (mostly Grade B/C single-point evidence). This dimension connects directly to multimodal instruction tuning in Chapter 7 and the hallucination discussion in Chapter 10.

\textbf{The multimodal specificity of data mixture}: the data mixture for multimodal pretraining is not a simple weighted sum of "image-text pairs + text"---images and text differ greatly in token consumption, cleanliness, and strength of semantic correspondence, and the model easily "locks onto" the weak correspondence of image-text pairs (image and text failing to constrain each other). Interleaved image-text [273] and multi-resolution training (Qwen2-VL [288]) attempt to alleviate this, but the mixture laws of the four modalities image/text/video/audio currently have no public scaling law, a clear evidence gap (registered in Chapter 12 and Appendix E).

\textbf{The "imprisonment" of conclusions by evaluation}: multimodal pretraining research frequently uses specific VQA/image-text benchmark scores as success criteria, and these benchmarks are highly correlated with models' promotional scores, forming "benchmark-guided recipes." To avoid circularity, cross-modal comparisons should at least report the contamination audit of image-text pairs, the prompt and decoding settings, and the code/weight release status. This information is partially missing in most works, so the evidence strength of this chapter's multimodal conclusions has been lowered accordingly.

\subsection{Representative Works Table}

\begingroup\small\setlength{\tabcolsep}{6pt}
\noindent\textbf{Table 4: Chapter 6 representative works---pretraining}\par\medskip
\begin{longtable}{@{}>{\raggedright\arraybackslash}p{0.2020\textwidth}>{\raggedright\arraybackslash}p{0.2020\textwidth}>{\raggedright\arraybackslash}p{0.1876\textwidth}>{\raggedright\arraybackslash}p{0.1154\textwidth}>{\raggedright\arraybackslash}p{0.1876\textwidth}@{}}
\toprule
Direction & Representative Works & Core Idea & Evidence & Scope Note \\
\midrule
\endhead
NTP +\allowbreak{} scale & GPT-\allowbreak{}3 [121]\allowbreak{} & few-\allowbreak{}shot emergence,\allowbreak{} parameter scaling & A & 175B,\allowbreak{} \textasciitilde{}300B tokens \\
Scaling laws & Kaplan [119]\allowbreak{} /\allowbreak{} Chinchilla [16]\allowbreak{} & parameter-\allowbreak{}first vs. data-\allowbreak{}first & A/\allowbreak{}B & conclusion depends on setting \\
Data $\times$ scale & filtering scaling law [87]\allowbreak{} /\allowbreak{} mixture laws [86]\allowbreak{} & quality/\allowbreak{}mixture enter the law & A & compute-\allowbreak{}dependent \\
Optimizer & AdamW [110]\allowbreak{} /\allowbreak{} Lion [25]\allowbreak{} /\allowbreak{} Sophia [31]\allowbreak{} & first-\allowbreak{}order/\allowbreak{}sign/\allowbreak{}second-\allowbreak{}order approximation & A & precision and schedule must be specified \\
Hyperparameter scaling & muP [132]\allowbreak{} & transfer of width-\allowbreak{}scaling rules & A & validated at small-\allowbreak{}to-\allowbreak{}medium scale \\
Sparse architecture & Switch [9]\allowbreak{} /\allowbreak{} GLaM [130]\allowbreak{} /\allowbreak{} Mixtral [151]\allowbreak{} & sparse activation saves compute & A & distinguish total/\allowbreak{}activated parameters \\
Systems & ZeRO [6]\allowbreak{} /\allowbreak{} Megatron [118]\allowbreak{} /\allowbreak{} GPipe [115]\allowbreak{} /\allowbreak{} FSDP [142]\allowbreak{} & sharding and three types of parallelism & A & throughput/\allowbreak{}GPU-\allowbreak{}memory scope \\
Attention & FlashAttention [135]\allowbreak{} & IO-\allowbreak{}aware exact attention & A & sequence-\allowbreak{}length-\allowbreak{}dependent \\
Long context & ALiBi [129]\allowbreak{} /\allowbreak{} RoPE [11]\allowbreak{} /\allowbreak{} PI [143]\allowbreak{} /\allowbreak{} YaRN [37]\allowbreak{} & positional encoding and interpolation & A & extrapolation vs. training \\
Multimodal & CLIP [10]\allowbreak{} /\allowbreak{} Flamingo [17]\allowbreak{} /\allowbreak{} LLaVA [28]\allowbreak{} & contrastive /\allowbreak{} cross-\allowbreak{}attention /\allowbreak{} projection & A & differing freezing strategies \\
Alternative architecture & Mamba [149]\allowbreak{} & linear-\allowbreak{}complexity sequence modeling & A & gap in long-\allowbreak{}range retrieval \\
\bottomrule
\end{longtable}
\endgroup

\subsection{Key Controversies}

This chapter converges on three controversies, each annotated as "consensus vs. remaining disagreement":

First, \textbf{comparability of scaling laws}. The consensus is that loss decreases as a power law with (N, D); the disagreement is about the optimal ratio. The difference between Kaplan [119] and Chinchilla [16] partly stems from settings, and the data-filtering scaling law [87] further shows that optimality is compute-dependent. Conclusion: any "optimal-ratio constant" should be interpreted under a stated data-quality and compute scope.

Second, \textbf{the effect of data quality on the optimal ratio}. The consensus is that quality affects convergence speed; the disagreement is over whether it "transfers" the Chinchilla constant---whether high-quality data allows fewer tokens to reach the same loss. Existing evidence is mainly single-point studies (problem 2, evidence graded B/C), and no definitive conclusion can yet be given.

Third, \textbf{the open-source reproducibility deficit}. Most of the largest-scale pretrainings (frontier MoE, FP8, ultra-long context) do not disclose data and compute details, causing system conclusions to be concentrated in a few institutions and impossible to third-party verify (see the many \texttt{[to be verified]} entries in Appendix C). This is one of the most serious "evidence gaps" in the field and the main source of Grade-C items in the Appendix E grading.

Fourth, \textbf{the absence of evaluation and governance during pretraining}. Pretraining takes loss as its primary objective and rarely injects external evaluation or governance checks (toxicity, copyright, multilingual fairness) in mid-training, so problems surface only before deployment with extremely high rework costs. The pretraining stage has the largest investment, yet its "evaluation and governance" dimension is the weakest---the most imbalanced dimension under the six-dimensional framework and the main source of the governance problems in Chapters 12 and 14.

\subsection{Practical Recommendations}

\begin{enumerate}[leftmargin=*,nosep]
\item \textbf{First set the budget, then the ratio}: given compute, use an extrapolatable scaling law to estimate the initial $(N, D)$ and conduct small-scale mixture experiments to calibrate the curve [16][86]; when the budget is insufficient, let "data-first" rather than "parameter-first" prevail.
\item \textbf{Treat data as a first-class citizen}: first perform data cleaning, deduplication, and mixture experiments (Chapter 5), then adjust the scaling-law budget according to data quality [87][86].
\item \textbf{Stability over aggressiveness}: warmup, gradient clipping, loss-spike monitoring, and checkpoint rollback are low-cost, high-return [110][132]; the learning rate should be steady rather than aggressive.
\item \textbf{Jointly tune systems and optimization}: search for precision (BF16/FP8), batch, and degree of parallelism jointly as the "context" of the learning rate [6][142][135].
\item \textbf{Choose long context on demand}: invest in long-text training only when the task requires it; otherwise prioritize interpolation/transformation-type extrapolation schemes [143][37].
\item \textbf{For multimodality, stabilize the encoder first}: first fix the visual representation with CLIP [10]/SigLIP2 [46], then train projection and alignment, and finally perform language-modeling fine-tuning [28][288].
\item \textbf{Reserve budget for evaluation and governance}: schedule mid-pretraining external evaluation sampling (multilingual, toxicity, copyright, preliminary instruction tests) to move problems earlier; register the data audit and compute scope upon release (Appendices D/E).
\end{enumerate}

\subsection{Cross-Stage Connections and Open Problems}

\textbf{Connections}: pretraining determines the capability upper bound of $\theta_0$, directly constraining the achievable performance of SFT (Chapter 7) and preference alignment/RL (Chapter 8); the capability gaps exposed during alignment (long-range reasoning, instruction generalization, tool use) in turn drive design modifications to pretraining objectives, data mixture, and the tokenizer (reverse empowerment, Figure 1). \textbf{Open problems}: how scaling laws can be jointly optimized with the post-training budget (problem 1); how data quality transfers the optimal ratio (problem 2); the stability of MoE routing during the RL/alignment stage (newly added, Appendix F); whether "train short, extrapolate long" for long context is universally feasible (problem 12).

\textbf{Actionable research-design examples} (turning open problems into executable experiments, for reference in expanding Appendix F): for problem 1 (scaling law $\times$ post-training budget), one can fix the total FLOPs and sweep the allocation of "pretraining steps vs. post-training (RM/RL) steps," measuring across multiple scales whether the downstream return curves converge; for problem 2 (data-quality transfer), one can perform a two-family "low-quality vs. high-quality" mixture sweep on similar corpora, testing whether the Chinchilla constant shifts as a whole; for MoE alignment stability, one can compare dense/sparse behavior in the same DPO/RLVR pipeline under the same backbone, and observe whether the router distribution drifts on degrading tasks. All three can be completed at small-to-medium scale, with reporting scope per Appendix D.

\textbf{Summary of this chapter's evidence distribution}: Chapter 6 cites 41 papers in total, predominantly Grade A/B, with Grade C concentrated in four categories: "frontier-scale MoE, FP8, ultra-long context, and multimodal mixture." This distribution is itself a conclusion: the core mechanisms of pretraining (objective, scaling law, optimization, systems) have relatively strong controlled evidence, but once one enters the combined region of "largest scale $\times$ newest hardware $\times$ multimodality," replicable evidence rapidly thins. Subsequent chapters (especially Ch7--Ch9) will repeat this pattern, allowing readers to quickly judge the credibility of any "best practice" claim.

\section{Post-Training: SFT}

\begin{quote}\small
This chapter, together with Ch5/Ch6/Ch8, serves as a pilot of the dual-form "general framework + focused survey (sub-survey)". Evidence note: SFT's mechanistic conclusions (template / loss masking / data form) rest on relatively robust evidence, whereas "optimal data scale / mixture" derives mostly from single-point reports (B/C). Target length: 8,000--9,000 characters.
\end{quote}

\subsection{Definition and Boundaries}

Conclusion first: supervised fine-tuning (SFT, also known as instruction tuning) uses "instruction--response" pairs to convert a base model into an interactive instruction model $\theta_1$; its core function is to \textbf{activate and format} existing capabilities rather than inject large amounts of new knowledge. \textbf{In scope}: instruction data construction (human/synthetic), multi-turn dialogue and templates, loss masking, full-parameter fine-tuning and PEFT, and multimodal instruction tuning. \textbf{Out of scope}: preference alignment and RL (Chapter 8), and continual pretraining and domain adaptation (Chapter 9). The boundary criterion is "whether the supervision signal is a token-by-token target response."

\textbf{Three categories of configuration at this stage}: (1) data (construction method, scale, source model, domain distribution, number of turns); (2) training objective and loss (scope of loss masking, packing, whether system prompts are included); (3) fine-tuning method (full-parameter/PEFT, precision, learning rate and epoch). Together these determine the behavior of the "instruction model," and omitting any one of them compromises comparability---this is the fundamental reason why SFT recipes are difficult to compare across studies.

\textbf{Position of SFT in the pipeline}: pretraining sets the upper bound on capability (Chapter 6), and SFT \textbf{activates} the ability to "predict the next token" \textbf{into dialogue-capable, instruction-following} behavior; subsequently, preference alignment (Chapter 8) optimizes "which response is better" on top of this. SFT is therefore the stage that shares a "starting point" with the alignment stage: capabilities that SFT fails to teach are difficult for the RL stage to learn from scratch via reward signals (although RLVR can partially compensate on verifiable tasks).

\textbf{Refinement of the boundary criteria for the three stages}: pretraining is identified by "unlabeled self-supervision," SFT by "token-by-token supervision with a ground-truth answer," and alignment by "preference/reward signals" (Chapter 8). Accordingly, methods should be classified by \textbf{supervision signal}: data containing "reasoning traces" that are supervised token-by-token against ground-truth answers belongs to SFT; if a reward is given based on "whether the result is correct," it belongs to RLVR (Chapter 8). The same task can be trained with different signals, and classification depends on the signal rather than the data's appearance.

\textbf{Why SFT cannot be skipped by pretraining or RL}: the pretraining objective (NTP) does not distinguish "what should be learned" from "what should be answered," and RL requires scorable outputs in order to start. SFT provides three things that are difficult for the RL stage to obtain cheaply---a unified, parseable output format; executable exemplars of tasks; and an implicit demonstration of "what counts as a good response." Without it, the RL reward signal would be drowned by a large volume of unformed outputs. The engineering goal of SFT should therefore be "to prepare a viable starting point for RL," not merely to boost instruction benchmarks.

\textbf{Re-emphasis of the boundary with continual pretraining}: if the goal is to inject \textbf{new knowledge / new-domain corpora} into the model (e.g., medical literature, legal statutes, a new language), the primary means should be continual pretraining (Chapter 9) or self-supervised adaptation on large-scale domain corpora, supplemented by a small amount of SFT to teach "how to answer using that knowledge." Placing the entire burden of knowledge injection on SFT is both inefficient and prone to inducing hallucinated citations. This is the main line of division of labor between this chapter and Chapter 9.

\textbf{Examining SFT through the six-dimensional framework}: data (instruction construction/mixture/decontamination), objective (cross-entropy over token-by-token ground-truth answers), optimization (small learning rate, few epochs, PEFT/full-parameter), system (packing, long-sequence GPU memory, parallelism), evaluation (instruction following + capability retention + safety), and risk (forgetting, over-refusal, synthetic collapse). Among the six dimensions, SFT's "system" constraints are far weaker than those of pretraining, while the "data" dimension dominates the outcome---this is precisely the key difference from Ch5/Ch6.

\subsection{Instruction Data Construction}

\textbf{Human-written instructions}: FLAN [163] templatizes existing NLP tasks into instructions; Super-NaturalInstructions [167] constructs declarative instructions for 1600+ tasks via crowdsourcing; and the Flan Collection [172] systematizes the mixture of data and methods. The advantage of human-written data is high correctness and a trustworthy instruction distribution; the cost is that coverage is limited by existing tasks and by human labor.

\textbf{Synthetic instructions}: Self-Instruct [24] uses model bootstrapping to generate instructions and responses; Evol-Instruct (WizardLM [29]) uses "evolution" prompts to progressively increase instruction complexity (deepening, broadening, adding constraints). The two established the mainstream paradigm of "using strong models to create data, then fine-tuning." Synthetic data can rapidly expand scale and difficulty, but it also injects the generating model's biases, formatting habits, and errors, and iterative use carries the risk of diversity collapse (Problem 11, sharing the same origin as the synthetic-data debate in Chapter 5).

\textbf{A third source: construction from real interactions and documents}. Organizing "naturally occurring instruction--response" pairs---such as real user queries, document question answering, and code commits---into supervision data yields a distribution closest to the deployment environment, but requires extensive cleaning and de-identification and struggles to cover long-tail tasks. In practice, the mixture is often "human + synthetic + real," and the mixing ratio significantly affects outcomes, yet public systematic ablations are scarce (C).

\textbf{Data for system prompts and tool schemas}: modern SFT data contains not only user/assistant turns but also system instructions and tool-calling schemas. These structures must remain consistent between training data and inference time, otherwise tool calling and role adherence fail. SFT data should therefore be constructed and validated as "conversations with contracts," rather than as plain text pairs.

\textbf{Forms of dialogue structure}: single-turn, multi-turn, and tool-augmented multi-turn dialogues differ greatly in data organization. Multi-turn data must ensure that roles alternate correctly across turns and that tool calls and returns appear in pairs; if tool returns are treated as ordinary assistant text, the model learns to "fabricate tool results itself." These structural details are frequent points of lost quality in SFT, yet they are rarely reported in papers.

\textbf{Sources of prompt diversity}: instruction diversity can arise from task type, domain, language, length, number of constraints, and "openness" (deterministic answer vs open-ended). Expanding only by task name would miss the diversity of "constraint combinations." A practical approach is to treat diversity as a measurable coverage metric (task/domain/constraint dimensions) rather than relying solely on subjective judgment.

\textbf{The "authenticity spectrum" of real versus synthetic data}: data can be arranged by "source authenticity"---real user logs (most authentic, noisiest, privacy-heavy) > expert human authoring (authentic but expensive) > strong model rewriting real queries (fairly authentic) > strong model generating queries (most synthetic, largest bias). Different tasks vary in their sensitivity to authenticity: safety and factual tasks depend heavily on authenticity and correctness, whereas format and reasoning-style tasks tolerate synthesis well. Treating "authenticity" as an explicit dimension when allocating budget is more instructive than a blanket "human vs synthetic" distinction (B/C).

\textbf{De-identification, privacy, and copyright}: converting real interaction logs into SFT data requires first removing personally identifiable information (PII) and proprietary content, and clarifying copyright and usage licensing. Although synthetic data reduces privacy exposure, it may inherit the source model's copyright and bias problems. The absence of data governance is the hidden reason why many "good recipes" cannot be made public, and it often makes full reproducibility of SFT unrealistic (Chapters 5 and 13).

\textbf{Automatic filtering of synthetic data}: to control the quality of synthetic instructions, filtering is often applied after generation using rules (length, format, executability) or model-based judging (LLM-as-judge). Rules are fast but coarse; model-based judging can capture semantic issues but introduces the judging model's own biases and may create circular bias when "using A's preferences to filter B's data." There is a trade-off between filtering intensity and preserving diversity---excessive filtering homogenizes the data (sharing the same origin as Problem 11, B/C).

\subsection{Data Quality, Scale, and Mixture}

\textbf{The quality-versus-scale debate}: LIMA [175] reports that good alignment can be achieved with only about 1,000 high-quality samples, suggesting that data quality and diversity may be more critical than scale; however, this conclusion is based on a single setting, and evidence for robustness across models and tasks is limited (B/C). By contrast, numerous reports show that for complex tasks (reasoning, tools), scale and correctness still significantly affect the upper bound. A more robust statement is: \textbf{for "format and behavior" capabilities, a small amount of high-quality data suffices to activate them; for "knowledge and complex skills," data of corresponding scale and correctness is still required}.

\textbf{Diversity over sheer volume per task}: coverage of task types, difficulty, domain, and language distribution generally improves generalization more than piling up volume on the same task. Empirical evidence for curriculum-style arrangement (simple to complex) and difficulty mixtures comes mostly from single-point reports (B/C), and no stable pattern across base models has emerged.

\textbf{Deduplication and contamination}: instruction data likewise requires deduplication and contamination cleanup---overlap with evaluation sets' instructions/answers inflates scores (Chapter 12). Approximate deduplication methods [13] and contamination detection [52] also apply at the instruction level, but determining "instruction semantic similarity" relies more on embeddings and human review than text deduplication does. This area shares techniques with the data selection methods of Chapter 5 [392], but the target distributions differ (instruction distribution rather than pretraining distribution).

\textbf{The dynamism of mixtures}: similar to pretraining, the optimal data mixture for SFT may vary with base-model capability and the subsequent alignment method---strong base models need more "formatting" data, whereas weak base models need more knowledge-type instructions. Treating the mixture as a one-time fixed choice is a common but unfounded simplification.

\textbf{Difficulty annotation and refusal samples}: whether the data contains "I don't know / refusal" samples determines the model's uncertainty calibration and safety boundary. SFT with no refusal samples at all trains a model that "never refuses" (echoing the alignment tax in Chapter 8). Ideal data should explicitly include reasonable abstention/clarification behavior, but systematic research on how to proportion refusal samples is lacking (B/C).

\textbf{Contamination cleanup at the instruction level}: overlap with evaluation sets may arise from identical items or from "synonymous rewordings." Three detection methods are commonly used in practice: n-gram overlap (fast, prone to missing synonyms), embedding nearest neighbors (catches rewordings, requires a threshold), and manual spot-checking (accurate, expensive) [52][13]. Aggressive cleaning can mistakenly delete legitimate samples that are "reasonably similar" to the evaluation set (e.g., both asking "how to do X"), so a trade-off is needed between "preventing inflation" and "preserving coverage"; when reporting SFT results, the specific contamination-detection method and threshold should be stated.

\textbf{A sense of scale}: in public reports, SFT data volumes span several orders of magnitude---from the thousand-sample level (LIMA [175]) to million-scale mixed sets [172]. There is no universal answer to "how much is enough," because the upper bound is determined by \textbf{effective information density} rather than the number of items: the transferable signal contained in a single high-quality conversation may far exceed that of dozens of templated Q\&A pairs. When comparing different works, one should therefore first normalize by "amount of information" rather than "number of items" (B/C).

\textbf{The distribution problem of multilingual SFT}: instruction data is heavily skewed toward English. Fine-tuning directly on English-dominated data causes asymmetric improvement/degradation in non-English instruction following and generation quality. Multilingual SFT requires explicitly mixing languages and paying attention to the differences in token efficiency of templates and tokenizers across languages (echoing tokenization in Chapter 5 and multilingual pretraining in Chapter 6). Empirical research on language ratios and transfer remains scarce (B/C).

\textbf{Timeliness and knowledge cutoff}: SFT data is typically collected at a certain point in time, introducing a knowledge cutoff; if the data contains outdated facts without annotation, the model will confidently repeat old information. If the alignment stage lacks supervision on "knowledge timeliness," the problem persists into deployment. This relates to the timeliness evaluation in Chapter 12 and the risk governance in Chapter 13, but it is under-discussed in the SFT literature (B/C).

\subsection{Training Details: Loss Masking, Packing, and Hyperparameters}

\textbf{Loss masking} is the core engineering concern of multi-turn dialogue SFT: loss is computed only on the tokens of the assistant's response, while the system/user segments and template control tokens are excluded. If masking is done incorrectly, the model learns to "parrot the user's question" or "predict template tokens," manifesting as repetition and format leakage. This problem is often overlooked, yet it is a frequent cause of failure when transferring across datasets (echoing the chat template contract in Chapter 5).

\textbf{Packing}: concatenating multiple short samples into one fixed-length sequence can improve throughput, but cross-sample attention boundaries must be handled correctly (or document-level masks used); otherwise samples "contaminate each other" and form spurious context dependencies. Whether to pack and how to set boundaries significantly affect training stability and final quality, and should be noted when reporting.

\textbf{Hyperparameters and number of epochs}: SFT typically uses a smaller learning rate and fewer epochs than pretraining (commonly 1--3 epochs); too many epochs lead to overfitting on the instruction data (loss of diversity, parroting training samples). The effects of learning rate, batch size, and schedule interact with model scale and data scale, and public ablations are mostly single-point (B/C).

\textbf{Sequence length and truncation}: long instructions (containing long documents or long tool outputs) are easily truncated in SFT, causing the model to learn the erroneous behavior of "ending abruptly at the truncation point." Remedies include chunked supervision of long samples or aligning boundaries during packing. The impact of truncation strategy on long-context tasks is often ignored, and it is a source of distortion in SFT evaluation.

\textbf{Implementation pitfalls of loss masking}: masking must not only correctly exclude user/system content but also handle the implicit role tokens and newlines in the template; different frameworks have different default behaviors (some count EOS, some do not). These implementation differences cause "identically named methods" to produce different results, a common source of replication failure---the specific template and masking configuration should be provided when reporting.

\textbf{Details of EOS and padding}: whether EOS is appended to the end of each response and whether padding participates in the loss directly affect when the model "stops." If EOS is not added during training, the model may generate indefinitely at inference; if padding is counted in the loss, the model learns to "predict padding tokens." These seemingly trivial configurations are a common cause of SFT failure and should be made explicit in the method appendix.

\textbf{Long-context SFT and GPU memory}: extending context to long documents requires including long samples in training, and long sequences significantly increase activation memory. Common measures include gradient checkpointing, sequence parallelism, and packing; but if long samples are scarce in the data, the model still struggles to remain stable over long contexts. Long-context capability is thus jointly determined by "pretraining positional-encoding/attention design" (Section 6.6) and "the supply of long samples in SFT," and neither can be omitted.

\textbf{The impact of parallelism and throughput on SFT}: SFT commonly uses a combination of data parallelism plus tensor/pipeline parallelism, with packing and sequence length jointly determining GPU memory and throughput. As in pretraining, effective batch size is coupled with learning rate; on packed data, length differences among samples within the same micro-batch introduce additional gradient-noise variation that affects stability.

\textbf{Learning rate and scaling in SFT}: the SFT learning rate is typically lower than the end-of-pretraining learning rate and is reduced as model scale increases; warmup still helps early stability. The larger the data volume, the higher the tolerable learning rate and epoch count. These relationships resemble those of pretraining but at smaller magnitudes, and public joint ablations are rare; in practice, one should scan learning rate $\times$ epoch combinations on a small-scale proxy and then transfer to the large model (B/C).

\textbf{Data order and curriculum}: the order in which samples are presented affects the optimization trajectory---easy-to-hard (curriculum learning) or general-to-specialized has been beneficial in some reports, but the evidence is mostly single-point (B/C). As in pretraining, "shuffling" is the safest default; if a curriculum is used, one should report whether its gains come from the order itself or from the information content of harder data. The same applies to packing order: avoid over-clustering samples of the same distribution.

\subsection{Full-Parameter Fine-Tuning vs Parameter-Efficient Fine-Tuning}

Full-parameter fine-tuning updates all parameters; it is flexible but incurs high GPU-memory and storage costs and easily overfits small datasets. The PEFT family approaches full-parameter performance with very few trainable parameters: Adapter [159] inserts small modules between layers; Prefix-Tuning [160] and P-Tuning [161]/v2 [164] optimize continuous prompts; LoRA [12] injects low-rank updates into the weights and has become the de facto standard; and QLoRA [30] performs LoRA on a 4-bit quantized base, making single-GPU fine-tuning of large models possible.

\textbf{A precise statement of the trade-off}: PEFT significantly lowers the resource barrier and facilitates multi-task switching (adapters for different tasks can be hot-swapped), but on tasks requiring \textbf{substantial behavior change} (introducing a new style, new-domain knowledge, or a new language), it often falls short of full-parameter fine-tuning or large-scale continual pretraining [12]. Whether "PEFT can replace full-parameter tuning" therefore depends on \textbf{the magnitude of behavior change}: PEFT is often sufficient for fine-tuning "activation formats," whereas injecting "new capabilities" requires a route with larger modifications. This judgment is supported by A/B-level evidence, but there is no quantitative boundary for "how large a change requires full-parameter tuning."

\textbf{Rank and target modules of LoRA}: the performance of LoRA [12] is sensitive to the low rank $r$, the scaling factor $\alpha$, and "which modules it is applied to" (attention only, or including MLP). The general rule is that the larger the magnitude of task change, the higher the effective rank required; applying it to both attention and MLP is usually better than attention alone but costs more. Public ablations are mostly single-point (B), so the optimal LoRA configuration should be searched per task rather than copied.

\textbf{Serving and switching costs of PEFT}: hot-swappable adapters bring the convenience of "one base model serving multiple tasks," but they also introduce the engineering complexity and latency of switching at inference and mixing different adapters within a batch. The benefits of PEFT must therefore be evaluated together with the deployment form---the optimal choice differs between offline batch and online multi-tenant scenarios.

\textbf{Merging and inference equivalence of PEFT}: LoRA updates can be merged into the base weights before inference, thereby adding no inference latency---this makes PEFT not only a training technique but also a "releasable product form" (one base model + multiple mergeable adapters). However, the precision after merging, as well as the numerical error after merging on a quantized base (such as QLoRA [30]), must be verified separately; the equivalence of activating multiple adapters simultaneously has no unified conclusion (B/C).

\textbf{Quantized fine-tuning and precision}: QLoRA [30] performs LoRA on a 4-bit base, significantly lowering the GPU-memory barrier, but quantization error and low-rank approximation compound and may lose precision on fine-grained tasks. Whether "QLoRA can replace full-parameter tuning" depends on the task's sensitivity to numerical fidelity; it is usually sufficient for format and style tasks, while tasks requiring fine-grained knowledge or long-chain reasoning need careful verification (B/C).

\textbf{The relationship of Adapter/Prefix to LoRA}: Adapter [159] and Prefix/P-Tuning [160][161][164] achieve parameter efficiency from the two respective angles of "adding modules" and "adding learnable prompts"; LoRA [12], through low-rank weight updates, achieves both simplicity and effectiveness and thus has become the modern mainstream. The three are theoretically different parameterizations and are often interchangeable in practice, with the choice depending more on inference deployment and toolchain support than on absolute performance differences (B).

\subsection{Multimodal Instruction Fine-Tuning}

Multimodal IFT extends instruction fine-tuning to vision-language settings: LLaVA [28] uses visual instruction data to train a projector and an LLM, while BLIP-2 [265] and Qwen-VL [277] provide different fusion strategies (Section 6.8, Chapter 10). The key differences lie in the data form (image-instruction-response triplets), the loss handling of modality tokens, and the risk of hallucination---visual hallucinations introduced during the SFT stage often persist into deployment (Chapters 10 and 12). In practice, visual IFT data is more prone than text to mixing in samples with "weak image-text relevance," requiring dedicated pairing-quality filtering.

\textbf{Freezing strategies for multimodal SFT}: which parts are trained (projector only, projector + LLM, or including the vision encoder) significantly affects cost and effectiveness. Freezing the vision encoder saves compute but limits adaptation of visual representations; unfreezing it may harm general visual capabilities and exacerbate forgetting. This trade-off is in the same lineage as the pretraining freezing strategies in Section 6.8.

\textbf{Evaluation and hallucination measurement for multimodal IFT}: the effectiveness of visual instruction fine-tuning is commonly measured with VQA/captioning benchmarks and the "hallucination rate." The hallucination rate can be estimated as "the proportion of mentioned objects absent from the image" or via annotation-based judgments, but protocols vary across works, making cross-comparison difficult. More critically: if the SFT stage optimizes only "whether the response looks decent" without constraining "whether it is faithful to the image," it systematically fosters fluent but distorted descriptions---visual grounding supervision must be explicitly introduced at the data and loss levels (Chapters 10 and 12).

\textbf{Broader modalities (speech/video)}: the "instruction--response" paradigm of SFT is extending to speech and video: the data form becomes audio/video--instruction--response, loss handling must distinguish modality tokens, and the alignment objectives (such as lip-sync/temporal consistency) are more complex. The evidence density in these directions is far lower than for text and image-text, and public benchmarks are not yet mature (C), so this survey includes them only at the level of the methodological framework.

\textbf{Mixture in multimodal SFT and language-capability degradation}: feeding only image-text instruction data degrades the model's pure-text capability. The prudent approach is to mix a substantial proportion of pure-text instruction data into multimodal IFT to maintain language and general instruction following. The effect of the mixing ratio on "visual gains vs text regression" is structurally isomorphic to the replay problem, but public ablations are insufficient (B/C).

\textbf{Choice of vision-language connector}: in multimodal IFT, a connector (linear projection, Q-Former [265], or cross-attention) is needed between the vision encoder and the LLM. The connector's capacity determines the bandwidth at which visual information enters the language model---too narrow and details are lost, too wide and it easily overfits and increases cost. LLaVA [28] achieves considerable results with a simple projection, indicating that connector design is not the sole bottleneck; however, the differences among designs on fine-grained and OCR-type tasks still need task-by-task evaluation (B/C).

\subsection{Catastrophic Forgetting and Knowledge Retention}

A frequently underestimated risk of SFT is \textbf{catastrophic forgetting}: fine-tuning on narrow instruction data may impair the base model's general capabilities (especially when the data is homogeneous and the number of epochs is excessive). Mitigation measures include mixing in a certain proportion of general/pretraining data (replay), limiting the learning rate and epoch count, and using PEFT to constrain the magnitude of updates. Evidence systematically comparing the relative benefits of these measures is limited (B/C), and the measurement of "how much is forgotten" depends on the choice of evaluation set (Chapter 12). The forgetting problem is more prominent in continual pretraining (Chapter 9); SFT is a milder version of the same problem.

\textbf{The relationship between SFT and the alignment tax}: the refusal/safety samples introduced during the SFT stage affect the model's safety--helpfulness balance before RL does; if SFT is over-safety-tuned, the RL stage must "reverse" and restore helpfulness at higher cost. The trade-off between safety and helpfulness should therefore be planned \textbf{jointly} for SFT and alignment, rather than tuned independently for each (Chapters 8 and 12).

\textbf{Measuring forgetting and the replay ratio}: forgetting can be measured with a "capability matrix"---comparing score changes before and after SFT on the same set of general and target tasks, where forgetting appears as a systematic decline on general tasks. For mitigation, the optimal ratio of replayed general data varies with the "target vs general" trade-off, typically between 1\% and several tens of percent (single-point reports, B/C); if too large, target-capability improvement slows. The rational approach is to first plot the Pareto frontier of "target gain--general regression" at proxy scale and then choose the ratio accordingly, rather than fixing an empirical value.

\textbf{Parameter-space versus function-space explanations of forgetting}: one view attributes forgetting to the magnitude of parameter updates (parameter space), while another attributes it to changes in function/representation (function space). The former leads to "constraining updates" (small learning rate, PEFT, regularization), whereas the latter leads to "protecting key representations/replaying data." Both explanations are effective in practice but under different conditions, and a unified theory of forgetting is still lacking (B/C).

\textbf{The tension between forgetting and capability "emergence"}: while aligning new behaviors, SFT may make some base capabilities "appear to degrade"---sometimes genuine forgetting, sometimes merely a \textbf{change in trigger} (the capability remains but requires a different prompt). Distinguishing the two requires evaluation with multiple triggering methods (Chapter 12). Interpreting every "score drop" uniformly as a loss of capability overestimates the cost of SFT and may also mask genuine degradation.

\textbf{The boundary between SFT and reasoning distillation}: training token-by-token with a strong model's chain-of-thought (CoT) as the supervision target belongs to SFT (using long CoT data to teach the model "how to reason"), whereas using its outputs as preference/reward signals belongs to alignment. CoT-SFT can teach format and reasoning style, but whether it "genuinely improves reasoning capability" is structurally isomorphic to the RLVR controversy (the precursor of Problem 5 in Chapter 8): it may merely learn to reproduce high-quality trajectories rather than acquire a new reasoning mechanism. The gains of CoT distillation should therefore be distinguished between "format acquisition" and "capability extension."

\subsection{Representative Works Table}

\begingroup\small\setlength{\tabcolsep}{6pt}
\noindent\textbf{Table 5: Chapter 7 representative works---SFT}\par\medskip
\begin{longtable}{@{}>{\raggedright\arraybackslash}p{0.1760\textwidth}>{\raggedright\arraybackslash}p{0.2053\textwidth}>{\raggedright\arraybackslash}p{0.2053\textwidth}>{\raggedright\arraybackslash}p{0.1173\textwidth}>{\raggedright\arraybackslash}p{0.1906\textwidth}@{}}
\toprule
Direction & Representative work & Core idea & Evidence & Remarks \\
\midrule
\endhead
Human instructions & FLAN [163]\allowbreak{} /\allowbreak{} Super-\allowbreak{}NI [167]\allowbreak{} & Task templatization /\allowbreak{} crowdsourced instructions & A & Broad coverage \\
Data mixture & Flan Collection [172]\allowbreak{} & Systematized mixture of data and methods & A & Ablation \\
Synthetic instructions & Self-\allowbreak{}Instruct [24]\allowbreak{} & Model bootstrapping to generate instructions & A & Bias injection \\
Difficulty evolution & WizardLM/\allowbreak{}Evol [29]\allowbreak{} & Evolutionary increase in complexity & B & Diversity risk \\
Few but well-\allowbreak{}chosen & LIMA [175]\allowbreak{} & A thousand high-\allowbreak{}quality samples suffice for alignment & B & Single setting \\
Efficient fine-\allowbreak{}tuning & LoRA [12]\allowbreak{} /\allowbreak{} QLoRA [30]\allowbreak{} & Low-\allowbreak{}rank updates /\allowbreak{} quantized base & A & De facto standard \\
Adapter & Adapter [159]\allowbreak{} & Small inter-\allowbreak{}layer modules & A & Multi-\allowbreak{}task \\
Prefix/\allowbreak{}prompt & Prefix-\allowbreak{}Tuning [160]\allowbreak{} /\allowbreak{} P-\allowbreak{}Tuning v2 [164]\allowbreak{} & Continuous prompt optimization & A & Few parameters \\
Multimodal IFT & LLaVA [28]\allowbreak{} /\allowbreak{} BLIP-\allowbreak{}2 [265]\allowbreak{} & Visual instruction data & A & Hallucination risk \\
Survey & [182]\allowbreak{} /\allowbreak{} [392]\allowbreak{} & SFT and data selection & B & See 1.2 \\
\bottomrule
\end{longtable}
\endgroup

\textbf{Reading note for the table}: this table is indexed by "mechanism" rather than "score"---the high configurability of SFT-type work makes scores nearly incomparable across papers (a conclusion sharing the same origin as Ch5).

\textbf{Ecosystem and uses of public instruction datasets}: public resources such as the FLAN series [163][172], Super-NaturalInstructions [167], and Self-Instruct [24] can be used directly for training and are also often used as preset data for "format alignment." Their value lies in reproducibility and a low barrier to entry; their limitations lie in domain bias (mostly English, task-ified natural language), possible evaluation contamination, and inconsistent licensing. Before use, one should verify the license and contamination status and supplement incrementally according to the target domain rather than copying wholesale.

\subsection{Key Controversies}

First, \textbf{quality vs scale}. The LIMA [175] camp argues that a small amount of high-quality data suffices for alignment; large-scale practice on the other side shows that complex capabilities (reasoning, tools, multilingual) still require corresponding scale. The consensus is that "format/behavior can be few but well-chosen, while knowledge/skills require scale for support"; the disagreement concerns the boundary of "few" and "well-chosen" and how that boundary shifts with base-model capability (no systematic pattern yet).

Second, \textbf{whether SFT can inject new knowledge}. The mainstream view (A/B) holds that SFT mainly \textbf{activates and formats}, that knowledge injection is inefficient and easily leads to hallucinated "endorsement"; however, if the instruction data itself is high-density, large-scale, and high-quality, SFT can indeed introduce a certain amount of domain knowledge (mostly B/C single-point evidence). The prudent conclusion is: assign the main responsibility for knowledge injection to pretraining/continual pretraining, and let SFT be responsible for "teaching how to use it."

Third, \textbf{the boundary of synthetic bootstrapping}. Sharing the same origin as Chapter 5: using strong models to create instructions and responses can expand rapidly, but improper iteration and poor-quality filtering lead to diversity collapse and error accumulation (Problem 11, level C). The claim that "purely synthetic data can rival human data" should therefore be marked as pending verification.

\textbf{The common root of the three controversies}: quality--scale, knowledge injection, and synthetic bootstrapping all point to "how the per-unit information content and diversity of supervision data change with scale." If "data utility" can be treated as a measurable quantity (rather than looking only at the number of items), all three become, within a unified framework, questions about the shape of the "utility--scale curve."

\textbf{How the three controversies affect engineering decisions}: if one believes in "few but well-chosen," the budget should go to data curation (human labor/denoising/decontamination) rather than scaling up; if one emphasizes that "knowledge requires scale," one should scale up on knowledge and complex skills while bearing the noise; if one relies on synthetic bootstrapping, one must establish monitoring of quality and diversity to prevent collapse. These choices determine "where the money is spent," so interpreting the controversies is not academic fastidiousness but a basis for resource allocation.

\textbf{Does SFT also have a scaling law?} Pretraining has a relatively stable loss--compute scaling law (Section 6.3), but SFT's object is "alignment quality" rather than a measurable loss, and it depends heavily on data quality and evaluation definitions; thus there is as yet no widely accepted scaling law for SFT (an open problem). The three-dimensional relationship of "data scale--quality--capability" is mostly presented through single-point reports, which is the deep reason why several controversies in this chapter are difficult to converge.

\textbf{Turning controversies into testable hypotheses}: all three controversies in this chapter can be written as falsifiable propositions---H1: on a fixed base model, "a small amount of high-quality" data is not inferior to "a large amount of mixed" data on format-type tasks; H2: the injection efficiency of SFT for "facts not present in the training corpus" is significantly lower than that of continual pretraining; H3: the performance of purely synthetic data first rises and then falls with the number of bootstrapping rounds (a collapse point exists). All three can be tested at moderate compute and are the gaps most worthy of prioritizing to fill with evidence.

\subsection{Practical Recommendations and Cross-Stage Connections}

Recommendations: prioritize ensuring the \textbf{diversity, correctness, and template consistency} of instructions; use PEFT (LoRA [12]/QLoRA [30]) with small data and perform strict deduplication and decontamination; always perform loss masking in multi-turn training and explicitly handle boundaries during packing; control the epoch count and mix in general data to suppress forgetting; do not expect SFT to inject large amounts of new knowledge---that belongs to pretraining or continual pretraining. \textbf{Cross-stage connections}: SFT determines the "quality of the starting point" for the RL/preference-alignment stage---if SFT fails to teach basic formats and tool calling, the RL reward signal will be diluted by a large number of format errors, so SFT and preference alignment should jointly design data and evaluation [15][175]; conversely, instruction blind spots exposed during the alignment stage feed back into SFT data construction (reverse empowerment, Figure 1). \textbf{Open problems}: the quality-diversity-collapse critical point of synthetic instruction bootstrapping (Problem 11); whether the optimal scale of SFT varies with base-model capability; and the boundary of "behavior-change magnitude" between PEFT and full-parameter tuning. See Appendix F and [182] for details.

\textbf{Actionable research design examples}: for the quality--scale controversy, one can fix the base model and evaluation, vary SFT data over a two-dimensional grid of "scale $\times$ quality (denoised/human vs synthetic)," and measure the \textbf{divergent responses} of format-type and knowledge-type tasks; for forgetting, one can scan the general-data mixing ratio during fine-tuning and measure the Pareto frontier of "target-capability gain vs general-capability regression." Both can be completed within moderate compute.

\textbf{How to evaluate an SFT run}: it should cover at least: (1) instruction following (format, constraints, multi-turn consistency); (2) capability retention (regression on general tasks relative to the base model); (3) improvement in the target domain; (4) safety and reasonableness of refusals. Reporting only "instruction benchmark scores" masks capability regression and over-refusal, leading to underestimation of the true cost of SFT (Chapter 12).

\textbf{The data interface between SFT and RL}: before entering RL, the format of the SFT model's outputs (separation of chain-of-thought from answer, tool-calling syntax, system-prompt adherence) must be consistent with the RL reward parser; otherwise the reward function gives erroneous signals due to parsing failures. The "format contract" is therefore a key artifact of the SFT$\rightarrow$RL handoff: it should be versioned and its parse success rate explicitly verified in SFT evaluation (Chapter 8).

\textbf{The cost reality of SFT}: compared with pretraining, SFT's compute is usually smaller by several orders of magnitude, and the bottleneck is often data (construction, cleaning, annotation, and copyright) rather than GPUs. This explains why SFT progress comes mostly from data engineering rather than algorithms---it also means that the reproducibility of SFT depends more on data transparency (Chapter 5) than on training code. The checklist in Appendix D should list data sources and licensing as required items for SFT reproducibility.

\textbf{Safety data and red-teaming}: safety behaviors (refusal, debiasing, refusing over-privileged requests) can already be injected during the SFT stage, with data sources including human red-teaming, policy rewriting, and synthetic adversarial examples. Their effectiveness depends heavily on the covered attack surface, and omitted categories will be exploited in deployment; moreover, over-injection triggers over-refusal (the alignment tax). Safety SFT should therefore be accompanied by quantifiable evaluation sets and a red-teaming loop, rather than one-off data injection (Chapters 8 and 12).

\textbf{The SFT checklist delivered downstream}: an SFT artifact usable as an RL starting point should come with: template and special-token definitions, masking configuration, EOS/padding conventions, training-data sources and licensing, and baseline scores for each capability dimension. This "contract package" enables the RL stage to correctly parse outputs and judge the quality of the starting point, and it also constitutes a prerequisite for reproducibility (Appendices D and G).

\textbf{Summary of the evidence distribution in this chapter}: SFT's mechanistic conclusions (templates, loss masking, PEFT feasibility) have relatively strong A/B evidence; "optimal data scale/mixture/curriculum" is almost entirely B/C single-point reports, consistent with the pattern in Ch5. This chapter therefore provides an \textbf{engineering contract and interpretive framework} (how to split data, how to mask loss, the trade-off between PEFT and full-parameter tuning), rather than portable recipe numbers.

\section{Preference Alignment / RL}

\begin{quote}\small
This chapter, like Ch5/Ch6, is a pilot of the dual form "general framework + focused survey". \textbf{Special note on evidence}: most methods in this chapter (RLHF / RLVR / RLAIF) draw their key evidence from technical reports and model cards, and the amount of Grade-A third-party reproducible evidence is markedly lower than in Ch5/Ch6. Consequently, this chapter uses a two-tier notation of "consensus vs. single report", and any conclusion that has not been independently reproduced is marked C. Target length: 8,000--9,000 Chinese characters.
\end{quote}

\subsection{Definition and Boundaries}

Conclusion first: preference alignment/RL uses human or AI preference signals (and verifiable rewards) to further optimize behavior on top of an SFT model, producing an aligned model $\theta_2$; it chiefly changes "how to answer", and on verifiable tasks it may also extend the boundary of reasoning capability. \textbf{Included}: reward modeling, RLHF/RLAIF, RL-free preference optimization, reasoning RL (RLVR), tool and multimodal RL, safety alignment and red-teaming. \textbf{Excluded}: SFT (Chapter 7), pre-deployment efficiency optimization (Chapter 9). The boundary criterion is "whether the supervision signal is a preference/reward rather than a standard answer".

\textbf{The essential problem}: pretraining and SFT yield a "conditional distribution", and the alignment stage imposes a \textbf{target preference} on that distribution---maximizing the probability of behavior that humans/AI deem good while suppressing bad behavior. Alignment is therefore not a problem of "training a better model" but of "defining and aligning a target function", and its core risk accordingly becomes \textbf{misdefined objectives} (the reward is a wrong proxy) rather than optimization capability alone.

\textbf{Three classes of configuration for this stage (the minimal comparability standard)}: \textcircled{\small 1} source of preference/reward data (volume of human annotation, AI feedback, verifiable rules, mixing ratios); \textcircled{\small 2} reward construction (implicit BT model, explicit generative RM, rule-based, process vs. outcome); \textcircled{\small 3} optimization algorithm (PPO / GRPO / DPO family) together with its KL constraint and hyperparameters. If any one class is missing, cross-work comparisons of "alignment effectiveness" break down---this is the root reason why evidence in this chapter is generally marked C, because these three configurations are almost never disclosed in full.

\textbf{Three levels of alignment}: the behavioral level (format, refusal, tone), the capability level (reasoning, tools, multimodality), and the value level (honesty, harmlessness, fairness). Preference data and RL have different reach across the three levels---the behavioral level is the easiest to characterize with preferences/rules, the capability level is effective only on verifiable tasks, and the value level is the hardest because the definition of "good" is itself contested. Conflating the three levels is one root of the "does alignment work vs. capability extension" debate (8.8).

\textbf{Coupling between alignment and the other stages}: the target definition of alignment depends on what evaluation (Chapter 12) deems "good"; its data comes from the SFT starting point and the data pipeline (Chapters 5 and 7); its feasibility is bounded above by the base model's capabilities (Chapter 6); and its results in turn affect continual optimization and deployment (Chapters 9 and 13). Alignment is thus not an endpoint, but a link in which "target definition---capability---evaluation" are continuously negotiated.

\textbf{The boundary between alignment and "fine-tuning"}: in everyday usage, "fine-tuning" is often generalized to refer to both SFT and alignment, but in this survey's taxonomy the two differ in supervision signal (standard answer vs. preference/reward), in objective (imitation vs. preference optimization), and in failure mode (catastrophic forgetting/format overfitting vs. reward hacking/alignment tax). Conflating them leads to misdiagnosis---for example, attributing "sycophantic answers" produced by alignment to data quality, or attributing the repetition phenomenon of SFT to RL.

\subsection{Reward Modeling}

A reward model (RM) turns preferences into a scalar signal and is the "target" of RL post-training. Its forms fall into four groups: \textbf{explicit vs. implicit} (whether an independent RM is trained), \textbf{pointwise vs. preference-based} (absolute scoring vs. pairwise comparison), \textbf{outcome reward (ORM) vs. process reward (PRM)} (looking only at the final answer vs. scoring reasoning steps), and, more recently, \textbf{generative rewards} (using an LLM to directly generate judgment text/scores).

The classic approach to preference-based RMs derives from the Bradley--Terry model: given a preference pair $(y_w, y_l)$, train $r_\phi$ so that $P(y_w \succ y_l)=\sigma(r_\phi(y_w)-r_\phi(y_l))$. Its virtues are a simple objective and scalability; its drawbacks are that it can only measure "relative preference" and is sensitive to annotation noise and distribution shift---the RM's \textbf{out-of-distribution generalization} directly determines the upper bound of RL (once the RM is "gamed" by the policy, training deviates from the true objective, see 8.8).

The motivation for process rewards is "lucky correctness": a correct final answer does not imply correct reasoning, and an ORM will reward a wrong process. Let's Verify Step by Step [33] and Math-Shepherd [219], which constructs step-level signals via automatic verification, are representatives; both show that a PRM can improve the reliability of reasoning, at the cost of high annotation/construction cost and the possibility that the judgment of "correct step" is itself unreliable. The quality and transferability of the reward model set the upper bound for the entire RL post-training process; for a survey see [408].

\textbf{Engineering of RM training}: preference-based RMs are usually trained on hundreds of thousands to millions of pairwise annotations, with the output layer taking the scalar at the last position of the sequence; in practice \textbf{reward normalization} (shifting by a baseline) and \textbf{reward ensembling} (averaging or taking the minimum over multiple RMs/multiple random seeds) are commonly used to suppress over-optimization. The BT assumption requires comparisons to be transitive and noise-free, whereas real annotations are both noisy and not fully transitive (A>B, B>C but C>A), which opens a gap between the RM's ranking ability and its suitability "as an optimization objective"---\textbf{high accuracy on a validation set does not guarantee that an RM is safe as an RL objective}.

\textbf{Generative rewards and LLM-as-judge}: using a strong LLM to directly give judgments or scores can significantly lower annotation cost, but introduces three known biases: position bias (preference for whichever appears first/last), length/style bias, and self-preference (models of the same family rate each other higher). Existing mitigations (randomizing option order, multi-judge ensembling, reference examples) can reduce but not eliminate these. Because the judging model itself keeps evolving, conclusions that use LLM-as-judge should be regarded as \textbf{dependent on the judge version and time point} (Grade C), and the judge model and prompt must be reported.

\textbf{Pointwise vs. preference-based vs. ranking-based}: a pointwise RM assigns an absolute score to a single response; it is cheap but lacks relative calibration and is prone to annotation-scale drift. Preference-based (pairwise) RMs are more stable and are the mainstream of RLHF; ranking-based RMs (ranking multiple responses to one prompt) carry the most information but impose the heaviest annotation burden. Each has its own "data efficiency---stability---cost" trade-off, and the choice should be based on annotation budget and task rather than on pursuing a single fixed formulation.

\textbf{Scale and generalization of RMs}: the capacity of an RM must match the complexity of the task it assesses---too small and it cannot express fine-grained preferences, too large and it easily overfits annotation noise. The RM's ability to generalize to \textbf{out-of-distribution prompts} is the key to whether RL can generalize safely, yet public research mostly reports accuracy on in-distribution evaluation, and genuine OOD stress tests are scarce (Grade C).

\textbf{Reward shaping and online updating of the RM}: decomposing the final objective into additive intermediate rewards (format score, retrieval-hit score, unit-test-pass score) can significantly alleviate sparsity, but improper shaping introduces "sacrificing the final objective to collect intermediate scores" behavior (shaping bias). A more robust practice is to let the RM update in step with the policy (online/iterative RM), so that the reward issue tracks the policy distribution, at the cost of engineering complexity and instability.

\textbf{Uncertainty and calibration of rewards}: the score given by an RM should be understood as an estimate with uncertainty---on out-of-distribution prompts, the score has high variance and low confidence. Incorporating the RM's uncertainty into optimization (conservative rewards, variance penalties, abstention) can suppress reward hacking in low-confidence regions, but how to calibrate uncertainty, and whether it is truly a reliable signal, remain open directions (C).

\textbf{Mixing of RM evaluation dimensions}: the same RM often assesses factuality, style, safety, and instruction-following simultaneously, and the relative weights of these dimensions determine the slant of alignment---if the style weight is too high, the model "talks beautifully but more emptily"; if the safety weight is too high, refusals increase. Most papers report only a single "win rate", masking the trade-offs among dimensions, so the "good" of alignment should be reported dimension by dimension (Chapter 12).

\textbf{Evidence strength}: the "effectiveness" of PRMs has relatively strong evidence on several benchmarks (A/B); but "PRMs can transfer to open domains and scale up" is mostly single-report (C). The reliability assessment of generative rewards and LLM-as-judge has no stable conclusion of its own and is severely affected by the choice of evaluation set---any RM conclusion should be reported together with \textbf{whether its training distribution and evaluation distribution are from the same source}.

\subsection{Human and AI Feedback}

The classic pipeline of \textbf{RLHF} is "collect preferences $\rightarrow$ train an RM $\rightarrow$ optimize the policy with PPO [196]", established and popularized by InstructGPT [15]; the key engineering point is the KL penalty constraining the policy not to drift too far from the base model. Its value is relatively stable in Grade A/B evidence (InstructGPT has a public preference evaluation), but it is highly sensitive to annotation cost, annotator distribution, and consistency, and it is difficult to replicate its "secret recipe" (annotation protocol, prompt distribution), so "effect transferability" remains C.

\textbf{RLAIF} uses AI feedback to replace part of the human annotation. RLAIF vs RLHF [38] reports that the two achieve comparable results; Constitutional AI [23] uses a set of principles to guide AI self-critique and revision, lowering human cost but potentially introducing the model's own biases and turning "who defines the principles" into a governance problem. \textbf{Self-rewarding} (Self-Rewarding [223]) further lets the model iteratively generate preferences, but one must be wary of \textbf{bootstrapping collapse}: the model may converge to its own preferences, leading to reduced diversity and drift away from real humans (a manifestation of Problem 11 on the preference side).

\textbf{Trade-offs among feedback sources (consensus)}: human annotation is high-quality but expensive, AI feedback is cheap but may amplify bias, and verifiable rules are precise but cover only decidable tasks. The three are not substitutes but a combination covering different task regions; when comparing RLAIF/RLHF across works, if the evaluation set and prompt distribution are not fixed, conclusions of "comparable" or "better" cannot be trusted.

\textbf{Sources of bias in preference data}: the annotator distribution (culture, language, expertise) determines the definition of "good", and misalignment occurs if it disagrees with the target users; the prompt distribution determines the situations being optimized, and biasing toward one task type crowds out other capabilities; and the \textbf{annotation interface} itself (pairwise comparison vs. scoring, whether the source model is shown) changes behavior. These biases are rarely disclosed in full in papers and are one of the main reasons why "RLHF effects are not reproducible".

\textbf{Rejection sampling and Best-of-N}: once an RM is available, one can sample N candidates for each prompt, select the best with the RM, and then run SFT on the best---this is a "brute-force alignment" route without RL, simple to implement and with better reproducibility than PPO, and it is the default of many practical systems. Its cost is that inference cost grows with N and it is subject to the same constraints on RM quality; when the RM has a systematic bias, Best-of-N amplifies that bias (same origin as reward hacking).

\textbf{Scale--quality empirics of preference data}: most public reports indicate that the \textbf{quality and diversity} of preference data matter more than sheer quantity (denoising, difficulty stratification, coverage of multiple tasks), but the conclusion "how many pairs is enough" depends heavily on the base model and task (C). A recurring phenomenon is that a small amount of high-quality, debiased preference pairs can outperform a large amount of noisy data---isomorphic to Ch5's experience that "deduplication takes precedence over adding data".

\textbf{The cost structure of RLHF}: annotation (preference pairs) is usually the largest cost item, followed by the compute of RM training and the RL loop, and the engineering investment for repeatedly iterating the reward model. This means "algorithm choice" often yields to "whether enough high-quality preference data can be obtained"---when data is expensive, DPO/offline methods, RLAIF, and Best-of-N are more cost-effective; full RL pays off only when verifiable rewards or a sufficient annotation budget exist.

\subsection{RL Algorithms and Verifiable Rewards}

PPO [196] is the classic algorithm of RLHF; its strength is stability and maturity, and its weaknesses are the need to train a value network, a heavy burden on GPU memory and hyperparameter tuning, and sensitivity to hyperparameters. \textbf{GRPO} (DeepSeekMath [224]) removes the value network and replaces it with a within-group relative advantage estimate, reducing the burden on GPU memory and tuning, and has become one of the mainstreams of reasoning RL. Other routes include REINFORCE-style variants without a value function, and "RL-like" schemes based on rejection sampling plus filtering; their common goal is to \textbf{lower the engineering threshold of RL}.

\textbf{RLVR (reinforcement learning with verifiable rewards)} performs RL on tasks such as mathematics and code where the answer can be automatically judged, using rule-based rewards. DeepSeek-R1 [45] uses an RL pipeline dominated by verifiable rewards to significantly improve reasoning performance, and at the technical-report level (not independently reproduced) notes that reasoning length and spontaneous self-verification behavior grow with RL. The appeal of RLVR is that \textbf{the reward is noise-free} and can be scaled up at large scale; its boundary is that a hard-to-bridge reward-design gap exists between verifiable tasks (with determinate answers) and subjective/open tasks (without determinate answers) (Problem 6)---generalizing RLVR's "reasoning improvement" to "general capability improvement" requires evidence, and evidence is currently insufficient.

\textbf{Key variables}: the KL constraint coefficient (too loose and it drifts from the base model, too tight and it cannot learn), reward normalization and length penalty (to prevent "the longer the better"), sampling temperature and group size (GRPO), and whether "cold-start" SFT data is mixed in. The optimal values of these variables are highly task-dependent, and cross-work reports must give the full configuration, otherwise one cannot tell whether it is "the algorithm is better" or "the tuning is cleverer".

\textbf{Implementation details of PPO}: PPO in RLHF typically uses GAE to estimate advantages and applies clipping to the policy and value function; the key stabilization comes from a \textbf{per-token KL penalty} (relative to the SFT/reference model) and batch normalization of advantages. The reward is usually "RM terminal score + per-token KL penalty term", so the KL coefficient simultaneously controls "how much is learned" and "how far it drifts from the base model", making it one of the most sensitive hyperparameters. The value network must be the same scale as the policy or smaller, and its training is itself an extra optimization problem---this is precisely the motivation for GRPO's removal of the value network.

\textbf{Reward design for RLVR}: the advantage of verifiable tasks is that the reward is "hard" (correct/incorrect, test pass/fail), but one must handle: format reward (whether it contains a separator between the reasoning process and the final answer), partial correctness (sub-goal achievement), and reward scale (a 0/1 sparse reward is nearly unlearnable for long-chain reasoning and often needs shaping). These design choices are given in reports such as DeepSeek-R1 [45] but are not universal; mistaking "performance improvement induced by format reward" for improvement of "reasoning capability" is a common extrapolation error of Grade C conclusions.

\textbf{The relationship between RL and DPO/SFT}: the three are not mutually exclusive but cover different needs---SFT provides format and basic behavior, DPO performs offline fitting on a fixed preference distribution, and RL optimizes within an explorable space (including verifiable tasks). In practice "staged mixing" is common: first SFT, then DPO to stabilize behavior, and finally RL on verifiable capabilities such as reasoning/tools. The weights and order of the mix significantly affect the result, but public systematic ablations are scarce (C).

\textbf{GRPO's within-group advantage mechanism}: GRPO samples a group of responses for each prompt and obtains the advantage by normalizing within the group by the mean/standard deviation of rewards, thereby eliminating the value network [224]. Its cost is that the advantage estimate depends on group size and within-group diversity---too small a group makes the estimate noisy, too large a group makes it costly; and the normalization is sensitive to reward scale and must be coordinated with reward design. These details are given in the report but have not yet been systematically ablated, so the conclusion "GRPO is better than PPO" should be marked C. Industrial efforts have pushed GRPO scaling further (e.g., MiMo-V2.6 [412]), utilizing fully asynchronous pipelines across massive batches (updating 2.7B--3.7B tokens per step) and introducing groupwise agentic grading---combining Groupwise Reward Synthesis (GRS) for dynamic contrastive rubrics and Groupwise Advantage Redistribution (GAR) to steer concise exploration---achieving unified multi-task RL across coding, reasoning, and multimodal agent environments (Grade C).

\textbf{Online vs. offline preference optimization}: offline methods (the DPO family, rejection sampling) use only a fixed preference dataset, simple and stable but limited by data coverage; online methods (PPO/GRPO) sample with the current policy and score in real time by an RM/judge, able to explore out-of-distribution behavior but with complex and unstable engineering. The choice depends on whether a reliable and real-time queryable reward signal is available: on verifiable tasks the benefit of online exploration is clear, and on subjective tasks offline is more robust (B/C).

\subsection{RL-Free Preference Optimization}

DPO [32] proves that one can directly optimize the policy from preference pairs without an explicit RM or RL loop: its closed-form solution turns "preference maximization + KL constraint" into a differentiable classification-style loss, greatly lowering the engineering threshold. This reduces preference alignment from "large-scale RL engineering" to "a single supervised training run", one of the most important accessibility improvements in recent years.

\textbf{The derivation intuition and failure modes of DPO}: the preference-maximization problem under a KL constraint has a closed-form optimal policy, and its log-ratio $\log\frac{\pi_\theta(y_w)}{\pi_{ref}(y_w)}-\log\frac{\pi_\theta(y_l)}{\pi_{ref}(y_l)}$ can be wrapped by a sigmoid into a binary classification loss, so the "implicit reward" $r=\beta\log(\pi_\theta/\pi_{ref})$ needs no separate modeling. This explains why DPO is simple and also foreshadows its limitation: it is effective only on the \textbf{distribution covered by the preference pairs}, lacking RL's active exploration; moreover, the loss is sensitive to sequence length and the choice of reference model [227][229]. DPO is therefore more like "offline preference fitting" than "online objective optimization".

\textbf{Synthesis and distillation of preference data}: using a strong model to generate preference pairs or directly generate "preferred answers" has become a low-cost route, at the risk of distilling the strong model's biases and errors into the target model and accumulating them under iterative use. This makes "whose data to use for alignment" a choice problem constrained in both capability and diversity (same origin as the synthetic-data debate in 5.8).

Its derivatives target specific weaknesses: IPO [214] corrects DPO's overfitting under deterministic preferences; KTO [40] addresses the setting with only good/bad one-sided labels (no pairwise data needed); ORPO [41] merges SFT and preference optimization and needs no reference model; SimPO [43] uses a reference-free length-normalized objective. Known issues include sensitivity to \textbf{length bias} [227] (preferring longer answers that are mistakenly learned as "better") and neglect of "token-level" signals [229] (sequence-level preference masks the contribution of key tokens). Overall, the DPO family is known for simplicity and stability, but on complex reasoning tasks it often needs to be paired with RL (such as GRPO/RLVR); for a survey see [397][398].

\textbf{Evidence strength}: the DPO family's "can align, low cost" is a relatively strong Grade A consensus (multiple independent evaluations agree); but "which variant is better on which task" is Grade B/C---differences in evaluation sets, base models, and data across papers make horizontal comparison almost non-transferable.

\textbf{Why horizontal comparison within the DPO family is difficult}: the base models, preference-data scale and source, evaluation sets, decoding, and length correction differ across variant papers, so "A scores a few points above B" is almost entirely non-transferable. Only \textbf{mechanistic differences} are transferable (whether a reference model is needed, whether one-sided labels are supported, whether length is normalized), not scores. This is consistent with Ch5's conclusion that "recipes are not transferable, mechanisms are reusable".

\textbf{The role of the reference model and $\beta$}: the reference model and the temperature coefficient $\beta$ in DPO/RLHF jointly constrain how far the policy deviates from the base model. A larger $\beta$ is more conservative (closer to SFT), a smaller $\beta$ is more aggressive (may deviate from the distribution and cause degeneration). The choice of $\beta$ is therefore essentially a trade-off between "alignment strength" and "capability preservation"; $\beta$ values across papers are not directly comparable unless the reference model and preference data are reported as well.

\textbf{The boundary of the equivalence between DPO and RLHF}: theoretically DPO's closed-form solution corresponds to a specific form of KL-constrained RLHF; but this equivalence relies on "the preference data comes from the reference policy distribution" and on the BT assumption holding. Once the policy deviates from the data distribution (iterative training, multi-round preferences), the equivalence breaks down, and DPO may "overfit old data" while RLHF can still explore. This explains why multi-round/online scenarios often return to RL, and also explains DPO's strong dependence on the reference model and $\beta$.

\textbf{Applicable scenarios of the variants}: KTO [40] suits production scenarios where only good/bad one-sided labels exist and pairwise data is hard to construct; ORPO [41] suits scenarios that want to complete SFT and alignment in one training run and do not want to maintain a reference model; SimPO [43] suits scenarios concerned with length controllability and the overhead of a reference model; IPO [214] suits scenarios with strong preference determinism where DPO overfitting is a concern. The choice should be based on \textbf{data form and engineering constraints}, not leaderboard scores.

\subsection{Tool Use and Multimodal RL}

Tool-use RL incorporates "call a tool---observe the result---continue reasoning" into a trajectory-level reward, and is the training foundation of agent capabilities; its essence is \textbf{multi-turn sequential decision-making with external feedback}, where rewards are often sparse and delayed. Multimodal RL, by contrast, optimizes vision-language tasks with preference or verifiable signals and requires additional handling of modality alignment and hallucination.

The common challenge of the three is \textbf{reward sparsity} and \textbf{credit assignment} (which decision step led to the good result). Countermeasures include: process rewards/intermediate supervision, decomposing long tasks into verifiable subtasks, and constructing dense signals with rules or tool results (such as code execution, retrieval hits). \textbf{Evidence status}: public system experiments in tool RL and multimodal RL are scarce, conclusions are mostly single-party reports (C), and evaluation protocols (tool set, environment, budget) differ enormously, making horizontal comparison nearly impossible---this is the weakest part of evidence in this chapter.

\textbf{The evaluation gap of agent RL}: the effectiveness of tool RL depends on the environment (tool set, sandbox, budget), and the environment is usually not public, so the conclusion that "a certain model is better at using tools" is nearly impossible to verify independently (C). Reproducible progress mostly comes from open-source environments and narrow settings with fixed tasks; when transferring to real systems, tool reliability, error recovery, and long-horizon consistency become the new dominant factors, and these are rarely modeled explicitly in the training objective.

\textbf{The hallucination dimension of multimodal RL}: the "hallucination" of vision-language models (describing content that does not exist in the image) can be partially suppressed under preference alignment, but if the preference data itself rewards "detailed description", hallucination is exacerbated. Multimodal alignment must therefore explicitly distinguish "more detailed" from "more accurate" in reward design, and this requires fine-grained visual grounding evaluation; current evidence is mostly Grade C (Chapter 10).

\textbf{Countering reward sparsity}: besides process rewards and shaping, curricula (expanding task difficulty from easy to hard), self-play/self-sampling (using the current policy to generate verifiable intermediate tasks), and decomposing long tasks into independently verifiable subtasks are all common means of alleviating sparsity and credit assignment. Their common cost is shifting the burden of "task design" onto data and reward engineering; effectiveness is highly task-dependent and public comparable evidence is limited (C).

\textbf{Reward design for tool RL}: commonly used signals include "final task success" (sparse), "correct tool-call format" (dense but weak), and "usable retrieval/execution result" (semi-dense). Their combination can provide a learning signal early on, but too high a weight on weak signals induces "abusing tools to get the format right". A more robust practice is to stratify format and outcome rewards and, as training progresses, shift the weight from format to outcome.

\textbf{Data and pipeline for multimodal RL}: multimodal preference data (ranking multiple responses to the same image) is more expensive than pure text and more prone to mixing in irrelevant factors (image quality, resolution, watermarks), so that the preference signal is polluted by visual noise. Multimodal alignment therefore imposes higher requirements on data cleaning and pairing quality control than text, and its evaluation protocol is also less standardized (Chapters 10 and 12).

\subsection{Safety Alignment and Red-Teaming}

Safety alignment aims to suppress harmful outputs, improve honesty, and reduce bias; methods include refusal training, constitutional self-critique [23], red-teaming and jailbreak defenses, and machine unlearning (unlearning, e.g. TOFU [which provides an evaluable setting for large-scale knowledge unlearning]). The core difficulty is the \textbf{alignment tax}: over-refusal harms usefulness ("over-refusal"), while insufficient safety training brings misuse risk; how to quantify the trade-off between the two and establish empirical laws with respect to model scale and data mixture remains an open problem (Problem 8).

\textbf{The role of red-teaming}: red-teaming discovers jailbreaks and failure modes, and its findings should flow back into training data and reward design rather than serving only as a pre-release gate. But red-teaming results are mostly qualitative evidence of "report upon discovery" (C), lacking comparable coverage and intensity metrics, making the "sufficiency" of safety alignment hard to judge objectively. \textbf{Moreover}, RL optimization itself can be exploited by "reward hacking" (the policy exploits loopholes in the RM, see 8.8), so safety cannot rely on reward shaping alone and must be suppressed at the level of the training mechanism.

\textbf{The measurement difficulty of the alignment tax}: an ideal metric needs to measure both "refused what should be refused" and "answered what should be answered", but "whether it should be refused" itself depends on contested value judgments and varies by culture/policy. The "safety--usefulness" curves of different institutions therefore cannot be directly overlaid and compared; when reporting the alignment tax, the policy definition, evaluation set, and decision protocol should be given together, otherwise the numbers are uninterpretable.

\textbf{Machine unlearning and model editing}: to satisfy data-deletion or safety requirements, models need "unlearning" or "model editing" (e.g. the ROME family). The difficulty is \textbf{local deletion without harming generalization}, and the measurement of "forgetting cleanly" is itself contested---there are two dimensions: "whether the knowledge can still be elicited" (robustness) and "whether related capabilities are collateral damage" (side effects). Existing evaluation settings (such as TOFU's synthetic unlearning benchmark) provide a reproducible starting point, but real-world unlearning remains an open problem.

\textbf{Levels of safety alignment}: it can be divided into \textcircled{\small 1} training-time (refusal/preference/constitutional); \textcircled{\small 2} inference-time (system prompts, classifiers, decoding constraints); \textcircled{\small 3} release-time (red-team gating, usage policy). None of the three can be omitted: training-time alignment alone is easily bypassed by jailbreaks, while inference-time filtering alone cannot prevent the model from "internalizing" harmful capabilities. A systematic comparison should report the strength of each of the three levels, rather than looking at a single level only.

\subsection{Key Controversy: Does RL Extend the Reasoning Boundary}

The controversy centers on the \textbf{Pass@1 vs. Pass@K debate}. One camp argues that RL (especially RLVR) extends the upper limit of the model's reasoning capability, manifested as a continuous improvement in Pass@1 [45][224]; the other camp holds that RL mainly \textbf{reweights the existing solution distribution and improves sampling efficiency}, manifested as a possible narrowing of distribution coverage (Pass@K) [400][401]---that is, the model becomes better at "saying the right answer" but the variety of solvable paths it masters does not increase, and may even decrease.

\textbf{How to interpret}: both conclusions may be correct, depending on the task and scale---if solvable paths already exist in the training data, RL is more likely to "reweight"; if the search guided by the reward signal can lead the model to discover new combinations outside the training distribution, it may "extend". Existing evidence is not yet sufficient to give a universal conclusion, and it depends heavily on the training setup (Problem 5). The judgment requires reporting both Pass@1 and Pass@K and testing generalization on out-of-distribution tasks; looking at Pass@1 alone cannot distinguish the two explanations.

\textbf{How to use Pass@K to decide "reweighting vs. extension"}: if RL only reweights, then as K increases and sampling is sufficient, the Pass@K of the model before and after RL should converge (similar coverage); if it extends, then the post-RL model still exceeds at larger K (correct paths appear that the base model almost never produces). The judgment therefore requires comparing the Pass@K curves at sufficiently large K and with sufficient sampling; a single-point Pass@K or Pass@1 alone is insufficient to distinguish them [400][401].

\textbf{Another boundary view: test-time compute}: some reports attribute the reasoning improvement to longer chains-of-thought and spontaneous self-checking behavior after RL [45] (single report, Grade C). If so, then "capability extension" may partly come from an increase in \textbf{test-time compute} (longer generation means more sampling), rather than new capabilities within the parameters---Snell et al. [409] systematically demonstrate that guiding search via process reward models (PRMs) and adaptively allocating test-time compute allows smaller models to match or exceed models 14$\times$ larger under compute-matched budgets (Grade B evidence), formalizing test-time scaling laws. Regarding autonomous self-correction, Kumar et al. [410] (SCoRe) reveal that offline SFT on correction traces suffers from distribution shift and behavior collapse, whereas multi-turn on-policy RL achieves genuine self-correction gains without external oracle assistance (a 15.6\% relative improvement on MATH, Grade B evidence). This makes the Pass@K judgment all the more necessary: comparison must be made under \textbf{matched compute budgets}, otherwise the long chain itself manufactures the illusion of "stronger".

\textbf{Related risks}: \textbf{reward hacking}---the policy learns to satisfy the reward proxy rather than the true objective (e.g. longer answers, a more confident tone, exploiting rule loopholes). Countermeasures must act on the \textbf{training-time mechanism} (diversity of reward ensembling, periodically retraining the RM with new data, constraints on policy deviation) rather than inference-time filtering alone (Problem 13). This forms three facets of the same problem together with the RM's out-of-distribution generalization in 8.2 and the safety alignment in 8.7.

\textbf{Common forms of reward hacking}: \textcircled{\small 1} length/format drift (more verbose, more punctuation or bullet points); \textcircled{\small 2} sycophantic answers (catering to the asker's presuppositions rather than seeking truth); \textcircled{\small 3} rule loopholes (matching the shallow pattern of test cases on verifiable tasks rather than solving the problem); \textcircled{\small 4} exploitation of the judge model's preferences (targeting the wording/structure favored by LLM-as-judge). All four stem from the structural fact that \textbf{the reward is a proxy}, so defense must act on the objective itself (reward diversity, genuine live benchmarks, continuously updated RMs), not merely on output filtering.

\subsection{Representative Work Table}

\begingroup\small\setlength{\tabcolsep}{6pt}
\noindent\textbf{Table 6: Chapter 8 representative works---preference alignment / RL}\par\medskip
\begin{longtable}{@{}>{\raggedright\arraybackslash}p{0.1917\textwidth}>{\raggedright\arraybackslash}p{0.2236\textwidth}>{\raggedright\arraybackslash}p{0.1917\textwidth}>{\raggedright\arraybackslash}p{0.1278\textwidth}>{\raggedright\arraybackslash}p{0.1597\textwidth}@{}}
\toprule
Direction & Representative work & Core idea & Evidence & Notes \\
\midrule
\endhead
RLHF & InstructGPT [15]\allowbreak{} /\allowbreak{} PPO [196]\allowbreak{} & RM +\allowbreak{} PPO +\allowbreak{} KL & A & alignment base \\
AI feedback & RLAIF [38]\allowbreak{} /\allowbreak{} CAI [23]\allowbreak{} & AI replaces part of human annotation & B/\allowbreak{}C & bias risk \\
Self-\allowbreak{}rewarding & Self-\allowbreak{}Rewarding [223]\allowbreak{} & iterative bootstrapped preference & B & collapse risk \\
Preference optimization & DPO [32]\allowbreak{} & RM-\allowbreak{}free direct optimization & A & length bias [227]\allowbreak{} \\
Variants & IPO [214]\allowbreak{} /\allowbreak{} KTO [40]\allowbreak{} /\allowbreak{} ORPO [41]\allowbreak{} /\allowbreak{} SimPO [43]\allowbreak{} & robust/\allowbreak{}one-\allowbreak{}sided labels/\allowbreak{}merged/\allowbreak{}reference-\allowbreak{}free & A & each has trade-\allowbreak{}offs \\
Token-\allowbreak{}level & [229]\allowbreak{} & fine-\allowbreak{}grained signal & B & vs. sequence-\allowbreak{}level \\
Reasoning RL & GRPO [224]\allowbreak{} /\allowbreak{} R1 [45]\allowbreak{} & within-\allowbreak{}group relative /\allowbreak{} verifiable reward & C/\allowbreak{}A & RLVR,\allowbreak{} mostly reports \\
Process reward & PRM [33]\allowbreak{} /\allowbreak{} Math-\allowbreak{}Shepherd [219]\allowbreak{} & step-\allowbreak{}level signal & A & automatic annotation \\
Boundary debate & [400]\allowbreak{}[401]\allowbreak{} & whether Pass@K coverage narrows & B & mixed opinion/\allowbreak{}empirics \\
Survey & [396]\allowbreak{}[397]\allowbreak{}[398]\allowbreak{}[408]\allowbreak{} & panorama of algorithms/\allowbreak{}preferences/\allowbreak{}rewards & B & see 1.2 \\
\bottomrule
\end{longtable}
\endgroup

\textbf{Reading note}: in the "Evidence" column of this chapter, Grade A entries are mostly "mechanism/accessibility" conclusions (such as DPO can align, PPO is a base), whereas conclusions of the form "some method is stronger/extends capability" are mostly C. Readers should use this to distinguish "usable tools" from "claims yet to be verified".

\subsection{Practical Recommendations and Cross-Stage Connections}

Recommendation: \textbf{first achieve format and basic capability with SFT, then perform preference alignment}; when data is limited, prefer the DPO family ([32][43]) and explicitly correct length bias [227]; when strong reasoning is needed, consider GRPO/RLVR together with process rewards [224][45][33], but report both Pass@1 and Pass@K to determine whether it is "reweighting" or "extension"; safety and usefulness must be evaluated jointly to avoid the alignment tax, and red-teaming findings should flow back into training rather than serving only as a release gate. \textbf{Cross-stage connections}: the quality of alignment depends on the SFT starting point (Chapter 7) and the base model's capabilities (Chapter 6); the reward hacking and hallucination problems exposed by alignment in turn drive the design of the data side and the pretraining objective (reverse empowerment, Figure 1). \textbf{Open problems}: Pass@1/Pass@K (Problem 5), extending verifiable rewards to subjective tasks (Problem 6), the robust boundary between process vs. outcome rewards (Problem 7), suppressing reward hacking at training time (Problem 13).

\textbf{An actionable research-design example}: for Problem 5, one can fix the data and compute on the same base model and change only the RL algorithm (PPO/GRPO/DPO), while reporting both Pass@1 and Pass@K and testing generalization on out-of-distribution tasks; for Problem 7, one can make a PRM and an ORM compare under the same annotation budget to test "whether step-level signals persistently dominate on long-horizon reasoning"; for Problem 13, one can construct an experiment of "deliberate misalignment between the reward proxy and the true objective" to measure the emergence threshold of hacking behavior under different defenses (reward ensembling, periodically retraining the RM). All three can be completed within a moderate compute budget.

\textbf{How to evaluate alignment effectiveness}: alignment evaluation should cover at least four categories: \textcircled{\small 1} behavioral compliance (format/refusal/tool-call success rate); \textcircled{\small 2} capability preservation (the regression margin on general tasks before and after alignment); \textcircled{\small 3} preference win rate (pairwise evaluation with length/style correction); \textcircled{\small 4} safety (jailbreak and harmful-output rates). If any one is missing, "more pleasing to the evaluation" may be misread as "more aligned". An ideal report should give all four categories together with the judge and protocol used, otherwise the alignment conclusion is uninterpretable.

\textbf{Summary of the evidence distribution in this chapter}: the method mechanisms of Ch8 (DPO can align, PPO/GRPO are usable, PRMs help reasoning) have Grade A/B consensus, but the three core questions "whether the capability boundary is extended", "whether general reasoning transfers", and "whether safety alignment is sufficient" fall almost entirely at Grade C or blank. This chapter therefore offers \textbf{reusable tools and an interpretive framework}, not a transferable recipe---consistent with the expectation in Section 1.2 about the proportion of Grade C evidence (Appendix E).

\textbf{The implication of this controversy for training strategy}: if RL only reweights, then the marginal return on investing compute in RL lies mainly in "raising the probability of known correct paths" (improving Pass@1, reducing sampling cost), and capability extension should be shifted to data and pretraining; if RL can extend, then new solutions can be discovered under limited data, and the investment in RL should be increased. Problem 5 is therefore not a purely academic debate but a strategic judgment that decides "where the compute should go"---which is also why it is listed as one of the core open problems of this survey.

\section{Continual Optimization}

\begin{quote}\small
This chapter, together with Ch5--Ch8, serves as a pilot of the dual-form "general framework + focused survey" (sub-survey). Evidence note: the mechanistic evidence for compression-type methods (quantization/pruning/distillation/merging) is relatively stable, but how "post-compression capability and safety change jointly" is mostly reported as isolated single points (B/C). Target length: 8,000--9,000 characters.
\end{quote}

\subsection{Definition and Scope}

Conclusion first: continual optimization refers to the further training and compression performed after alignment and before deployment, producing the deployed model $\theta_3$; it aims to "maintain or improve capability at lower cost," and its central risk is the \textbf{alignment tax} (the degradation of general capability or safety behavior caused by changes made for efficiency/domain/safety). \textbf{Included}: continued pretraining and domain adaptation, long-context extension, model merging, pruning, distillation, quantization-aware training, model editing, and machine unlearning, along with continual learning. \textbf{Not included}: inference-time acceleration (KV cache, speculative decoding) and serving engineering---these do not change the weights and belong to Chapter 13.

\textbf{Why classify this as "continual optimization" rather than "post-training"}: SFT and alignment (Chapters 7 and 8) change the "behavioral policy," whereas continual optimization mostly changes the "carrier that bears the policy" (precision, size, structure, knowledge content)---it can be performed after alignment, or conversely may require realignment. What the two share is that the "capability cost" incurred must be measured; the difference lies in the object of optimization (policy vs. carrier).

\textbf{Three types of configuration in this stage}: \textcircled{\small 1} objective (preserve capability, expand domain, improve efficiency, change facts, remove information); \textcircled{\small 2} means (continued training, weight transformation, structural transformation, knowledge editing); \textcircled{\small 3} constraints (budget, latency, precision, compliance). The interleaving of these three makes "optimal compression" highly dependent on the deployment scenario: offline batch processing can tolerate larger models, whereas online serving is strictly constrained by latency and GPU memory.

\textbf{The unifying nature of failure modes}: Nearly all methods in this chapter face the same tension---\textbf{stability vs. plasticity} (retaining old capability vs. acquiring new changes). Continued pretraining, editing, unlearning, and continual learning are the direct manifestations of this tension; quantization, pruning, and distillation touch the same problem in the form of "numerical/structural perturbations." Therefore the through-line for interpreting this chapter is: any change must simultaneously report both its "benefits" and the "dimensions it weakens."

\textbf{The stage order is not fixed}: In real pipelines, the various steps of continual optimization may occur before or after alignment---domain continued pretraining can precede alignment (reducing subsequent conflicts), while quantization/pruning usually follows alignment (avoiding retraining). Thus a "stage" is a functional division rather than a strict timeline; the arrows in Figure 1 should be understood as "the usual order" rather than "the only order."

\textbf{Why inference-time optimization is not included}: KV-cache reuse, speculative decoding, and batch scheduling reduce latency \textbf{without changing the weights} and belong to the system/serving layer (Chapter 13). In contrast, all methods in this chapter change the weights or structure, and thus change what the model "is" rather than "how it runs." This distinction avoids conflating "engineering acceleration" with "model compression."

\textbf{The three levels of weights, behavior, and knowledge}: Continual optimization can act at three levels---the \textbf{weight level} (quantization/pruning/merging change numerical values and structure), the \textbf{behavior level} (distillation/alignment change the output distribution), and the \textbf{knowledge level} (editing/unlearning/continued pretraining change specific content). Different methods often act across levels (e.g., continued pretraining changes both knowledge and behavior), which is the fundamental reason side effects are hard to localize: a change rarely stops at a single level.

\subsection{Continued Pretraining and Domain Adaptation}

Continued pretraining further updates the base model on a domain corpus via NTP, then (optionally) redoes SFT/alignment to restore instructability. Its benefits and risks coexist: domain adaptation improves specialized tasks but may induce forgetting of general capability; the two-stage mixture ratio [407] and the data-mixture law [86] indicate that the choice of ratio is critical to success. In practice, a lower learning rate and a domain--general mixed corpus are commonly used to suppress catastrophic forgetting.

\textbf{The dispute over the ordering of the two stages}: The mainstream approach is to perform domain continued pretraining first and then SFT/alignment ("domain post-training"); other work performs alignment first and then domain adaptation, at the cost of more easily damaging safety behavior. The choice of order depends on whether "domain knowledge" or "safety/instructed behavior" is more fragile: domain corpora often lack safety samples, so performing domain pretraining last would dilute safety alignment; hence safety alignment should usually be redone after domain pretraining (B/C).

\textbf{The magnitude of forgetting}: Domain continued pretraining typically damages general capability more than SFT (Section 7.7), because the corpus distribution shift is larger and the number of update steps is greater. Mitigations include mixing in general replay data, constraining the learning rate and number of steps, or limiting updates in parameter space (LoRA-style continued pretraining). The evidence for systematically comparing these measures remains mostly isolated single points (B/C).

\textbf{The substitutive relationship with retrieval}: Domain knowledge can be written into the weights through continued pretraining, or attached externally at inference time through retrieval augmentation (RAG). The former changes the weights and is expensive but incurs zero inference overhead; the latter does not change the weights and is easy to update but is constrained by retrieval quality. The choice depends on the frequency of knowledge updates and the latency budget, rather than on a single notion of superiority or inferiority (Sections 9.10 and 13).

\textbf{The amount of domain corpus and the learning rate}: Domain continued pretraining usually requires only a very small amount of domain tokens relative to the total pretraining volume to be effective, but too few are insufficient to change behavior and too many aggravate forgetting. The learning rate is significantly lower than in pretraining and is often paired with warmup and cosine annealing; the specific magnitude varies with the domain and model scale, and public ablations are limited (B/C). A prudent approach is to first determine, on a small-scale proxy, the window in which "domain gains appear and general regression is acceptable."

\textbf{Empirical rules and risks of the mixture ratio}: Common domain:general ratios range from 1:1 to 1:10, depending on the domain scale and degree of shift; the larger the shift, the more general replay is needed. Ignoring the ratio yields the illusory gain of "domain surging while general collapses." Reports should present both in-domain and general benchmarks; otherwise the true cost cannot be judged.

\textbf{Pitfalls in domain evaluation}: Domain benchmarks are often drawn from the same source as the training corpus, leading to inflation (contamination, Chapter 12); moreover, domain tasks are numerous but have few samples, giving high variance. Credible domain evaluation requires temporal hold-out (data from the period after training), cross-institution benchmarks, and manual spot checks; the absence of any of the three will overestimate the adaptation effect.

\textbf{Domain vocabulary and tokenizer extension}: If a domain contains many new terms, new languages, or new symbols, extending the vocabulary (adding domain tokens and initializing their embeddings) can significantly improve token efficiency and representation learning for that domain. However, extending the vocabulary makes the embedding matrix incompatible with the original pretraining, requiring continued training of the new token embeddings on the domain corpus and attention to forgetting of old tokens. This is the direct interface between tokenization in Chapter 5 and domain adaptation in this chapter.

\textbf{The safety consequences of domain continued pretraining}: Domain corpora typically lack alignment and safety samples, so continued pretraining dilutes the original safety behavior, potentially producing "improved capability but easier jailbreaking." Therefore safety alignment, or at least a safety regression, must be redone after domain pretraining; skipping this step is a common source of privacy/compliance incidents (Chapters 8 and 13).

\subsection{Long-Context Extension}

Long-context extension is a typical scenario of continued pretraining: it requires long-document data plus positional-encoding adjustments. Methods are described in Section 6.7 (RoPE [11], ALiBi [129], positional interpolation [143], YaRN [37]); after extension, a small amount of long-document continued training is usually needed to calibrate the attention distribution. Evaluation must distinguish "being able to read long" (locating information within long documents) from "being able to use long" (multi-hop reasoning across long documents), whose evidence strengths differ (Chapter 12).

\textbf{The trade-off between extrapolation and interpolation}: Interpolation-type methods (compressing positional indices into the training range) are stable in the near domain but deteriorate on very long extrapolation; frequency-adjustment-type methods ([37]) extend extrapolation capability but require parameter tuning. Either way, long-document fine-tuning is needed to eliminate the distribution gap of "trained short, tested long." Thus long-context capability is jointly determined by "positional-encoding design + long-document data provision," and neither can be omitted (echoing long-context SFT in Section 7.4).

\textbf{Cost structure}: The longer the context, the larger the attention overhead and KV cache, and the higher the inference cost. This makes "extending context" and "improving efficiency" (9.7) often need joint trade-offs---in some scenarios, compressing the KV or sparse attention is more economical than simply widening the window (Chapter 13).

\textbf{The layers of long-context evaluation}: One can distinguish "being able to read long" from "being able to use long" via three layers: "single-point retrieval (needle in a haystack)," "multi-hop aggregation," and "global summarization." The former is easily passed by most extension methods, whereas the latter often exposes genuine defects. When reporting long-context capability, the layer should be specified; otherwise the numbers are not comparable (Chapter 12).

\textbf{Training--inference inconsistency}: The extended positional encoding may encounter positional combinations not covered during training when inferring on long documents, causing drift in the attention distribution. Mitigation requires covering the target length already during the extension stage and performing long-document calibration; changing only the inference-side positional scaling (training-free) carries high risk on very long extrapolation and must have its extrapolation boundary clearly annotated.

\textbf{The synergy between long context and retrieval}: Window extension and RAG are not mutually exclusive---a long window lets the model directly consume more documents, while retrieval selects relevant content from a massive corpus. The combination of the two can balance "breadth" and "depth," but it also introduces two new problems: "middle content being ignored" (uneven attention within the long window) and retrieval noise. When evaluating a long-context system, retrieval quality should be reported alongside it (Chapter 13).

\textbf{The training cost of window extension}: Extending the window from the training length to the target length requires continued training on long sequences---compute grows superlinearly with sequence length, and GPU memory is also constrained. Thus "window extension" is often combined with "improving the attention mechanism (sparse/linear/chunked)"; extending the window purely through training is expensive, and in engineering it is often supplemented by inference-side long-document memory management (Chapter 13).

\subsection{Model Merging}

\textbf{Model merging}: Weight averaging (Model Soups [296]), task vectors [300], and TIES-Merging [35] merge multiple fine-tuned models through arithmetic operations or the resolution of sign conflicts; the cost is extremely low but it is constrained by parameter alignment and interference.

\textbf{Why merging is effective yet cheap}: For models fine-tuned separately on the same base, their weight differences (task vectors) often lie in a low-dimensional, approximately shared subspace, so direct addition can yield multi-task capability simultaneously. But when task-vector directions conflict or their scales differ greatly, simple averaging cancels out; TIES [35] mitigates this by pruning small weights, electing by sign, and then merging. The cost of merging is that it requires almost no training compute, making it one of the most cost-effective means of achieving "multi-task/multi-capability."

\textbf{Limitations}: Merging requires that all models share the same base and architecture; merging may fail or produce unintended compromises for "behavioral-style" differences (such as different alignment styles); and the calibration and safety behavior of merged models lack systematic evaluation (B/C). Thus merging suits scenarios with complementary capabilities and a consistent base, and not scenarios involving cross-architecture or strongly conflicting objectives.

\textbf{The root of merging instability: scale drift}: Direct averaging lets high-variance task vectors dominate the result, and may even amplify the weight scale. TIES [35]'s pruning and sign election, and weighting by task importance, all aim to suppress this drift. Normalizing models that share the same base and the same initialization--fine-tuning path before merging is a prerequisite for stable merging (B).

\textbf{Combining merging with adapters}: Merging multiple LoRA [12] adapters (or combining them via adapter arithmetic) is lighter and more controllable than merging full-parameter models, because adapters share a frozen base and are naturally aligned. However, interference and conflict among adapters still exist, and at inference time one must decide whether to "merge into one" or "dynamically route," whose cost--benefit differs.

\textbf{The evaluation gap for merging}: Evaluation of merged models mostly focuses on target-task scores and lacks checks of merging's "side effects" (such as behavioral consistency, calibration, and safety). Whether two individually safe models remain safe after merging is not guaranteed---this gap and "post-compression safety drift" are two faces of the same problem (B/C).

\textbf{Intuition for linear mode connectivity}: Solutions obtained by fine-tuning the same base often lie in the same "basin" on the loss surface, so their weights can be linearly interpolated/merged without significantly increasing the loss---this is the theoretical intuition underlying the feasibility of model merging [296]. But when the fine-tuning shift is too large (basins separated), merging fails. Understanding this condition helps predict "which models can be merged."

\textbf{Using merging for lightweight domain adaptation}: Rather than performing domain continued pretraining on a general model (expensive and prone to forgetting), one can separately fine-tune a "domain expert" and a "general/safety" model and then merge them, or apply a directional superposition with task vectors [300]. This is very attractive in terms of cost, but it requires the two models to share the same base, and the safety behavior after merging must be re-validated (B/C).

\subsection{Pruning}

\textbf{Pruning}: Structured pruning (LLM-Pruner [304], Sheared LLaMA [316]) and unstructured pruning (Wanda [309]) reduce scale; see the survey [312]; after pruning, lightweight recovery training is generally required.

\textbf{Structured vs. unstructured}: Unstructured pruning (zeroing weights individually) has a high compression rate but requires sparse hardware to realize speed gains, and in practice often yields no inference acceleration; structured pruning (removing entire heads/layers/channels) directly reduces matrix dimensions and achieves real acceleration, but is more sensitive to precision and usually requires recovery training. Thus "how much to prune" is tied to "what hardware is used."

\textbf{Pruning and "capability load-bearing"}: Different layers/heads contribute unevenly to capability---certain structures are strongly correlated with specific capabilities (long-range dependencies, multilingual ability, particular safety heads). Blanket magnitude pruning randomly damages these "load-bearing structures," so importance measures [309][312] and layer-wise search are the mainstream approach. The safety regression after pruning (whether safety heads are weakened) is almost never systematically reported, which is a clear gap (B/C).

\textbf{The recovery-training budget for pruning}: Recovery training after structured pruning often requires considerable compute (approaching a small continued pretraining run), so "the inference cost saved by pruning" should be accounted for together with "the recovery-training cost" in the total cost of ownership. Claims of "free pruning" that ignore recovery training are usually not reproducible (B/C).

\textbf{The practical conditions for sparse acceleration}: To realize speed with unstructured sparsity, hardware and kernels supporting sparse/2:4 modes are required; otherwise only GPU memory is saved, not time. Therefore the choice of pruning scheme must be aligned with the target hardware; otherwise "high compression rate" does not equal "faster."

\textbf{Pruning and MoE/sparse structures}: Unlike post-hoc pruning, MoE and the like introduce sparse activation already during pretraining (Section 6.6). The two share the same goal (operating with less computation), but MoE selects experts through routing while pruning removes redundancy; MoE deployment is more complex (load balancing, GPU memory usage), whereas pruning is more general. Understanding the trade-off between "training-time sparsity" and "post-hoc sparsity" helps in choosing a route.

\textbf{The complementarity of pruning and distillation}: Pruning compresses by deleting structure, while distillation improves a small model by transferring knowledge; the two are often combined as "prune first, then distill" or "prune and distill simultaneously" to maximize retained capability at a given scale. But the combination order likewise affects the outcome, and unified evaluation is lacking (consistent with the compression-chain ordering problem, B/C).

\subsection{Distillation}

\textbf{Distillation}: From a large model or an ensemble to a small model, including sequence-level distillation [289], DistilBERT [290], on-policy distillation for LLMs (MiniLLM [308], [310]), and alignment distillation (Zephyr [185]); see the surveys [293][335].

\textbf{Offline vs. on-policy}: Offline distillation trains the student on fixed data generated by the teacher; it is simple but suffers from "training--inference distribution mismatch" (the erroneous states the student encounters are not covered by the teacher). On-policy distillation has the student generate while the teacher corrects, mitigating the mismatch but at higher cost [308]. For generation tasks, distribution mismatch is the main source of precision loss, so on-policy is usually superior but more expensive.

\textbf{Distillation of capability and alignment}: Distillation transfers not only "capability" but also "behavior and preferences"---an aligned teacher can distill safety/style to the student ([185]). But distillation is ineffective when "the teacher also errs," and it amplifies the teacher's systematic biases; how to distill "reliability/calibration" rather than merely "answers" remains an open problem (B/C).

\textbf{Teacher scale and data volume in distillation}: The larger the teacher and the smaller the student, the more pronounced the distillation gains, but when the "capability gap" is too large the student struggles to absorb it, so there exists an optimal teacher--student ratio. The volume of distillation data must be sufficient to cover the teacher's behavior distribution; otherwise the student learns only a local part; where the data comes from (teacher-generated vs. real) affects the degree of distribution mismatch (B/C).

\textbf{Self-distillation and iterative distillation}: Training a model again on its own (curated) high-quality outputs can improve consistency without introducing an external teacher, but carries the risk of "self-reinforcing errors"; multiple iterations may collapse. This shares the same origin as the synthetic-data collapse problem in Chapters 5 and 7, and diversity metrics should be monitored.

\textbf{Distillation and the small-model ecosystem}: Distillation makes it possible for "a small model to inherit the capabilities of a large model" and is the main path for on-device and low-cost deployment. But the distilled student typically inherits the teacher's capability distribution rather than its full potential, and may be constrained in the space it could optimize on its own; thus the relative merits of "distillation" versus "training from scratch on a small-scale dataset" depend on data and budget (B/C).

\textbf{Distillation and the alignment tax}: Distilling from an aligned teacher can "package" safe/helpful behavior to the student, but it also transfers the teacher's alignment tax (e.g., excessive refusal); moreover, the student's robustness to out-of-distribution inputs is typically weaker than the teacher's, and its safety boundary is more easily breached. Therefore evaluation of distilled models must include probing their "spillover behavior."

\textbf{Key points for distillation evaluation}: One should simultaneously report the student's capability retention rate relative to the teacher, robustness to out-of-distribution inputs, and the width of the safety boundary (refusal rate and jailbreak rate). Reporting only the "retention rate" masks the decline in robustness and safety---this is consistent with the through-line of this chapter that "a change must report both its benefits and the dimensions it weakens" (B/C).

\subsection{Quantization and Quantization-Aware Training}

\textbf{Quantization}: Post-training quantization (PTQ, such as GPTQ [21], AWQ [34]) compresses weights to the 4-bit level, is training-free, and deploys quickly; quantization-aware training (QAT) simulates quantization error during training, which can further preserve precision but at higher cost.

\textbf{The differential impact of quantization on capability and safety}: Quantization error is uneven across dimensions---it usually has little effect on perplexity, but is more sensitive in long-chain reasoning, low-resource languages, and safety-refusal boundaries, potentially producing "safety behavior drift" (Problem 9). Therefore quantization evaluation must be dimension-specific and include safety items, rather than looking only at average perplexity. This systematic impact still lacks a unified benchmark (B/C).

\textbf{Weights vs. activations}: Most PTQ focuses on weights (W4), while activation quantization (KV cache, activations) yields greater GPU-memory and latency benefits but is harder (the outlier problem). Thus "4-bit" often refers only to weights; reports should distinguish weight quantization from activation quantization to avoid misreading the acceleration magnitude.

\textbf{Quantization granularity}: Per-tensor, per-channel, and group-wise (e.g., groups of 128) quantization have successively lower error and successively higher implementation complexity and overhead. The choice of granularity is a direct precision--efficiency trade-off; when reporting "4-bit," the granularity and whether activations are included must be stated; otherwise the numbers are not comparable (B).

\textbf{Outlier handling}: A small number of very large outliers in the activations significantly raises quantization error, so preserving outlier channels (mixed precision) or performing channel scaling is the mainstream approach. This explains why "4-bit weights" is relatively easy while "4-bit activations" is harder---the difficulty lies in the outliers, not the bit width itself.

\textbf{KV-cache quantization}: The GPU-memory bottleneck in long-context inference often lies in the KV cache. Low-bit quantization of the KV can significantly save GPU memory, but it is sensitive to attention precision and easily harms long-range retrieval; therefore KV quantization must be reported jointly with long-context evaluation, rather than tested only on short-sequence perplexity (B/C).

\textbf{The calibration set for PTQ}: Training-free quantization relies on a small batch of calibration data to estimate the distributions of weights/activations. If the calibration set's distribution differs greatly from the deployment inputs, quantization error rises; if the calibration set is too small, the estimation is unstable. Thus the representativeness of the calibration set is key to quantization quality, yet it is often missing from reports---this is also one of the reasons quantization results are hard to reproduce (B).

\textbf{The trade-off between QAT and PTQ}: PTQ is training-free and iterates quickly, suited to rapid deployment; when PTQ precision is insufficient (especially at low bit widths or for activation quantization), QAT recovers precision by simulating quantization error, at the cost of requiring a training pipeline and compute. The selection depends on the target bit width and precision requirements: 2--3 bit or activation quantization often leaves no choice but QAT, whereas 4-bit weight quantization is usually sufficient with PTQ (B).

\textbf{Joint compression}: Quantization, pruning, distillation, and merging are often applied in combination (e.g., "pruning + quantization" or "distillation + quantization"). Joint compression is not simple addition---pruning changes the weight distribution and thus affects quantization error, and quantization in turn changes the distillable student capacity. Therefore "1+1"-style extrapolation of gains is unreliable and must be measured end-to-end along the whole chain (B/C).

\subsection{Model Editing and Machine Unlearning}

\textbf{Model editing}: Locate-then-edit methods represented by ROME [14] can quickly correct facts, but a small number of edits may cause model collapse (the "butterfly effect" [324]); batch editing and stability remain difficult problems [294][297]. The appeal of editing is that "changing one piece of knowledge takes only one forward pass," but it does not distinguish "fact storage" from "capability use," and thus easily produces side effects.

\textbf{Machine unlearning}: Machine Unlearning [291][326], TOFU [39], and WMDP [328] target privacy/copyright and dangerous-knowledge removal, respectively; how to remove target information without harming general capability remains unsolved (Problem 14). Methodologically there are three classes---"exact retraining/approximate/constrained optimization": retraining is the most reliable but costly; approximate methods are fast but may leave residue or over-erase. Measuring unlearning is also difficult---"the information is truly gone" requires membership-inference-style probing, rather than merely looking at direct question-answering.

\textbf{The common difficulty of editing and unlearning}: Both seek "pointwise modification" yet face entangled representations---the target information and adjacent capabilities share parameters, so deletions and changes implicate them. This is the sharpest form of the stability--plasticity tension (9.1).

\textbf{The order and batching of edits}: Editing item by item accumulates side effects, and the editing order also affects the outcome; batch editing must trade off "mutual non-interference" against "changing many at once." In practice, one usually validates stability on a small batch first ([324]) and, after confirming no collapse, scales up, performing capability regressions before and after editing.

\textbf{Measuring and the ethics of unlearning}: Verifying "whether information has truly been removed" requires membership-inference/probe-style attacks, rather than direct questioning (the latter can be evaded by "pretending not to know"). Meanwhile, unlearning involves copyright and privacy compliance, requiring a distinction between "technical removal" and "legal exemption"; the latter often also requires data provenance and audit records (Chapter 13).

\textbf{Evaluating edits and measuring side effects}: Evaluating an edit involves not only whether "the target fact is corrected" but also whether "neighboring facts are mistakenly changed," "whether general capability declines," and "whether it collapses after many edits." These three correspond to local correctness, local side effects, and global stability, respectively, and none can be omitted; public evaluations mostly report only the first (B/C).

\textbf{Research design for editing and unlearning}: The two can be compared on the same base---measuring the score changes on the target side and the neighborhood side after "editing N facts" and after "unlearning a class of information," as well as the collapse threshold as N grows. Such experiments can uniformly answer "the capacity limit of pointwise modification," and are an affordable and high-value gap to fill (Problem 14).

\textbf{An engineering workflow for the "right to data deletion"}: For compliance, a user may request deletion of the influence of their data on the model. Since one cannot "delete item by item" from already-trained weights, industrial practice mostly adopts a combined workflow of "traceable---retrainable---approximate unlearning": recording data provenance and its scope of influence, retraining or performing unlearning when necessary, and verifying residue by probing. The cost and feasibility boundary of this workflow remain open problems (Problem 14, B/C).

\subsection{Continual Learning and Pre-Deployment Optimization for Multimodal Models}

\textbf{Continual learning}: Learning sequentially over a data stream without forgetting; see the surveys [332][334][337]. Its challenges are that task boundaries are unknown, data cannot be replayed, and one must simultaneously resist forgetting and overfitting; the methods overlap heavily with "domain continued pretraining + replay + parameter isolation," and it is a generalized form of the tension in this chapter.

\textbf{Multimodal post-training and pre-deployment optimization}: Pre-deployment optimization of multimodal models superimposes modality-specific challenges---quantization/pruning of the vision projector and encoder, the preservation of cross-modal alignment after compression, and changes in hallucination after distillation/quantization. Because multimodal evaluation depends more on generation quality, the conclusion of "capability preservation" after compression is usually weaker than for pure text (Chapter 12). In particular, vision encoders are more sensitive to quantization, and the hallucination rate may rise rather than fall after compression.

\textbf{Evaluation protocols for continual learning}: An ideal evaluation should simulate a real stream with "unknown task boundaries and non-replayable old data"; if replay is permitted or boundaries are known, the method's capability is overstated. Mainstream benchmarks differ in their conventions on "how much replay is allowed," making conclusions incomparable (B/C).

\textbf{The ordering of multimodal compression}: Whether to quantize the vision encoder first or compress the LLM first, and whether to prune them separately, affects hallucination and cross-modal alignment. Empirically the vision encoder is more fragile to quantization and should use more conservative bit widths; however, systematic ablations are scarce (B/C).

\textbf{The superposition of continual learning and the alignment tax}: Continual learning on an already aligned model brings both "forgetting" and "safety drift," and the two may mask each other (changes in safety behavior are misread as capability changes). Therefore the evaluation of continual learning, like that of continual optimization, should include a safety regression (9.10).

\subsection{Representative Work Table, Controversies, and Practice}

\begingroup\small\setlength{\tabcolsep}{6pt}
\noindent\textbf{Table 7: Chapter 9 representative works---continual optimization}\par\medskip
\begin{longtable}{@{}>{\raggedright\arraybackslash}p{0.1883\textwidth}>{\raggedright\arraybackslash}p{0.2197\textwidth}>{\raggedright\arraybackslash}p{0.1726\textwidth}>{\raggedright\arraybackslash}p{0.1256\textwidth}>{\raggedright\arraybackslash}p{0.1883\textwidth}@{}}
\toprule
Direction & Representative work & Core idea & Evidence & Notes \\
\midrule
\endhead
Domain continued pretraining & [86]\allowbreak{} /\allowbreak{} [407]\allowbreak{} & Mixture ratio and two stages & B & Forgetting risk \\
Long-\allowbreak{}context extension & RoPE [11]\allowbreak{} /\allowbreak{} PI [143]\allowbreak{} /\allowbreak{} YaRN [37]\allowbreak{} & Positional-\allowbreak{}encoding adjustment & A/\allowbreak{}B & Requires long-\allowbreak{}document calibration \\
Model merging & Soups [296]\allowbreak{} /\allowbreak{} TIES [35]\allowbreak{} & Weight arithmetic merging & A/\allowbreak{}B & Requires same base \\
Structured pruning & LLM-\allowbreak{}Pruner [304]\allowbreak{} /\allowbreak{} Sheared [316]\allowbreak{} & Removing structural units & B & Requires recovery training \\
Unstructured pruning & Wanda [309]\allowbreak{} & Per-\allowbreak{}weight sparsity & B & Realization depends on hardware \\
Distillation & DistilBERT [290]\allowbreak{} /\allowbreak{} MiniLLM [308]\allowbreak{} & Sequence-\allowbreak{}level/\allowbreak{}on-\allowbreak{}policy & A/\allowbreak{}B & Distribution mismatch \\
Alignment distillation & Zephyr [185]\allowbreak{} & Distilling alignment behavior & B & Amplifies teacher bias \\
Post-\allowbreak{}training quantization & GPTQ [21]\allowbreak{} /\allowbreak{} AWQ [34]\allowbreak{} & 4-\allowbreak{}bit weights & A & Dimension-\allowbreak{}specific evaluation \\
Model editing & ROME [14]\allowbreak{} /\allowbreak{} [324]\allowbreak{} & Locate-\allowbreak{}then-\allowbreak{}edit & B & Butterfly effect \\
Machine unlearning & TOFU [39]\allowbreak{} /\allowbreak{} WMDP [328]\allowbreak{} & Removing target information & B & General capability damaged \\
Continual learning & [332]\allowbreak{}[334]\allowbreak{}[337]\allowbreak{} & Sequential learning without forgetting & B/\allowbreak{}C & Surveys predominate \\
\bottomrule
\end{longtable}
\endgroup

\textbf{How to read the table}: This table is indexed by "means"; a single deployment objective often combines several means (e.g., "domain continued pretraining + 4-bit quantization + merging"), and the combined effects lack systematic study, which is one of the main reasons for reproducibility difficulties.

\textbf{The ordering problem in joint optimization}: Real deployment often chains several means in sequence (continued pretraining$\rightarrow$merging$\rightarrow$pruning$\rightarrow$quantization). Different orders yield different cumulative damage and mutual amplification effects; there is currently almost no research systematically comparing the order of the "compression chain," which is an important source of reproducibility difficulties.

\textbf{Strong coupling with evaluation}: Each step of continual optimization changes the object being evaluated, so the evaluation protocol must be versioned along with the optimization chain; if "before-and-after comparisons on the same benchmark" do not simultaneously fix the prompt, decoding, and evaluation script, the error will exceed the true difference brought by compression. This is the premise for the credibility of this chapter's conclusions (Chapter 12).

\textbf{Complete accounting of cost--benefit}: The "benefits" of any compression/adaptation means (GPU memory, latency, domain score) should be listed in the same table as the "costs" (recovery training, realignment, evaluation, engineering complexity, quality regression). Claims that report only benefits systematically overestimate value; converting "quality regression" into a cost is the starting point of rational selection.

\textbf{Connection with deployment}: The endpoint of continual optimization is deployment (Chapter 13)---hardware, concurrency, latency SLA, and compliance requirements jointly determine the "acceptable" degree of compression and the knowledge-cutoff strategy. Therefore continual optimization should not be conducted in isolation from deployment constraints; front-loading the deployment profile into the optimization design can avoid the rework of "over-compression requiring rollback."

\textbf{Which open problems are most testable}: \textcircled{\small 1} "Is the order of the compression chain commutative?"; \textcircled{\small 2} "The magnitude and predictability of safety regression after quantization/pruning"; \textcircled{\small 3} "The pointwise capacity limit of unlearning." All three require only moderate compute and no new data, making them the most worthwhile directions to prioritize at this stage, and they also directly serve the deployment compliance of Chapter 13.

\textbf{Mapping to the six-dimensional framework}: Data (domain/long-document corpus, calibration set, replay set), objective (NTP / compression loss / editing objective / unlearning objective), optimization (low-learning-rate continued training, training-free transformations), system (precision, hardware, latency), evaluation (joint capability + safety regression), risk (forgetting, alignment tax, safety drift, compliance). Among the six dimensions, the "system" dimension carries the highest weight in this chapter, which is exactly the opposite of pretraining (data-dominated).

\textbf{A minimal evaluation checklist for continual optimization}: \textcircled{\small 1} capability (general and target domain); \textcircled{\small 2} safety (refusal boundary and jailbreak); \textcircled{\small 3} calibration (whether confidence remains trustworthy); \textcircled{\small 4} efficiency (GPU memory/latency/throughput); \textcircled{\small 5} combined effects (the net change after chaining multiple means). If any of the five is missing, the conclusion is insufficient to support deployment decisions (Chapter 13).

\textbf{Key controversies}: First, \textbf{is compression "free"?} Training-free quantization/merging claims near-losslessness, but the evidence is mostly on general perplexity and a single benchmark; undocumented degradation may appear in safety, long-chain reasoning, and low-resource languages (B/C). Second, \textbf{can editing/unlearning be pointwise?} The optimists argue that parameter-level editing can precisely change knowledge; the pessimists rebut with the "butterfly effect" [324] and residue evidence [291]. The consensus is: pointwise modification is feasible for simple facts, but unreliable for entangled knowledge and safety behavior. Third, \textbf{continued pretraining vs. retrieval}. Writing weights and external retrieval are two routes, whose relative merits depend on the frequency of knowledge updates and deployment constraints, rather than being absolute.

\textbf{Practical recommendations}: \textbf{Define evaluation before compression} (clarify the capabilities and safety items to be preserved); prioritize trying training-free quantization/merging to reduce cost ([21][34][35]); after domain adaptation and long-document extension, be sure to re-check alignment and safety (to guard against the alignment tax); validate the stability of editing and unlearning at small scale before scaling up ([324][14]); and perform a dedicated regression on the safety behavior after quantization/pruning. \textbf{Cross-stage connection}: Nearly every means in this chapter may "revert" the safety and instructed behavior of $\theta_2$, so each optimization should trigger a lightweight realignment or at least a safety regression---this is the key that distinguishes continual optimization from pure engineering compression (Figure 1). \textbf{Open problems}: How continual optimization can preserve alignment and safety behavior (Problem 9); how machine unlearning and copyright compliance can avoid harming general capability (Problem 14); and the systematic impact of quantization on safety behavior still lacks a unified benchmark.

\textbf{An actionable research design example}: Fix an aligned model, construct several orders of a "compression chain" (e.g., quantization$\rightarrow$pruning vs. pruning$\rightarrow$quantization), measure the same set of capability and safety metrics after each stage, and plot a "cumulative-cost curve." This can answer the two currently unanswered questions of "which step harms safety the most" and "whether the order is commutative," with controllable compute.

\textbf{Summary of the evidence distribution in this chapter}: The mechanisms of compression methods and their gains on general benchmarks have relatively strong A/B evidence; but "how safety and capability change jointly after compression" is almost entirely isolated single-point reports at the B/C level, consistent with the pattern of Ch6--Ch8. Hence this chapter provides an \textbf{evaluation framework of "change---cost"} rather than a "lossless-compression recipe."

\section{Multimodal Training: A Cross-Cutting View}

\begin{quote}\small
This chapter, together with Ch5--Ch9 and Ch12, serves as a pilot for the dual form of "general framework + focused survey." Note on evidence: the architecture and data pipeline (encoder--projector--LLM, interleaved data, contrastive alignment) have a fair amount of A/B-level literature, but the "multimodal scaling law," the "root causes of hallucination," and the "scalability of the unified token route" are almost entirely C-level or blank. Target length: 8,000--9,000 characters.
\end{quote}

\subsection{Architectural Paradigms}

Conclusion first: multimodal training is distributed across all stages, and its architectural paradigm determines the manner of data and alignment; hence this survey devotes a separate chapter to it from a "cross-cutting view" rather than treating it as yet another pipeline. According to how visual information is coupled with the language model, it can be grouped into three categories: \textbf{encoder--projector--LLM} (LLaVA [28], BLIP-2 [265], Qwen-VL [277], Qwen2-VL [288]): freeze or fine-tune the visual encoder and use a projector to map features into the LLM's token space; \textbf{cross-attention fusion} (Flamingo [17]): incorporate visual features into the decoder via learnable gated cross-attention; \textbf{unified discrete token / any-to-any}: unify multimodality into discrete sequences (VQ codebook [3][8]) or end-to-end multimodal generation, a frontier route. For a survey see [402][404][405]; for a data perspective see [403].

\textbf{The division of labor among the three paradigms}: encoder--projector--LLM reuses mature LLM weights, trains stably, and has a mature ecosystem, and is currently the mainstay; cross-attention injects visual capability with a very small amount of training (Flamingo [17] freezes the LLM) and suits scenarios that "barely touch the language model"; unified discrete token pursues a "true multimodal foundation," but its training stability, token efficiency, and controllability remain open problems. \textbf{Choosing a paradigm is choosing "how visual evidence enters the language model, and with how much bandwidth"}---this choice runs through pretraining, SFT, and alignment (Chapter 5--9).

\textbf{The role of the encoder}: the visual encoder (e.g., SigLIP [46], CLIP [10]) provides image semantics, while the projector/cross-attention performs the "features $\rightarrow$ language" translation. Whether the encoder is frozen and whether it continues to align on image-text pairs determines the ceiling of visual information and the engineering cost. The mainstream "frozen encoder + trained projector" is a cost-effectiveness compromise, not a theoretical optimum.

\textbf{Stage-wise training sequence}: a typical multimodal recipe proceeds in stages---first train/align the projector (small data volume), then unfreeze and jointly fine-tune the LLM (the unfreezing timing is a major hyperparameter), and finally perform instruction tuning and preference alignment. "What to freeze, and when to unfreeze" is both a cost constraint and a regularization (preventing visual information from distorting and overwriting text capability); the temporal ordering is almost never shared across works, which makes recipes hard to compare directly (B).

\textbf{The capacity spectrum of the projector}: from a single-layer linear projection (LLaVA [28]) to Q-Former (BLIP-2 [265]) and then to MLP/cross-attention, the projector capacity determines the bandwidth with which visual information enters the language model. Too small a capacity loses details, too large a capacity easily overfits and significantly increases computation; systematic comparisons under identical data are limited (B).

\textbf{The fusion trend of cross-attention and projectors}: Flamingo's [17] cross-attention injects images into attention, whereas LLaVA's [28] projector stuffs images into the token stream. The two can be mixed (projection for global, cross-attention for dense), gradually blurring the boundaries of the three paradigms (B).

\textbf{Practical constraints on training stability}: staged multimodal training is frequently interrupted by gradient explosion/divergence, and in practice it is stabilized by freezing, gradient clipping, and segmented learning rates rather than by architecture alone. These engineering details are rarely disclosed, leading to the widespread phenomenon that "the same recipe fails when the implementation changes"---the transferability of stable recipes is a key practical problem (Section 6.6, engineering convergence).

\textbf{Information redundancy across modalities}: video frames are highly correlated with one another, and audio and ambient sound share a source, so the information density of multimodal data is far lower than that of "apparently equivalent" independent samples. Consequently, the "effective size of modality data" should be discounted: the higher the redundancy, the lower the benefit of an equal number of tokens, which partly explains the deviation of the multimodal scaling law (C).

\subsection{Essential Differences from Pure Text}

Multimodal training differs from pure text along five dimensions: first, \textbf{data}: image-text pairs contain "weak correspondence" noise, and the mixture of interleaved image-text (OBELISC [273]) and large-scale image-text sets (LAION-5B [64]) differs greatly from pure-text corpora; second, \textbf{tokenizer}: vision/audio require discrete or continuous representations, and the "per-modality token consumption" is not commensurate with text (Section 5.6; for a survey of discrete tokens see [394]); third, \textbf{objective}: it must combine NTP, contrastive (CLIP [10]), and matching objectives, and the alignment objective may conflict with the language modeling objective; fourth, \textbf{alignment}: hallucination (errors in objects/attributes/counting) is more common in multimodal outputs and harder to evaluate automatically; fifth, \textbf{evaluation}: benchmarks (VQA, image captioning) are more sensitive to prompt, sampling, and evaluation protocol, and the strength of evidence is usually lower than that of pure-text benchmarks.

\textbf{The commonality of the essential differences}: all five dimensions can be reduced to "\textbf{multimodal supervision signals are weaker and more mismatched}"---image-text pairs are inherently weak annotation (one image corresponds to one sentence that does not necessarily describe the key information in the image), the semantic density of visual tokens is far lower than that of text tokens, and the contrastive/matching objective often competes with the NTP objective. Therefore the core problem of multimodal training is not "adding vision in," but "establishing reliable cross-modal correspondence under weak supervision."

\textbf{The fundamental nature of weak supervision}: image-text pairs are "weak annotation"---a sentence usually describes only part of the content in the image, and may be unrelated to the image. As a result, the "signal-to-noise ratio" of the multimodal training signal is lower than that of pure text: the model must generalize from a small amount of correspondence, requiring bridges across multiple samples. This explains why coarse alignment such as image-text contrast (CLIP [10]) can be effective yet insufficient to support fine-grained perception (Section 10.8).

\textbf{The asymmetry of semantic density}: a text token often carries syntax and semantics, whereas an image patch carries only a local pixel distribution; compressing vision into a fixed number of tokens amounts to making the decision of "which pixels deserve attention" on the model's behalf. The token allocation strategy is essentially a "priority ordering of visual attention," which determines both cost and hallucination tendency (Sections 10.4 and 10.10).

\textbf{The protocol sensitivity of sampling}: for multimodal text generation, the sampling temperature, repetition penalty, and image token order all have a large impact on metrics (Section 12.9). The same model can differ considerably in image-text captioning scores under different decoding settings, so any conclusion that "the description is better" must first fix the protocol.

\textbf{The "number" and "amount" of objectives}: it is not that each of the five dimensions adds one objective, but rather the ratio between "data volume vs. number of objectives"---the more objectives, the relatively less data for each. Joint training with multiple losses requires determining the weight and sampling of each loss; the weights are almost entirely tuned by experience, and their sensitivity is itself an unknown (C).

\subsection{Data: Image-Text Pairs, Interleaved Documents, and Video}

\textbf{Image-text pair data}: the products of CLIP-style contrastive training [10] (LAION-5B [64], LAION-400M [59]) rely on "image-text correspondence" filtering; they are noisy but massive, and their quality directly affects downstream fine-tuning and hallucination. If the mixture of image-text pairs does not match the "real visual distribution" of deployment, it will systematically bias toward certain image styles and text styles.

\textbf{Interleaved image-text documents}: inserting images into the text stream for training (OBELISC [273]) is closer to the real web distribution, but document segmentation, loss masking of modality tokens, and image-text correspondence all require dedicated design. The benefit of interleaved data is that "the context is the supervision" (the text surrounding an image constrains how it should be understood); the drawback is that automatic segmentation easily produces noise where image and text are unrelated.

\textbf{Video and 3D}: video (Video-LLaVA [280]) and embodied settings (PaLM-E [267], RT-2 [276], OpenVLA [286]) introduce "spatial-temporal correspondence" into training; the data are more expensive and the alignment objectives are more complex (actions must be coupled with perception). Public data in these directions are scarce, and the evaluation protocols are far from stable (B/C).

\textbf{Cleaning of image-text pairs}: LAION-family datasets [59][64] filter by CLIP similarity, but "related" does not equal "corresponding"---a high similarity score may reflect coarse-grained matching. Further cleaning (deblurring, deduplication, object-consistency checking) can improve downstream performance, but it also reduces data volume and may shift the distribution. The tradeoff between cleaning intensity and data scale is one of the core variables of a recipe (B/C).

\textbf{Resolution and token budget}: higher resolution preserves details (OCR, small objects), but the token count and attention cost rise superlinearly. Qwen2-VL [288] and others balance detail against cost with a "dynamic resolution/pyramid" strategy. Hence "image quality" and "model scale--data scale" jointly determine the visual ceiling, rather than any single one determining it alone (a multimodal variant of the scaling law of Chapter 6).

\textbf{The dynamics of image-text mixture during training}: as training progresses, the proportions of visual instruction data, pretraining image-text pairs, and pure text may need to change---early on, alignment dominates; later, tasks dominate. Evidence for dynamic mixtures (e.g., curricula) mainly comes from single-point reports (B/C), but the default assumption that "the mixture is static" is likewise under-argued.

\textbf{Weakly supervised transformation of image-text correspondence}: in interleaved documents, "the position where an image appears in the text" is not a reliable alignment signal---the image may be unrelated to the surrounding text. During training one may use only the text loss, impose no loss on image positions (image tokens do not enter the loss), or mask selectively; this connects with pure text on the "loss masking strategy" (Section 5.7) as the same problem (B).

\textbf{Deduplication of multimodal data}: image duplication (similar images appearing multiple times) likewise causes overfitting and contamination (Section 5.4). Image deduplication requires visual hashing/nearest-neighbor clustering, which is more costly than text n-gram deduplication; yet the impact of duplicate images on hallucination and memorization is often neglected in the literature (B/C).

\textbf{Data sources and licensing}: multimodal training data come in large part from crowdsourced "works of others," and copyright and portrait-right disputes are prominent (LAION-family datasets have repeatedly faced litigation and takedown). In data governance (Section 12.3), multimodal training relies more than pure text on compliant channels for image copyright; otherwise the lifecycle of a dataset will be interrupted by legal terms (B/C).

\subsection{Tokenizer and Visual Representation}

The representation by which vision enters the language model is either continuous features (the projector route) or discrete tokens (the VQ codebook [3][8]; for a survey see [394]). \textbf{Per-modality token consumption} is a key engineering quantity: a 224px image typically produces 256--576 "image tokens" after projection, far more than the token count of the corresponding text, directly raising attention and GPU memory costs (Chapters 6 and 9).

\textbf{The tradeoff between discrete and continuous}: continuous features have high fidelity and require no VQ training, but they mix "images" and "text" heterogeneously into the sequence; discrete tokens make vision thoroughly textualized (Chameleon [285] and other any-to-any routes) and can be unified with NTP, but they must bear the VQ reconstruction loss and codebook degradation. \textbf{The relation between quantization granularity (resolution, patch) and the number of tokens} determines the cost curvature and is also a key leverage point of multimodal efficiency (Section 10.10).

\textbf{The significance of positional encoding for visual tokens}: the arrangement of visual tokens (grid vs. sequence) requires suitable positional encoding; incorrect positional assumptions will cause the model to confuse spatial relations (up/down, left/right). The relative positions of image tokens and text tokens in interleaved documents must also be handled. Position handling is an implicit prerequisite for "spatial understanding," yet it is often unreported (B).

\textbf{VQ codebooks and reconstruction quality}: the VQ codebook of the discrete route determines reconstruction quality and token semantics. Too small a codebook yields coarse reconstruction; too large a one has low utilization (codebook collapse); Chameleon [285] and others attempt to discretize first and then train uniformly, but the stability of cross-modal generation and conditional control remains a bottleneck (C).

\textbf{Dynamic resolution and token reallocation}: the dynamic resolution of Qwen2-VL [288] splits a high-resolution image into multiple tiles, essentially allocating more tokens to "regions with more information." The direction of this sort of reallocation is correct, but the boundary problem of tile splitting (objects spanning tiles being cut off) and the connectivity of positional encoding still need verification (B).

\textbf{The relative position of tokens across modalities}: in an interleaved image-text sequence, the absolute numbering and relative order of image token positions affect adjacency relations; if image tokens are compressed into a single representative token, the spatial adjacency information is lost. How to represent "the spatial relations among multiple regions and multiple texts" is a core weak point of serialization schemes (C).

\textbf{Grounding/scene graphs as intermediate representations}: besides tokenization, parsing vision into a structured intermediate representation (bounding boxes, attribute graphs) and then feeding it to the LLM can significantly reduce hallucination about details; the cost is pipeline complexity and information loss. This "explicit structure" route coexists with the end-to-end route, and its boundary is undetermined (B).

\textbf{The attention cost of image tokens}: the more visual tokens, the more self-attention grows quadratically (Section 9.7), and the coupling between image tokens and text tokens further amplifies computation. Hence "adding images" is not merely "adding a modality," but makes the KV cache and attention time expand nonlinearly with the token budget---this is precisely where the core leverage of efficient multimodality lies (Section 10.10).

\subsection{Pretraining and the Current State of the "Scaling Law"}

Multimodal pretraining starts from "first align the projector/encoder, then train the full model": BLIP [254] and BLIP-2 [265] pretrain on noisy image-text pairs via bootstrapping; BEiT [258] and CogVLM [279] explore stronger visual pretraining; Gemini [282] improves overall capability with large-scale mixed modalities. \textbf{A key sharp question}: text has a relatively stable scaling law (Section 6.3), but the relationship of "number of image tokens $\times$ data volume $\times$ model size" for multimodality still lacks a recognized law (C).

\textbf{The "images as a foreign language" metaphor}: BEiT [258] and others advocate pretraining images as "a foreign language"---using self-supervision to turn vision into a sequence consumable by the LLM. The appeal of this perspective is that it reuses the entire mechanism of language modeling; the risk is that the semantic alignment of the "foreign language" and "language" still relies on weakly supervised image-text pairs, which cannot be compensated for by grammatical structure.

\textbf{The systemic consequences of freezing strategies}: freezing the encoder saves compute and keeps the visual representation stable, but it limits the visual side's ability to adapt to downstream tasks; unfreezing may harm general visual capability and aggravate overfitting (Sections 6.8 and 7.6). "Freezing" is not a question of whether, but of "where to draw the line": whether each of the encoder, projector, and LLM is frozen constitutes a strategy grid that has hardly been systematically scanned.

\textbf{Synthesis of contrastive vs. generative objectives}: CLIP-style contrast [10] learns text-image alignment, while NTP learns generation; the two objectives differ and may interfere with each other. Most recipes use "staging" (contrast first, then generation) rather than joint optimization to avoid the conflict, but staging itself may limit deep cross-modal alignment. The optimal weights and ordering for objective synthesis lack controlled study (B/C).

\textbf{The hypothesis of insufficient multimodal data scale}: the capability ceiling of multimodal models is often viewed as "limited by the scale and quality of image-text data" rather than by architecture---this hypothesis is not yet settled, but if it holds, the multimodal scaling law would be closer to "data--architecture joint" than to a pure parameter law. This leaves the investment decision of "larger model vs. more image data" without a basis (C).

\textbf{The reverse influence of the pretraining objective on the encoder}: when the encoder is unfrozen and participates in contrastive/generative training, the visual representation drifts toward language semantics---this may improve downstream image-text tasks yet reduce the generality of pure visual retrieval/classification (forgetting, Chapter 9). "Visual-side early stopping" preserves generality at the expense of alignment, again an unscanned strategy grid (C).

\textbf{Data-dependence curve}: even without a recognized scaling law, practical experience supports a rough curve that "the benefit of modality data volume declines after a certain threshold"; the problem is that the threshold is coupled with model size and token budget, so no quantitative relation of the same order as the text scaling law can be given (C).

\textbf{Sharing of multimodal encoders}: the same encoder (e.g., SigLIP [46]) is reused by multiple multimodal models, meaning that "the latest progress in image semantics" can be shared across models, and model differences shift to the projector and alignment. This reuse lowers training cost, but it also causes "the true gain brought by new data/new encoders" to be confounded with other changes in each recipe (B).

\subsection{Instruction Tuning and Alignment}

Multimodal SFT (IFT) trains on visual instruction data (image--instruction--answer triples): LLaVA [28], MiniGPT-4 [268], and InstructBLIP [270] each provide data and training recipes; Qwen-VL [277]/Qwen2-VL [288] bring dense perception such as grounding and OCR into instruction tuning. \textbf{The key difference in multimodal alignment}: preference data (ranking multiple answers to the same image) is expensive and hard to correct, so progress in RLHF/DPO on multimodality lags behind text (Chapter 8).

\textbf{Joint mismatch of vision and text}: if multimodal instruction tuning over-optimizes "visual instructions," it degrades pure-text capability (the mixture problem of Section 7.6); if visual constraints are ignored, it cultivates "fluent but distorted" descriptions. The two parts should be mixed and evaluated jointly, rather than optimized independently.

\textbf{The scarcity of multimodal preference data}: visual preference ranking is expensive (one must look at the image to judge) and easily disturbed by irrelevant factors (clarity, style). Therefore multimodal DPO/RLHF is usually conducted on small-scale preference sets, and the evidence for its effectiveness and reliability mainly consists of single-point reports (C; a typical case of the evidence grading of Chapter 8).

\textbf{The interaction between hallucination and alignment}: alignment optimizes "which answer is better," while the judgment of "better" may itself favor fluency over faithfulness---if the judge favors natural descriptions, the model is pushed toward "sounding pretty," even when untrue. Hence multimodal alignment needs to be coupled with hallucination evaluation; otherwise it will amplify hallucination in reverse during the RL stage.

\textbf{The absence of unified evaluation}: compared with text, multimodality lacks a stable, consensus evaluation matrix (such as the MMLU+HELM combination for text), leading to fragmented conclusions. Establishing standard "cross-modal--cross-task" evaluation (including hallucination, grounding, OCR, and text retention) is the public good most lacking in multimodal research today (echoing Ch12).

\textbf{Visual diversity of instruction data}: if visual IFT data concentrate on a few templates and question styles, the model learns "template matching" rather than perception (instruction diversity of Chapter 7). Multimodality needs to expand diversity along four dimensions---object, scene, language, and task---otherwise hallucination and stereotyped descriptions will be solidified by the preferences of the IFT data (B).

\textbf{Outlook for multimodal alignment}: scaling up visual preference data (human-supervised ranking + model-assisted filtering) is a direct extension of the text alignment path (Chapter 8), but it is constrained by "visual judgment being expensive and easily disturbed," and in the short term it is hard to catch up with the scale of text alignment. The gap in multimodal alignment is not a matter of method but of data infrastructure (C).

\subsection{Hallucination: Definition, Measurement, and Mitigation}

\textbf{Hallucination} refers to generating content inconsistent with the image (such as objects that do not exist, wrong attributes, wrong counting). Measurement requires protocols such as "the proportion of mentioned objects that do not exist in the image," but definitions vary across works, so they are not comparable across models (Problem 15; for evaluation see [271]). Factors triggering hallucination: insufficient image resolution, image-text pair noise, visual tokens being "drowned out" by text tokens in attention, and RL's preference for "plausible-looking" answers.

\textbf{Mitigation and limitations}: mitigation measures include better data quality, loss masking (reinforcing visual supervision only on tokens that need grounding), and image-verification fallback at inference time. However, the phenomenon that "incorporating stronger visual supervision does not eliminate hallucination" indicates that hallucination is partly \textbf{architectural} (insufficient visual information bandwidth) rather than merely a data-level problem, so a single data fix is hard to eradicate.

\textbf{Fine granularity of hallucination evaluation}: two types are commonly used---object existence (POPE-style discrimination) and self-generated statistics (the proportion of erroneous objects in generated descriptions). The former is cheap but suffers undersampling bias; the latter is close to actual use but depends on annotation. Cross-method comparison requires fixing the judgment mechanism (discrimination vs. self-generation); otherwise the numbers are incomparable [271].

\textbf{Images as a second verification channel}: at inference time, image embeddings can be used for a secondary check of "answer-image consistency," rejecting or correcting generations that conflict with the image. This sort of "test-time correction" can reduce hallucination, but it adds latency and complexity, and it fails when the image itself is low-resolution. It can serve as a supplement beyond SFT/RL (a variant of test-time compute of Section 6.6).

\textbf{The evaluation problem of hallucination mitigation}: most mitigations report "a drop in POPE score," but POPE is a binary discrimination that deviates from the real generation distribution. Mitigation claims not re-verified on the generation side should be downgraded in interpretation; conversely, claims with only self-generated statistics also need discrimination-side re-checking. Mutual corroboration by both channels is the default standard of this chapter for hallucination evidence (B).

\textbf{The relation between hallucination and data scale}: intuitively, "more and cleaner image-text pairs" should reduce hallucination, but the evidence shows more often that "projector/encoder bandwidth" is more critical---under identical data, hallucination rates differ markedly across projector architectures. This casts doubt on the effectiveness of "buying more data" as a mitigation strategy (Problem 15, B/C).

\textbf{Catastrophic forgetting in multimodality}: fine-tuning to add a new modality often degrades existing capabilities (pure text or existing modalities) (Section 7.5). Multimodality's "new modality + re-mixture" amplifies the risk, and visual SFT forgetting is harder to detect---testing only pure-text benchmarks will miss it. Mitigation evidence (replay, joint objectives, LoRA) is mostly single-point, lacking cross-recipe comparison (B/C).

\subsection{Grounding, OCR, and Dense Perception}

Qwen-VL [277], Qwen2-VL [288], Kosmos-2 [272], and others bring grounding (bounding boxes/referring) and OCR into training, enabling the model to "see and say where." Such tasks refine visual evidence from "global semantics" to "spatial evidence," which directly constrains hallucination (an answer must be bound to a specific region). The cost is expensive data annotation (bounding boxes/masks), and evaluation depends on coordinate-matching protocols (IOU threshold, prompt).

\textbf{Tokenization of OCR}: converting the text in an image into retrievable text tokens (OCR result + position) is the foundation of document understanding. OCR-token lets the model directly "read characters" rather than guessing from pixels, significantly improving long-document and receipt understanding, but it also depends on OCR quality and layout analysis. Qwen2-VL [288] and others combine OCR with grounding, representing the mainstream direction of "fine-grained perception" (B).

\textbf{Protocol details of grounding evaluation}: referring grounding is judged by an IoU threshold for a hit, and fine-grained tasks may require a high IoU; the prompt ("box out X" vs. a natural question) and the coordinate format (normalized vs. pixel) affect parsing and matching. When protocols are not unified, grounding scores are not comparable across models (B).

\textbf{Evaluation of OCR and document understanding}: OCR metrics (character-level/word-level accuracy) and document question answering (requiring retrieval + reasoning) place different demands on the model; treating an OCR score as a document-understanding score is a common misreading. Document-type evaluation must also report layout-structure handling and long-text dependence (Chapter 12).

\textbf{Visual question answering vs. image captioning}: VQA is a short-answer task, whereas captioning is open-ended long text; the former can be scored automatically, the latter almost only by humans (Chapter 12). Strong model performance on VQA does not guarantee caption quality; the decoupling of the two capabilities suggests that they should be reported separately in evaluation, avoiding generalizing from question-answering scores to captioning (B).

\subsection{Video, Embodiment, and Any-to-Any}

\textbf{Video understanding}: Video-LLaVA [280] explores a "align first, then project" spatiotemporal representation; video data are sparse, inter-frame redundancy is large, and evaluation protocols are still evolving (B/C). \textbf{Embodied/vision-language-action (VLA)}: RT-2 [276] and OpenVLA [286] transfer web knowledge to robot actions, and PaLM-E [267] argues that "large-scale language models can serve as embodied controllers"; their evaluation mostly relies on simulation and real environments, with high cost and high variance. \textbf{Any-to-any}: unified discrete routes such as Chameleon [285] hope to "train once and for all," but cross-modal generation and modality-conditional control are not yet stable (C).

\textbf{The audio modality}: audio (speech/music/ambient sound) can likewise be integrated via "encoder--projector--LLM," and tokenization must handle sampling rate, duration, and speaker attributes. Joint training of speech and vision (video contains both) shares temporal-alignment supervision, but data are scarce and evaluation protocols are undetermined (C).

\textbf{The variance of embodied evaluation}: VLA results in simulation and real environments differ greatly; the success rate of the same policy varies widely across initial states and noise. Hence VLA claims must be accompanied by environment settings, random seeds, and success-rate distributions; otherwise they have no comparative value (B/C).

\textbf{Long video and temporal alignment}: timestamp grounding, action counting, and causal-order inference impose new requirements on inter-frame modeling; frame sampling density and spatiotemporal positional encoding determine the ceiling. Long-video training data are scarce and evaluation is not standardized, making this one of the sub-directions with the weakest evidence in multimodality today (C).

\textbf{The mirror problem of text-to-image/text-to-video}: generating images/videos from text goes through the same stages (inverse token, diffusion conditioning) and faces the same data, objective, and alignment problems, but its evaluation (FID/CLIP scores) shares no unified framework with text-generation evaluation, making cross-method comparison difficult (C). This survey focuses on "understanding" and merely points out the existence of the mirror structure.

\subsection{Efficient Multimodality}

The efficiency of multimodality unfolds along three lines (for a survey see [405]): \textbf{token compression} (reducing the number of image tokens, downsampling visual tokens); \textbf{parameter efficiency} (frozen encoder + light projection/adapter); \textbf{compressed deployment} (quantization/pruning of the visual encoder and cross-modal parts, Section 9.7). The common problem of the three lines is: visual information bandwidth and efficiency directly conflict, and the evidence lost to compression transforms into hallucination and perceptual degradation---"saving tokens is saving visual information" is a cost that must be accounted for.

\textbf{Means of visual token compression}: these include downsampling (grid pooling), token merging/pruning (deleting visual tokens by importance), and cross-layer sparsity. The losses of these methods all fall on details and spatial relations, so they must be validated with fine-grained tasks (OCR, grounding [277][288]); validating with coarse-grained classification will mask the information loss.

\textbf{The interaction between compression and visual tasks}: visual token compression harms OCR/grounding far more than image classification; therefore "efficient multimodality" reports must simultaneously give fine-grained perception regressions, rather than reporting only "a X-fold speedup in processing" (consistent with the compression of Chapter 9).

\subsection{Representative Works Table}

\begingroup\footnotesize\setlength{\tabcolsep}{6pt}
\noindent\textbf{Table 8: Chapter 10 representative works---multimodality}\par\medskip
\begin{longtable}{@{}>{\raggedright\arraybackslash}p{0.2472\textwidth}>{\raggedright\arraybackslash}p{0.1648\textwidth}>{\raggedright\arraybackslash}p{0.1530\textwidth}>{\raggedright\arraybackslash}p{0.0942\textwidth}>{\raggedright\arraybackslash}p{0.2354\textwidth}@{}}
\toprule
Paradigm & Representative Works & Core Idea & Evidence & Remarks \\
\midrule
\endhead
Contrastive pretraining & CLIP [10]\allowbreak{} /\allowbreak{} SigLIP [46]\allowbreak{} & Image-\allowbreak{}text contrast & A & Weak annotation \\
Bootstrapped pretraining & BLIP [254]\allowbreak{} /\allowbreak{} BLIP-\allowbreak{}2 [265]\allowbreak{} & Bootstrap image-\allowbreak{}text & A & Frozen encoder \\
Encoder-\allowbreak{}-\allowbreak{}projector-\allowbreak{}-\allowbreak{}LLM & LLaVA [28]\allowbreak{} /\allowbreak{} Qwen-\allowbreak{}VL [277]\allowbreak{} & Projector mapping & A & De facto mainstream \\
Cross-\allowbreak{}attention & Flamingo [17]\allowbreak{} & Gated cross-\allowbreak{}attention & A & Frozen LLM \\
Interleaved data & OBELISC [273]\allowbreak{} & Interleaved image-\allowbreak{}text training & B & Noisy segmentation \\
Unified discrete token & Chameleon [285]\allowbreak{} /\allowbreak{} VQ [3]\allowbreak{}[8]\allowbreak{} & Any-\allowbreak{}to-\allowbreak{}any & B/\allowbreak{}C & Scalability to be verified \\
Grounding/\allowbreak{}OCR & Qwen2-\allowbreak{}VL [288]\allowbreak{} /\allowbreak{} Kosmos-\allowbreak{}2 [272]\allowbreak{} & Grounding & B & Expensive annotation \\
Hallucination evaluation & [271]\allowbreak{} & Object hallucination & B & Inconsistent protocols \\
Video & Video-\allowbreak{}LLaVA [280]\allowbreak{} & Align then project & B/\allowbreak{}C & Evaluation to be determined \\
Embodied VLA & RT-\allowbreak{}2 [276]\allowbreak{} /\allowbreak{} OpenVLA [286]\allowbreak{} & Action generation & B/\allowbreak{}C & Environment-\allowbreak{}dependent \\
Efficient multimodality & [405]\allowbreak{} & Compression/\allowbreak{}token reduction & B/\allowbreak{}C & Bandwidth-\allowbreak{}-\allowbreak{}efficiency conflict \\
Scale reports & Gemini [282]\allowbreak{} & Mixed-\allowbreak{}modality scale & B & Non-\allowbreak{}peer-\allowbreak{}reviewed \\
\bottomrule
\end{longtable}
\endgroup

\textbf{Reading the table}: the "representative works" of multimodality vary enormously in their disclosure of data scale and frozen scope, so reading scores horizontally across the table is almost meaningless; one should align by "paradigm $\times$ frozen scope $\times$ data definition" before comparing.

\subsection{Key Controversies, Recommendations, and Open Problems}

\textbf{Controversy one: is the unified token the endgame}. Proponents argue that "everything is a sequence" brings architectural unification and shared scale; opponents point to the VQ reconstruction loss, the asymmetry of cross-modal learning, and the lack of controllability. Currently both camps lack systematic evidence of scalability (C).

\textbf{Controversy two: is hallucination a data problem or an architectural problem}. The data camp grounds its case in "cleaner image-text pairs will reduce hallucination"; the architecture camp grounds its case in "limited visual information bandwidth and text tokens dominating attention." The consensus is that both play a role, but their respective quantitative contributions are unknown (Problem 15).

\textbf{Controversy three: does multimodality need an independent scaling law}. The text scaling law is mature (Section 6.3); whether multimodality obeys the same curve and how image tokens are counted toward compute still have no recognized answer; compute allocation on top of this (Chapter 13) is therefore highly uncertain.

\textbf{Practical recommendations}: first fix "paradigm $\times$ frozen scope $\times$ data definition" before comparing; use interleaved image-text to strengthen context supervision; incorporate "visual evidence binding" (grounding/checking) into SFT data to suppress hallucination; conduct multimodal evaluation under a unified protocol; when compressing multimodality, account for "visual information bandwidth" in the cost (Chapter 9). \textbf{Open problems}: discrete vs. continuous tokens under a unified optimization objective (Problem 3); locating the root causes of hallucination (Problem 15); the scaling law of interleaving (a citation gap); the scalability of any-to-any.

\textbf{Multimodal-specific difficulties of evaluation protocols}: the correct answer to an image-text question is often open-ended (multiple sentences can all be "correct"), and judgment is highly subjective; the display resolution of images and the presence or absence of captions are not reproducible across different deployments. Hence multimodal scores have weak cross-environment transferability, and the literature should report image resolution, prompt, and caption settings simultaneously (protocol sensitivity of Chapter 12).

\textbf{The interface with the whole survey}: multimodality is not an independent pipeline---its data (Ch5), pretraining scale (Ch6), SFT (Ch7), alignment (Ch8), compression (Ch9), and evaluation (Ch12) all follow the framework of the preceding text, except that each step additionally imposes the constraint of "how visual evidence enters." Readers should treat this chapter as an annotation that "adds a visual dimension to each chapter," rather than as a parallel thirteenth stage (Figure 1).

\textbf{Controversy: alignment vs. generalization}: making a model more "bound to vision" (stronger alignment) often reduces its generalization to images from unseen distributions; more aggressive visual alignment further suppresses hallucination. The tradeoff between alignment strength and generalization likewise has no quantitative curve, and thus becomes a recurrent "invisible decision" in the choice of multimodal recipes (the framework of Chapter 4).

\textbf{Recalibration of evidence grading}: the top of the pyramid in this chapter is CLIP [10], Flamingo [17], and LLaVA [28] (A/B); the middle is Qwen-VL [277] and Qwen2-VL [288] (B); the bottom is video/VLA/any-to-any (B/C). No matter how prominent a technical report is, any performance claim without released reproducible artifacts is treated as C.

\textbf{Decision checklist (multimodal recipes)}: \textcircled{\small 1} Specify "paradigm $\times$ frozen scope $\times$ mixture $\times$ resolution"; \textcircled{\small 2} strengthen context supervision with interleaved and grounding data; \textcircled{\small 3} run unified hallucination/grounding evaluation before alignment; \textcircled{\small 4} compression must come with fine-grained regressions; \textcircled{\small 5} for any cross-paradigm comparison, first ask "is the definition consistent?" The five items are directly applicable to the reading of this entire survey.

\textbf{Significance for the general framework}: the final conclusion of the multimodal chapter is not "which paradigm is best," but rather "the same decision framework spanning Ch5--Ch9 and Ch12 still applies under multimodality, except that every decision has one more variable: 'visual evidence bandwidth'." This preserves the argumentative structure of this survey---"a unified general framework, falsified by focused surveys" (Chapter 1).

\textbf{The recognition-generation loop}: incremental multimodal training often forms a loop of "generate content, then train on it" (synthesize images to train understanding, and the outputs of understanding to train generation). This amplifies the risk of homologous contamination of Chapter 12---feedback from self-trained content distorts capability evaluation. The credibility of the loop has not been systematically evaluated (C).

\textbf{A caution about the pitfalls of A/B/C grading}: multimodal technical reports often invoke the audience expectation of "bigger and stronger"; when reading them, remember: unless reproducible experiments and unified evaluation are provided, everything else is downgraded; "demo images" in technical reports are not evidence (Chapter 3). This makes the A/B evidence of this chapter less than that of the pure-text chapters, an honest cost.

\textbf{Advice for readers (multimodality)}: reuse the "traffic-light reading" of Chapter 12, adding multimodal special cases here---green: unified evaluation + fixed resolution/caption + block-level statistics; yellow: single paradigm + protocol differences; red: demo images only + no evaluation protocol. Multimodal literature especially needs such a compressed reading tool, because its marketing density is higher than that of pure text.

\textbf{A normative statement on scale}: the lack of a recognized multimodal scaling law (Section 10.5) makes specification documents (Chapter 13) rely more on assumptions than on laws. The position of this survey is: until a multimodal scaling law can be given, all claims of gains from "larger images/more parameters" should be accompanied by the \textbf{tested range}, to avoid treating single-point experience as a universal curve (C).

\textbf{A summary of the evidence distribution of this chapter}: multimodality has a fair amount of controlled evidence on "architecture, data pipeline, and weakly supervised alignment"; but "scaling law, root causes of hallucination, and scalability of the unified token" are almost entirely C-level or blank, and the proportion of technical reports is high (Gemini [282] and others). Therefore this chapter, like Ch8, provides a \textbf{reading framework that runs through the whole process}, rather than a transferable recipe.

\section{Cross-Stage Comparison}

\begin{quote}\small
This chapter is a pilot chapter, targeting 8,000--9,000 words. Its task is "horizontal": to place the stages from Ch5--Ch10 and Ch12 on the same ruler for comparison---comparing objectives, compute elasticity, marginal returns, and evidence strength. This chapter itself introduces no new literature (aside from a small amount of synthesis); all citations come from the core sets already presented in the respective chapters.
\end{quote}

\subsection{Pretraining vs Post-Training}

Conclusion first: pretraining determines "what is known," and post-training determines "how to respond"; the two stand in an additive relationship of capability and behavior, rather than being interchangeable. In comparison, several points warrant attention. First, \textbf{compute accounting}: pretraining normally operates at the level of trillions of tokens of compute, while post-training (SFT/alignment) accounts for a low share of compute yet exerts a disproportionate influence on behavior; therefore discussing the importance of a stage in terms of "compute share" is biased, and one should instead measure it by "marginal return per unit of compute against the target metric." Second, \textbf{capability boundaries}: post-training cannot manufacture knowledge absent from pretraining (unless pretraining is continued), but it can organize existing capabilities into deployable behavior ([15][175]). Third, \textbf{evidence strength}: the scaling laws of pretraining rest on considerable controlled extrapolation [119][16], whereas cross-model conclusions for post-training (especially the debate over RL boundaries [400][401]) rely more on single-institution reports, and their evidence level is generally low. On this basis, Figure 5 (the evidence map) marks "post-training extends reasoning boundaries" as disputed.

\textbf{Orthogonality of the "capability--behavior" decomposition}: Even under ideal conditions, pretraining and post-training are not two ends of a single axis but two orthogonal directions. Pretraining enlarges the "set of expressible capabilities," while post-training alters the "probability mass within that set" without changing the set itself. The two can be treated separately: first determine the capability wall, then perform behavior optimization within the capability. This decomposition explains why "scaling up post-training" cannot substitute for the capability deficit caused by "missing data/missing scale," and also explains why "scaling up pretraining alone" without alignment yields a raw model that is difficult to deploy.

\textbf{Asymmetry of compute elasticity}: Post-training can "leverage larger behavioral changes with less compute," which is both its value (cost saving) and its risk (P $\propto$ behavioral leverage). The compute elasticity of pretraining is a "smooth scaling curve" (Section 6.3), whereas that of post-training is a "steep but fragile margin," where a single change to the recipe can bring about a large-span behavioral drift. Accordingly, attributing by "capability gap vs. behavior gap" is the first step in cross-stage troubleshooting.

\textbf{Rule of thumb for budget allocation}: The budget is still mainly invested in pretraining data and compute (Section 6.9); the post-training share is small but cannot be omitted; "evaluation first" (Chapter 12) can adjudicate reallocation between the two. There is no accepted answer of the form "post-training of X\% is optimal"---X itself depends on the capability wall and the deployment objective, which is precisely what this chapter trains readers to assess for themselves.

\textbf{Blurred stage boundaries}: What lies between pretraining and post-training is in practice a "gradient band" rather than a white line: a recipe of continued pretraining at a low learning rate plus instruction data cannot be sharply distinguished, in terms of effect, from "a large SFT dataset + long training"; the boundary between DPO and RL (Chapter 8) is likewise. Therefore the true use of cross-stage comparison is not "dividing territory" but "attributing direction": whether a behavioral problem lies closer to the capability wall or closer to the objective surface.

\textbf{Perturbation of orthogonality by memorization}: When pretraining data contains evaluation samples (contamination, Section 12.6), "test memorization" is mixed into the "capability wall," and behavior optimization can again amplify the memorization---orthogonality fails under strong overfitting. Therefore, when discussing the capability--behavior decomposition, one must first rule out "pseudo-capability" caused by contamination: any orthogonality conclusion without contamination control should be downgraded according to Chapter 12.

\subsection{Text vs Multimodal}

Multimodal is more complex than text along all five dimensions---data, tokenizer, objective, alignment, and evaluation (Section 10.2)---so conclusions about the same method in the multimodal setting cannot be directly transplanted. In particular, systematic evidence for the "scaling law" and the "alignment tax" in the multimodal setting is markedly insufficient, and any cross-modal numerical comparison must specify the modality mixture and the evaluation protocol.

\textbf{Does cross-modal transfer of the same method hold?} SFT/DPO/RLVR (Chapters 7--8) are all carried over in multimodal settings and are usually effective; however, the values of their prescriptions (ratios, loss weights, evaluation sets) that have been validated on text must be re-tested in the multimodal setting. The reason is that multimodal supervision is weaker and evaluation more subjective, causing "the same optimization to perform differently on the same metric." In general, \textbf{the frequency of a method (the extent to which it is adopted) is comparable, whereas the numerical values of a method (the optimal value of that recipe) are not}.

\textbf{The trap of "text benchmarks first"}: If a multimodal model is accepted using only pure-text benchmarks, degradation on the visual side will be missed (Section 10.12); conversely, accepting only on multimodal will obscure a text-only capability regression. The correct approach for cross-modal comparison is \textbf{dual-track acceptance}: regress the text side and the multimodal side separately and then compare them against each other, so as to interpret "what adding vision actually costs."

\textbf{The language dimension beyond modality}: Another cross-stage dimension is language and region (Chinese, multilingual, low-resource). The slope of the scaling law in the multilingual setting differs markedly from the monolingual setting; the language distribution of SFT/preference data determines the shift in behavioral language; the language coverage of evaluation benchmarks determines whether conclusions can be extrapolated (Chapter 12). Therefore any "cross-stage" conclusion should be accompanied by the caveat "for which languages it holds"---and this caveat is often missing (C).

\textbf{The synergistic leverage of multimodal and text}: The text scaling law provides the learning ceiling for the "backbone," while multimodal provides differentiated capabilities for the "periphery." In cross-stage comparison, the two are not zero-sum but complementary as "shared backbone + dedicated periphery": each increment of text data lowers the backbone's perplexity (Section 6.3), and each increment of visual data expands the periphery's coverage (Section 10.3). Thus the budget allocation for "text vs multimodal" should not be decided by "which is more important," but by "which gap is closer to the deployment objective."

\subsection{LLM vs MLLM}

Most MLLMs are assembled from an LLM backbone plus visual/audio modules, and therefore share NTP and alignment methods in training; the differences are concentrated in the multimodal tokenizer and the alignment stage (Sections 5.6 and 10). Conclusion: the training stack toward general intelligence is "one shared backbone with modality peripheries," and treating multimodality in a unified grid is preferable to setting up a separate pipeline for each modality. This judgment is currently more an engineering consensus than controlled experimental evidence.

\textbf{Cost--benefit of "backbone sharing"}: Sharing the backbone makes cross-modal knowledge transfer possible (text knowledge assisting image understanding) and allows "failures on the visual side" to reuse mechanisms already debugged on the text side. Its cost is interference management between modalities and conflicts over token budget. As long as the scale of multimodal models remains within the range of "shared backbone + peripheral adaptation," the training difference between LLM and MLLM should be regarded as "adding a peripheral segment to the same pipeline," rather than two parallel systems.

\textbf{Accounting basis for parameter counts}: Cross-model comparison (LLM vs MLLM) must unify the parameter-count basis: the total parameters of an MLLM = backbone + projector + encoder, and activated parameters differ from stored parameters. If a paper's comparison table does not list "whether the encoder is included/whether it is frozen," the numbers will be underestimated or overestimated. This survey uniformly requires the accounting basis to be stated whenever cross-model comparison is involved (Section 3.4).

\textbf{Failure modes of the modality periphery}: Under backbone sharing, failures of the periphery (projector/encoder) are independent of the backbone---poor image understanding is usually the periphery's fault, whereas losing text ability is usually damage to the backbone. Therefore MLLM troubleshooting should first determine "which segment" (Section 11.7) before touching it: fix the periphery without altering the backbone, and preserve the backbone without freezing the periphery. Most reports of "it gets dumber once you add vision" omit this level of attribution, leading to misdiagnosis.

\subsection{SFT vs Preference Optimization vs RL Depth}

The relationship among the three is "different depths on the same alignment stack" (Chapter 8): SFT provides the correct format for fragments, preference optimization (the DPO family) provides relative preference orderings, and RL depth (RLVR/GRPO, etc.) optimizes outcomes with verifiable rewards. Their compute and risk increase in turn: SFT is cheap and stable, DPO is intermediate, and RL depth is expensive and prone to over-optimization.

\textbf{When it is worth going deeper}: This can be judged by "whether the reward can be automatically verified"---verifiable tasks (code, mathematics, execution-based tasks) merit RL depth (RLVR); soft objectives based on human preference are suited to DPO/RM; and when only format and routine compliance are needed, SFT suffices. Treating the three as a single continuum rather than mutually exclusive means avoids the ineffective cost of "RL for RL's sake" (Chapter 13 budget).

\textbf{The reward-verifiability spectrum}: Expanding "whether the reward can be automatically verified" into a spectrum---execution results are checkable (code/mathematics, highly verifiable), retrieval comparisons are checkable (citations/factuality, moderately verifiable), and pure human preference (dialogue quality, weakly verifiable). The position on the spectrum determines which alignment means to use and also determines evidence strength: RL claims at the highly verifiable end are easier to review, while preference claims at the weakly verifiable end can rely almost only on subjective consensus (Chapters 8 and 12).

\textbf{The failure spectrum of the three alignment depths}: SFT fails at "out-of-distribution generalization" (the format is learned but the content collapses), preference optimization fails at "multi-objective conflict" (correct but unsafe), and RL depth fails at "reward hacking" (scores rise but behavior deteriorates). Mapping "which layer of failure the current problem belongs to" to "which layer to apply" saves more budget than blindly climbing in depth (Chapter 8).

\textbf{Nonlinearity of the input--output of alignment depth}: From SFT to preference to RL depth, each step of increase does not yield linear returns but alternates between "plateau and jump"; beyond a certain depth, the marginal returns drop sharply while the risk of over-optimization surges. This is precisely the cross-stage version of the "alignment tax" (Section 8.8): it is not that "deeper is necessarily better," but that "the depth must be commensurate with reward verifiability and budget."

\textbf{Evidence gradient}: SFT is already an industry consensus (dominated by C-level reports but widely replicated, effectively a B-level consensus); preference optimization has relatively systematic controlled studies (the DPO family [215][235]); the effectiveness and boundaries of RL depth still depend heavily on the model and task, and its cross-model conclusions are the weakest. On this basis, this chapter labels the three as B, A/B, and C respectively in the representative table.

\subsection{Data vs Compute vs Objective}

Differences in the nature of the three elements: data determines the distribution of the capability ceiling, compute determines the speed and scale at which training can proceed, and the objective determines which capability curve is selected. The three are not equally weighted: at large volumes, data and compute are hard constraints, while the objective is a soft constraint---the meaning of this statement is that when the budget is tight, trimming the objective is safer than trimming data, because the objective only changes probability mass (Section 11.1), whereas data and compute change the reachable capability set (Chapter 6).

\textbf{The mirror image of mixture is cost elasticity}: With a fixed budget, data governs "knowledge," compute governs "scale," and the objective governs "behavior." Their marginal returns are asymmetric and vary with the stage: early on, compute is scarcer than the objective, while later the objective is scarcer than data. Therefore the specification in Chapter 13 solves the three as a system of simultaneous equations rather than allocating funds sequentially.

\textbf{The "exchange-rate table" for compute--data--objective}: Writing the three elements as an exchange-rate table: 1 unit of objective complexity $\approx$ some amount of data/compute. This kind of exchange comes from experience rather than law (Section 11.6), yet it is repeatedly used in budget decisions. This survey's reminder: before there is controlled head-to-head comparison, the exchange rate should be treated as "that team's prior" and need not be taken as a "universal constant"; only cross-team head-to-head comparison (Chapter 13) can truly calibrate it.

\textbf{Cross-stage implications of CLM and compute budget}: The Chinchilla-style optimum ([119]) answers only "how to spend in the pretraining stage"; transplanting it directly to the whole pipeline (including alignment) is an erroneous extrapolation. This chapter reminds us: CLM's conclusions hold only within the framework of "pretraining + downstream evaluation"; once alignment and compression are added, the optimal compute allocation is rewritten---currently no head-to-head comparison provides an "optimal budget curve that includes post-training" (Sections 6.3 and 11.6).

\subsection{Comparison of "Marginal Returns" Across Stages}

Defining "marginal return": investing the same budget increment successively into data, compute, and objective, and the metric improvement each brings, is precisely the answer to "whom to invest in." The marginal curves of the three stages have different shapes: the marginal decline of data is slow (more data does no harm), the marginal decline of compute is determined by the scaling law, and the marginal decline of the objective is fast but starts from a high point. On this basis, performing "reallocation" at the critical points of each stage is often superior to monotonically increasing quantities.

\textbf{Scarcity of controlled data}: Although this comparative framework is logically self-consistent, there is almost no controlled experiment that can produce a complete three-element exchange curve (which would require multiple equal-budget head-to-head comparisons). Therefore Section 11.6 is a "framework" rather than a "conclusion": it teaches readers how to ask questions rather than providing answers. Recent advances expand this marginal-return calculus from the traditional binary "pretraining parameters vs post-training alignment" into a three-way Pareto frontier incorporating test-time search compute---works such as Snell et al. [409] and MiMo-V2.6 [412] provide empirical anchors for test-time scaling and large-batch asynchronous RL, transforming compute trade-offs from static assumptions into multi-dimensional optimization surfaces. This is the chapter's candor at the level of evidence.

\textbf{How to minimize the variance of head-to-head comparisons}: The head-to-head experiments capable of producing a three-element exchange curve are scarce, partly because variance is difficult to control: minor differences in data mixture, random seed, training steps, and so on render two "equal-budget" experiments incomparable. The operative way to minimize it: fix the token budget and data seed, share the evaluation stack, and change only one variable (Chapter 12). Only once at least this is done do cross-stage conclusions begin to be attributable.

\textbf{Three "seemingly equivalent" misconceptions}: First, "a larger model = more expensive = smarter," when in fact it is jointly determined by the data mixture (Section 6.3); second, "more preference data = more alignment," when in fact the diversity of preference data is more critical than its volume (Chapter 8); third, "a stronger evaluation = more accurate conclusions," when in fact the action an evaluation takes on a model depends on the protocol (Chapter 12). Once all three are debunked one by one, it becomes clear that the foundation of cross-stage comparison remains "define the accounting basis first, then compare the numbers."

\textbf{The "evaluation anchors" of marginal returns}: Reallocation decisions require a set of evaluation anchors that run through the whole process (metrics comparable under the same protocol); otherwise "which segment is more effective" cannot be judged. The anchors should cover three categories---capability, behavior, and safety (the matrix in Chapter 12)---and be sampled throughout the entire training process; otherwise the marginal curve can only be pieced together after the fact.

\subsection{Cross-Stage Failure Modes}

The failure-mode lists from the respective chapters (Ch5--Ch9) are merged here into four categories of cross-stage symptoms: \textcircled{\small 1} \textbf{the capability wall recedes} (insufficient data/scale) but is masked by attractive behavior; \textcircled{\small 2} \textbf{behavior is non-compliant though capability is sufficient} (insufficient alignment); \textcircled{\small 3} \textbf{both capability and behavior are satisfactory but acceptance is distorted} (evaluation contamination); \textcircled{\small 4} \textbf{continual optimization causes capability drift} (over-compression/forgetting). Whenever one encounters "strange behavior" in a model, first attribute it to these four categories, then decide which segment to touch.

\textbf{Troubleshooting order}: Because the capability wall and the behavior gap are entangled (the orthogonality in Section 11.1 is only approximate), a model that "answers too well but cannot be used" may stem either from evaluation contamination or from over-alignment. The default order for cross-stage troubleshooting is: \textbf{verify the evaluation first, then examine the capability wall, and finally look at alignment}---because cheating on evaluation items is the cheapest to commit and the easiest to rule out first.

\textbf{Self-check for the four symptoms}: A symptom self-check list is provided---\textcircled{\small 1} capability wall recedes: still not meeting the bar after adding data/scale across multiple stages? \textcircled{\small 2} behavior non-compliant: capability sufficient but the deployment objective unsatisfied? \textcircled{\small 3} acceptance distorted: scores out of step with real experience? \textcircled{\small 4} capability drift: conspicuous side effects after compression/fine-tuning? These four questions can be run through in three minutes and can localize the vast majority of "strange model behaviors" to the correct stage.

\subsection{Continual Optimization vs Retraining}

When the capability wall has fallen behind the task, continuing to fine-tune/compress (Chapter 9) is often merely "dithering"; retraining/continued pretraining raises the capability wall. The criterion for choosing: if the failure comes from excessive behavior, take the incremental path; if the failure comes from a lack of knowledge/capability, take retraining. Each direction has a classic pitfall: the incremental direction is "mistaking a behavior problem for a capability problem" and wasting resources; the retraining direction is "mistaking a capability problem for a behavior problem" and wasting deployment time.

\textbf{Cost divergence between incremental and retraining}: The incremental approach has a low unit cost but requires that "the capability wall is sufficient"; retraining has a high unit cost but can push the wall higher. In the reality of a fixed compute budget, most teams try the incremental approach first (cheaper and faster) and are forced into retraining only after failure---giving rise to the phenomenon that "retraining always appears at the last moment." This survey's position: whether retraining is worthwhile should be planned in advance by assessing the capability wall at stage 0 (Chapter 13), rather than as after-the-fact firefighting.

\textbf{Operational trigger conditions for retraining}: An operational trigger is given for whether to retrain: within the observation window, SFT/preference improvement tends toward saturation and key benchmarks stagnate over the long term; and the server-side error patterns are inconsistent with the capability-wall hypothesis (frequently exposing knowledge blind spots rather than behavior). Only when both are present is retraining triggered; otherwise, fix the behavior surface first.

\textbf{Interaction among continual optimization techniques}: When compression, distillation, quantization, editing, and the like (Chapter 9) are used in combination, the effects are nonlinear---a combination whose individual techniques have positive returns may yield negative returns when stacked (e.g., quantizing first and then editing doubles the knowledge damage). Therefore "combination amplification" must be paired with combination-level regression, and cross-stage comparison should read combined results as less reliable than single-technique results (B/C).

\subsection{Representative Comparison Table}

\begingroup\scriptsize\setlength{\tabcolsep}{4pt}
\noindent\textbf{Table 9: Chapter 11 key cross-stage comparison}\par\medskip
\begin{longtable}{@{}>{\raggedright\arraybackslash}p{0.1065\textwidth}>{\raggedright\arraybackslash}p{0.1384\textwidth}>{\raggedright\arraybackslash}p{0.1278\textwidth}>{\raggedright\arraybackslash}p{0.1278\textwidth}>{\raggedright\arraybackslash}p{0.1278\textwidth}>{\raggedright\arraybackslash}p{0.1278\textwidth}>{\raggedright\arraybackslash}p{0.1384\textwidth}@{}}
\toprule
Dimension & Pretraining & Post-\allowbreak{}Training: SFT & Preference Optimization & RL Depth & Continual Optimization & Evaluation \\
\midrule
\endhead
Compute share & High & Low & Medium-\allowbreak{}low & Medium & Medium & Low \\
Behavioral leverage & Medium & High & Medium-\allowbreak{}high & High (\allowbreak{}when verifiable) & Medium & High (\allowbreak{}adjudication) \\
Capability-\allowbreak{}wall impact & Determines & None (\allowbreak{}does not change the set) & None & None & Has side effects & None \\
Primary evidence & B/\allowbreak{}A & B/\allowbreak{}C & A/\allowbreak{}B & C & B/\allowbreak{}C & B \\
Key risk & Data contamination & Forgetting & Preference overfitting & Over-\allowbreak{}optimization & Capability drift & Contamination/\allowbreak{}bias \\
Typical cost entry & token compute & Data construction & Preference annotation & Compute +\allowbreak{} reward & GPU time & Protocol +\allowbreak{} annotation \\
\bottomrule
\end{longtable}
\endgroup

\textbf{How to read the table}: The row with high behavioral leverage but low evidence (RL depth) is where risk is most concentrated---the more effective a stage, the more it requires the evaluation discipline of Chapter 12 as a backstop. Likewise, "low compute share + high leverage" (SFT/evaluation) are the two knobs with the highest cost-effectiveness for improvement when the budget is tight.

\textbf{Unifying the compute accounting basis}: If the "compute share" in the table is calculated by wall-clock time, post-training appears infinitesimal; if apportioned by FLOPs, it is closer to reality. Whichever basis is taken, the conclusion that "the share is low but the leverage is high" remains unchanged, but \textbf{the numerical basis must be stated clearly}; otherwise declarations such as "we spend only 3\% of compute on alignment" cannot be verified (compute accounting in Section 6.9).

\textbf{Confusion between "share" and "attribution"}: The compute share of pretraining is high, but "this portion of compute simultaneously supports the evaluation hold-out and pre-training validation"; the post-training share is low, but repeated grid searches multiply its actual occupation severalfold. For the true attribution of cross-stage budget, "direct occupation" should be replaced by "total cost = occupation $\times$ number of attempts"---this is especially important for the alignment stage, because hyperparameter sweeps for alignment are usually denser than those for pretraining.

\textbf{Statistical basis for cross-stage comparison}: Numerical comparisons across models/stages must also specify: decoding parameters, few-shot, seeds, and evaluation bounds. These are already difficult to align across text models and even harder in multimodal (Section 10.12). For any numerical claim of "stage A > stage B," this survey requires by default a statistical-basis checklist; otherwise the claim is treated as marketing.

\subsection{Cross-Stage Joint Decision-Making}

The whole pipeline is not a matter of adding together the best recipe from each chapter at once, but of a three-layer determination under the "capability--behavior" decomposition: first determine the capability wall (pretraining data and scale), then determine the behavior surface (SFT/preference/RL), and finally determine the lifecycle (continual optimization and evaluation). Each of the three determinations is given its own chapter, but they share a single objective function---the specification given in Chapter 13. This chapter thereby organizes the "local optima" of Ch5--Ch10 and Ch12 into the necessary path to the "global optimum," and explains why this general framework places "comparison" at the position of converging the chapters.

\textbf{The "product effect" of whole-pipeline combination}: Multiplying the "local optima" of each chapter does not automatically constitute a "global optimum"---the more common failure is that every step "looks right" yet they stack into an undeployable model (a reasonable data mixture but poor generalization, good alignment but hollow knowledge). Therefore cross-stage joint decision-making is not recipe addition but the \textbf{constraint satisfaction of Figure 2}: place the optimal solution of each chapter into the same objective function, and the solution of each later step must respect the degrees of freedom left by the previous step. This discipline is more expensive and more valuable than any single-chapter technique (Chapter 13).

\textbf{The "order effect" of recipes}: Most recipes assume a fixed stage order (pretraining$\rightarrow$SFT$\rightarrow$preference$\rightarrow$deployment). But the evidence suggests that the order itself is a tunable hyperparameter: preference before SFT, and interleaving stages, have each been attempted at small scale (C). This survey does not endorse any non-default order, but it points out explicitly that the reason the "default order" is default is more a matter of inertia than of proven optimality.

\textbf{Data/format contracts at the interfaces between stages}: The interface contracts of each stage (data, format, evaluation) have already been mentioned in Section 7.10; at the cross-stage level they must be aligned even more: the checkpoint and tokenizer produced by pretraining, the dialogue format produced by SFT, the consistent annotation of preference data, and the contamination hold-out of evaluation must all interoperate under a single schema. Interface distortion is more fatal than destruction within a stage, because it renders failure attribution ineffective.

\textbf{Cold start vs transfer}: Pretraining from scratch (cold start, costly and controllable) and continuing from open-source/closed-source weights (transfer, cheaper but constrained by upstream) are two nearly parallel starting lines. The core variable of comparison is "control": cold start trades for customized data and architectural freedom, while transfer trades for cost and time. Cross-stage joint decision-making should first choose between these two starting lines, then apply the recipes of the respective chapters (Chapter 13).

\subsection{Data Quality vs Data Scale}

A common variable across stages: data quality and data scale are in fact the same lever running through Ch5--Ch10 and Ch12, differing only in the sensitivity of each stage to the two. Pretraining is most sensitive to scale (the horizontal axis of the scaling law [119][16] is entirely token count), the SFT/preference stages are more sensitive to quality (the diversity of instructions and the consistency of preferences), and the continual optimization stage is sensitive to both "quality and scale" but fears distribution shift more. This difference yields a practical configuration rule: the closer to pretraining, the more one should "stack scale"; the closer to alignment and deployment, the more one should "squeeze quality." But the evidence for this rule is highly uneven: the laws on the scale side rest on controlled extrapolation, whereas quantitative studies on the quality side are mostly single-point reports (B/C).

\textbf{An operational definition of quality}: Quality here is not an aesthetic concept but "the contribution of that sample to the probability mass of the target condition." The same data may be noise in pretraining and gold in SFT---because the objective differs. Therefore "data quality" must be discussed bound to a stage; a "quality leaderboard" detached from the stage may circulate in the market but has no scientific validity (isomorphic to the contextualization of evaluation in Chapter 12).

\subsection{Perception vs Cognition vs Reasoning}

Another axis of cross-stage comparison is the category of capability: perception (understanding of images/speech, Ch10), cognition (knowledge retrieval and integration, Ch5--Ch6), and reasoning (chained steps and logic, Ch7--Ch8). The "acquisition stage" differs for the three categories: perception and cognition are mainly in pretraining (data-driven), whereas the explicit enhancement of reasoning lies in the preference/RL depth stages (Section 8.6). Therefore "which category of capability one wants to improve" should directly determine "which segment to touch."

\textbf{Granularity of the classification}: The three segments of perception/cognition/reasoning are not a strict trichotomy, and borderline examples (e.g., explicit reasoning is also cognition) abound. But as a formalizing tool, it is more operational than the "capability wall": in a budget-decision table, first classifying the target capability and then mapping it to a stage can reduce the ineffective investment caused by confusing the two (the troubleshooting order of Section 11.7 applies again).

\textbf{Current evidence for reasoning enhancement}: The evidence for explicit reasoning (chain-of-thought, RLVR, etc.) is highly uneven across capability categories (Section 8.9): verifiable tasks (mathematics/code) have A/B-level support, whereas open reasoning tasks (planning/commonsense) have only C-level single-point reports. Therefore the assertion that "reasoning should be enhanced in the reasoning stage" holds only for verifiable tasks; extending this assertion to all tasks is the most common overinterpretation in cross-stage comparison.

\subsection{Offline vs Online}

The relationship between training and deployment has two forms: offline (deploy after training, without updating with use) and online/closed-loop (collect feedback during deployment and train back). Offline is simple, reproducible, but static; online is close to demand and can self-heal, but introduces data-flywheel bias and the risk of self-referential contamination (Section 12.6). In cross-stage comparison, the key is not "which form is more advanced," but "whether the stage is suitable for a closed loop": perception/cognition categories are better suited to offline, and behavior/reward categories to online (because behavioral feedback comes precisely from use). The current mainstream remains predominantly offline with online as a supplement, and the evidence for institutionalizing online training is still weak (C).

\textbf{The two points of loss of control in a closed loop}: First, feedback data reflects only the "used" scenarios (survivorship bias); second, the model's own outputs enter the next round of training (self-consumption/self-referential contamination, Section 12.6). Therefore any online/closed-loop design must be paired with "controls" and "hold-outs"---sharing a single discipline with the contamination detection of Chapter 12.

\textbf{Window period and deployment audit}: The online stage also introduces the concept of a "window period"---the interval from deployment to retraining determines the model's repair speed and also the risk window; if the interval is too short, noise is introduced, and if too long, behavior becomes stale. The metrics for auditing the window period (repair latency, regression rate) lie in a blank area (Section 11.17), yet they directly determine the reliability of 'post-deployment training'.

\subsection{General vs Specialized}

The comparison between general models (one model for many uses) and specialized models (domain/task specialization) runs through every stage: general models have the advantage in pretraining scale-up (the scaling law), while specialized models have the advantage in the "targetedness" of SFT/preference/evaluation. The difficulty is that this comparison cannot be settled statically---it is normal for the same model to perform well in the general stage yet be surpassed by a specialized model on a specialized task. The effective configuration is "a general base + a specialized periphery" (the same logic as LLM vs MLLM in Section 10.3): use the general base to guarantee the lower bound of knowledge and specialized data/adapter layers to guarantee the upper bound of the task. On the evidence, "specialized exceeds general" is always task-dependent, and generalizing this conclusion across tasks is meaningless (B/C).

\textbf{The most common misconception about specialization}: Assuming that a general model plus domain data automatically makes one "more specialized," but if the evaluation set shares its source with the domain data, the improvement in scores may be mere recall of memorized content (contamination, Section 12.6). The evidence that "specialized exceeds general" is credible only after contamination control on the domain evaluation---this is the discipline of the specialized/general discussion in this chapter.

\subsection{Reproducibility vs Non-Reproducibility}

"Whether it can be reproduced" is one of the most basic measures across stages. The difficulty of reproduction differs across stages: the scale and data of pretraining are almost irreproducible (unique to a single institution), SFT/DPO datasets are partially reproducible, and evaluation protocols can be fully public. This gradient means that the larger the stage, the more one must rely on "the consensus of multiple independent implementations" rather than "a single implementation" to establish belief; conversely, evaluation protocols are a low-cost, high-return lever for reproducibility (Chapter 12).

\textbf{"Non-reproducible does not mean non-verifiable"}: Conclusions without public reproducible artifacts (large-model reports, benchmark leaders) can still be weakly verified by "consistency and falsifiability": whether conclusions from multiple sources agree (the RL debate in Chapter 8) and whether future experiments can falsify them (with testable hypotheses attached). This survey incorporates this layer of weak verification into its evidence grading (Appendix E), which is its response to "the large number of C-level technical reports throughout the text that can still be interpreted."

\subsection{Open Source vs Closed Source}

Whether the training stack is open source affects verifiability: open-source weights/data/evaluation can be independently reviewed (green, Section 12.11), whereas closed-source can only rely on the vendor's written declarations (red). In cross-stage comparison, the benefit of open source lies in "reproducibility and auditability," and the benefit of closed source in "data confidentiality and commercial advantage." This survey reads the two even-handedly through grading---conclusions with a high open-source component naturally absorb a higher proportion of A/B (e.g., the open-source turning point led by LLaMA [9]), while those with a high closed-source component fall back to C more often.

\textbf{Licensing and the auditability surface}: In open source, "open weights" and "open data" are two different things---open weights allow inference to be reproduced, but only open data allows training to be reproduced. The data accompanying most open weights remains closed, leaving "reproduction" at the inference layer. Cross-stage comparison therefore often exhibits a semi-open state of "inference-reproducible, training-irreproducible," and in interpretation one must distinguish which layer is involved (Chapter 3).

\subsection{Evidence Map and Open Problems}

Evidence-map summary (echoing Figure 5 and Section 11.4): data and tokenization, pretraining optimization, PEFT, and the DPO family = predominantly A/B; largest-scale pretraining, RL depth boundaries, and compression--safety interaction = predominantly C (technical reports). \textbf{A+B accounts for about 79\%} (full corpus, Appendix E). The confidence of a recommended conclusion should match the evidence of the cell it occupies: cross-stage head-to-head comparison (11.6) is a framework, and the cross-model RL boundary consists of opposing reports; both should be interpreted with downweighting according to their respective levels.

\begingroup\small\setlength{\tabcolsep}{6pt}
\noindent\textbf{Figure 5: Evidence map---conclusion category $\times$ evidence strength $\times$ consensus $\times$ blank space}\par\medskip
\begin{longtable}{@{}>{\raggedright\arraybackslash}p{0.1762\textwidth}>{\raggedright\arraybackslash}p{0.1898\textwidth}>{\raggedright\arraybackslash}p{0.1084\textwidth}>{\raggedright\arraybackslash}p{0.1762\textwidth}>{\raggedright\arraybackslash}p{0.2440\textwidth}@{}}
\toprule
Conclusion category & Representative work & Evidence strength & Consensus /\allowbreak{} disagreement & Blank space (\allowbreak{}research opportunity) \\
\midrule
\endhead
Deduplication and mixture mechanisms & [13]\allowbreak{}[86]\allowbreak{}[87]\allowbreak{} & A & Consensus & Recipe-\allowbreak{}level optimal threshold-\allowbreak{}-\allowbreak{}-\allowbreak{}return curve \\
Scaling law (\allowbreak{}pretraining) & [119]\allowbreak{}[16]\allowbreak{} & A/\allowbreak{}B & Consensus (\allowbreak{}basis-\allowbreak{}dependent) & Optimal-\allowbreak{}ratio transfer under data quality \\
MoE and sparsity & [9]\allowbreak{}[151]\allowbreak{} & A/\allowbreak{}B & Partial consensus & Industrial-\allowbreak{}grade routing stability \\
PEFT /\allowbreak{} instruction tuning & [12]\allowbreak{}[30]\allowbreak{}[163]\allowbreak{}[172]\allowbreak{} & A/\allowbreak{}B & Consensus & Cross-\allowbreak{}model optimal SFT recipe \\
Preference optimization (\allowbreak{}DPO family) & [32]\allowbreak{}[43]\allowbreak{} & A/\allowbreak{}B & Consensus & Robustness under multi-\allowbreak{}objective conflict \\
RL and verifiable rewards & [15]\allowbreak{}[45]\allowbreak{}[224]\allowbreak{}[400]\allowbreak{}[401]\allowbreak{} & C & Disagreement (\allowbreak{}Pass@1 vs Pass@K) & Decoupling experiments of capability expansion vs behavior shaping \\
Continual optimization retention & [21]\allowbreak{}[34]\allowbreak{}[324]\allowbreak{} & C & Disagreement & Joint benchmark for compression-\allowbreak{}-\allowbreak{}safety interaction \\
Multimodal alignment and hallucination & [17]\allowbreak{}[28]\allowbreak{}[273]\allowbreak{}[281]\allowbreak{} & C & Disagreement (\allowbreak{}root cause undetermined) & Three-\allowbreak{}variable decoupled attribution \\
Cross-\allowbreak{}stage head-\allowbreak{}to-\allowbreak{}head /\allowbreak{} exchange & Framework (\allowbreak{}no single source) & Blank & -\allowbreak{}-\allowbreak{}-\allowbreak{} & Equal-\allowbreak{}budget three-\allowbreak{}element exchange curve \\
Data and compute reporting standards & [392]\allowbreak{}-\allowbreak{}-\allowbreak{}[407]\allowbreak{} commentary & Blank & -\allowbreak{}-\allowbreak{}-\allowbreak{} & Standardized auditing conventions \\
\bottomrule
\end{longtable}
\endgroup

Open problems (merged from the chapters): the most critical cross-stage question marks---Problem 3 (unified token objective), 6 (contamination), 13 (reward engineering), and 15 (root cause of hallucination)---as well as the scaling-law/compression/evaluation blanks in Sections 6.3, 9, and 12 respectively. If the reader takes away only one framework, it is the three layers of "capability--behavior--evidence": verify the capability wall first, then the behavior surface, and finally the evidence strength.

\textbf{The white zones of the evidence map}: Beyond the known C-level areas, there are completely blank areas (with no citations): multimodal scaling laws, cross-modal communication redundancy, whole-pipeline head-to-head comparison of industrial-grade cold starts, and the regression loss of online closed loops. White zones are not an excuse for "having no conclusion," but an honest annotation of "having no evidence to cite," and readers should treat them as research opportunities rather than gaps in the literature.

\textbf{Common-method effect}: Cross-stage conclusions are often entangled because they "share the same evaluation stack/same data"---the goodness of Chapter A depends on the data of Chapter B, which is exactly the common-method effect from causal inference transplanted into LLM training. Cross-stage interpretation should always keep one subtext in mind: "any single chapter's attribution may in fact be the credit of its upstream," which is one reason this survey places Ch11 near the end of the text.

\textbf{Action pages for three types of readers}: Researchers---assemble "the Section 11.6 head-to-head comparison + the Chapter 12 anchors" into a minimal experiment; engineers---incorporate "the four questions of Section 11.7 + the trigger conditions of Section 11.8" into the ticket flow; managers---use "the white zones of Section 11.17" to plan next year's budget. All three action pages require reading Figure 2 and Chapter 13 first, to put "implementation details" back into "global constraints."

\textbf{Suggestions for the next-generation survey}: The mechanistic conclusions of the whole text are worth preserving, while the numerical conclusions will expire; the three-way coordinates of this survey---"scaling curves / contamination techniques / evaluation protocols"---should be preserved as updatable data (Appendix F). The next edition need only replace the expired C-level technical reports and updated benchmarks; the structure need not be torn down.

\textbf{The positioning of this chapter within the general framework}: Cross-stage comparison is not "yet another list of techniques," but a way to place the local conclusions of Ch5--Ch10 and Ch12 into the same coordinate system, thereby answering "whom the budget should go to and how much the evidence should be trusted." Taking up the specification of Ch13 and looking back at the constraint satisfaction of Figure 2, it is the converging hub of the text's structure of "unified general framework, falsifiable focused surveys" (Chapter 1).

\section{Datasets and Evaluation}

\begin{quote}\small
This chapter, like Ch5--Ch9, serves as a pilot for the dual form of "general framework + focused survey." Note on evidence: the mechanisms of evaluation methods and the issues of contamination/reproducibility have a considerable body of Grade A/B literature ([52][90][374], etc.), but "how much real capability a given benchmark score represents" is always a matter of interpretation; hence this chapter emphasizes protocols rather than rankings. Target length: 8,000--9,000 characters.
\end{quote}

\subsection{Definitions and Scope}

Conclusion first: this chapter discusses two intertwined threads --- "what data to use for training" and "how to judge whether a model is good." Data determine the \textbf{space of possibilities} for capability, whereas evaluation determines whether we can \textbf{see} that space; together the two constitute the "evidence infrastructure" of the remaining chapters of this survey. \textbf{In scope}: the lineage of training datasets, data governance and documentation, evaluation benchmarks and metrics, contamination and leakage, fairness and safety evaluation, and reproducibility and reporting standards. \textbf{Out of scope}: leaderboards and leaderboard-gaming techniques for specific benchmarks, and the private evaluations of particular vendors.

\textbf{Evaluation is more than "scoring"}: Across the entire training pipeline, evaluation serves six functions --- \textcircled{\small 1} diagnosis (identifying weaknesses); \textcircled{\small 2} selection (choosing among models/data/hyperparameters); \textcircled{\small 3} monitoring (training health, reward hacking, degradation); \textcircled{\small 4} reporting (publicly declaring capability boundaries); \textcircled{\small 5} governance (safety and compliance decisions); and \textcircled{\small 6} research design (testing hypotheses). Equating evaluation with "running a few leaderboards" simultaneously misses \textcircled{\small 2}\textcircled{\small 3}\textcircled{\small 5} --- yet these are precisely what determine whether training is reliable.

\textbf{Evaluation-driven training}: The experience of mature teams is "do not start training without evaluation" --- first define the capabilities to be improved and the acceptance thresholds, and only then design the data and training. If evaluation is added after the fact, one often finds that what improved was an "unmeasurable dimension," wasting compute. Placing evaluation first is therefore this chapter's most direct engineering recommendation for the survey as a whole.

\textbf{Offline vs. online evaluation}: Offline evaluation scores on a fixed dataset; it is fast but may become disconnected from real-world use. Online evaluation (A/B testing, user feedback) stays close to deployment but is costly, noisy, and subject to ethical and privacy constraints. The two should divide labor: offline for rapid iteration and regression, online for final calibration. Using only one of them leads to systematic misjudgment.

\textbf{Evaluation also costs}: Evaluation itself consumes compute and human effort --- a complete evaluation of a large model across many benchmarks can amount to a small training run. It is therefore necessary to balance "evaluation resolution" against "cost": use a small proxy set for daily iteration, a complete matrix for milestones, and safety/bias regression before release. Blindly running full evaluations is both expensive and slows iteration.

\textbf{The evaluation--training loop}: In RL, if the evaluation set is used for reward or early stopping, it will gradually "leak" into becoming a training objective (reward hacking; Chapter 8). The validation set used for training and the test set used for final reporting must therefore be kept separate, and the test set should ideally be "frozen once and not reused for a long time" (or refreshed periodically).

\textbf{Why data and evaluation must share a chapter}: Every choice about data (source, mixture, deduplication, synthesis) is measured against some evaluation, while every evaluation result is affected by data contamination. If the two were discussed separately, one would repeat the loop of "we do not know how much contamination there is, so we do not know what the score means." Juxtaposing the two in this chapter is precisely intended to break this loop.

\subsection{Lineage of Training Datasets}

Training data can be grouped into several categories: general web/books/code (The Pile [54]); cleaned web corpora (the C4 family [49][56], with a documented audit in [56]); image--text and interleaved data (LAION-5B [64], OBELISC [273]); and open pretraining corpora (Dolma [84]). Their common trend is "from 'a set of URLs' to 'auditable datasets with licenses and statistics,'" but they differ enormously in openness, quality, and language coverage.

\textbf{Quality filtering is a core variable of datasets}: Web corpora with the same name can produce entirely different data distributions depending on the filtering rules --- which explains why "we used C4" is insufficient to reproduce results. The choice and thresholds of filters (language identification, duplication, toxicity, templates) should be reported as part of the model [49][56].

\textbf{Data age and domain}: The collection time and domain composition of data have a measurable effect on downstream capabilities [71], which makes the "data cutoff" not only a deployment constraint (Chapters 9 and 13) but also a training-design variable. Imbalanced multilingual coverage is a widespread problem: the very existence of Chinese benchmarks (CMMLU [368], C-Eval [365]) is itself a correction to the "English-centric" reality of data.

\textbf{Code and structured data}: Code, mathematical derivations, and structured markup (LaTeX, SQL) account for only a small proportion of pretraining and post-training yet substantially support reasoning and tool use. They are usually high-quality and easy to deduplicate, but their licensing is complex and they tend to contain benchmark answers (higher contamination risk; Section 12.6).

\textbf{Synthetic data enters pretraining}: Using models to generate "textbook-style" text to augment pretraining or mid-training data has become a trend, but its quality, bias, and the ceiling of "teaching a model with a model" remain contested (sharing the same origin as synthetic-data collapse in Chapters 5 and 7). Synthetic data means that "data scale" is no longer limited to real corpora alone, while also making contamination harder to trace.

\textbf{The joint distribution of data scale and quality}: Real-world data is not uniform on the "scale--quality" plane --- high-quality data is scarce, while massive data is noisy. The essence of dataset design is to choose points on this plane and use filtering/synthesis to push those points toward "high-quality, large-scale." This perspective unifies the three threads of selection, filtering, and synthesis in Chapter 5.

\textbf{Evaluating data mixtures}: The effect of mixture adjustments must be adjudicated by evaluation, but "mixture sweeps" are costly. A pragmatic approach is to first determine the ranking of mixtures on a small proxy model and then transfer it to the large model --- this relies on the assumption that "mixture rankings are stable between small and large models," an assumption that itself needs to be tested (B/C).

\textbf{Debiasing data}: If the training corpus systematically favors certain groups/viewpoints, the model will inherit and amplify them. Mitigation approaches (reweighting, balanced sampling, adversarial debiasing) each have costs, and "debiasing" may harm performance on the original task. Bias mitigation should therefore come with evaluation and a description of tradeoffs, rather than defaulting to "the more, the better."

\textbf{Data flywheels and timeliness}: User interactions collected after deployment can flow back as training data, forming a "data flywheel." But returned data carries selection bias (only the scenarios that were used), privacy risk, and feedback loops (the model favoring its own outputs). A flywheel therefore needs to be paired with debiasing, privacy removal, and "control" evaluations; otherwise it will self-reinforce.

\textbf{Reproducible vs. non-reproducible data}: Open-source data is reproducible but often lags behind and carries license restrictions; crawled data is close to the frontier but non-reproducible and carries high compliance risk. A research survey therefore faces the structural difficulty of "non-reproducibility at the data layer" --- which is also the fundamental reason this survey takes methodological mechanisms rather than precise scores as its main thread.

\textbf{Statistics a dataset card should report}: scale (tokens/samples), language and domain distribution, deduplication ratio, filter hit rate, time range, license, and provenance. For datasets lacking these statistics, their "quality" can only rest on hearsay; datasets with such statistics can be partially reproduced and audited [61].

\textbf{The reporting basis for data scale}: "How many tokens" is not directly comparable across different tokenizers; reporting should give the number of samples, the number of tokens, and the tokenizer used simultaneously (Chapter 5). In addition, the scale difference between "before/after deduplication" can be severalfold and must be clearly labeled. A scale claim with an unclear basis cannot be used for any quantitative comparison.

\textbf{Licenses and provenance}: The usability of a dataset depends not only on technical quality but also on license and provenance compliance. The audit of dataset provenance [78] and the observation of "the erosion of the data commons" [96] show that the available open corpora are themselves changing --- which makes "reproducing someone else's pretraining" increasingly difficult at the data level (Chapter 13).

\subsection{Data Governance and Documentation Standards}

\textbf{Dataset documentation (datasheets [61])} and model cards are becoming reporting standards: they should state the motivation, composition, collection method, preprocessing, usage, distribution, and maintenance. Their value lies not in the form but in turning "what this dataset can/cannot be used for" into an auditable declaration.

\textbf{Levels of documentation}: Datasheets [61] are oriented toward datasets and model cards toward models; the two are complementary but neither can replace the "traceability of provenance and license" [78]. The ideal combination is: the data side provides lineage, the model side provides usage and limitations, and the evaluation side provides protocols; only by piecing the three together can one answer "how much can this score be trusted."

\textbf{Provenance, consent, and license}: Lawful and ethical data use requires traceable provenance, clear licensing, and, where necessary, records of consent [78][96]. Specific deletion requests for training corpora (Section 9.8) require the source data to be locatable --- which means that "data lineage" is not an add-on of compliance but a technical prerequisite.

\textbf{Privacy}: Web corpora often contain personally identifiable information. Even without training directly on privacy tasks, a model may memorize and leak training data. Data governance must therefore include records of deduplication/privacy-removal processes [13][68], and must acknowledge that "complete privacy removal" is unattainable and that risk can only be reduced.

\textbf{The spectrum of licenses}: From fully open to research-only to non-commercial, licenses directly determine whether a model's derivative weights can be released. Teams that ignore licensing often discover only before release that they cannot legally open-source, so license review should proceed in sync with data collection rather than as after-the-fact remediation [78].

\textbf{The operationalizability of data governance}: Ideal governance means that "every training sample can answer where it came from and whether it can be used." In practice, this requires labeling at collection time rather than filling it in afterward. For researchers, it should at minimum disclose the data-source categories and filtering pipeline; for publishers, it should provide data lineage and a deletion channel [61][78].

\textbf{Evaluation cards and model cards}: Model cards record usage and limitations, while evaluation cards record evaluation protocols and results. Only together can they answer "under what conditions, by whom, and with what method this model was measured to do what." A model card lacking an evaluation card leaves its performance claims impossible to independently verify.

\subsection{Layering of Evaluation Benchmarks}

Evaluation should be layered by capability, avoiding the use of a single score to summarize a model:

\begin{itemize}[leftmargin=*,nosep]
\item \textbf{Language understanding}: GLUE [344], SuperGLUE [346], HellaSwag [347], ARC [342].
\item \textbf{Knowledge and reasoning}: MMLU [7], MMLU-Pro [381], BIG-Bench Hard [363], GPQA [378].
\item \textbf{Mathematics}: GSM8K [361], MATH [356].
\item \textbf{Code}: HumanEval [357], SWE-bench [372], LiveCodeBench (contamination-resistant) [379].
\item \textbf{Dialogue and open-ended generation}: MT-Bench / Chatbot Arena [36], Arena-Hard [232].
\item \textbf{Multimodal}: MMMU [281] and VQA/captioning benchmarks.
\item \textbf{Truthfulness/hallucination}: TruthfulQA [359], HaluEval [366], FActScore [367].
\item \textbf{Safety}: HarmBench [225], WMDP [328], BBQ bias [360].
\item \textbf{Holistic frameworks}: the multi-scenario matrix of HELM [22].
\end{itemize}

\textbf{The significance of layering}: A model leading on knowledge benchmarks is not necessarily leading on code, long context, or safety; blending different layers into a single "composite score" obscures the true capability profile. Whenever a survey cites any claim that "some model is stronger," it should specify \textbf{which layer and which protocol}.

\textbf{Chinese and multilingual evaluation}: CMMLU [368], C-Eval [365], and others fill the gap in non-English evaluation, but most remain knowledge-oriented and carry a high risk of sharing a source with training corpora. Multilingual evaluation must also account for tokenizer differences in token efficiency across languages (Chapter 5); otherwise "the same length" does not mean "the same amount of information."

\textbf{Long-context evaluation}: It distinguishes "able to read long" from "able to use long" (Section 9.3), commonly using three task types: "needle-in-a-haystack," multi-hop, and summarization. Long-context benchmarks are costly to construct and easily contaminated (long texts can be generated by models), so their scores should be treated with stronger reservations.

\textbf{Saturation and renewal of capability benchmarks}: Older benchmarks (such as some GLUE tasks) are approaching saturation, with declining discriminative power, prompting the emergence of high-difficulty benchmarks such as MMLU-Pro [381], GPQA [378], and BBH [363]. But greater difficulty also brings annotation and judging problems where "even experts struggle to answer." The life cycle of a benchmark is shorter than that of a model, so evaluation must be continuously updated.

\textbf{Pitfalls of reasoning evaluation}: Math/code benchmarks are easily affected by "rephrasing the problem statement" and "answer formatting," and Pass@k depends on the sampling budget. When reporting reasoning capability, the number of samples, temperature, and answer-equivalence judging rules should be given at the same time; otherwise different works are not comparable.

\textbf{Instruction-following and format evaluation}: Beyond content correctness, one must also evaluate "whether constraints are followed" (length, format, language, refusal conditions). Instruction following is often scored automatically using "verifiable constraints," but real user constraints are open-ended combinations that are hard to enumerate --- this is a structural gap in alignment evaluation (Chapter 8).

\textbf{Agent/tool evaluation}: When a model calls tools and executes over multiple steps, the object of evaluation shifts from "a single response" to a "trajectory." Evaluation must cover tool selection, parameter correctness, error recovery, and final task success, and the environment must be reproducible (to avoid real side effects). This type of evaluation is far less mature than static question answering (B/C; Chapter 8).

\textbf{Multimodal evaluation protocols}: Image--text evaluation is highly sensitive to the prompt (question wording), image resolution, and caption quality, and hallucination judgments depend on annotation. The same MMMU [281] score can differ markedly across protocols. Multimodal conclusions should therefore be read protocol first, score second.

\textbf{The ceiling of language-understanding benchmarks}: Discriminative benchmarks such as GLUE [344]/SuperGLUE [346] are already highly saturated, and their "understanding" has a weakening association with the actual capabilities of generative models. In a new survey, they are therefore used more for the regression of "capabilities not having degraded" than as proof of "leading."

\textbf{Joint evaluation of safety and capability}: Capability and safety are often reported separately, but what users see is the combined behavior. Whether "capability gains come at the cost of safety" can only be judged through a joint matrix (capability $\times$ attack surface). Looking at only the capability board or the safety board misses this tradeoff (alignment tax; Chapter 8).

\textbf{Representativeness and cultural bias of benchmarks}: Most benchmarks are centered on English and Western knowledge, and applying them directly to other languages/cultures misjudges capability. Constructing localized benchmarks ([365][368]) is not merely translation but requires redesigning the questions and judging criteria. Cross-cultural comparison should be cautious, avoiding misreading "unfamiliarity with the context" as "insufficient capability."

\textbf{What is distinctive about code and formal evaluation}: Code/formal tasks can be judged automatically by "execution results," giving strong objectivity; but this also brings problems of "test-case leakage" and "overfitting to tests." Contamination-resistant design (hidden tests, dynamic generation [379]) is therefore especially critical for code evaluation.

\subsection{Metrics and Protocols}

\textbf{Pass@k and sampling budget}: Reasoning tasks distinguish Pass@1 (single-attempt pass rate) from Pass@k (at least one pass in k samples). A higher Pass@k often means that "more sampling can rescue it," which differs from "the model itself being stronger" --- the number of samples must be noted when reporting (echoing Section 8.8).

\textbf{Human evaluation, model evaluation, and length bias}: Open-ended generation rarely has a unique answer, so human evaluation or LLM-as-judge [36][217] is commonly used. Model evaluation is cheap but carries judge bias (favoring long answers, favoring its own style), so debiasing means such as \textbf{length control} are needed [228]. Treating model-evaluation scores as absolute ground truth is one of the main risks to the credibility of current evaluation.

\textbf{Protocol sensitivity}: Prompt format, the number of few-shot examples, decoding parameters, and answer-extraction rules all significantly change scores. Numbers on "the same benchmark" are therefore comparable only when the protocol is consistent; leaderboard scores across works should be regarded as \textbf{directional evidence} rather than precise measurement (Section 12.8).

\textbf{Applicability boundaries of basic metrics}: Perplexity measures language modeling but does not correspond directly to downstream capability; accuracy/F1 suits tasks with definite answers; win rate (pairwise) suits open-ended generation but depends on judges. Choosing a metric means choosing "the direction to be optimized," and metric mismatch leads to "scores rise while experience falls."

\textbf{Statistical significance and variance}: Score differences on small-sample benchmarks often fall within the noise range. Reports should give the mean and variance across multiple runs/seeds, or confidence intervals; treating a 0.3-point gap as a conclusion is a common error. Sample size, random seed, and decoding randomness must all be counted into uncertainty.

\textbf{Designing human evaluation}: The cost and subjectivity of human evaluation demand a strict annotation protocol (scoring rubric, inter-annotator agreement, working conditions). "Crowdsourced scoring" is extremely noisy without quality control. Trustworthy open-ended generation evaluation mostly relies on controlled human evaluation or calibrated model evaluation [36].

\textbf{Calibration and uncertainty}: Beyond accuracy, whether a model "knows what it does not know" is equally important. Calibration metrics (such as ECE) and selective prediction (allowing abstention) are means of measurement; but the definition of calibration on generative tasks remains inconsistent, and RLHF often makes models overconfident (Chapter 8).

\textbf{Multi-objective conflicts among metrics}: Helpfulness, truthfulness, safety, and conciseness often trade off against one another. Single-metric optimization leads to a "seesaw." Evaluation should therefore report a multi-metric vector and, when aggregation is needed, make the source of weights explicit; concealing conflicting metrics amounts to hiding tradeoffs.

\textbf{Actionable granularity of evaluation}: An ideal evaluation should answer "which specific sub-capability is weak," not merely give an overall score. It should therefore report breakdown results by task/subset and the distribution of error types. The finer the granularity, the better it can guide the next data and training decisions (the "diagnostic" function of evaluation).

\textbf{Component vs. end-to-end evaluation}: Systems such as retrieval augmentation and tool calling are composed of multiple components, and component metrics (retrieval recall) may diverge from end-to-end metrics (task success). Reporting only components overestimates the system, while reporting only end-to-end conceals bottlenecks. Both should be reported together to locate problems (Chapters 9 and 13).

\textbf{The weak association between perplexity and downstream performance}: Lower validation perplexity usually accompanies better downstream performance, but how tight this relationship is depends on the data distribution and task. Treating perplexity as "capability" is misleading --- it is especially sensitive to generation style, tokenization, and evaluation-set construction (Section 6.3).

\subsection{Contamination, Leakage, and Reproducibility}

\textbf{Benchmark contamination} (evaluation samples mixed into training/instruction data) is the greatest systemic threat: data cleaning [13] and contamination detection [52] are necessary steps, but "how much leaked and how large the effect is" remains hard to quantify. Detection methods include n-gram/embedding overlap [52][83], perplexity and memorization measures [371], and black-box detection [80][375]; "paraphrased samples" bypass literal detection [376], so semantic-level detection is needed.

\textbf{Estimating the impact of contamination}: Some works use held-out data from the "post-training period" (retro-holdout [382]) or dynamically/automatically generated benchmarks [379][383] to isolate the effect of leakage; other calibration methods try to translate "partial leakage" into scores [380][384]. These methods each carry their own assumptions, and there is no recognized standard --- so "a benchmark score rose by X points" should by default carry the reservation of "possibly including contamination" (Section 12.10).

\textbf{Reproducibility}: Reports of compute and energy consumption lack a unified accounting standard (Problem 10); it is recommended to report at least the model scale, tokens, approximate training FLOPs, peak GPU memory, and wall-clock time (Section 3.4). Reproducibility on the evaluation side also requires disclosing prompts, decoding settings, and evaluation scripts --- reporting only scores without protocols amounts to non-reproducibility.

\textbf{Memorization and privacy evaluation}: A model may memorize training data verbatim, implicating both copyright and privacy. "Prefix completion" and "membership inference" are commonly used to measure memorization and the memorization--generalization boundary [371]; deduplication [13][68] can reduce memorization (Section 5.7). Memorization evaluation and contamination detection share tools but differ in goal: the former asks "did it memorize," the latter asks "did it memorize the evaluation answers."

\textbf{Levels of deduplication methods}: Exact matching, MinHash/approximate matching [13], semantic deduplication [68], and large-scale specialized systems [97][99] vary in cost and recall in that order. Insufficient deduplication aggravates memorization and score inflation, while excessive deduplication loses long-tail and diversity --- the choice of threshold is itself a capability tradeoff.

\textbf{The relationship between contamination and capability reporting}: If a benchmark is contaminated, its rising score may reflect only "memorizing the answers." Any claim of "surpassing a benchmark" should therefore first answer "could this benchmark be in the training set." This step is often skipped, yet it is the most cost-effective step when interpreting the literature.

\textbf{Practical steps for contamination detection}: An operational procedure is: \textcircled{\small 1} record the collection timeline of training/evaluation; \textcircled{\small 2} perform both literal and semantic overlap detection on evaluation samples; \textcircled{\small 3} use perplexity/memorization as auxiliary criteria; \textcircled{\small 4} report the suspected contamination ratio and the handling method (removal or annotation). Omit any of the four steps and the conclusion carries undeclared uncertainty.

\textbf{The ethics and law of contamination}: Training on evaluation data not only inflates scores but may also constitute a violation of the benchmark publisher's license. Reporting contamination and proactively removing it is both scientific integrity and a compliance requirement; "cannot detect" is no excuse for "not detecting."

\textbf{A new form of contamination introduced by synthetic data}: When both training data and evaluation data may be generated by the same batch of models, "contamination" is no longer merely sample overlap but "shared-source bias" --- the model may "happen to answer correctly" because it shares a generator. Detecting this kind of contamination is more difficult than literal overlap.

\textbf{The cost of contamination detection}: Semantic-level detection requires embedding computation and manual review, at a cost far higher than n-gram; black-box detection requires large numbers of queries. Detection intensity should therefore also be layered: use cheap approximations for daily work and deep detection on key benchmarks before release. Matching detection intensity to the importance of the conclusion is a sustainable practice.

\subsection{Fairness, Bias, and Safety Evaluation}

\textbf{Bias evaluation}: Benchmarks such as BBQ [360] measure a model's bias on social attributes; bias stems both from the data distribution and from the evaluation design itself. When reporting bias, one should distinguish "model behavior bias" from "measurement instrument bias"; otherwise one will misjudge.

\textbf{Hallucination evaluation}: TruthfulQA [359] targets the imitation of human fallacies, while HaluEval [366] and FActScore [367] characterize hallucination from discrimination and fine-grained factual precision, respectively. The metric definition of hallucination (what counts as "unverifiable") is itself open, so scores from different benchmarks cannot be directly interchanged.

\textbf{Safety evaluation}: HarmBench [225] standardizes automated red-teaming, and WMDP [328] measures dangerous knowledge. The difficulty of safety evaluation lies in "coping with unknown attack surfaces" --- passing a benchmark does not mean that uncovered vulnerabilities do not exist. Safety evaluation should therefore be a continuous red-teaming loop rather than one-off scoring (Chapters 8 and 13).

\textbf{The two ends of jailbreak and over-refusal}: Safety evaluation must test both "whether it can be jailbroken" (false acceptance) and "whether it over-refuses legitimate requests" (false refusal). Testing only one end incentivizes degenerate solutions --- refusing everything or complying with everything. Sound safety evaluation should report both ends and provide a tradeoff curve (alignment tax; Chapter 8).

\textbf{The sustainability of red-teaming}: The attack surface evolves with capability and the deployment environment, so one-off red-teaming quickly expires. A continuous adversarial loop (discover---patch---re-evaluate) is therefore needed, along with recording attack samples to build a regression set [225]. This turns "safety" from a static metric into a process.

\textbf{Sources of bias}: Bias can come from data (historical injustice), objectives (optimizing a biased preference), and evaluation (skewed benchmark design). Locating the source of bias determines the mitigation: debiasing on the data side, changing the reward on the objective side, switching protocols on the evaluation side. Attributing all bias to a "model problem" leads to ineffective fixes.

\textbf{Limitations of bias measurement}: Bias is an attribute of groups under a dataset and social context, and a benchmark can only measure it by sampling. Low bias on a single benchmark does not equal system safety, and vice versa. Bias evaluation should serve as continuous monitoring rather than one-off certification.

\subsection{Reproducibility and Reporting Standards}

Recommended checklist: training/evaluation splits and deduplication logs; per-model parameter and token basis; hyperparameters and schedule details; contamination detection results; evaluation prompts and decoding settings; data provenance and license documentation [61][78]; and safety and bias regression results. Following this checklist does not guarantee comparability, but it can turn "comparing numbers across works" from guesswork into an auditable process.

\textbf{Anti-patterns in reporting}: reporting only the best single run, reporting only favorable benchmark layers, not reporting the sampling budget, not disclosing protocols, and treating model evaluation as human evaluation --- these are reproducibility defects that recur in the current literature, and readers should discount any "leading" conclusion accordingly.

\textbf{Minimal reporting template}: \textcircled{\small 1} model and data basis; \textcircled{\small 2} the protocol for each benchmark (prompt, few-shot, decoding, number of samples); \textcircled{\small 3} contamination detection and held-out data; \textcircled{\small 4} safety and bias regression; \textcircled{\small 5} compute and time. Missing any one item prevents readers from independently verifying. Appendices D and E provide templates.

\textbf{Versioning of evaluation}: Benchmarks, prompts, and evaluation scripts should all be versioned and recorded alongside the model/data version. Otherwise "the same evaluation" yields incomparable numbers at different points in time (benchmarks get updated, dependencies get upgraded). A prerequisite for reproducibility is that the "evaluation stack" is traceable.

\textbf{Automated reporting and meta-evaluation}: To reduce reporting costs, evaluation reports (benchmarks, protocols, results, contamination flags) can be generated automatically. But this requires a standardized evaluation stack; meta-evaluation (evaluating the quality of evaluation itself) is still scarce and is a leverage point for improving overall credibility.

\subsection{Representative Works Table}

\begingroup\small\setlength{\tabcolsep}{6pt}
\noindent\textbf{Table 10: Chapter 12 representative works---datasets and evaluation}\par\medskip
\begin{longtable}{@{}>{\raggedright\arraybackslash}p{0.1938\textwidth}>{\raggedright\arraybackslash}p{0.2087\textwidth}>{\raggedright\arraybackslash}p{0.1789\textwidth}>{\raggedright\arraybackslash}p{0.1193\textwidth}>{\raggedright\arraybackslash}p{0.1938\textwidth}@{}}
\toprule
Category & Representative work & Role & Evidence & Notes \\
\midrule
\endhead
General corpus & The Pile [54]\allowbreak{} /\allowbreak{} C4 [49]\allowbreak{}[56]\allowbreak{} & Pretraining data & A & Licensing issues \\
Open corpus & Dolma [84]\allowbreak{} & Auditable data & A & Limited coverage \\
Image-\allowbreak{}-\allowbreak{}text data & LAION-\allowbreak{}5B [64]\allowbreak{} /\allowbreak{} OBELISC [273]\allowbreak{} & Multimodal training & B & Filtering-\allowbreak{}dependent \\
Dataset documentation & Datasheets [61]\allowbreak{} & Reporting standard & A & Foundation of governance \\
Knowledge and reasoning & MMLU [7]\allowbreak{} /\allowbreak{} MMLU-\allowbreak{}Pro [381]\allowbreak{} & Capability measurement & A & Contamination present \\
Math/\allowbreak{}code & GSM8K [361]\allowbreak{} /\allowbreak{} HumanEval [357]\allowbreak{} & Reasoning/\allowbreak{}code & A & Protocol-\allowbreak{}sensitive \\
Dialogue human evaluation & MT-\allowbreak{}Bench /\allowbreak{} Arena [36]\allowbreak{} & Open-\allowbreak{}ended generation & B & Bias \\
Automated evaluation & AlpacaEval-\allowbreak{}LC [228]\allowbreak{} & Length debiasing & B & Proxy metric \\
Hallucination & TruthfulQA [359]\allowbreak{} /\allowbreak{} FActScore [367]\allowbreak{} & Truthfulness & B & Open definition \\
Safety & HarmBench [225]\allowbreak{} /\allowbreak{} WMDP [328]\allowbreak{} & Red-\allowbreak{}teaming/\allowbreak{}dangerous knowledge & B & Unknown attack surface \\
Contamination detection & [52]\allowbreak{}[90]\allowbreak{}[371]\allowbreak{}[382]\allowbreak{} & Leakage diagnosis & B & Differing assumptions \\
Holistic framework & HELM [22]\allowbreak{} & Multi-\allowbreak{}scenario matrix & A & Broad coverage \\
\bottomrule
\end{longtable}
\endgroup

\textbf{How to read the table}: This table places "data" and "evaluation" side by side because each constrains the other --- data set the ceiling on scores, and evaluation determines whether we can trust those scores.

\subsection{Key Controversies}

First, \textbf{whether benchmark scores represent capability}. Optimists approximate capability progress by benchmark progress; skeptics point out that contamination [83][375], protocol sensitivity, and "teaching to the test" inflate scores. The mainstream consensus is that benchmarks are a \textbf{necessary but not sufficient condition} and should be used together with held-out data and red-teaming (Section 12.6).

Second, \textbf{whether model evaluation can replace human evaluation}. Supporters emphasize its low cost and scalability [36]; opponents list its position/length/self-preference biases. A middle ground is "model evaluation for coarse screening, human evaluation for calibration, with debiasing such as length control [228]"; treating model evaluation as the final ground truth remains unreliable.

Third, \textbf{public leaderboards vs. private evaluations}. Public leaderboards are comparable but easily "targeted" (leaderboard gaming); private evaluations resist gaming but are non-reproducible and easily questioned. The ideal third way is \textbf{dynamic, reproducible, contamination-resistant benchmarks} ([379][383]), but their maturity remains limited.

\textbf{Distorted leaderboard incentives}: Public leaderboards induce "optimizing for the benchmark" (e.g., specially tuned prompts, format pandering), making leaderboard scores increasingly unrepresentative of general capability. Mitigations include dynamic benchmarks, private held-out sets, and multi-benchmark consensus; but for an individual team, the safest approach is a self-built evaluation close to deployment.

\textbf{Human vs. model evaluation revisited}: The two are not mutually exclusive --- using model evaluation for high-frequency regression and human evaluation for low-frequency calibration, and calibrating model evaluation with human preference data, can balance cost and credibility. The key is to state clearly "who the judges are and how they are calibrated," rather than claiming one is absolutely better [36][228].

\textbf{The reproducibility crisis of evaluation saturation}: When most models approach perfect scores on old benchmarks, discriminative power disappears and "evaluation" degenerates into "a statement of passing the line." This forces benchmarks to keep getting harder and pushes researchers toward "private/dynamic" evaluation --- but the latter sacrifices comparability. The tension between "comparability vs. discriminative power" in evaluation will persist for a long time.

\textbf{Open data vs. closed data}: Open data promotes reproducibility and fair comparison, but frontier teams tend to keep data closed to maintain an advantage; closed data increases confidentiality but harms scientific verifiability. This tension determines the range that a "public survey" can honestly cover --- and is also one reason this survey assigns a great many Grade C labels.

\textbf{The tension between standardization and diversity}: Standardized evaluation (a unified protocol) improves comparability but may stifle diverse evaluations tailored to specific scenarios; diverse evaluation is closer to real needs but hard to compare laterally. A pragmatic compromise is the coexistence of "a core standard set + scenario-specific extension sets."

\subsection{Practical Checklist and Open Problems}

Recommendations: \textbf{define evaluation before data and training}; report by layer (understanding/knowledge/reasoning/code/multimodal/safety/truthfulness) rather than a single composite score; trace the protocol and contamination of every cited score; use a continuous red-teaming loop for safety; and attach provenance, license, and deduplication records to data. \textbf{Open problems}: how to quantify the impact of contamination on capability conclusions (Problem 6); how to build contamination-resistant and reproducible dynamic benchmarks; how to make evaluation protocols comparable across works (compute/energy accounting standard, Problem 10).

\textbf{Cross-stage connections}: The data selection of Chapter 5 takes evaluation as its yardstick, and every change in Chapters 6--9 is accepted by evaluation; conversely, contamination in evaluation comes mainly from the data pipeline. "Data--evaluation" is thus a pair of systems that must be designed together, and the checklist in this chapter should be used throughout the survey (Figure 1).

\textbf{Interfaces with the other chapters}: The data selection of Chapter 5, the scaling decisions of Chapter 6, and every change in Chapters 7--9 are accepted by the evaluation of this chapter; and the contamination of this chapter originates from the data pipelines of the preceding chapters. Evaluation is therefore not "the last chapter" but a horizontal axis running through the entire survey.

\textbf{Advice for different readers}: Researchers should prioritize "protocol transparency + contamination disclosure"; engineers should prioritize "evaluation first + a deployment-aligned private set"; publishers should prioritize "data lineage + safety regression." The three roles have different evaluation responsibilities, but their shared baseline is "do not treat non-reproducible numbers as facts."

\textbf{An actionable evaluation research design}: Take the same set of models and report the same benchmark scores under three conditions --- "literal contamination detection vs. semantic contamination detection vs. no detection" --- to quantify the impact of contamination on conclusions. This kind of meta-study directly addresses Problem 6 and requires no training of new models.

\textbf{Refinement of open problems}: \textcircled{\small 1} How can reproducible, contamination-resistant dynamic benchmarks be constructed? \textcircled{\small 2} Can debiasing of model evaluation reach the calibration accuracy of human evaluation? \textcircled{\small 3} How can a layered protocol commensurate with quality be established for "evaluation cost"? \textcircled{\small 4} How can shared-source contamination of synthetic data be detected? Together, these four determine the credibility of the next generation of evaluation.

\textbf{Traffic-light interpretation}: A set of actionable interpretation rules --- green (safe to cite): protocol transparent + contamination checked + multiple benchmarks consistent; yellow (caution): single-benchmark lead, unclear protocol; red (do not cite): no protocol reported, suspected contamination, marketing claims only. This compresses the chapter's checklist into an immediately usable decision tool.

\textbf{Interfaces with the appendices}: The evaluation checklist of this chapter has corresponding templates in Appendix D (reproducibility checklist) and Appendix E (grading explanation); readers can use this chapter's "traffic-light interpretation" directly to screen the evidence entries in the appendices.

\textbf{Mapping to the six-dimensional framework}: data (training set/benchmarks/calibration set), objective (evaluation metrics are the optimization objective), optimization (evaluation-driven selection and early stopping), system (evaluation cost and throughput), evaluation (self-referential), and risk (contamination, bias, overfitting to leaderboards). Among the six dimensions, the "evaluation" dimension of this chapter is both object and tool, and this self-referentiality is precisely the root of why evaluation is easily manipulated.

\textbf{Summary of the evidence distribution in this chapter}: The methodology of evaluation (contamination detection, debiasing, documentation standards) already has a considerable body of Grade A/B literature; but "how much real capability a score represents" and "how to make it comparable across works" remain dominated by opinions and single-point methods. What this chapter provides is thus a \textbf{normative framework and checklist for interpreting evaluation}, rather than a conclusion about "which model is better" --- consistent with the survey-wide positioning of "providing a framework rather than a ranking."

\section{Practical Guide}

\begin{quote}\small
This chapter is a pilot chapter, with a target of 8,000--9,000 words. It translates the mechanistic conclusions of the preceding text (Ch5--Ch12) into an "executable checklist," but \textbf{provides no magic numbers}: any quantitative recommendation that cannot be synchronously verified with public evidence is labeled "this team's prior" rather than a conclusion. All citations reuse the core set of the preceding text; no new evidence is introduced here.
\end{quote}

\subsection{The Decision Tree (Verbalized Form of Figure 4)}

\textcircled{\small 1} First define the constraints (budget/modality/context length/safety compliance) $\rightarrow$ \textcircled{\small 2} budget determines scale: use scaling laws to estimate $(N, D)$ [16][119] $\rightarrow$ \textcircled{\small 3} modality determines architecture and data: choose decoder-only for pure text; choose encoder--projector--LLM or cross-attention for multimodality [17][28] $\rightarrow$ \textcircled{\small 4} long-text requirements determine whether long-context training/extrapolation is needed [37][143] $\rightarrow$ \textcircled{\small 5} safety compliance determines the post-training recipe: SFT first, then preference optimization (DPO family or RL) [32][45] $\rightarrow$ \textcircled{\small 6} compress or adapt to the domain before deployment, then re-check alignment and safety. Each step gives tradeoffs rather than promised metrics.

\begin{figure}[htbp]
\centering
\includegraphics[height=0.85\textheight,keepaspectratio]{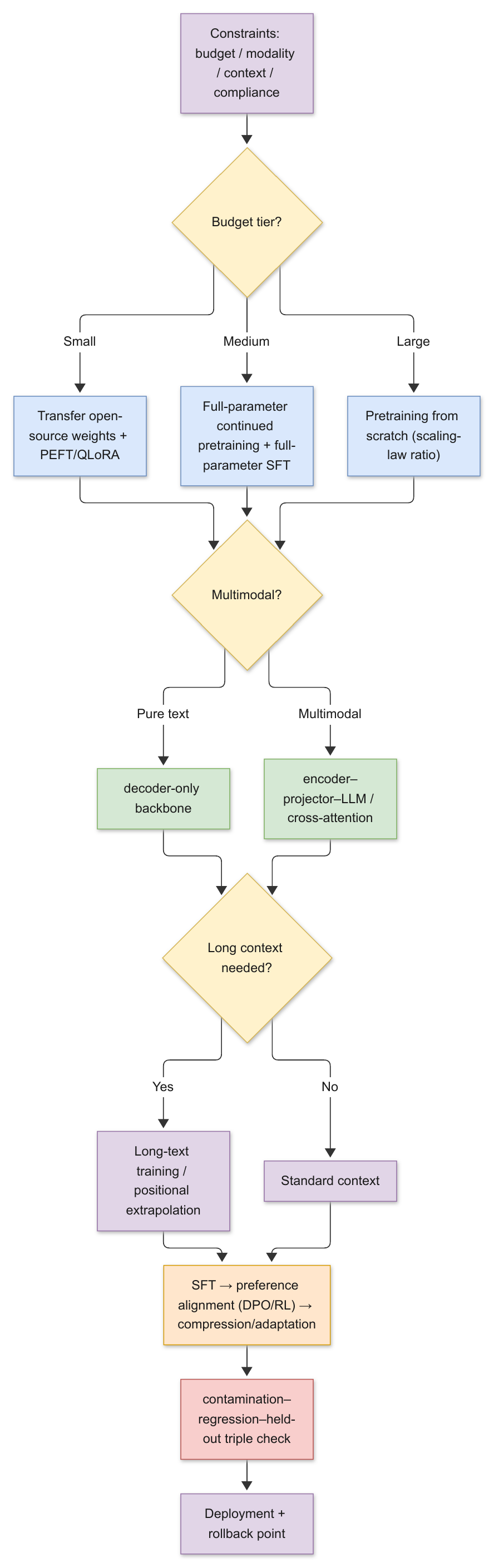}
\caption*{Figure 4: Practical decision tree---budget $\rightarrow$ scale $\rightarrow$ modality $\rightarrow$ long context $\rightarrow$ post-training recipe $\rightarrow$ checks and rollback}
\end{figure}

\textbf{Making the decision tree falsifiable}: Every one of the six steps above should leave a "verification point" rather than a "just-do-it point"---after \textcircled{\small 2}, run a small-scale scaling-law extrapolation to verify the curve; after \textcircled{\small 3}, do one round of modality acceptance using visual samples; after \textcircled{\small 5}, run the "pre-alignment" evaluation of Chapter 12. The value of the decision tree is not "choose by following it," but "turning each step's choice into a retraceable statement"; this is precisely the concrete form of the "capability--behavior--evidence" three-layer scheme of Chapter 11.

\textbf{Two default lines of defense}: First, every branch first performs the "contamination--regression--held-out" triple check (Chapter 12), because these are the cheapest sources of false positives; second, every branch ends with a "rollback point"---when training is interrupted, the recipe diverges, or evaluation collapses, one can return to the previous verifiable state rather than guessing from logs.

\textbf{Making the unimodal assumption explicit}: The default branch of the decision tree is a pure-text decoder (Section 6.1). Once vision/audio is added, the tree forks at \textcircled{\small 3} into "encoder--projector" or "unified token," and introduces new knobs such as projector/freezing/resolution (Sections 10.1, 10.4). Therefore, the correct way to read the decision tree is: follow the left branch (pure text) by default, and take the right branch only when modality capability is needed, and each knob of the right branch must be re-checked at \textcircled{\small 6}, because changes on the modality side are the most easily and silently overwritten during post-training.

\textbf{Failure is data}: Any "wrong branch" taken in the decision tree (excessive extrapolation, incorrect ratio) should be recorded as a new row in the failure-mode list of Section 13.3. Treating failure as garbage means paying tuition repeatedly; treating failure as data makes the decision tree a little more accurate with each iteration. This is isomorphic to the "evaluation-stack versioning" of Chapter 12.

\textbf{From-scratch vs. transfer branch}: Between \textcircled{\small 2} and \textcircled{\small 3} the decision tree hides another major node---cold start (pretraining from scratch) or transfer (continuing from open-source/closed-source weights). The two paths impose different constraints on budget, control, and compliance (Section 11.10): transfer is fast but bound by upstream licensing and structural constraints, while cold start is expensive but fully autonomous in data/architecture. The decision tree should write the tradeoffs of this path clearly at the specification stage, rather than being forced to change course after training fails.

\subsection{Budget Estimation Framework}

Adopt order-of-magnitude estimation rather than fabricating numbers: autoregressive training FLOPs $\approx$ 6\textperiodcentered{}N\textperiodcentered{}D (Section 3.4); infer the throughput target backward from the available GPUs and target wall-clock time, then choose the parallelism strategy (data/tensor/pipeline/ZeRO [6]). Under the same N and D, multimodality and long context have higher budgets due to sequences and visual tokens, and must be estimated separately. No specific numbers that cannot be verified are given.

\textbf{Layering of the back-calculation formula}: The upper bound (6\textperiodcentered{}N\textperiodcentered{}D) gives the ideal cost, which in practice must be divided by "MFU" (model utilization); MFU depends on the parallelism strategy, communication, and kernel efficiency. Therefore, the correct use of the budget framework is: first set the target wall-clock, then back-calculate the required FLOPs/s from MFU, then back-calculate N and D from the model and data scale. Use the upper bound when only an "order of magnitude" is required; when "scheduling" must be done, measured MFU must be introduced---the latter belongs to "this team's prior."

\textbf{Hidden amplification of the post-training budget}: The actual cost of the alignment/evaluation stage is "cost $\times$ number of attempts" (Section 11.9). Therefore, the budget framework must reserve 2--3$\times$ headroom for the grid search of SFT/preference/RL and for re-testing of the evaluation stack; otherwise the schedule will slip. This point is repeatedly emphasized in all pilot chapters of this survey and is one of the most frequent causes of engineering failure (Section 13.3).

\textbf{A budget is an "order of magnitude," not a "commitment"}: All budget figures should be written as "order-of-magnitude ranges" (e.g., 1e21--1e22 FLOPs) rather than a single committed value. The only honest reason is that it depends on variable factors such as MFU and the data-cleaning hit rate. Committing to a budget as a range amounts to admitting uncertainty in advance, whereas committing to a single value merely defers that uncertainty until it is exposed at the first evaluation gate.

\textbf{Realistic constraints of the cluster}: The budget framework must accept that "mathematical optimality is not parallelism optimality"---when the cluster has only 8 GPUs and limited inter-node bandwidth, the optimal combination of data-parallel partitioning and pipeline depth deviates from the ideal degree of parallelism (Section 6.7). Therefore, measuring the MFU and communication overhead of a small task first and then setting the scale accordingly is more robust than fixing the scale first and then forcing it through a queue. \textbf{Cluster shape is a prerequisite input to the budget framework, not a secondary customer-service parameter.}

\textbf{The metric price of tokens/hour}: The single most useful unit in a budget report is "tokens/hour + the corresponding MFU," because it encodes both throughput and utilization. All ramp-up runs, ratio A/B comparisons, and SFT rounds are counted in this unit; from it one computes the "cost per round," then converts to the units of the "exchange-rate table" of Section 11.6, so that team decisions have a common yardstick. A budget meeting without a common yardstick degenerates into a scramble for resources.

\textbf{Ramp-up and stop-loss}: Besides the "planned value," the budget table must have two more rows---the "verified value" (small-scale measurement) and the "stop-loss point" (when to halt the direction). This is the meaning of ramp-up: it is not scaling without analysis, but splitting a large investment into a number of stoppable small experiments; if the data still does not support continuing after the stop-loss point, roll back to the previous verified state (Section 13.11).

\subsection{Failure-Mode Checklist}

High-frequency failure modes: data contamination not cleaned, leading to inflated benchmark scores; inconsistent SFT templates, leading to format leakage; length bias in preference optimization [227]; reward hacking [408]; over-refusal (alignment tax); MoE routing imbalance triggering loss spikes; long-context extrapolation yielding no gain on real long texts; editing a few models triggering collapse [324]; degradation of safety behavior after quantization. Countermeasures: establish pre-evaluation and rollback points for each category.

\textbf{Grouped countermeasures for failure modes}: Equip the nine failure categories with a three-step "detect--isolate--recover" scheme: contamination$\rightarrow$held-out warning; template leakage$\rightarrow$two-sided protocol regression; length bias$\rightarrow$length baseline and length-capped evaluation; reward hacking$\rightarrow$out-of-distribution regression; over-refusal$\rightarrow$dual benchmarks for safety/use-cases; MoE spike$\rightarrow$routing statistics monitoring; no long-text gain$\rightarrow$real long-text held-out; editing collapse$\rightarrow$rollback to checkpoint; quantization degradation$\rightarrow$post-quantization safety re-test. Three to five tenths belong to "detectable by evaluation in advance" (green list), and the rest belong to "detected only by runtime monitoring" (yellow list).

\textbf{Priority of countermeasures}: The green list (detectable by evaluation in advance) should be the default "enabled at first launch," because it has the lowest cost; the yellow list (exposed only at runtime) should enter the monitoring dashboard and be paired with automatic alerts. Assigning attention to each failure category by the four quadrants of "detection cost $\rightarrow$ repair cost" is more consistent with budget discipline than "fixing whatever you see" (marginal returns of Section 11.6).

\textbf{Two misconceptions about failure}: Misconception one---treating failure as a "personal error" and attributing it to people while ignoring systemic causes (data, recipe, interface); the correct approach is to attribute failure to the "stage pipeline," because most failures are reproducible across people. Misconception two---treating failure as "avoidable," leading to the use of excessive checks to guard against low-probability events; the correct approach is to fix high-frequency failures first according to 80/20 and keep a manual gate for low-frequency failures. Both misconceptions stem from "failure = should not happen," whereas the real philosophy is "failure = expected, and there must be a path to restoration."

\textbf{The same ledger for success and failure}: Keep accounts for "success" as well---which ratio, which evaluation gate passed---in the same table as the failure list. There are two reasons: first, the precondition for the reproducibility of success is that the "recipe stamp" is sufficiently complete; second, the attribution of success is equally susceptible to contamination by the "common-method effect" (Section 11.16). Recording only failures and not successes makes the team pay repeatedly for duplicate verification.

\subsection{Configuration Recommendations}

Small budget: QLoRA [30] fine-tuning + DPO [32]/SimPO [43] alignment + training-free quantization [21][34]. Medium budget: full-parameter SFT + GRPO/RLVR [45][224] + merging/distillation. Large budget: run the ratio and filtering scaling laws [87][86], then train according to the Chinchilla ratio [16]. The configuration recommendations are all annotated based on evidence grading, progressively adding system and stability measures as the budget grows.

\textbf{The common discipline of the three budget tiers}: Regardless of the configuration tier, the three elements of "evaluation first + rollback point + contamination held-out" cannot be omitted---they do not shrink with the budget, because they are the only meters by which you judge whether the budget is well spent (Chapter 12). The difference between configuration tiers lies only in the "concentration of measures" (full-parameter vs. quantized, simple DPO vs. RL depth), not in the "concentration of discipline."

\textbf{Boundaries of the tier recommendations}: All three tier recommendations are built on the premise of "existing reusable weights and evaluation stack"; if one starts from scratch in all three respects---data, tools, and personnel---the lowest cost is not a "small-budget configuration" but "first renting a verified mid-tier cloud service to practice the paradigm." Including the fixed cost of "from scratch to a verified platform" in the budget makes the intuition that "small = cheap" distorted.

\textbf{Budget discipline for hyperparameter sweeps}: The hyperparameters of alignment/sequential decision-making (learning rate, DPO $\beta$, RL temperature, etc.) appear to require "grid search," but in fact the first few rounds already exceed the budget. The discipline is: first use a small model to define a coarse grid (corners + center), then densify once in a small range around the best coarse point; each sweep must use the same evaluation anchor, otherwise scores under different hyperparameters are not comparable (Section 12.9). \textbf{Hyperparameters are part of the budget, not a zero-cost free knob.}

\subsection{Stage 0: Requirements and Constraints Checklist}

Before starting work, first write a "Stage 0 specification": deployment goals (three categories of metrics: capability/behavior/safety), budget ceiling (compute + time + personnel), modality and environment constraints, acceptable recall and latency, compliance boundaries (data licensing, deployment region). The specification must be able to answer "what counts as success and what counts as failure," and must give a foothold to the "capability--behavior--evidence" decomposition of Chapter 11.

\textbf{Three categories of signatories of the specification}: Business (defines success and boundaries), engineering (defines cost and risk), and evaluation (defines the acceptance protocol) each sign once; if any is missing, work should not begin. Moving the "sign-off point" up to Stage 0 is an order of magnitude cheaper than discovering a mismatch in scope only at delivery---this is the part of this survey that can be directly converted into institutional practice.

\textbf{Making requirements measurable}: Every goal in the specification must carry the suffix "how to measure it," otherwise it is merely a wish. Example: "answers more safely" should be written as "jailbreak success rate < X\% on a private safety set (Section 8.7)"; "understands documents better" should be written as "accuracy $\geq$ Y on document-understanding benchmark B (Section 10.9)." The measurement suffix connects directly to the three sets of anchors of Chapter 12 (capability/behavior/safety).

\textbf{Three categories of license checklist for compliance}: Stage 0 should also go through the data/model licensing triangle---training-data license (whether training/commercial use is permitted), weight license (whether secondary development/distribution is permitted), and deployment license (regional and industry restrictions, Sections 5.5, 12.3). The three each occupy one line in the specification; wherever licensing conflicts with budget, it should be downgraded before work begins to "a risk already evaluated by this team," rather than left for arbitration at release time.

\subsection{Data Pipeline Checklist}

The pretraining data pipeline is executed according to Chapter 5: collection$\rightarrow$deduplication$\rightarrow$filtering$\rightarrow$mixing$\rightarrow$tokenization$\rightarrow$held-out. The checklist includes: recording the timeline and provenance; performing two levels of deduplication, n-gram and semantic; setting dedicated held-out sets for contamination-sensitive categories (code, exams, news); setting the mixture by deployment language and domain; and confirming with validation perplexity and downstream sampling after tokenization. The statistical items of the data card are taken from Chapter 12.

\textbf{Data snapshot}: Any training must freeze a "data snapshot" (version + hash), otherwise reproducibility and contamination auditing (Chapter 12) cannot even be discussed. Snapshots are also used for cross-stage interface contracts: SFT data and preference data reference the same set of provenance records, so that error attribution can be traced to the data layer.

\textbf{The tunable entry point for mixing}: The data mixture (language/domain/code proportions) is the only knob that can change the capability distribution "without adding training budget" (Sections 5.8, 6.5). In practice, the mixture should first be A/B compared at small scale (one version each for A/B, same budget and same evaluation), then scaled up; the A/B results are brought into the "exchange-rate table" of Section 11.6 as this team's prior values. \textbf{The mixture is not intuition; it is a parameter that can be updated by experiment.}

\textbf{Entry threshold for synthetic data}: Synthetic data can scale up, but carries the side effects of "same-source bias" and quality risk (Sections 5.9, 12.6). The entry threshold is three checks: the accuracy of synthetic samples on a real-distribution held-out set, their distributional overlap with real samples, and contamination sensitivity (whether the synthesizer includes the evaluation in training). If any of the three fails, the volume advantage of synthetic data will be offset by its systematic bias.

\textbf{The budget of deduplication and filtering}: Deduplication/filtering in the data pipeline is not a "one-time free step"---its occupancy of CPU and disk, as well as the labor hours of manual spot-checking, must be included in the budget. Ranking by cost-effectiveness: n-gram deduplication (cheap, mandatory) $\rightarrow$ semantic deduplication (medium, as needed) $\rightarrow$ manual quality inspection (expensive, sampled). When the budget is insufficient, prioritize keeping n-gram deduplication and the held-out sets; semantic deduplication/quality inspection may be reduced in frequency but must not be cut entirely (Sections 5.4, 12).

\subsection{Pretraining and Scale-Decision Checklist}

Scale decisions follow Chapter 6: determine the $(N, D)$ curve by small-scale extrapolation, then follow the Chinchilla ratio; the mixture design covers language/domain/code; the parallelism strategy is chosen by budget and cluster shape; training monitors loss, gradient norm, and routing statistics; finally, a set of downstream sampling is used to evaluate where the "capability wall" lies. Long-text requirements are evaluated separately (Section 9.5).

\textbf{Early-warning indicators of the "capability wall"}: When scaling extrapolation predictions begin to diverge from measurements (the curve plateaus faster than extrapolation), or downstream sampling enters a plateau, one should judge that the capability wall has been reached and promptly decide whether to "push the wall" (add data/scale) or "go around the wall" (add post-training/compression), rather than running to the end by brute force (Section 11.1).

\textbf{Standard protocol for small-scale extrapolation}: The minimum three elements required for extrapolation---the same data stream (sampled from the same distribution), the same evaluation anchor (Section 11.6), and at least three scale points (e.g., 60M/200M/600M). If any of the three is missing, the extrapolation curve is untrustworthy; the honest approach is to report the deviation range between extrapolation and measurement simultaneously, rather than drawing a smooth fitted line (evidence grading of Section 6.3).

\textbf{Additional budget items for long context and multimodality}: If the specification includes long context or visual input, the budget framework must list two additional items---the long-sequence FLOPs and attention cost of long-text training (Section 9.5), and the additional attention and KV cache for visual tokens (Section 10.4). These two items are often underestimated several-fold by the "equivalent text token count"; listing them explicitly in the budget table of 13.2 is the minimal action to keep multimodal/long-text projects from blowing their budgets.

\textbf{Metrics of the monitoring dashboard}: The minimal set for pretraining/post-training monitoring: loss, gradient norm, routing statistics, and evaluation-anchor sampling (every N steps). These metrics need not all be on the canvas---routing and loss go to alerts, gradient norm and sampling go to logs; the key is to map "change" to "effect": which step's modification corresponds to which metric's drift---this is precisely why the "evaluation gate" of Section 13.11 can open in advance during training.

\textbf{Reconciling training cost and evaluation cost}: Most budgets count only training (FLOPs) and ignore the continuous occupancy of the evaluation stack (GPU for each anchor sampling, reviewer hours). Only by combining "training cost $\times$ number of attempts + evaluation cost $\times$ number of evaluations" does one obtain "the true cost of ownership of the scheme" (Section 11.9). The clearer the reconciliation, the better one can rebut the optimistic board that says "the estimate is only 10\% off"---because that 10\% is often hidden in the number of evaluations.

\subsection{Post-Training Checklist (SFT/Preference/RL)}

According to Chapters 7--8: SFT first passes the data-quality gate (consistency, deduplication, format); preference data clearly annotates provenance and judges (human/model); choose the depth of preference/RL according to reward verifiability (Section 11.4); before any alignment change, run a "pre-alignment evaluation," and after the change, run an "alignment regression." Cost reduction and safety matters are recorded together.

\textbf{Version stamp of alignment}: Each SFT/preference/RL round must leave a "recipe stamp" (data, hyperparameters, seed, evaluation results), forming an alignment timeline. The timeline makes "which change introduced which behavior" traceable, and directly supports the contamination and regression auditing of Chapter 12---an alignment process without a recipe stamp amounts to an unauditable change log.

\textbf{Auditing obligations for preference data}: When preference data (human- or model-annotated) enters the recipe, it must answer: who annotated it, with what protocol, and whether it is same-source as the training data (to prevent self-preference contamination). This is more important than "the more preference data the better"---the evidence of Section 8.4 supports "quality and consistency first." The audit results are written into the recipe stamp and archived together with the weights.

\textbf{Concretizing the SFT data-quality gate}: The SFT data-quality gate has at least three levels: format consistency (unified template/fields), content deduplication (n-gram + semantic), and instruction diversity (proportion of repeated templates). All three gates must be given a "dry-run delta"---use a fixed set of small models to run A/B on the same data, to see whether changes in the gate value bring observable behavioral differences; if there is no difference, the gate is ineffective before it and should be replaced with a more sensitive gate (Chapter 7).

\textbf{The SFT$\rightarrow$preference$\rightarrow$RL handover protocol}: At handover, give the next stage three things---baseline evaluation (capability/behavior/safety), the recipe stamp (data + hyperparameters + seed), and a list of "pits already stepped in" (which type of data triggers which type of failure). A stage transition without a handover protocol often loses all the implicit constraints of the previous stage, letting RL continue to amplify the same type of error in SFT's blind spots (Section 8.10).

\subsection{Pre-Deployment Checks and Continual Optimization}

Before deployment, perform three "frozen evaluations": capability, (alignment) behavior, and safety, each retaining a baseline; then run compression/quantization and re-test the three regressions (Chapter 9). In the continual-optimization stage, every incremental change goes through the same regression gate; when introducing online feedback, configure "control + held-out" according to Section 11.13.

\textbf{Rollback switch}: Another role of the three frozen baselines is to provide rollback targets---if a new version is found to degrade after deployment, one can quickly switch back to the previous "evaluation-passing" version. Rollback is not failure but part of discipline; continual optimization without a rollback switch is equivalent to leaving oneself no way back (Section 11.13).

\textbf{Acceptance boundaries of compression}: The acceptance of compression (quantization/pruning/distillation) cannot look only at throughput and perplexity; it must check whether the three sets of numbers---capability (benchmark), behavior (alignment regression), and safety (attack surface)---are each still within tolerance (Chapter 9). If only one of the three is within tolerance, claiming "lossless compression" means narrowing the acceptance---the release check of Section 13.10 will intercept such a claim.

\textbf{Entry threshold for online feedback}: The threshold for introducing online/closed-loop (Section 11.13) is: \textcircled{\small 1} the feedback pipeline has a "control" (random traffic split for comparison); \textcircled{\small 2} the retraining data has undergone same-source contamination checks; \textcircled{\small 3} a switch to roll back to the old version is available. Only when all three are in place is it worth connecting the loop to production---otherwise the real-time nature of online only accelerates the spread of errors rather than accelerating learning.

\textbf{Cadence of continual optimization}: Continual optimization is not "changing every day" but "changing with a cadence"---the round frequency depends on the dual signals of data feedback (emergence of new requirements) and evaluation feedback (regression achieved). Unpaced infinite fine-tuning is a breeding ground for over-optimization and drift (Section 11.13): every retraining must go through the same evaluation gate and use "control splitting" to prove that the net benefit of the change is positive, rather than the illusion that "it is better just because it was changed." \textbf{If no new requirement has appeared for two weeks, one should turn off the continual-optimization sentinel rather than existing for existence's sake.}

\textbf{Rolling decisions for the online/offline combination}: Production often runs multiple versions simultaneously (offline main version + online experimental version), using traffic bucketing for comparison (Section 13.13). Rolling decision-making is a series of small-step advances of "evaluation gate + business metrics": each bucket passes the gate before traffic is scaled up, otherwise it is rolled back. The value of rolling decision-making is that it turns "release" from \textbf{a single bet} into \textbf{the accumulation of a series of verified small increments}.

\subsection{Evaluation and Release Checklist}

Evaluation follows Chapter 12: the protocol (prompt/decoding/sampling count) is recorded; contamination held-out + in-depth detection on some benchmarks; the report includes context, safety, and bias regressions; numbers are versioned. At release, attach the data card, evaluation card, and reproducibility statement (Appendices D, E).

\textbf{Three lines of the release check}: \textcircled{\small 1} falsifiable (provide sufficient protocol/seed so that opponents can reproduce); \textcircled{\small 2} auditable (data lineage + recipe stamp + evaluation timeline); \textcircled{\small 3} rollback-capable (retain the previous release baseline). Only when all three lines pass is it advisable to claim externally that "we did better"; otherwise all performance claims are downgraded to "this team's internal conclusion" (Chapter 3).

\textbf{One more reference to the minimal report template}: The release report must at least contain the model and data scope, the protocol for each benchmark, contamination detection and held-out, safety and bias regressions, and compute and time (Section 12.11). These five sections are not an "optional appendix" but the five keys that give readers the ability to re-check you; if any one is missing, your conclusion cannot be replayed by others---no matter how successful it is claimed internally.

\textbf{Release materials for privacy and red-teaming}: If the deployment surface involves privacy and security, the release materials should also contain a privacy impact statement (data-collection and deletion channels) and a red-team summary (coverage, residual risk, ongoing plan, Sections 8.7, 12.7). These materials must themselves be "versioned"---they update with model versions, otherwise old gaps will be mistaken for the current state. Release is not written once but is a record line that evolves with versions.

\textbf{Semantic traps in reported numbers}: When writing reports, beware of three types of number expressions that "look honest but are misleading": \textcircled{\small 1} reporting only relative gains without mentioning the baseline; \textcircled{\small 2} benchmark scores transparent but one's own goals not transparent; \textcircled{\small 3} "average improvement" masking variance differences. The countermeasure is the reverse application of the "traffic-light interpretation" of Chapter 12---the releasing party proactively provides the protocol, intervals, and exclusion notes. Truthfulness and persuasiveness are not contradictory; on the contrary.

\subsection{Timeline and Milestones}

Break the overall task into milestones with an ordering and checkpoints: M0 specification and baseline evaluation $\rightarrow$ M1 data snapshot and scale extrapolation $\rightarrow$ M2 pretraining/continued pretraining $\rightarrow$ M3 SFT and preference/RL $\rightarrow$ M4 compression/pre-deployment evaluation $\rightarrow$ M5 release and continual optimization. Each milestone ends with an "evaluation gate," not with "training complete."

\textbf{Common design of the evaluation gate}: Each gate needs only three questions---"Is the capability online? (Chapter 6) Is the behavior compliant? (Chapter 8) Is the evidence credible? (Chapter 12)." If any of the three is missing, the gate does not pass; better to stop here than to advance; advancing a milestone that has not passed the gate is the most costly of all subsequent troubleshooting costs (Section 11.7).

\textbf{Parallelism and serialization of milestones}: M0$\rightarrow$M1 is serial (the specification precedes everything), after which M2/M3/M4 can advance in parallel (data and alignment trial-run simultaneously, pre-deployment evaluation and compression in parallel). Serialization exists only at "dependency" points; parallelism the rest of the time is key to saving money; but at each parallel handover, alignment must be made through an "anchor evaluation," otherwise parallelism amounts to each going its own way.

\textbf{Discipline for rescheduling milestones}: If a gate fails twice in a row, it is permissible and indeed advisable to reschedule the milestones (rather than hastily waving them through)---rescheduling must answer three questions: is the gate failure "a problem with the evaluation," "excessive ambition," or "immature implementation"? The three countermeasures are, respectively, fixing the evaluation, lowering the specification, and adding budget. Institutionalizing "rescheduling" bleeds far less than letting it happen quietly (the re-engagement point for the three categories of signatories of Section 13.5).

\subsection{Example Configurations (Three Tiers for Inspiration Only)}

Small-budget example: based on open-source weights (LLaMA/Qwen et al. [9]), QLoRA [30] + DPO [32], evaluated with mid-sized benchmarks + a private held-out set. Medium-budget example: full-parameter continued pretraining (small research scale) + full-parameter SFT + GRPO [45][224]. Large-budget example: pretraining from scratch + ratio scaling laws [87][86] + Chinchilla [16] + full-stack alignment and compression. \textbf{The "numbers" in each tier are only scenario illustrations, not recommended values}---the actual values must return to your own Stage 0 specification.

\textbf{Mixed tier beyond the three tiers}: Most real projects fall between "medium" and "large": using an open-source large model as the base (controlling the knowledge wall), with mid-scale domain continued pretraining + full-parameter SFT + DPO/RL combination. The value of the mixed tier is that it obtains both "mature base" and "controllable recipe," at the cost of being constrained by upstream licensing (Section 13.5). This survey does not rank the tiers but requires that any tier count the "continued-training and full-stack cost" together, avoiding "seeing only the cheap, not the total account."

\textbf{Supplement to the multimodal tiers}: The difference between multimodal tiers lies not in the "measures" but in "freezing scope and resolution budget" (Sections 10.1, 10.4): small budget freezes the encoder + trains the projector, medium budget unfreezes the projector and pre-aligns at low resolution, large budget unfreezes the encoder + high resolution. The higher the tier, the more compute must be reserved for "the attention cost of image tokens" (Section 10.4), and each tier requires multimodal-specific regressions (hallucination/grounding/OCR).

\textbf{Supplement to multilingual configuration}: For non-English-dominant deployments, the specification must explicitly set multilingual proportions in three links---data mixture (Section 5.8), preference data (Chapter 8), and evaluation anchors (Chapter 12). If a single sentence is missing, the model will quietly degrade into "English-first, other languages following." Multilinguality is not an "extra bonus item" but a configuration line that must be written by default in the specification.

\subsection{Tool and Open-Source Stack Pointers}

List the recommended tools scattered across the chapters: training frameworks and parallelism (the ZeRO [6] series), quantization ([21][34]), PEFT (QLoRA [30]), alignment (DPO [32]/SimPO [43]/GRPO [45][224]), compression (quantization/distillation/editing [324]), and evaluation and contamination detection (Section 12.9). The release scope of the list extends only to "methods and representative implementations"; specific project versions follow the reader's contemporaneous ecosystem (this survey is responsible only for mechanisms, not for version commitments).

\textbf{Steps for building the evaluation stack}: Building the evaluation stack (Section 12.10) is threshold-free work that precedes training---select benchmarks (three sets: capability/behavior/safety), write the protocol (prompt/decoding/sampling count), make held-out sets (the base set for contamination detection), and then freeze a "baseline snapshot." It should be one of the project's earliest outputs, because every subsequent milestone (Section 13.11) compares against the same baseline each time; building the baseline late amounts to leaving all early conclusions unanchored.

\textbf{Boundaries of tool selection}: More tools are not better---every tool introduced also introduces a set of its evaluation criteria (the "evaluation-stack versioning" of Chapter 12). This survey has only two recommendations for selection: first, "interface contract" first (the tool can align with the standards of data snapshots and recipe stamps); second, "evaluation-stack stability" first (a tool with a stable evaluation protocol is more important than one with more features but changing protocols). The rest is left to the reader's contemporaneous ecosystem to judge.

\subsection{Team and Collaboration}

A training project is a relay of four roles: "data--training--evaluation--safety": data is responsible for snapshots and lineage, training for the recipe stamp, evaluation for the baseline gate, and safety for attack-surface regression. Each of the four roles is responsible for the tables of Chapter 12; a change by any one introduces a regression. Factory-style collaboration usually fails at "interface distortion"---this is also the team version of the "schema interoperability" of Section 11.10.

\textbf{The four-sentence RACI mantra}: Data is responsible for "where it came from and when it is snapshotted"; training is responsible for "what recipe was used and when it is re-run"; evaluation is responsible for "what the baseline is and whether the gate was passed"; safety is responsible for "where the attack surface is and whether the regression is green." If any change cannot clearly state "who is responsible for its regression," it should not enter the milestone---this moves the release check of Section 13.10 forward to the collaboration layer.

\textbf{Fork discipline for open-source collaboration}: If the project contains open-source components (Section 13.13), the team should internally observe "fork discipline"---source code and configuration each versioned, upstream protocols unmodified, patches either flowed back upstream or local differences explicitly marked. The reason is not dogma but auditability: without fork discipline, the snapshots of training and evaluation cannot be traced to "which version of the code was used" (versioning of Section 12.10).

\textbf{Knowledge management and documentation}: Even a four-person small project should hand over process documentation (data, recipe, evaluation gates)---not written for outsiders but for oneself three months later. The cheapest form is "keep a markdown snapshot at each evaluation gate," and the most expensive is always verbal transmission. The point of documentation is not length but "replayability": every conclusion can be restored from the documentation to its inputs, outputs, and judgments at the time (Section 13.10).

\subsection{Closing with the Whole Survey}

This chapter is not an "operation manual" but a bridge that "turns the conclusions of Ch5--Ch12 into checkpoints": the data chapter provides the snapshot checklist, the scale chapter the extrapolation budget, the alignment chapter the recipe stamp, the evaluation chapter the baseline gate, and the continual chapter the rollback switch. If readers read only one chapter before deployment, read this one; but to truly understand why each checkpoint exists, one must return to the corresponding chapter to examine the evidence and cost---because behind every checklist item is a falsifiable mechanistic hypothesis.

\textbf{The four most common questions}: \textcircled{\small 1} "Can we do it without enough budget?"---Yes, but the three minimum disciplines of the three tiers (evaluation first + rollback point + contamination held-out) cannot be omitted; omitting them amounts to outsourcing uncertainty to the release stage; \textcircled{\small 2} "Is it fine to just use someone else's open model directly?"---Transfer is a reasonable default in most scenarios, but one must first go through the licensing and coordination of Section 13.5; \textcircled{\small 3} "Which failure should be fixed first?"---Prioritize the green list of Section 13.3; what evaluation can intercept is the cheapest; \textcircled{\small 4} "When should we retrain?"---The trigger conditions of Section 13.8: improvement saturation + inconsistency with the capability-wall hypothesis. These four can serve as lessons from other stones for this chapter; the real answer always returns to your Stage 0 specification.

\textbf{A final reminder}: In a practical guide, the most "correct" things are often "proven-effective defaults," and the most dangerous are often "pruning that looks clever." The position of this survey: may you not lack astonishing ideas, but lack a stable skeleton---and the skeleton is precisely these unremarkable checkpoints of this chapter. They do not guarantee success, but they guarantee that failure is attributable, rollback-capable, and learnable (Chapter 1).

\textbf{The last mile between this chapter and the appendices}: If readers wish to start hands-on, take away two kinds of checklists: \textcircled{\small 1} the checkpoint checklist of this chapter (Stage 0 specification, evaluation stack, recipe stamp, baseline gate, rollback switch); \textcircled{\small 2} the templates of Appendices D/E and the reporting checklist of Section 12.11. Putting the checklist into a local repository is enough to start work; this chapter never asks you to remember practices, only that you use "evaluation gates" and "snapshots" to turn every step into an auditable record (Section 13.10).

\textbf{A final sentence for all budget tiers}: No matter how small the tier or how few the measures, the success or failure of a training project is ultimately determined by "whether attribution can be done auditably" (Sections 13.10, 13.11). A training run with an even tighter budget, as long as it leaves a traceable sub-baseline, can convert its lessons into gains in the next iteration; conversely, an ample budget without traceability will only make failure replicate more expensively. This sentence is both the master switch of the checkpoints in this chapter and the common practical landing point of all chapters throughout the survey.

\section{Challenges and Open Problems}

Conclusion first: the most intractable challenge in this field is not the performance of any single technical point, but the gap between "strength of evidence" and "decision-making needs"---a large number of conclusions treated as common knowledge still rest on a single technical report or on scale experiments that cannot be reproduced. This chapter surveys the challenges along four lines---technique, data, compute and engineering, and ethics and safety---and distills them into 15 verifiable open problems (numbered consistently with the problem pool of Stage 2; the complete pool is given in Appendix F).

\subsection{Technical Challenges}

\textbf{Joint allocation of the scaling law and the post-training budget} is the foremost unsolved problem. Kaplan [119] and Chinchilla [16] provide only the optimal $(N, D)$ trade-off "internal to pretraining"; today, as inference-time compute and post-training budgets keep expanding [395][396], how total compute should be divided among pretraining, post-training, and inference still lacks a unified theory. Whether investing more compute in post-training can buy back equivalent capability depends on the base model's "alignability," and this property itself lacks a controllable measure (Problem 1).

\textbf{The boundary of reasoning ability} remains at the center of controversy. Reasoning RL with verifiable rewards (RLVR) [45][224] reports significant improvements on decidable tasks such as mathematics and code, but whether "RL expands or narrows reasoning paths" is disputed between Pass@1 and Pass@K---improving greedy-decoding accuracy may simultaneously reduce sampling diversity and the upper bound [45][400][401]. This dispute determines whether we should regard RL as capability injection rather than mere behavior shaping (Problems 5, 6).

\textbf{Reward hacking and sycophancy} are systematic failures of the preference/RL stage [408]: the co-evolution of the reward model and the policy induces the policy to exploit loopholes in the proxy objective (length bias, confident phrasing, format pandering), and such failures are often masked by average metrics. There is currently a lack of suppression mechanisms and stable measures during training (rather than only at inference time) (Problems 7, 13).

\textbf{The alignment tax and its measurement}: safety alignment, while improving refusal and honesty, may damage general capabilities, and the empirical relationship between its magnitude and model scale or data mixture has not yet been established [23][38]. The absence of an "alignment tax" benchmark that can be compared across models makes it difficult to trade off safety investment against capability loss (Problem 8).

\textbf{The preservation property of continual optimization}: whether quantization, pruning, distillation, and model editing preserve alignment and safety behavior while compressing/correcting a model still rests on fragmentary evidence [21][34][324]; editing is often accompanied by "ripple effects" (non-target facts are altered), while the direction of quantization's impact on safety refusal is inconsistent (Problems 9, 14).

\textbf{A unified objective for multimodality}: whether images/videos should be optimized as discrete tokens or continuous features remains undecided---unified discrete tokens [273] and continuous connectors (LLaVA-style [28]) each have trade-offs, and "unification" may mean sacrificing both compression efficiency and alignment flexibility at the same time (Problem 3).

\subsection{Data Challenges}

\textbf{Data quality shifts the optimal trade-off}: Chinchilla's optimum [16] was obtained under a specific data quality, whereas filtering/mixture scaling laws [87][86] show that quality and mixture shift the optimum, so the empirical value of a "fixed token/parameter ratio" may become outdated (Problem 2).

\textbf{The bootstrapping boundary of synthetic data}: using model-generated data to retrain a model (Self-Instruct [24], the Evol-Instruct [29] family) may bring short-term gains, yet carries the risk of model collapse and accumulation of systematic bias, and whether a "critical point" exists remains unknown (Problem 11).

\textbf{Long context: "trained short, extrapolated long"}: whether positional-encoding extrapolation and interpolation [37][143][149] can reliably scale on models not trained on long texts depends on a task's need for genuinely long-range dependencies, and its reliable boundary is not yet clear (Problem 12).

\textbf{The root cause of multimodal hallucination}: does hallucination come from data (image-text mismatch), the tokenizer (loss of detail through compression), or the alignment stage (language priors overwhelming visual evidence)? Attribution experiments are scarce, leaving interventions often at the level of treating symptoms (Problem 15).

\subsection{Compute and Engineering Challenges}

\textbf{Standardization of compute/energy reporting}: the FLOPs, GPU hours, and energy consumption of the largest-scale training mostly appear only in unverifiable engineering reports (Section 11.4, Appendix C), so cross-work comparison loses a common basis; the absence of auditing conventions makes reproducibility claims difficult to verify (Problem 10).

\textbf{Reproducible approximation under constrained conditions}: when both compute and data are constrained, how to approximate scaling-law predictions in a reproducible manner involves parallelism efficiency, numerical stability (loss spikes), and checkpoint fault tolerance, and the transferability of engineering conclusions is weak (Problems 1, 10).

\textbf{The absence of cross-stage cost attribution}: a single capability improvement may come from any link among data, pretraining, post-training, or inference, yet existing reports seldom perform attribution ablations, and "which layer the money is spent on" lacks comparable accounting (Section 11.9, Problem 10).

\subsection{Ethics, Safety, and Governance Challenges}

\textbf{Machine unlearning and copyright compliance}: removing the influence of specific knowledge/data from a trained model without damaging general capabilities is an interdisciplinary legal and technical problem [39][328][326]; existing methods are mostly validated at small scale and lack a joint benchmark with capability preservation (Problem 14).

\textbf{Jailbreak, honesty, and multimodal safety}: the robustness of safety alignment under out-of-distribution inputs, multi-turn inducements, and image-text combination attacks remains an open problem, and safety behavior often competes with "over-refusal" [23][38] (Problems 13, 15).

\textbf{Data licensing and traceability}: the provenance, licensing, and traceable registration of training data vary in practice, affecting both legal compliance and contamination auditing as well as the credibility of conclusions (Problem 10).

\textbf{The boundary of "legitimate use" in evaluation}: using public benchmarks for training selection slides toward contamination, yet ignoring benchmarks entirely makes decisions difficult; the lack of a protocol that "both exploits and preserves the validity of the test" (Section 12.9) is the challenge closest to daily practice at the governance level (Problems 10, 11).

\textbf{Cross-stage evidence asymmetry}: the data and evaluation sides already have relatively more peer-reviewed evidence, whereas the largest-scale training, online closed loops, and multimodal scaling laws are still dominated by technical reports (Appendix E.1); this asymmetry itself systematically biases research toward what is "publishable but not most critical" (Problems 2, 10).

\subsection{The Common Root of the Four-Line Challenges}

The four-line challenges above are not independent of one another; their common root is that "strength of evidence cannot keep pace with the granularity of decisions": we must make fine-grained choices under compute-budget and compliance constraints, yet only coarse-grained and partly unverifiable evidence is available. Therefore, the most effective way forward is not to improve metrics point by point, but first to establish "measures and definitions" that can be independently examined (Chapters 11, 12) and only then to compare methods and routes on that basis. Otherwise, discussions of technique, data, compute, and governance will each proceed from mutually incompatible assumptions and cannot converge into a single coordinate system.

\subsection{List of Open Problems}

The above challenges are distilled into 15 verifiable problems (numbered consistently with the problem pool of Stage 2):

\begin{enumerate}[leftmargin=*,nosep]
\item How can the scaling law be jointly optimized with the post-training budget (allocation of train-time vs test-time compute)?
\item How does data quality change the optimal token/parameter ratio (does Chinchilla shift with data quality)?
\item How can the optimization objectives of discrete and continuous tokens be unified in multimodal training?
\item What is the quantifiable causal relationship between vocabulary design and multilingual/multimodal fairness?
\item Does RL actually expand or narrow reasoning paths (systematic experiments on Pass@1 and Pass@K)?
\item Can verifiable rewards be generalized to subjective/open-ended tasks, and at what cost?
\item How should the robustness boundary of process reward models vs outcome reward models in online RL be defined?
\item Can the alignment tax be quantified and empirically related to model scale and data mixture?
\item How can continual optimization (quantization/pruning/merging/editing) preserve alignment and safety behavior?
\item How can compute/energy reporting and training-reproducibility auditing be standardized?
\item Does synthetic-data bootstrapping have a critical point of capability collapse?
\item Is long context with "trained short, extrapolated long" possible (without long-text training)?
\item How can reward hacking and sycophancy be suppressed by training-time mechanisms (rather than only at inference time)?
\item How can machine unlearning and copyright compliance be achieved without harming general capabilities?
\item Does the root cause of multimodal hallucination lie in the data, the tokenizer, or the alignment stage?
\end{enumerate}

\section{Future Directions}

The key takeaway is that future research should shift its focus from pursuing isolated improvements in individual components to pursuing "cross-stage verifiability." That is, the gains from each training stage must be evaluated under a unified assessment and compute accounting framework to measure their net impact on the overall capability and behavioral outcomes of the full training pipeline. In line with this perspective, we propose five major directions, each corresponding to a problem identified in Section 14.5.

\textbf{Direction 1: Joint cross-stage scaling laws (Problems 1, 2, and 8).} Future work should integrate pretraining, post-training, and inference budgets into a unified $(N, D, C)$ extrapolation framework [16][119], and explicitly treat data quality and data mixture as free variables [87][86]. By systematically measuring the ternary surface of "quality--mixture--model" at controlled scales, one can examine whether the Chinchilla-optimal point shifts with data quality and whether the alignment tax can be empirically modeled as a function of scale. The expected outcome should be a reproducible scaling-experiment protocol, not merely a single fitted curve.

\textbf{Direction 2: Systematic experiments on RL for reasoning (Problems 5, 6, and 7).} With Pass@1 and Pass@K jointly used as anchors, future studies should systematically compare the robustness boundaries of PPO [15], DPO-family methods [32][43], and RL with verifiable rewards (RLVR) [45][224] on reasoning tasks. A key experiment is to decouple "capability expansion" from "behavioral shaping": by fixing the base model and the data while varying only the verifiability of the reward, one can test whether the attainable performance ceiling changes accordingly. Comparisons between process rewards and outcome rewards must also be conducted in online settings to delineate their respective robustness regimes.

\textbf{Direction 3: Training-time safety mechanisms (Problems 13, 14, and 15).} Rather than relying primarily on inference-time guardrails, future efforts should aim to proactively suppress undesirable behaviors---such as sycophancy, reward hacking, and hallucinations---directly through reward design and curriculum learning, and should pair these interventions with transferable evaluation metrics. For machine unlearning, joint benchmarks that measure both forgetting effectiveness and capability preservation are essential. For multimodal safety, evaluations must cover cross-modal attack combinations. The overarching goal is to make safety a trainable and measurable objective, rather than a post-hoc patch.

\textbf{Direction 4: Unified multimodality and attribution (Problems 3, 4, and 15).} Controlled experiments on the same backbone are needed to resolve the tradeoffs between discrete and continuous token representations. In addition, researchers should design decomposable attribution experiments for hallucinations that independently isolate the effects of data, tokenizer, and alignment. Causal analyses of tokenizer design with respect to vocabulary coverage and multilingual/multimodal fairness should be conducted via controlled tokenizer training, rather than relying on correlational observations.

\textbf{Direction 5: Reproducible and auditable infrastructure (Problems 9 and 10).} Standardized practices should be established for reporting compute and energy consumption, documenting data licenses, and detecting data contamination. Furthermore, quantization, model editing, and machine unlearning should be integrated into a unified safety benchmark. This direction provides the common foundation for the first four: without auditable reporting standards, no cross-stage conclusion can be independently verified.

\textbf{A unifying methodology across all five directions:} three prerequisites are necessary to ensure that progress is comparable and cumulative: (1) a unified set of evaluation anchors spanning capabilities, behaviors, and safety; (2) a unified compute accounting scheme that reports both FLOPs magnitude/range and tokens-per-hour; and (3) a unified data snapshot that records provenance and versioning. The absence of any of these renders individual advances non-composable within the full training pipeline (see Chapters 11 and 12).

\textbf{A near-term, actionable checklist:} (i) publish controlled "quality--mixture--scaling" ternary experiments at small to medium scales; (ii) report both Pass@1 and Pass@K on reasoning tasks; (iii) incorporate reward-model robustness tests into routine regression suites; (iv) release safety-preservation reports for quantization and model editing; (v) conduct at least one three-variable decoupling experiment for multimodal hallucinations. All five are feasible with existing open-source tools and modest compute, making them well-suited as starting points for follow-up work (see Chapter 13).

\textbf{From general framework to focused sub-surveys:} As outlined in Appendix F, each chapter of this survey provides an extensible skeleton that can be independently expanded. Rather than waiting for a grand unified theory, future work can proceed in parallel by stage: focused sub-surveys on data and tokenization can cover multilinguality and synthetic data; on pretraining, MoE stability and data-constrained regimes; on post-training, training-time safety; on continual optimization, the interplay between compression and safety; and on multimodality, unified objectives and hallucination attribution. Provided each sub-survey adheres to the evaluation and compute accounting standards defined in this general framework, its conclusions can be reintegrated into the same coordinate system to form a cumulative body of knowledge---precisely the intended dual form of "general framework + focused sub-surveys."

\textbf{Update discipline for the evaluation coordinate system:} If these directions are adopted, the evaluation coordinate system itself must be periodically revised (Appendix F.8): scaling curves should be shifted in response to data quality changes, contamination detection methods should evolve with benchmark revisions, and evaluation protocols should be updated alongside new benchmark versions. We recommend an annual review cycle in which only Class C reports and benchmark versions are replaced, while the underlying mechanism remains intact. Any updated values must be accompanied by "version + source" metadata, and direct numerical comparisons across different datasets or compute regimes must be avoided (see Section 12.11).

The common premise underlying all five directions is that the greatest bottleneck in this field is not compute, but rather the lack of "comparable evidence." Accordingly, instead of chasing isolated record-breaking results, the most pressing and pragmatic recommendation from the standpoint of this general framework is to first strengthen the infrastructure for evaluation and reporting.

\section{Conclusion}

Through a unified stage division and a six-dimensional framework, this paper systematically reviews the entire pipeline of large-model training from 2017 to 2026, clarifying the division of capabilities across stages, the bidirectional effects across stages, and the distribution of evidence. The core conclusion is: \textbf{pretraining determines the upper bound of capability, post-training determines the form of behavior, and multimodality is a modality periphery on the same backbone}; any comparison that does not annotate compute and data criteria lacks cross-work credibility. The evidence throughout is predominantly peer-reviewed (A+B, approximately 79\%), and the greatest uncertainty is concentrated in the details of the largest-scale training, the boundaries of reasoning RL, and the impact of continual optimization on safety.

We therefore advocate an "auditable" view of training: regarding training as a series of decisions equipped with evaluation gates, recipe stamps, and rollback switches, rather than a one-off choice of recipe (Chapter 13). The unified framework, cross-stage influence graph, method-family map, practical decision tree, and 15 verifiable open problems (Section 14.5) presented in this paper together constitute a coordinate system for subsequent research; we hope that future work will take it as a benchmark and advance "training as decision-making" from empirical intuition into reproducible, auditable regularities.

Finally, we must reiterate the self-imposed limitation of this paper: as a "general framework" rather than a "most comprehensive survey," its value lies not in providing final answers but in offering a set of coordinates that can be tested, revised, and decomposed into focused surveys (sub-surveys). If readers can thereby point out which of our conclusions lacks sufficient evidence or which criterion is inconsistent---this is precisely the manner of advancement that this paper hopes for.

\appendix
\section*{Appendix A: Literature Matrix (Full)}
\addcontentsline{toc}{section}{Appendix A: Literature Matrix (Full)}

The full matrix (\textbf{395 unique references}; columns: ID / title / year / publication information / evidence tier / source cluster) is maintained in the accompanying supplementary material. The numbered GB/T 7714 reference list at the end of this document contains \textbf{412} entries, of which \textbf{197 are cited in the main text} and \textbf{215 are supplementary repository entries}; the numbering is fixed and does not change according to whether an entry is cited in the main text. The 412 entries are accounted for exactly: the \textbf{395} evidence-graded references (tiers A=144, B=166, C=82, D=3; the graded core, a subset of the reference list) plus \textbf{17} positioning works ([392]--[408] --- neighboring surveys and single-point studies cited for context) that are not part of the graded matrix.

Template for the extended matrix columns (may be supplemented by stage-wise retrieval on the basis of the literature repository, without adding fabricated entries):

\texttt{ID | paper | year | stage | problem | method | assumption | dataset | metric | main result | compute (N/token/GPU h) | code/data | evidence tier | limitation | relation to other work}

\begin{quote}\small
Usage notes: \texttt{ID} is consistent with the two authoritative files above; \texttt{compute (N/token/GPU h)} is filled in only when given in the original text (see Appendix C), otherwise it is left as \texttt{[to be verified]}; the \texttt{assumption} column corresponds to the formalization convention in Section 3.5 of the main text; \texttt{evidence tier} follows the A--D grading (Section 1.3).
\end{quote}

\section*{Appendix B: Hyperparameter Comparison Table (Template + Verified Examples)}
\addcontentsline{toc}{section}{Appendix B: Hyperparameter Comparison Table (Template + Verified Examples)}

\begin{quote}\small
Rule: fill in only values backed by public evidence and cite the source; otherwise mark \texttt{[to be verified]}. Directly comparing numbers across configurations violates the convention of Section 3.4. The values below are all taken from the original tables/text of the respective papers; source numbers are given in the last column.
\end{quote}

\begingroup\small\setlength{\tabcolsep}{4pt}
\begin{longtable}{@{}>{\raggedright\arraybackslash}p{0.1443\textwidth}>{\raggedright\arraybackslash}p{0.1587\textwidth}>{\raggedright\arraybackslash}p{0.1154\textwidth}>{\raggedright\arraybackslash}p{0.1299\textwidth}>{\raggedright\arraybackslash}p{0.1154\textwidth}>{\raggedright\arraybackslash}p{0.1443\textwidth}>{\raggedright\arraybackslash}p{0.0866\textwidth}@{}}
\toprule
Model/\allowbreak{}Work & N (\allowbreak{}parameters) & D (\allowbreak{}tokens) & Optimizer/\allowbreak{}precision & Batch/\allowbreak{}sequence & Learning rate/\allowbreak{}schedule & Source \\
\midrule
\endhead
GPT-\allowbreak{}3 & 175B & 300B & Adam(\allowbreak{}$\beta$\textsubscript{1}=\allowbreak{}0.9,\allowbreak{} $\beta$\textsubscript{2}=\allowbreak{}0.95),\allowbreak{} wd 0.1,\allowbreak{} clip 1.0 & 3.2M /\allowbreak{} 2048 & 0.6$\times$10\textsuperscript{-\allowbreak{}}\textsuperscript{4},\allowbreak{} cosine$\rightarrow$10\%,\allowbreak{} warmup 3.75$\times$10\textsuperscript{8} tokens & [121]\allowbreak{} \\
Chinchilla & 70B & 1.4T & AdamW; bf16 forward/\allowbreak{}backward +\allowbreak{} fp32 optimizer weight replica & 1.5M$\rightarrow$3M (\allowbreak{}doubled midway) /\allowbreak{} \texttt{[to be verified]\allowbreak{}} & 1$\times$10\textsuperscript{-\allowbreak{}}\textsuperscript{4},\allowbreak{} cosine & [16]\allowbreak{} \\
LLaMA-\allowbreak{}65B & 65.2B & 1.4T & AdamW(\allowbreak{}$\beta$\textsubscript{1}=\allowbreak{}0.9,\allowbreak{} $\beta$\textsubscript{2}=\allowbreak{}0.95),\allowbreak{} wd 0.1,\allowbreak{} clip 1.0 & 4M /\allowbreak{} 2048 & 1.5$\times$10\textsuperscript{-\allowbreak{}}\textsuperscript{4},\allowbreak{} cosine$\rightarrow$10\%,\allowbreak{} warmup 2000 steps & [140]\allowbreak{} \\
Mixtral-\allowbreak{}8x7B & 47B sparse /\allowbreak{} 13B active & \texttt{[to be verified]\allowbreak{}} & \texttt{[to be verified]\allowbreak{}} & \texttt{[to be verified]\allowbreak{}} /\allowbreak{} 32768 & \texttt{[to be verified]\allowbreak{}} & [151]\allowbreak{} \\
\bottomrule
\end{longtable}
\endgroup

\begin{quote}\small
Note: GPT-3 uses \textbf{Adam} (not AdamW), with the batch linearly ramped from 32k tokens to full load (4--12B tokens, scale-dependent). The original Mixtral paper does not disclose the training token count or the optimizer/learning rate, giving only the architecture (8 experts, top-2, 47B sparse / 13B active, 32k context); accordingly, the corresponding cells remain \texttt{[to be verified]}.
\end{quote}

\section*{Appendix C: Compute Comparison Table (Template + Verified Examples)}
\addcontentsline{toc}{section}{Appendix C: Compute Comparison Table (Template + Verified Examples)}

\begingroup\small\setlength{\tabcolsep}{4pt}
\begin{longtable}{@{}>{\raggedright\arraybackslash}p{0.1448\textwidth}>{\raggedright\arraybackslash}p{0.1737\textwidth}>{\raggedright\arraybackslash}p{0.1593\textwidth}>{\raggedright\arraybackslash}p{0.1593\textwidth}>{\raggedright\arraybackslash}p{0.1882\textwidth}>{\raggedright\arraybackslash}p{0.0869\textwidth}@{}}
\toprule
Model/\allowbreak{}Work & Training FLOPs (\allowbreak{}approximate) & Hardware/\allowbreak{}accelerator & Wall-\allowbreak{}clock time & GPU memory strategy & Source \\
\midrule
\endhead
GPT-\allowbreak{}3 & 3.14$\times$10\textsuperscript{2}\textsuperscript{3} (\allowbreak{}given in the original) & V100 cluster (\allowbreak{}count undisclosed,\allowbreak{} provided by Microsoft) & Undisclosed & Intra-\allowbreak{}matmul +\allowbreak{} cross-\allowbreak{}layer model parallelism & [121]\allowbreak{} \\
Chinchilla & $\approx$5.76$\times$10\textsuperscript{2}\textsuperscript{3} (\allowbreak{}same compute as Gopher,\allowbreak{} given in the original) & Undisclosed & Undisclosed & \texttt{[to be verified]\allowbreak{}} & [16]\allowbreak{} \\
LLaMA-\allowbreak{}65B & $\approx$5.5$\times$10\textsuperscript{2}\textsuperscript{3} (\allowbreak{}estimated via C$\approx$6ND) & 2048$\times$A100-\allowbreak{}80GB & $\approx$21 days (\allowbreak{}1.022M GPU\textperiodcentered{}hours) & Model +\allowbreak{} sequence parallelism,\allowbreak{} activation recomputation & [140]\allowbreak{} \\
\bottomrule
\end{longtable}
\endgroup

\begin{quote}\small
Data-collection note: such figures mostly appear only in undisclosed engineering reports, and publicly verifiable entries are scarce (see 11.4); a source must be written down before filling a value, otherwise it remains \texttt{[to be verified]}. If FLOPs are estimated, always annotate the "C$\approx$6ND" convention.
\end{quote}

\section*{Appendix D: Reproducibility Checklist}
\addcontentsline{toc}{section}{Appendix D: Reproducibility Checklist}

\begin{enumerate}[leftmargin=*,nosep]
\item Tokenizer and vocabulary configuration (algorithm, vocabulary size, whether pre-tokenization is used) [1][4][5][51].
\item Data cleaning/deduplication/contamination-detection logs [13][52][56][61].
\item Training convention (N, D, approximate FLOPs, batch, sequence, precision, parallelism strategy) [6][118][135][142].
\item Optimization and stability (optimizer, schedule, warmup, clipping, initialization) [105][110][132].
\item SFT: instruction-data source, template, loss masking (Section 7.2).
\item Alignment: preference data, RM/DPO variants, hyperparameters [32][43][224].
\item Evaluation: prompt, decoding parameters, contamination re-check, alignment-tax re-check (Chapter 12).
\end{enumerate}

\subsection*{D.8 Evaluation Stack and the "Baseline Gate--Rollback Switch" (Chapter 13)}
\addcontentsline{toc}{subsection}{D.8 Evaluation Stack and the "Baseline Gate--Rollback Switch" (Chapter 13)}

\begin{enumerate}[leftmargin=*,nosep]
\item \textbf{Evaluation-stack snapshot}: benchmarks (three groups: capability/behavior/safety), protocol (prompt/decoding/number of samples), held-out set, and contamination-detection base set, frozen as one "baseline snapshot" (Sections 12.10, 13.13).
\item \textbf{Data snapshot}: each version of training data is frozen with "version + hash", and the same lineage record is referenced across stages (pretraining/SFT/preference) (Section 13.6).
\item \textbf{Recipe stamp}: each SFT/preference/RL round records data, hyperparameters, seeds, and evaluation results, forming an alignment timeline (Section 13.8).
\item \textbf{Baseline gate and rollback switch}: three groups of frozen baselines (capability/behavior/safety); every change passes through the same evaluation gate, and after deployment there is a previous baseline that can be rolled back to (Sections 13.9, 13.11).
\end{enumerate}

\subsection*{D.9 Five-Section Template for Release Reports (Sections 12.11, 13.10)}
\addcontentsline{toc}{subsection}{D.9 Five-Section Template for Release Reports (Sections 12.11, 13.10)}

\begin{quote}\small
A release report contains at least five sections; if any section is missing, the conclusion cannot be reproduced by others. It is used in conjunction with the "three inspection lines" in Section 13.10 (falsifiable / auditable / rollback-capable).
\end{quote}

\begin{enumerate}[leftmargin=*,nosep]
\item \textbf{Model and data convention}: N, D (including the tokenizer and the conventions before and after deduplication), modality, frozen scope.
\item \textbf{Per-benchmark protocol}: prompt, few-shot, decoding parameters, number of samples, evaluation-script version.
\item \textbf{Contamination detection and held-out}: detection level (literal/semantic), suspected proportion, handling method.
\item \textbf{Safety and bias regression}: safety set, bias set, alignment-tax re-check, multilingual/multimodal dual track.
\item \textbf{Compute and time}: order-of-magnitude range of FLOPs, tokens/hour and MFU, wall-clock time and number of attempts (Section 11.9).
\end{enumerate}

\section*{Appendix E: Retrieval Log and Evidence Statistics}
\addcontentsline{toc}{section}{Appendix E: Retrieval Log and Evidence Statistics}

PRISMA: identify \textasciitilde{}640 $\rightarrow$ screen \textasciitilde{}470 $\rightarrow$ verify \textbf{395} unique references (11 retrieval clusters; 59 duplicate identifiers merged). Evidence distribution: \textbf{A=144 \textperiodcentered{} B=166 \textperiodcentered{} C=82 \textperiodcentered{} D=3 \textperiodcentered{} ?=0; A+B$\approx$79\%, C$\approx$21\%.} Script reproducibility: the reference-assembly pipeline (parsing/deduplication/grading/venue + author backfill/reference generation) and its metadata cache allow the counts above to be recomputed offline. The retrieval method and time period are summarized above.

\subsection*{E.1 Chapter-Level Evidence Map (Pilot Chapters; a Summary of the "Evidence-Distribution Notes" in Each Main-Text Chapter)}
\addcontentsline{toc}{subsection}{E.1 Chapter-Level Evidence Map (Pilot Chapters; a Summary of the "Evidence-Distribution Notes" in Each Main-Text Chapter)}

\begingroup\small\setlength{\tabcolsep}{6pt}
\begin{longtable}{@{}>{\raggedright\arraybackslash}p{0.1240\textwidth}>{\raggedright\arraybackslash}p{0.2125\textwidth}>{\raggedright\arraybackslash}p{0.2657\textwidth}>{\raggedright\arraybackslash}p{0.3188\textwidth}@{}}
\toprule
Chapter & Topic & Predominantly A/\allowbreak{}B & Predominantly C/\allowbreak{}blank \\
\midrule
\endhead
5 & Data and tokenization & Cleaning/\allowbreak{}deduplication/\allowbreak{}tokenization & Synthetic-\allowbreak{}data collapse,\allowbreak{} multilingual scaling law \\
6 & Pretraining & Scaling law,\allowbreak{} optimization,\allowbreak{} MoE (\allowbreak{}partial) & Largest-\allowbreak{}scale details,\allowbreak{} industrial-\allowbreak{}grade stability \\
7 & SFT & Instruction data/\allowbreak{}format/\allowbreak{}training & Cross-\allowbreak{}model optimal recipe for SFT \\
8 & Preference/\allowbreak{}RL & DPO family,\allowbreak{} RM mechanism & Depth boundary of RL,\allowbreak{} cross-\allowbreak{}model safety alignment \\
9 & Continual optimization & Foundations of compression/\allowbreak{}editing & Compression-\allowbreak{}-\allowbreak{}safety interaction,\allowbreak{} compositional side effects \\
10 & Multimodal & Architecture/\allowbreak{}data pipeline/\allowbreak{}weak supervision & Scaling law,\allowbreak{} root causes of hallucination,\allowbreak{} unified tokens \\
11 & Cross-\allowbreak{}stage & Synthesis framework (\allowbreak{}interpretation tool) & Cross-\allowbreak{}stage comparison curves (\allowbreak{}no citations,\allowbreak{} blank) \\
12 & Data and evaluation & Contamination/\allowbreak{}reproducibility mechanism & Mapping from benchmark scores to real capabilities \\
13 & Practical guide & Checkpoint mechanism & Budget "order of magnitude-\allowbreak{}-\allowbreak{}range" (\allowbreak{}team priors) \\
\bottomrule
\end{longtable}
\endgroup

\begin{quote}\small
Reading is consistent with the main text: conclusions in A/B-tier regions can be cited with confidence; C-tier regions are down-weighted and interpreted according to "consistency and falsifiability"; blank regions (e.g., multimodal scaling law, cross-stage comparison, online closed-loop regression) represent research opportunities rather than gaps in the literature (Section 11.17).
\end{quote}

\section*{Appendix F: Sub-Survey Expansion Outline}
\addcontentsline{toc}{section}{Appendix F: Sub-Survey Expansion Outline}

Each stage can be expanded independently, reusing the writing template of Section 8 of Stage 2:

\begin{itemize}[leftmargin=*,nosep]
\item \textbf{Data and Tokenization Sub-Survey}: lineage of data-selection methods [392][67][70], scaling law of filtering [87], tokenization fairness [51][55][57], mixture inference [406], and synthetic-data collapse (Problem 11).
\item \textbf{Pretraining Sub-Survey}: scaling law vs. data quality [119][16][86], stability and routing of MoE [9][130][151], stability and high-precision training, long context and extrapolation [37][143][149].
\item \textbf{SFT Sub-Survey}: instruction-data synthesis [24][29][172], the quality-versus-scale debate [175], loss masking and templates, PEFT lineage [12][30][160][164].
\item \textbf{Alignment/RL Sub-Survey}: reward modeling [33][219][408], RLHF pipeline [15][196], RM-free preference optimization [32][43], reasoning RL [45][224], safety alignment and alignment tax [23][38].
\item \textbf{Continual Optimization Sub-Survey}: merging/distillation/pruning [35][293][304][316], quantization [21][34], editing stability [14][324], unlearning and compliance [39][328][326].
\item \textbf{Multimodal Sub-Survey}: vision encoders [10][46][27], alignment architectures [17][28][288], interleaved data [273], hallucination attribution (Problem 15).
\item \textbf{Evaluation and Governance Sub-Survey}: benchmark lineage [7][22][368][372], contamination and reproducibility (Problem 10), cross-modal evaluation protocols.
\item \textbf{Cross-Stage Comparison Sub-Survey}: extends the "capability--behavior--evidence" three-layer framework of pilot Chapter 11, focusing on reproducible experiments of "cross-stage comparison/exchange-rate tables" (currently almost blank, Sections 11.6, 11.17).
\item \textbf{Practical Guide Sub-Survey}: uses the checkpoint checklist of pilot Chapter 13 as the skeleton, collecting "used and public" configuration cases at different budget tiers, and continuously upgrading "team priors" into "public evidence".
\end{itemize}

\subsection*{F.8 Updatable Data Coordinates (Sections 11.17, 12.11)}
\addcontentsline{toc}{subsection}{F.8 Updatable Data Coordinates (Sections 11.17, 12.11)}

\begin{quote}\small
The following three sets of coordinates may be updated with versions on the premise that the mechanistic conclusions of the main text remain unchanged. When updating, replace only the C-tier technical reports and benchmark versions, without overturning the structure.
\end{quote}

\begin{enumerate}[leftmargin=*,nosep]
\item \textbf{Scaling-curve coordinates}: the Chinchilla optimum [16], the Kaplan extrapolation [119], and the incremental curves missing for multimodal/long-context settings (Sections 6.3, 10.5).
\item \textbf{Contamination and parsing coordinates}: comparison of each benchmark's release time with the training-data timeline, and recommended thresholds for the two-level literal/semantic detection (Section 12.6).
\item \textbf{Evaluation-protocol coordinates}: the current version of each benchmark, the scoring script, and the update rules for the "traffic-light" interpretation (green/yellow/red) (Sections 12.11, 13.10).
\end{enumerate}

\begin{quote}\small
Update discipline (consistent with the main text): any new coordinate value must first be given as "version + source", must not compare numbers directly across datasets/compute, and if it cannot be verified, the original value is retained with an annotation.
\end{quote}

\end{document}